\RequirePackage{silence}
\documentclass[twoside,openright,titlepage,numbers=noenddot,headinclude,footinclude,cleardoublepage=empty,abstract=on,BCOR=15mm,paper=a4,fontsize=11pt]{scrreprt}
\usepackage{afterpage}
\usepackage[linesnumbered,ruled,vlined]{algorithm2e}    
\newcommand\myemptypage{
    \null
    \thispagestyle{empty}
    \addtocounter{page}{-1}
    \newpage
    }

\PassOptionsToPackage{utf8}{inputenc}
  \usepackage{inputenc}

\PassOptionsToPackage{T1}{fontenc} 
  \usepackage{fontenc}

\PassOptionsToPackage{
  drafting=false,    
  tocaligned=false, 
  dottedtoc=true,  
  eulerchapternumbers=true, 
  linedheaders=false,       
  floatperchapter=true,     
  eulermath=false,  
  beramono=true,    
  palatino=true,    
}{classicthesis}

\newcommand{\myTitle}{Tailored GNN Architectures for Real-World Challenges\xspace}

\newcommand{\myName}{Yassine ABBAHADDOU\xspace}

\newcommand{\myFaculty}{Put data here\xspace}

\newcommand{\myUni}{École Polytechnique\xspace}

\newcommand{\myVersion}{\classicthesis}

\providecommand{\mLyX}{L\kern-.1667em\lower.25em\hbox{Y}\kern-.125emX\@}
\newcommand{\ie}{i.\,e.}

\newcommand{\eg}{e.\,g.}

\PassOptionsToPackage{american,french}{babel} 
    \usepackage{babel}

\usepackage{csquotes}
\PassOptionsToPackage{%
  backend=biber,bibencoding=utf8, 
  backend=bibtex8,bibencoding=ascii,%
  language=auto,%
  style=numeric-comp,%
  style=authoryear-comp, 
  bibstyle=authoryear,dashed=false, 
  sorting=nyt, 
  maxbibnames=10, 
  dashed=true,
  backref=true,%
  natbib=true 
}{biblatex}
\usepackage{biblatex}

  \usepackage{amsmath}

\usepackage{graphicx} %
\usepackage{scrhack} 
\usepackage{xspace} 
\PassOptionsToPackage{printonlyused,smaller}{acronym}
  \usepackage{acronym} 

\usepackage{tabularx} 
\usepackage{subcaption}

\usepackage{listings}
\usepackage{classicthesis}

\hypersetup{%
  colorlinks=true, linktocpage=true, pdfstartpage=3, pdfstartview=FitV,%
  breaklinks=true, pageanchor=true,%
  pdfpagemode=UseNone, %
  plainpages=false, bookmarksnumbered, bookmarksopen=true, bookmarksopenlevel=1,%
  hypertexnames=true, pdfhighlight=/O,
  urlcolor=CTurl, linkcolor=CTlink, citecolor=CTcitation, 
  pdftitle={\myTitle},%
  pdfauthor={\textcopyright\ \myName, \myUni, \myFaculty},%
  pdfsubject={},%
  pdfkeywords={},%
  pdfcreator={pdfLaTeX},%
  pdfproducer={LaTeX with hyperref and classicthesis}%
}

\makeatletter
\@ifpackageloaded{babel}%
  {%
    \addto\extrasamerican{%
    }%
    \addto\extrasngerman{%
    }%
    }{\relax}
\makeatother

\usepackage{makecell}

\usepackage{amssymb}
\usepackage{pifont}

\usepackage{tikz}
\usepackage{amsmath}
\usepackage{physics}

\usepackage{array}
\newcolumntype{L}{>{\arraybackslash}m{12cm}}

\usepackage[utf8]{inputenc}
\usepackage{helvet}

\usepackage{geometry}
\usepackage{xcolor}
\usepackage[absolute,overlay]{textpos}
\usepackage{graphicx}
\usepackage{lipsum}
\usepackage{hyperref}
\usepackage{array}
\usepackage{caption}
\usepackage{multicol}
\usepackage[french]{babel}

\label{form}

\newcommand{\PhDTitle}{Key Principles of Graph Machine Learning: Representation, Robustness, and Generalization} 	
\newcommand{\PhDname}{Yassine Abbahaddou}
\newcommand{\NNT}{2025IPPAX053} 	
\newcommand{\ecodoctitle}{l'\'Ecole Doctorale de l'Institut Polytechnique de Paris\\} 							
\newcommand{\ecodocacro}{ED IP Paris}	
\newcommand{\ecodocnum}{626 : } 	
\newcommand{\PhDspeciality}{Mathématiques et Informatique} 		
\newcommand{\PhDworkingplace}{l'\'Ecole Polytechnique} 					
\newcommand{\defenseplace}{Palaiseau} 	
\newcommand{\defensedate}{9 juillet 2025}

		\newcommand{\jurynameA}{Marc Lelarge}
\newcommand{\juryadressA}{Researcher, INRIA}
\newcommand{\juryroleA}{Rapporteur}
\newcommand{\jurynameE}{Thomas Gärtner}
\newcommand{\juryadressE}{Professor,  Technische Universität Wien}
\newcommand{\juryroleE}{Rapporteur}
\newcommand{\jurynameB}{Michalis Vazirgiannis}
\newcommand{\juryadressB}{Full Professor, \'Ecole Polytechnique (LIX)}
\newcommand{\juryroleB}{Directeur de thèse}
\newcommand{\jurynameD}{Johannes F. Lutzeyer}
\newcommand{\juryadressD}{Assistant Professor, \'Ecole Polytechnique (LIX) }
\newcommand{\juryroleD}{Co-encadrant de thèse}
\newcommand{\jurynameC}{Fragkiskos D. Malliaros}
\newcommand{\juryadressC}{Associate Professor, CentraleSupelec (CVN) }
\newcommand{\juryroleC}{Co-encadrant de thèse}
\newcommand{\jurynameF}{Céline Hudelot}
\newcommand{\juryadressF}{Professor, CentraleSupélec}
\newcommand{\juryroleF}{Presidente, Examinatrice}
\newcommand{\jurynameG}{Charlotte Laclau}
\newcommand{\juryadressG}{Professor, Télécom Paris}
\newcommand{\juryroleG}{Examinatrice}
\newcommand{\jurynameH}{Davide Bacciu}
\newcommand{\juryadressH}{Professor, Università di Pisa}
\newcommand{\juryroleH}{Examinateur}
\label{layout}
\hypersetup{
	pdfauthor={\PhDname},
	pdfsubject={Manuscrit de thèse de doctorat},
	pdftitle={\PhDTitle},
}

\usepackage{lettrine}
\usepackage{nicefrac}
\usepackage{multirow}

\usepackage{enumitem}
\usepackage{amsthm}
\usepackage{mdframed}
\usepackage{rotating}
\usepackage{float}
\usepackage{mathrsfs}
\usepackage{amsmath}
\usepackage{alphabeta}
\newtheorem{theorem}{Theorem}[section]
\newtheorem{proposition}[theorem]{Proposition}
\newtheorem{lemma}[theorem]{Lemma}

\newtheorem{definition}[theorem]{Definition}

\newtheorem{remark}[theorem]{Remark}
\usepackage[utf8]{inputenc}
\usepackage{arydshln}

\usepackage{amsmath}
\usepackage{amssymb} 
\usepackage{tikz} 

\newcommand*\circled[1]{\tikz[baseline=(char.base)]{
            \node[shape=circle,draw,inner sep=1pt] (char) {#1};}}

\DeclareMathOperator*{\argmax}{arg\,max}
\DeclareMathOperator*{\argmin}{arg\,min}

\begin{document}
\frenchspacing
\raggedbottom
\selectlanguage{american} 
\pagenumbering{roman}
\pagestyle{plain}



\thispagestyle{empty}

\color{black} \hfill \vfill \ecodocnum
\begin{textblock}{5}(0,0)
	\textblockcolour{black}
	\includegraphics [scale=0.95]{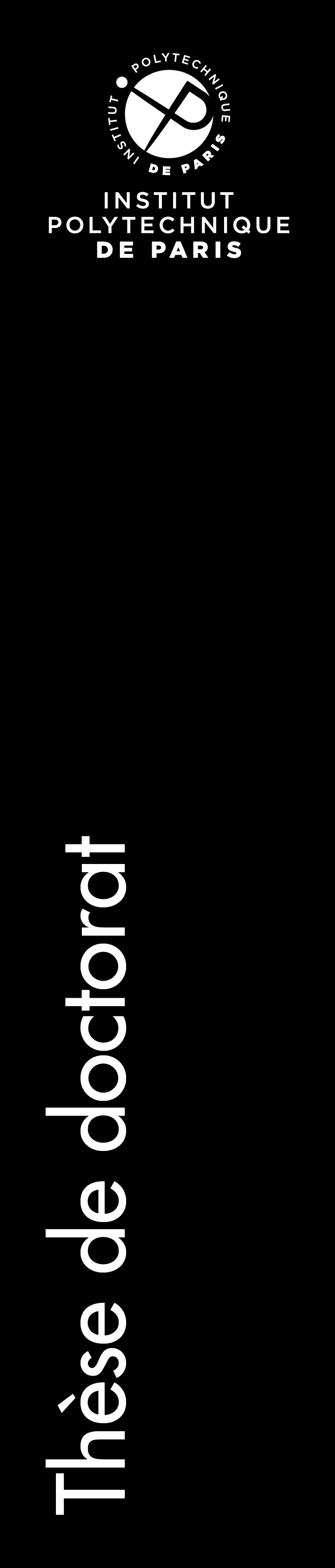}
	\vspace{300mm}
\end{textblock}

\begin{textblock}{1}(0.6,3)
	\Large{\rotatebox{90}{\color{white}{\textbf{NNT : \NNT}}}}
\end{textblock}



\begin{textblock}{10}(5.7,3)
	\textblockcolour{white}
	
	\color{black}
	\begin{flushright}
		\huge{\PhDTitle} \bigskip 
		\vfill
		\color{black} 
		\normalsize {Thèse de doctorat de l'Institut Polytechnique de Paris} \\
		préparée à \PhDworkingplace \\ \bigskip
		\vfill
		École doctorale n$^{\circ}$\ecodocnum ~\ecodoctitle ~(\ecodocacro)  \\
		
		\small{Spécialité de doctorat: \PhDspeciality} \bigskip 
		\vfill  
		\footnotesize{Thèse présentée et soutenue à \defenseplace, le \defensedate, par} \bigskip
		\Large{\textbf{\textsc{\PhDname}}} 
		\vfill
	\end{flushright}
	
	\color{black}
	\begin{flushleft}
		
		\small Composition du Jury :
	\end{flushleft}

	\small
	\newcolumntype{L}[1]{>{\raggedright\let\newline\\\arraybackslash\hspace{0pt}}m{#1}}
	\newcolumntype{R}[1]{>{\raggedleft\let\newline\\\arraybackslash\hspace{0pt}}lm{#1}}
	
	\label{jury} 																				
	\begin{flushleft}
	\begin{tabular}{@{} L{4.5cm} R{4.5cm}}
		\jurynameF  \\ \juryadressF & \juryroleF \\[5pt]
		\jurynameG  \\ \juryadressG & \juryroleG \\[5pt]
		\jurynameH  \\ \juryadressH & \juryroleH \\[5pt]
		\jurynameE  \\ \juryadressE & \juryroleE \\[5pt]
            \jurynameA  \\ \juryadressA & \juryroleA \\[5pt]
		\jurynameB  \\ \juryadressB & \juryroleB \\[5pt]
		\jurynameC  \\ \juryadressC & \juryroleC \\[5pt]
		\jurynameD  \\ \juryadressD & \juryroleD \\[5pt]

	\end{tabular} 
	\end{flushleft}   
\end{textblock}

\pdfbookmark[1]{Abstract}{Abstract}
\begingroup
\let\clearpage\relax
\let\cleardoublepage\relax
\let\cleardoublepage\relax

\chapter*{Abstract}
Graph Neural Networks (GNNs) have emerged as powerful tools for learning representations from structured data. Despite their growing popularity and success across various applications, GNNs encounter several challenges that limit their performance. in their generalization, robustness to adversarial perturbations, and the effectiveness of their representation learning capabilities. In this dissertation, I investigate these core aspects through three main contributions: (1) developing new representation learning techniques based on Graph Shift Operators (GSOs, aiming for enhanced performance across various contexts and applications, (2) introducing generalization-enhancing methods through graph data augmentation, and (3) developing more robust GNNs by leveraging orthonormalization techniques and noise-based defenses against adversarial attacks. By addressing these challenges, my work provides a more principled understanding of the limitations and potential of GNNs.
\endgroup

\vfill
\cleardoublepage
\begingroup
\let\clearpage\relax
\let\cleardoublepage\relax
\let\cleardoublepage\relax

\begin{otherlanguage}{french}
    \chapter*{Résumé}

Les réseaux de neurones de graphes (GNNs) sont devenus des outils puissants pour l’apprentissage de représentations à partir de données structurées. Malgré leur popularité croissante et leur succès dans diverses applications, les GNNs rencontrent plusieurs défis qui limitent leurs performances, notamment en termes de généralisation, de robustesse aux perturbations adversariales. Dans cette thèse, j’examine ces aspects fondamentaux à travers trois principales contributions.

La première contribution se concentre sur l'amélioration de l'apprentissage des représentations. J'introduis une nouvelle famille d'opérateurs de décalage de graphe (GSOs) appelés Opérateurs de Décalage de Graphe basés sur la Centralité (CGSOs). Contrairement aux GSOs traditionnels qui se basent principalement sur des informations locales comme le degré des nœuds, les CGSOs intègrent des métriques de centralité globale, telles que le PageRank, le k-core, pour normaliser la matrice d'adjacence. Cette approche enrichit la représentation du graphe avec une information structurelle globale tout en préservant sa sparsité,  un atout majeur pour l'efficacité calculatoire sur de grands réseaux. Une analyse spectrale approfondie de ces opérateurs révèle leurs propriétés fondamentales et leur impact sur la diffusion de l'information. De plus, je propose une seconde méthode, ADMP-GNN, qui s'attaque à la limitation des GNNs standards utilisant le même nombre d'operations pour tous les nœuds. En démontrant empiriquement que la profondeur optimale varie selon les caractéristiques locales des nœuds, cet approche ajuste dynamiquement le nombre de couches de propagation pour chaque nœud, améliorant ainsi la précision prédictive.

La deuxième contribution vise à améliorer la généralisation des GNNs, en particulier pour la classification de graphes face à des données limitées ou hors-distribution (OOD). Je développe d'abord un cadre théorique basé sur la complexité de Rademacher pour analyser l'impact de l'augmentation de données sur l'erreur de généralisation. Ce cadre montre que la maîtrise de la distance entre les représentations originales et augmentées est cruciale. Sur cette base, je propose GRATIN, une nouvelle technique d'augmentation qui opère dans l'espace latent des représentations de graphes. En utilisant des Modèles de Mélange Gaussien (GMMs) pour modéliser la distribution des graphes de chaque classe, GRATIN génère de nouveaux échantillons synthétiques qui enrichissent l'ensemble d'entraînement. Cette approche se distingue par son efficacité calculatoire et sa capacité à améliorer significativement la performance de généralisation.

La troisième contribution porte sur le développement de GNNs plus robustes. D'une part, j'établis un lien théorique entre la robustesse attendue d'un GNN et l'orthonormalité de ses matrices de poids, ce qui conduit à la proposition de GCORN (Graph Convolutional Orthonormal Robust Networks). Ce modèle, une variante robuste du GCN, intègre une contrainte d'orthonormalisation durant l'entraînement pour améliorer sa résistance aux attaques sur les attributs des nœuds. D'autre part, je propose RobustCRF, une méthode de défense post-hoc et agnostique du modèle, qui opère au stade de l'inférence. Cette approche innovante peut ainsi renforcer n'importe quel GNN pré-entraîné sans modification architecturale. En s'appuyant sur un Champ Aléatoire Conditionnel (CRF), cette technique affine les prédictions du modèle en exploitant l'hypothèse que des graphes structurellement similaires dans le voisinage d'une instance donnée devraient produire des prédictions similaires, corrigeant ainsi les erreurs induites par des perturbations adverses sans nécessiter un réentraînement.

En relevant ces défis, mon travail apporte une compréhension plus rigoureuse des limites et du potentiel des GNNs, tout en proposant des solutions pratiques et théoriquement fondées pour améliorer leur performance, leur généralisation et leur robustesse.

Au-delà des contributions méthodologiques, la thèse insiste sur la reproductibilité, des analyses d’ablation et une étude empirique sur des jeux de données hétérogènes (nœuds/graphes) pour éclairer les compromis expressivité–coût. Elle ouvre des pistes en bio-informatique, recommandation et sécurité, et discute des questions éthiques liées à la manipulation de graphes.
\end{otherlanguage}

\endgroup

\vfill
\cleardoublepage
\pdfbookmark[1]{Publications}{publications}
\chapter*{Publications}
\noindent
Several contributions presented in this dissertation relate to the following peer-reviewed articles:

\begin{itemize}
    \item Abbahaddou, Y., Ennadir, S., Lutzeyer, J. F., Vazirgiannis, M., \& Bostr{\"o}m, H. (2024). Bounding the Expected Robustness of Graph Neural Networks Subject to Node Feature Attacks. In \textit{The Twelfth International Conference on Learning Representations} (ICLR).

    \item Abbahaddou, Y., Malliaros, F. D., Lutzeyer, J. F., Aboussalah, A. M., \& Vazirgiannis, M. (2025). Graph Neural Network Generalization With Gaussian Mixture Model Based Augmentation. In \textit{The Forty-Second International Conference on Machine Learning} (ICML).

    \item Abbahaddou, Y., Ennadir, S., Lutzeyer, J. F., Vazirgiannis, M., \& Malliaros, F. D. (2024). Rethinking Robustness in Graph Neural Networks: A Post-Hoc Approach With Conditional Random Fields. \textit{arXiv preprint arXiv:2411.05399}.

    \item Abbahaddou, Y., Malliaros, F. D., Vazirgiannis, M., \& Lutzeyer, J. F. (2024). Centrality Graph Shift Operators for Graph Neural Networks. \textit{arXiv preprint arXiv:2411.04655}.
    
    \item Abbahaddou, Y., Malliaros, F. D., Lutzeyer, J. F., Aboussalah, A. M., \& Vazirgiannis, M. (2024). Gaussian Mixture Models Based Augmentation Enhances GNN Generalization. \textit{arXiv preprint arXiv:2411.08638}.

    \item Abbahaddou, Y., Ennadir, S., Lutzeyer, J. F., Malliaros, F. D., \& Vazirgiannis, M. (2024). Post-Hoc Robustness Enhancement in Graph Neural Networks with Conditional Random Fields. \textit{arXiv preprint arXiv:2411.05399}.
\end{itemize}


\begin{center}
    \noindent\rule{6cm}{0.4pt}  
    \\  
\end{center}

\noindent

The following publication is related, but is not extensively discussed in this thesis:

\begin{itemize}
\item Ennadir, S., Abbahaddou, Y., Lutzeyer, J. F., Vazirgiannis, M., \& Bostr{\"o}m, H. (2024). A Simple and Yet Fairly Effective Defense for Graph Neural Networks. \textit{Proceedings of the AAAI Conference on Artificial Intelligence}, \textbf{38}(19), 21063--21071.

\item Abbahaddou, Y., Lutzeyer, J., \& Vazirgiannis, M. (2023). Graph Neural Networks on Discriminative Graphs of Words. In \textit{NeurIPS 2023 Workshop: New Frontiers in Graph Learning}.

\item Ennadir, S., Smirnov, O., Abbahaddou, Y., Cao, L., \& Lutzeyer, J. F. (2025). Enhancing Graph Classification Robustness with Singular Pooling. In \textit{The Thirty-ninth Annual Conference on Neural Information Processing Systems} (NeurIPS).

\item Abbahaddou, Y., \& Aboussalah, A. M. (2025). A Geometry-Aware Metric for Mode Collapse in Multivariate Time Series Generative Models. In \textit{The Thirty-ninth Annual Conference on Neural Information Processing Systems} (NeurIPS).
\end{itemize}


\newcommand{\G}{\mathcal{G}}
\newcommand{\E}{\mathcal{E}}
\newcommand{\V}{\mathcal{V}}
\newcommand{\Y}{\mathcal{Y}}
\newcommand{\n}{n}
\newcommand{\m}{m}
\newcommand{\neigb}{\mathcal{N}}
\newcommand{\adj}{\mathbf{A}}
\newcommand{\X}{\mathbf{X}}
\newcommand{\K}{K}
\newcommand{\degr}{deg}
\newcommand{\numlayers}{L}
\newcommand{\llayer}{\ell}
\newcommand{\readout}{\texttt{READOUT}}
\newcommand{\mlp}{\texttt{MLP}}
\newcommand{\combine}{\texttt{COMBINE}}
\newcommand{\aggregate}{\texttt{AGGREGATE}}
\newcommand{\Gset}{\mathscr{G}}
\newcommand{\Lgso}{\mathbf{L}}
\newcommand{\Lsym}{\mathbf{L}_{\text{\textit{sym}}}}
\newcommand{\Lrw}{\mathbf{L}_{\text{\textit{rw}}}}
\newcommand{\Qgso}{\mathbf{Q}}
\newcommand{\Hgso}{\mathbf{H}}
\newcommand{\Atilde}{\mathbf{\hat{A}}}
\newcommand{\identity}{\mathbf{I}}
\newcommand{\D}{\mathbf{D}}
\newcommand{\cheeg}{h_C}
\newcommand{\smallestDegree}{\delta}
\newcommand{\largestDegree}{\Delta}
\newcommand{\shortedpathdist}{\delta_{\G}}
\newcommand{\diam}{Diam}
\newcommand{\core}{core}
\newcommand{\method}{GRATIN}
\newcommand{\eqnref}[1]{(\ref{#1})}

\

\pdfbookmark[1]{Acknowledgments}{acknowledgments}

\begingroup
\let\clearpage\relax
\let\cleardoublepage\relax
\let\cleardoublepage\relax
\chapter*{List of Acronyms and Symbols}
\textbf{Common Acronyms in Alphabetic Order}
\begin{itemize}
    \item \textbf{AMI:} Adjusted Mutual Information;
    \item \textbf{ARI:} Adjusted Rand Index;
    \item \textbf{BA:} Barabási–Albert Model;
    \item \textbf{GAT:} Graph Attention Networks;
    \item \textbf{GATv2:} Graph Attention Networks v2;
    \item \textbf{GCN:} Graph Convolutional Network;
    \item \textbf{GIN:} Graph Isomorphism Network;
    \item \textbf{GNN:} Graph Neural Network;
    \item \textbf{GSO:} Graph Shift Operator;
    \item \textbf{GSP:} Graph Signal Processing;
    \item \textbf{PNA:} Principal Neighbourhood Aggregation;
    \item \textbf{SBM:} Stochastic Block Model;
\end{itemize}

\textbf{Common Linear Algebra Symbols and Operations} For some dimension $p \in \mathbb{N}^{*}$, some vectors $a \in \mathbb{R}^{p}$ and $b \in \mathbb{R}^{p}$, and some square matrices $\mathbf{N} \in \mathbb{R}^{p\times p}$and $\mathbf{M} \in \mathbb{R}^{p\times p}$:
\begin{itemize}
    \item $\mathbb{1}_p\text{ or } \mathbb{1}:$ The $p-$dimensional vector containing $p$ entries all equal to 1;
    \item $a_i \in \mathbb{R}:$ The $i-$th element of $a$, for some $i \in \{1,\ldots, p\}$;
    \item $a^\top:$ The transposition of the vector $a$;
    \item $m_{i,j} \in \mathbb{R}$, $\mathbf{M}_{i,j} \in \mathbb{R}$,  or $\mathbf{M}(i,j) \in \mathbb{R}$ the $(i,j)-$th element of $\mathbf{M}$, for some $i,j \in \{1,\ldots, p\}$;
\end{itemize}

\textbf{Common Graph Representation Learning Symbols}
\begin{itemize}
    \item $\G = (\V,\E):$ A graph composed of a node set $\V$ and an edge set $\E$;
    \item $i \in \V:$ A node $i$ from $\G$;
    \item $(i,j)\in\E:$ An edge connecting node $i$ to node $j$ in $\G$;
    \item $\n \in \mathbb{N}^{*}$: The number of nodes in $\G$, i.e., $\n = |\V|$;
    \item $\m \in \mathbb{N}$: The number of edges in $\G$, i.e., $\m = |\E|$;
    \item $\neigb (i) \subseteq  V$: The neighborhood of a node $i \in \V$;
    \item $\identity \in \mathbb{R}^{\n \times \n}:$ The Identity matrix;
    \item $\adj \in [0,1]^{\n \times \n}:$ The Adjacency matrix of $\G$;
    \item $\D\in \mathbb{R}^{\n \times \n}:$ The degree matrix, i.e., a diagonal matrix defined as $\D_{ii} = \sum_{i=1}^n a_{ij}$;
    \item $\widetilde{\adj} \in \mathbb{R}^{\n \times \n}:$ The normalized adjacency matrix of $\G$, i.e., $\widetilde{\adj} = \D^{-1/2} \adj \D^{-1/2}$;
    \item $\Lgso \in \mathbb{R}^{\n \times \n}:$ The Unnormalised Laplacian matrix of $\G$;
    \item $\Qgso \in \mathbb{R}^{\n \times \n}:$ The Signless Laplacian matrix of $\G$;
    \item $\Lrw \in \mathbb{R}^{\n \times \n}:$ The Random-walk Normalised Laplacian of $\G$;
    \item $\Lsym \in \mathbb{R}^{\n \times \n}:$ The Symmetric Normalised Laplacian matrix of $\G$;
    \item $\Atilde \in \mathbb{R}^{\n \times \n}:$ The Normalised Adjacency matrix of $\G$;
    \item $\lambda_1(\Lrw):$ The Equidistribution radius of $\G$;
    \item $\varrho_r(H):$ The Normalized Spectral Gap of $\G$;    
    \item $\X_{\G} \in \mathbb{R}^{\n \times K}$, or simply $\X \in \mathbb{R}^{\n \times K}:$ The node feature matrix of the graph $\G$, obtained by stacking up all $x_i$ vectors;
    \item $\mathcal{E}_{dir}(\cdot):$ The Dirichlet Energy function;
    \item $\degr (i):$ The degree of a node $i$;
    \item $\degr^+(i):$ The out-degree of a node $i$;
    \item $\degr^-(i):$ The in-degree of a node $i$;
    \item $\smallestDegree (\G):$ The minimum degree of the graph $\G$;
    \item $\largestDegree (\G):$ The maximum degree of the graph $\G$;
    \item $h_C(\G):$ The Cheeger constant of the graph $\G$;
    \item $\gamma_h:$ The Homophily ratio of the graph $\G$;
    \item $\Gset:$ A set of multiple graphs;
    \item $\Gset_{train}:$ The set of training graphs;
    \item $\Gset_{val}:$ The set of validation graphs;
    \item $\Gset_{test}:$ The set of test graphs;
    \item $\numlayers :$ The number of message-passing layers in the GNN;
    \item $\V_{train}:$ The set of training nodes;
    \item $\V_{val}:$ The set of validation nodes;
    \item $\V_{test}:$ The set of test nodes;
    \item $\Y_{train}:$ The labels of the training nodes $\V_{train}\subset \V$;
    \item $\Y_{test}:$ The labels of the test nodes $\V_{test}\subset \V$;
    \item $\Y_{val}:$ The labels of the validation nodes $\V_{val}\subset \V$;
    \item $\Y:$ The labels space;
    \item $f: \Gset \to \mathcal{Y}:$ A graph function;
    \item $\readout:$ The Readout function in the GNN;
    \item $\combine:$ The Combine function in the GNN;
    \item $\aggregate:$ The Aggregate function in the GNN;
    \item $\mlp^{(\llayer)}:$ A multi-layer perceptron used at the $\llayer$-th layer of the GNN;

\end{itemize}

\endgroup

\pagestyle{scrheadings}
\pdfbookmark[1]{\contentsname}{tableofcontents}
\setcounter{tocdepth}{2} 
\setcounter{secnumdepth}{3} 
\manualmark
\markboth{\spacedlowsmallcaps{\contentsname}}{\spacedlowsmallcaps{\contentsname}}
\tableofcontents
\automark[section]{chapter}
\renewcommand{\chaptermark}[1]{\markboth{\spacedlowsmallcaps{#1}}{\spacedlowsmallcaps{#1}}}
\renewcommand{\sectionmark}[1]{\markright{\textsc{\thesection}\enspace\spacedlowsmallcaps{#1}}}
\clearpage
\begingroup
    \let\clearpage\relax
    \let\cleardoublepage\relax
    \pdfbookmark[1]{\listfigurename}{lof}
    \listoffigures

    \vspace{8ex}

    \pdfbookmark[1]{\listtablename}{lot}
    \listoftables

    \vspace{8ex}

    


    

\endgroup
\pagestyle{scrheadings}
\pagenumbering{arabic}
\part{Introduction}
\chapter{Introduction}
\lettrine[lines=3]{G}{\small{r}}aphs provide a natural representation of relationships and interactions between entities, making them a fundamental data structure in numerous domains. Formally, a graph consists of a set of nodes and edges that connect pairs of nodes. Each node can be associated with features, providing additional context or attributes to the graph structure. Graph-structured data arises in diverse applications, including social networks, where users and their relationships are modeled as graphs \citep{backstrom2011supervised}, biological networks representing molecular structures \citep{liu2017novel}, knowledge graphs encoding entities and relationships for semantic search \citep{galkintowards}, and transportation networks mapping roads, cities, and routes \citep{schichl2015predisaster}. Graphs are ubiquitous, capturing complex and high-dimensional relationships in data that cannot be modeled effectively with traditional Euclidean structures.

The need to extract meaningful insights from graph-structured data has led to the development of Graph Representation Learning techniques. These methods aim to derive low-dimensional embeddings that preserve the structural and feature-based properties of the graph. Unlike traditional machine learning approaches that operate on feature spaces, graph representation learning leverages the inherent topology of graphs to capture relationships and dependencies between nodes. Such embeddings are essential for downstream tasks like node classification, link prediction, and graph classification. These tasks facilitate applications such as community detection in social networks \citep{malliaros2013clustering}, recommendation systems \citep{son2017content}, and molecular property prediction \citep{stokes2020deep}.

To address the challenges of learning on graphs, researchers have developed Graph Neural Networks (GNNs), which extend traditional neural networks to operate directly on graph structures. GNNs build upon the principles of graph representation learning by incorporating both node features and graph connectivity into their computational models. This shift has enabled a more systematic approach to processing graph data compared to earlier heuristic-based methods. Specifically, GNNs propagate information through iterative message-passing mechanisms, allowing nodes to aggregate information from their neighbors. This iterative process mimics how information spreads through networks, making GNNs particularly effective for tasks requiring relational reasoning.

One prominent class of GNNs is Message Passing Neural Networks (MPNNs) \citep{Kipf:2017tc,xu2018how}, which operate in two stages: a message-passing stage where nodes exchange information with their neighbors, and an update stage where nodes refine their embeddings based on received messages. Popular architectures, such as Graph Convolutional Networks (GCNs), leverage convolutional operators to learn hierarchical features from graph data. These developments have significantly advanced the ability to learn rich representations from graphs, laying the foundation for more complex machine learning applications.

Graph Neural Networks have demonstrated remarkable success in both academic and industrial applications. In social media analysis, platforms like Twitter and LinkedIn use GNNs for recommendation systems and content filtering \citep{ying2018graph}. E-commerce platforms such as Amazon and Alibaba leverage GNNs for product recommendations \citep{wang2019knowledge}. In the pharmaceutical industry, GNNs aid in modeling molecular structures and predicting drug interactions, leading to the discovery of new antibiotics \citep{stokes2020deep}.

\section{The Challenges in GNNs}
Despite their success, GNNs face several challenges. One such challenge is over-smoothing, where repeated message passing leads to the loss of node identity. Scalability is another issue, as large-scale graphs require efficient processing techniques. Additionally, GNNs are vulnerable to adversarial attacks that manipulate graph structure or node features. There are also other important challenges, including robustness, generalization, and improving graph representation learning techniques. Addressing these challenges opens avenues for improving GNN architectures, making them more expressive, scalable, and resilient to adversarial perturbations.
\begin{itemize}
    \item Generalization refers to the ability of a GNN to perform effectively on unseen data or under distribution shifts that inevitably occur in real-world scenarios.
    \item Robustness addresses the resilience of GNNs to adversarial perturbations or random noise in both node features and graph structure, ensuring reliable performance even in malicious or unpredictable settings.
    \item Improved representation learning continues to push the boundaries of expressivity and interpretability, enabling GNNs to capture more relational information and complex structural patterns.
\end{itemize}

\section{Overview of Contributions}
This thesis explores the theoretical foundations and practical advancements in Graph Neural Networks (GNNs). By leveraging insights from graph geometry and spectral analysis, this research addresses key challenges such as robustness and generalization of GNNs, focusing on improving their  efficiency and performance in complex real-world applications.

\paragraph{Graph Representation Learning.} A major focus of my PhD research focuses on adapting the message-passing scheme in GNNs to align with the varying characteristics of different graphs and learning tasks. This work has resulted in two works. In one of them, I combined two key concepts, Graph Shift Operator (GSO) and node centrality criteria, to enhance the performance of GNNs. GSOs, such as the adjacency and graph Laplacian matrices, are a fundamental concept from graph signal processing, providing a way to understand how information propagates or ``shifts'' across a graph. A commonality of the most frequently used GSOs is their ability to encode purely local information in the graph, with the adjacency matrix encoding neighborhoods in the graph and the graph Laplacians relying on the node degree, a local centrality metric, to normalize the adjacency matrix. Building on this, I proposed and studied a new family of operators called Centrality Graph Shift Operators (CGSOs), which integrate the global position of nodes within the graph \citep{abbahaddou2024centrality}.  I performed a comprehensive spectral analysis to explore their fundamental properties, including eigenvalue structures and expansion behavior, examining how these operators influence information spread across the graph.  In the second work, I addressed the issue that standard GNNs use a fixed number of message-passing steps for all nodes, ignoring the fact that different nodes may require different depths of propagation. Through empirical analyses on real-world datasets and further validated by synthetic experiments, we observed that the optimal number of message-passing layers varies significantly depending on a node’s characteristics. To tackle this, I developed ADMP-GNN, a framework that dynamically adjusts the number of message-passing layers for each node. This approach applies to any model following the message-passing paradigm. Evaluations on node classification tasks demonstrate that ADMP-GNN outperforms baseline GNNs by tailoring the depth of message passing to the nodes’ unique needs.

\paragraph{GNN's Robustness.} Another significant aspect of my PhD research was devoted to study and improve GNNs' robustness against adversarial attacks \citep{abbahaddou2024bounding,ennadir2024simple,abbahaddou2024post}. Building on theoretical findings, I connected the expected robustness of GNNs to the orthonormality of their weight matrices and consequently propose an attack-independent and more robust variant of the GCN, called the Graph Convolutional Orthonormal Robust Networks (GCORNs) \citep{abbahaddou2024bounding}. To do so, I make use of an algorithm first, which we refer to as Bjorck Orthonormalization. In another project, we introduce NoisyGNNs \citep{ennadir2024simple}, a novel defense method that incorporates noise into the architecture of the underlying model. We establish a theoretical connection between noise injection and the enhancement of GNN robustness using Renyi Divergence, highlighting the effectiveness of our approach. And as most existing defense techniques primarily concentrate on the training phase of GNNs, involving adjustments to message passing architectures or pre-processing methods, I also proposed RobustCRF, a post-hoc approach aiming to enhance the robustness of GNNs at the inference stage using Conditional Random Fields \citep{abbahaddou2024post}.

\paragraph{GNN's Generalization.} Another interesting topic in GNN research is generalization. Despite their impressive capabilities, GNNs face significant challenges related to generalization, particularly when handling unseen or out-of-distribution (OOD) data \citep{guo2024investigating,li2022ood}. I introduced a novel approach, GRATIN, based on Gaussian Mixture Models (GMMs), for graph data augmentation that enhances both the generalization and robustness of GNNs \citep{GMMGDA}. Using the universal approximation property of GMMs, we can sample new graph representations to effectively control the upper bound of Rademacher Complexity, ensuring improved generalization of GNNs. 

These contributions collectively advance the theoretical understanding and practical implementation of GNNs, making them more robust, generalizable, and efficient for diverse applications.

\chapter{Preliminaries} \label{ch:preliminaries}
\lettrine[lines=3]{T}{\small{h}}is chapter provides an introduction to the key concepts and essential background information necessary for understanding this thesis. We begin by presenting the notation and fundamental characteristics of graphs, including core definitions such as nodes, edges, node degree, adjacency matrices, and random walks, which form the basis for subsequent discussions. Additionally, we explore graph-based concepts such as Graph Shift Operators (GSOs), which serve as representations of graphs and play a central role in spectral analysis, as well as node centrality measures, which are used to identify influential nodes within a graph. Furthermore, we introduce Graph Neural Networks (GNNs) and their applications, particularly in tasks related to node and graph classification. To evaluate our methods, we outline a variety of benchmarks encompassing both synthetic and real-world datasets. Collectively, these elements establish the groundwork for the methodologies and experiments presented in this thesis.

\section{Notation}

Let $\G = (\V,\E)$ be an undirected graph where $\V$ is the set of nodes and $\E$ is the set of edges. We will denote by $\n = |\V|$ and $\m = |\E|$ the number of nodes and number of edges, respectively.
Let $\mathcal{N}(j)$ denote the set of neighbors of a node $j \in \V$, \ie~ $\mathcal{N}(j) = \{ i \colon (j,i) \in \E \}$.
The degree $\degr (i)$ of a node $i \in \V$ is equal to its number of neighbors, \ie~  equal to $|\neigb(j)|$ for a node $j \in \V$.
A graph is commonly represented by its adjacency matrix $\adj \in [0,1]^{\n \times \n}$ where the $(i,j)$-th element $a_{i,j}$ of this matrix is equal to the weight of the edge between the $i$-th and $j$-th node of the graph and a weight of $0$ in case the edge does not exist.
In some settings, the nodes of a graph might be annotated with feature vectors. We use $\X \in \mathbb{R}^{\n \times \K}$ to denote the node features where $\K$ is the feature dimensionality. $\X$ is constructed by stacking up the feature vectors $x_i\in \mathbb{R}^{\K}$ for each node $i\in \V$. A \emph{path} between two nodes $i, j \in \V$ is a sequence of nodes $(i_0, i_1, \dots, i_k)$ such that $i_0 = i$, $i_k = j$, and $(i_{k^\prime-1}, i_{k^\prime}) \in \E$ for all $k\prime = 1, \dots, k$. A graph is said to be connected if there exists a path between every pair of nodes in the graph. A graph is said to be \emph{connected} if there exists a path between any pair of nodes in the graph. We call a graph \emph{non-empty} if it has at least one vertex (i.e., $\n \ge 1$), and we say it is \emph{finite} when its set of nodes (and hence edges) is finite in size. An \emph{isolated node} is a node with no neighbors, i.e., its degree is zero. We denote by 
\begin{equation}
    \smallestDegree (\G) = \min_{j \in \V}\{\degr(j)\}
\quad\text{and}\quad
\largestDegree (\G) = \max_{j \in \V}\{\degr(j)\},
\end{equation}
the smallest and largest degree of the graph, respectively.

\section{Graph Properties}

Graphs exhibit a wide range of characteristics and can be classified into different types based on their structural properties. Below, we briefly outline some common characteristics and graph types that are relevant in various context:
\subsection{Heterogeneity}

A \emph{heterogeneous graph}, also referred to as a \emph{multi-relational graph}, is defined as $\G = \bigl(\V,\E,\mathcal{T}_{\V},\mathcal{T}_{\E}\bigr),$ where $\mathcal{T}_V$ is the set of node types and $\mathcal{T}_E$ is the set of edge types. Each node $i \in \V$ belongs to exactly one type $c \in \mathcal{T}_{\V}$. Hence, we can write
      \begin{equation}
          \V = \bigcup_{c \in \mathcal{T}_{\V}} \V^c,
      \quad
      \V^c \cap \V^{c'} = \emptyset \text{ for } c \neq c'.
      \end{equation}

    Each edge \(e \in \E\) belongs to exactly one relation type \(r \in \mathcal{T}_E\). Accordingly,
  \begin{equation}
      \E = \bigcup_{r \in \mathcal{T}_E} \E^r,
      \quad
      \E^r \cap \E^{r'} = \emptyset \text{ for } r \neq r'.
  \end{equation}

Heterogeneity arises in many complex systems. For example, in knowledge graphs where heterogeneity arises from the presence of various types of entities and relationships \citep{li2024generalizing}. Likewise, in biological networks, heterogeneity emerges from the coexistence of multiple biological entities that interact through a variety of processes \citep{yu2022molecular}. As a result, one often constructs type-aware adjacency structures, \eg, separate adjacency matrices for each edge type $r \in \mathcal{T}_{\E}$.

In contrast, a \emph{homogeneous graph} consists of a single type of node and edge, making it structurally simpler and easier to process. For instance, social networks, where nodes typically represent individuals and edges denote friendships or connections, are often modeled as homogeneous graphs \citep{yang2019homogeneous}. While this simplicity facilitates computational efficiency, homogeneous graphs are inherently limited in their ability to represent the complexity of relationships. 

\subsection{Homophily}

\emph{Homophily} is the principle that \emph{similar nodes tend to connect with each other}. In many social and biological networks, entities sharing an attribute, \eg, labels or biological function, are more likely to form edges. Graph homophily is measured using the \emph{homophily ratio}. Formally, let there be a \emph{labeling function} $\phi: \V \rightarrow \mathcal{L}$ that assigns each node $i \in \V$ to a label $\phi(i)$ from a label set $\mathcal{L}$. We define the homophily ratio $H$ as:

\begin{equation}
    \gamma_h = \frac{\bigl|\{\,(i,j) \in \E : \phi(i) = \phi(j)\}\bigr|}{\m}.
\end{equation}

A value of $\gamma_h$ close to 1 indicates strong homophily, implying that most edges connect nodes having the same label. A value of $\gamma_h$ close to 0 indicates \emph{heterophily}, meaning most edges connect nodes with \emph{different} labels or attributes.

In tasks such as node classification, strong homophily often makes label propagation or graph-based semi-supervised learning more effective, since neighbors are likely to share labels \citep{platonov2024characterizing,luan2024graph}. This intuition is well illustrated by the widely used Label Propagation (LP) algorithm, which is commonly applied to node classification and serves as a simple yet effective baseline for Graph Neural Networks (GNNs) \citep{wang2020unifying}.  Specifically, labels are iteratively spread through the graph during label propagation, allowing unlabeled nodes to inherit the labels of their neighbors. When homophily is strong, the likelihood that neighboring nodes share the same label is high, making the propagation process more accurate and reliable.

\subsection{Sparsity}

A graph with $\n$ nodes has a maximum of $\frac{\n(\n-1)}{2}$ edges if it is an undirected, homogeneous and simple graph, \ie, with no self-loops. The graph $\G$ is said to be \emph{sparse} if the actual number of edges $\m = |\E|$ is much smaller than this maximum. In other words, $\m \;\ll\; \frac{\n(\n - 1)}{2}.$ One way to quantify sparsity is via the \emph{edge density}, defined as
\begin{equation}
    \rho = \frac{2\m}{\n(\n - 1)}.
\end{equation}
Sparsity significantly influences the computational complexity of graph algorithms. For example, many methods in graph signal processing, or graph neural networks, can be optimized when the graph is sparse. For example, sparse representations reduce the complexity of matrix-vector multiplications by processing only existing edges, avoiding redundant calculations over non-existent connections.

\subsection{Connectivity}

A graph is \emph{connected} if every pair of nodes is linked by some \emph{path} in $\G$. More precisely, $\G$ is connected if for any two nodes $u, v \in \V$ there exists a sequence of edges that starts at $u$ and ends at $v$. Formally:

\begin{equation}
    \forall i,j \in \V ~
\exists k \in \mathbb{N} ~\exists i_0,\ldots,i_k \in \V, ~~  (i_0=i, i_1), (i_1, i_2), \dots, (i_{k-1}, i_k = j) \in \E.
\end{equation}

If $\G$ is \emph{not connected}, it can be partitioned into \emph{connected components}, where each connected component $C \subseteq \V$ is a maximal set of nodes such that any two nodes within $C$ are connected by a path lying entirely in $C$. 

Connectivity in graphs has a high impact on GNN performances. On one hand, well-connected graphs enable features from labeled nodes to propagate effectively to unlabeled nodes. On the other hand, in disconnected graphs or graphs with isolated nodes, information exchange is limited to connected components, preventing the model from capturing global patterns and leading to fragmented representations that hinder learning and generalization \citep{abbahaddou2023graph}.
\\
\\
By accounting for heterogeneity, homophily, sparsity, and connectivity, one can better model and understand the diverse structures and behaviors observed in real-world graphs. Each of these properties influences the choice of algorithms and models, from conventional graph mining techniques to modern graph neural networks. In the next section, we will delve into walks and random walks, which are fundamental concepts in graph machine learning. 

\section{Graph Exploraton: Walks and Random Walks}
In this section, we introduce fundamental concepts related to graph
explorations, focusing on both deterministic and random walks. Walks provide a straightforward means of moving through the graph, while random walks incorporate probability distributions at each step, offering a powerful framework for studying and simulating various graph-based phenomena. We begin by defining and discussing walks, and then proceed to formalize random walks as Markov processes on graphs.
\subsection{Walks}
As defined in Definition \ref{def:walk}, a \emph{walk} in a graph is a sequence of nodes where each consecutive pair is connected by an edge. While walks can revisit nodes, a \emph{path} is a specific type of walk where all vertices are distinct (except possibly the start and end nodes in a cycle), ensuring no repetitions. 
\begin{definition}[Walk on a Finite Graph]\label{def:walk}
    Let $\G = (\V, \E)$ be an graph with node set $\V$ and edge set $\E$.  A \emph{walk} of length $T$ in $\G$ is a finite sequence of vertices $(i_0, i_1, \dots, i_t)$ such that $\bigl(i_i, i_{i+1}\bigr) \in \E$ for every $t = 0, 1, \dots, T-1$. If \(i_0 = i_T\), the walk is said to be \emph{closed}.
\end{definition}
The number of walks of length $t$ from a vertex $i$ to a vertex $j$ is precisely given by the $(i,j)$-th entry of $(\adj+\identity)^t$, where $\adj$  is the adjacency matrix of the graph. \emph{Normalized walks} are a variant of walks that account for differences in node degrees to balance contributions across the graph. For normalized walks, one typically uses the Normalised Adjacency matrix $\Atilde = \D^{-1/2}(\adj + \identity )\D^{-1/2}$; in these cases, the entries of $\Atilde^t$ give normalized number of walk from $i$ to $j$ in exactly $t$ steps.

Building on this concept of a walk, we can introduce a probability distribution at each step to define a \emph{random walk}. This stochastic perspective allows us to model processes where each successive move is chosen randomly, providing a powerful framework for studying and simulating diverse phenomena on graphs. Next, we introduce the formal definition of random walks.
\subsection{Random Walks}
Random walks form the foundation of many graph-based algorithms and analyses. They provide a probabilistic framework to model through a graph, where each step is determined by the connectivity of nodes. Random walks are particularly useful for studying graph properties, evaluating node importance, and designing algorithms for tasks such as node ranking, clustering, and semi-supervised learning \citep{yang2019homogeneous,malliaros2013clustering,chien2020adaptive}. We give the detailed mathematical definition of Random Walks in Definition \ref{def:random_walks}. 

\begin{definition}[Random Walk on a Finite Graph]\label{def:random_walks}
Let $\mathcal{G} = (\mathcal{V}, \mathcal{E})$ be a finite, undirected graph, and let $\shortedpathdist : \mathcal{V} \times \mathcal{V} \;\longrightarrow\; \mathbb{N}$ denote the \emph{shortest-path distance} between any two vertices, \ie, $\rho(i,j)$ is the minimum number of edges in a path connecting $i$ and $j$. A \emph{random walk} on $\mathcal{G}$ is a sequence of random variables $(I_t)_{t \geq 0}$ taking values in $\mathcal{V}$, such that:

\begin{enumerate}
  \item \textbf{Markov Property}: For all $t \geq 0$ and all vertices $i_0, i_1, \dots, i_t, k \in \mathcal{V}$ satisfying
  
    \begin{equation}
              \rho\bigl(i_{t^\prime}, i_{t^\prime+1}\bigr) \;\leq\; 1
          \quad \text{for} \quad 0 \leq t^\prime \leq t-1,
    \end{equation}
 
  we have
  
    \begin{equation}
            \mathbb{P}\bigl(I_{t+1} = k \,\big|\,(I_t, \dots, I_0) = (i_t, \dots, i_0)\bigr)
    \;=\; 
    \mathbb{P}\bigl(I_{t+1} = k \,\big|\; I_t = i_t\bigr).
    \end{equation}
    
  In other words, the probability of moving to $k$ at time $t+1$ depends only on the current vertex $i_t$, not on the earlier history.
  \item \textbf{Transition Probabilities}: For each vertex $v_t$, the probability of transitioning to a vertex $z$ at time $t+1$ is given by

\begin{equation}
      \mathbb{P}\bigl(I_{t+1} = k \,\big|\; I_t = i_t\bigr)
  \;=\;
  \begin{cases}
    0, & \text{if } \shortedpathdist(i_t, k) > 1, \\[6pt]
    \dfrac{a_{i_t, k}}{\degr(i_t)}, & \text{if } \shortedpathdist(i_t, k) \le 1.
  \end{cases}
\end{equation}

\end{enumerate}
\end{definition}

In other words, if $I_t$ is at vertex $i$, then $I_{t+1}$ is determined by moving to an adjacent vertex $k$, using a randomly and uniformly chosen edge connecting $j$ to $k$, independent of the past history of the walk. This includes the possibility that $I_{t+1} = j$, which may happen if there is a loop at $i$.

The distribution of the initial step $I_0$ of the walk is called the \textit{initial distribution}, characterized by the probabilities $\mathbb{P}(I_0 = i)$ for $i \in V$. If $I_0 = i_0$ almost surely, the random walk is said to start from $i_0$. The process $(I_t)$ forms a Markov chain with state space $V$ and transition matrix $P =\left\{ p_{j, k } \right\}_{j, k \in V}$ given by,
\begin{equation}
p_{j, k } =
\begin{cases}
0, & \text{if } \shortedpathdist(v, z) > 1, \\
\frac{a_{j,k}}{\degr(j)}, & \text{if } d_{\Gamma}(j, k) \leq 1,
\end{cases}
\end{equation}
where $a(v, z)$ represents the adjacency matrix coefficient.

A random walk on $\mathcal{G}$ admits a \emph{unique stationary distribution} $\pi$. For an undirected graph, it is well known that

\begin{equation}
    \pi(j) \;=\; \frac{\degr(j)}{\sum_{i \in \V} \degr(i)},
\end{equation}

meaning that in the long run, the probability of being at a node $v$ is proportional to its degree \citep{kowalski2019introduction,lovasz1993random}. This stationary behavior is essential for deriving analytical results that connect long-term graph exploration with the structural properties of the graph. It plays a key role in various applications, including graph-based ranking, community detection, and beyond \citep{page1999pagerank}.

\section{Graph Shift Operators}
In this section, we introduce \emph{Graph Shift Operators} (GSOs), 
a class of matrices that plays a central role in processing and 
analyzing graph-structured data. By encoding a graph’s connectivity 
through sparse, edge-based relationships, GSOs facilitate both 
theoretical insights into spectral properties and practical algorithms for tasks such as clustering, community detection, and spectral analysis. In particular, degree-normalized GSOs have received 
significant attention for their robustness and interpretability in 
applications ranging from signal processing on graphs to 
graph neural networks.
\subsection{Definition of Graph Shift Operators}
Understanding the structure and behavior of graph-structured data often requires efficient and insightful analysis tools. Graph Shift Operators (GSOs) play an important role in the analysis of graph-structured data. Degree-normalized GSOs, such as the Random-walk Normalized Laplacian \citep{modell2021spectral}, have been widely employed in spectral analysis and signal processing on graphs. These GSOs have many properties allowing great insight in the connectivity of nodes. Formally, a graph  $\G = (\V, \E)$, can be represented by an adjacency matrix $\mathbf{A} = [a_{i,j}] \in \mathbb{R}^{N\times N}$ where $a_{i,j} =1$ if $e_{i,j} \in \mathcal{E}$ and $a_{i,j} =0$ otherwise. Analyzing the spectra of the adjacency matrix provides information about the basic topological properties of the underlying graphs \citep{cvetkovic1980spectra}. For example, according to the Gerschgorin theorem, every eigenvalue of an adjacency matrix $\mathbf{A}$ lies in at least one of the circular discs with center $a_{ii}$ and radii $\sum_{i\leq j } |a_{ij}|$. Furthermore, the largest eigenvalue of $\mathbf{A}$  is an upper bound of the average degree and a lower bound of the largest degree \citep{cvetković_rowlinson_simić_2009,sarkar2018spectral}.

In addition to the spectral properties of the adjacency matrix, there are alternative graph representations that provide deep insights into the topology of the underlying graph. One often-used representation is the symmetrically normalized Laplacian matrix defined by $\Lsym = \mathbf{I}-\mathbf{D}^{-1/2}\mathbf{A}\mathbf{D}^{-1/2}$, where  $\D \in \mathbb{R}^{\n \times \n}$ is the degree matrix, i.e., a diagonal matrix defined as $\D_{ii} = \sum_{i=1}^n a_{ij}$. There are other graph representations with particularly interesting spectral properties, such as the random-walk Normalised Laplacian \citep{modell2021spectral} and the Signless Laplacian matrices \citep{cvetkovic2009towards}. All these graph representations belong to the family of \textit{Graph Shift Operators} (GSOs). Examples of such matrices appear in Table \ref{tab:classical_gso}. Note that the adjacency matrix $\adj$, the unnormalized Laplacian $\Lgso$, and other variations listed there all satisfy the structural criteria of GSOs, differing mainly by how they weight each edge.

\begin{definition}
Given an arbitrary graph $G = (\mathcal{V}, \mathcal{E})$, a \textit{Graph Shift Operator} $\mathbf{S}\in \mathbb{R}^{N\times N}$ is a matrix satisfying $\mathbf{S}_{ij} = 0$ for $i \neq j$ and $(i, j) \notin \mathcal{E}$ \citep{mateos2019connecting} and $\mathbf{S}_{ij} \neq 0$ for $i \neq j$ and $(i, j) \in \mathcal{E}$. 
\end{definition}

\begin{table}[t]
\caption{Graph Shift Operators.}
\label{tab:classical_gso}
\begin{center}
\resizebox{\textwidth}{!}{
\begin{tabular}{ll}
\toprule
Graph Shift Operator & Description when  $V=D$   \\
\midrule
 $ \adj$ & Adjacency matrix\\
 $\Lgso =\D-\adj$& Unnormalised Laplacian matrix \\
$\Qgso =\D+\adj$ & Signless Laplacian matrix\\
$\Lrw =\identity - \D^{-1}\adj$ & Random-walk Normalised Laplacian matrix\\
$\Lsym =\identity - \D^{-1/2}\adj \D^{-1/2}$ & Symmetric Normalised Laplacian matrix \\
$\Atilde = \D^{-1/2}(\adj + \identity )\D^{-1/2}$&Normalised Adjacency matrix \\
$ \Hgso =\D^{-1}\adj$& Mean Aggregation matrix\\
\bottomrule
\end{tabular}
}
\end{center}
\end{table}

In addition to the classical or fixed GSOs, parametrized GSOs are learned during the optimization process. These parametrized operators are a fundamental component of many modern GNN architectures and allow the model to adapt and capture complex patterns and relationships in the graph data. For example, the work of \textit{PGSO} \citep{Dasoulas2021LearningPG} parametrizes the space of commonly used GSOs leading to a learnable GSO that adapts to the dataset and learning task at hand.

\subsection{Spectral Clustering}
One important application of GSOs is spectral clustering \citep{von2007tutorial}, a technique for partitioning a graph into distinct clusters (or communities), leveraging the spectral properties of GSOs to partition graphs into meaningful clusters.  The process begins by computing GSO's eigenvectors. Nodes are then embedded into a lower-dimensional spectral space using the eigenvectors corresponding to the smallest non-zero eigenvalues. A clustering algorithm, such as $K$-means, is applied in this spectral space to group nodes, and the resulting clusters are mapped back to the original graph, revealing its community structure. Spectral clustering is widely used in practice. For instance, in social networks, it can be employed to detect user communities sharing similar interests or social ties \citep{wahl2015hierarchical}. In collaborative filtering and recommender systems, spectral clustering may help identify user or item groups with similar interaction patterns, providing a basis for improved recommendation mechanisms \citep{hoseini2012levelwise}. In Algorithm \ref{algo:spectral_clustering}, we outline the steps of the Spectral Clustering Algorithm using CGSOs. 
\begin{algorithm}[t]

\small
\textbf{Inputs: } Graph $\G$, A GSO $\Phi$, Number of clusters to retrieve $C.$\\
\begin{enumerate}
    \item Compute the eigenvalues $\{\lambda \}_{i=1}^\n$ and eigenvectors $\{u \}_{i=1}^\n$ of $\mathbf{\Phi}$;
    \item Consider only the eigenvectors $U\in \mathbb{R}^{N\times C}$ corresponding to the $C$ largest eigenvalues;
    \item Cluster rows of $\n$, corresponding to nodes in the graph,  using the $K$-Means algorithm to retrieve a node partition $\mathcal{P}$ with $C$ clusters;
    $$\mathcal{P} = \text{K-Means}(U, C ) $$
\end{enumerate}

\Return $\mathcal{P}$; 
\caption{Spectral Clustering using the Centrality GSOs}\label{algo:spectral_clustering}

\end{algorithm}

\subsection{Spectral Properties of GSOs}

Beyond clustering, studying the spectral properties of GSOs gives rise to important graph expansion inequalities. For example, the \textit{Cheeger constant} measures the \emph{bottleneck} of a graph, and Cheeger's inequality quantitatively links this bottleneck to the spectrum of the graph Laplacian. These relationships are crucial for understanding how well-connected a graph is. The relevant definitions and results are summarized below.

\begin{definition}[Cheeger constant]
Let $\G = (\V, \E)$ be a finite graph.  
\begin{enumerate}
    \item For any disjoint subsets of vertices $\V_1, \V_2 \subset \V$, we denote by $\E(\V_1, \V_2)$ the set of edges of $\G$ with one extremity in $\V_1$ and one extremity in $\V_2$,
    
    \begin{equation}
        \E(\V_1, \V_2) = \{(i,j) \in \E \mid \{i,j\} \cap \V_1 \neq \emptyset, \quad  \{i,j\}  \cap \V_2 \neq \emptyset \}.
    \end{equation}
    
    and we denote by $\mathcal{E}(V_1)$ the set $\mathcal{E}(V_1, V \setminus V_1)$ of edges with one extremity in $V_1$, and one outside $V_1$.

    \item The \textit{Cheeger constant} $h_C(\G)$ is defined by
    \begin{equation}
        h_C(\G) = \min \left\{ \frac{|\E(W)|}{|W|} \middle| \, W \subset V, \, W \neq \emptyset, \, |W| \leq \frac{1}{2}|\V| \right\},
    \end{equation}
    with the convention that $h_C(\G) = +\infty$ if $\G$ has at most one vertex.
\end{enumerate}
\end{definition}

\begin{proposition}[Discrete Cheeger inequality, \citep{cheeger1970lower}]\label{prelimsn_prop:cheeger}
Let $\G = (\V, \E)$ be a connected, non-empty, finite graph without isolated vertices. We have
\begin{equation}
    1 - \varrho_r(H) \leq \lambda_1(\Lrw) \leq \left( \frac{2\largestDegree (\G)}{\smallestDegree (\G)^2} \right) \cheeg(\G)
    \tag{3.32}
\end{equation}
where $\smallestDegree (\G)$ and $\largestDegree (\G)$ represent the minimum and maximum degree of the graph, $\lambda_1(\Lrw)$ is the normalized spectral gap, \ie the smallest non-zero eigenvalue of $\Lrw$, and $\varrho_r(H)$ is Equidistribution radius of $\G$, \ie the maximum of the absolute values $|\lambda|$ for $\lambda$ an eigenvalue of the Mean Aggregation matrix $\Hgso =\D^{-1}\adj$ which is different from $\pm 1$.
\end{proposition}

\begin{proposition}[Discrete Buser inequality, \citep{buser1982note}] \label{prelimsn_prop:buser}
Let $\G = (\V, \E)$ be a connected, non-empty, finite graph without isolated nodes. We have
\begin{equation}
    \cheeg(\G) \leq \largestDegree (\G)\sqrt{2\lambda_1(\Lrw) },
\end{equation}
where $\largestDegree (\G)$ represents the maximum degree of the graph, and $\lambda_1(\Lrw)$ is the Normalized Spectral Gap of $\G$, \ie the smallest non-zero eigenvalue of $\Lrw$.
\end{proposition}

\begin{proposition}[Spectral Properties of the Unnormalised Laplacian Matrix $\Lgso$]\label{prelimsn_prop:unnorm}
Let $\G = (\V, \E)$ be a finite graph. The Unnormalized Laplacian matrix $\Lgso$ of $\G$ exhibits the following key spectral properties:
\begin{enumerate}
    \item The multiplicity of the $0$ eigenvalue is equal to the number of connected components in $\G$ and the corresponding eigenvectors are indicator vectors establishing which vertex is an element of which connected component  \citep{von2007tutorial}.
    \item Let $\G$ be a connected graph with diameter $\diam(\G)$. Then, the Unnormalised Laplacian matrix $\Lgso$ of $\G$ has at least $\diam(\G)+1$ distinct eigenvalues \citep{brouwer2011spectra}. 
\end{enumerate}

\end{proposition}
The inequalities in Propositions~\ref{prelimsn_prop:cheeger}, \ref{prelimsn_prop:buser} and \ref{prelimsn_prop:unnorm} demonstrate how the graph's connectivity and expansion properties are intimately tied to the spectrum of the Random-walk Normalised Laplacian and the Unnormalised Laplacian matrices.  These inequalities underscore the importance of careful selection and design of GSOs in tasks such as clustering, community detection, and analysis of graphs.  For example, the celebrated \textit{Cheeger's Inequality} establishes a bound on the edge expansion of a graph via its spectrum \citep{cheeger2015lower}. We recall the concept of the \textit{Cheeger constant}, which quantifies how well-connected a graph is by measuring the ratio between the number of edges leaving a set of vertices and the size of that set.  The \textit{Cheeger Inequality} in Proposition \ref{prelimsn_prop:cheeger} shows how the \textit{smallest non-zero eigenvalue} of $\Lrw$ is bounded above and below by terms involving the Cheeger constant. By relating combinatorial quantities (\eg, the Cheeger constant) to spectral properties (\eg, eigenvalues of the Laplacian), we gain powerful tools for assessing and controlling how graph-structured data behaves under transformations induced by GSOs.

Overall, these theoretical underpinnings highlight why Graph Shift Operators are central to modern approaches for learning, analyzing, and processing data on graphs. By leveraging their spectral properties, researchers can design algorithms that better respect the structure of real-world networks, whether they are social, biological, or technological in nature.

\section{Node Centrality}

Node centrality measures the relative importance of a node within a graph based on its connectivity to other nodes. Node centrality has numerous real-world applications, including identifying influential nodes in social networks \citep{malliaros2016locating}, and ranking webpages in search engines \citep{brin1998anatomy}. It is also used in recommendation systems \citep{son2017content}, and traffic flow analysis \citep{borgatti2005centrality}, where understanding the central nodes helps optimize performance and decision-making. In addition to degree centrality, which provides a local measure based on a node's direct connections, various other centrality metrics have been proposed to capture different aspects of a node's importance. In this section, we explore several centrality measures, including the \emph{$k$-core}, \emph{PageRank}, and \emph{Walk Count}, each providing unique insights into the importance of nodes within a graph. These metrics serve as valuable tools for understanding the structural properties of nodes in complex networks. Below, we introduce three centrality measures that are used in this manuscript.

\subsection{$\boldsymbol{k}$-core} We leverage the $k$-core decomposition of a graph, which captures how well-connected nodes are within their neighborhood \citep{kcore-vldbj20}. The process of $k$-core decomposition involves iteratively removing vertices with degree less than $k$ until no such vertices remain. Mathematically, the $k$-core can be constructed using Algorithm \ref{algo:kcore}.

\begin{algorithm}[t]

\small
\textbf{Inputs: } Non-Empty Graph $\G$ with a diameter larger than 1.\\
\begin{enumerate}

\item \item \textbf{Initialization:} Set $\G_0 = \G$ and $k = 1$.

\item \textbf{Iterative Pruning:} At each iteration, construct the subgraph $\G_k$ by removing all vertices in $\V_{k-1}$ with degree less than $k$:
    \begin{equation}
        \G_{k} = \G_{k-1} \setminus \{ i \in \V_{k-1} \mid \deg_{\G_{k-1}}(i) < k \}
    \end{equation}
   where $\deg_{\G_{k-1}}(i)$  denotes the degree of vertex $i$ in the subgraph $G_{k-1}$.

\item \textbf{Termination:} Repeat the pruning process until $\G_k$ is either empty or no vertices with degree less than $k$ remain.  Increment $k$ at the end of each iteration:
$$k = k+1$$
\end{enumerate}

\Return $\mathcal{P}$; 
\caption{$k$-core Decomposition }
\label{algo:kcore}
\end{algorithm}

The core number $\core(i)$ of a vertex $i$ is the highest  $k$ for which $i$ is part of the $k$-core. Formally:
\begin{equation}
    core(i) = \max \{ k \mid i \in G_k \}.
\end{equation}

This core number reflects the robustness of a node's connectivity within the graph. Nodes with higher core numbers are considered more central as they reside in denser subgraphs, indicating strong interconnections with their neighbors.

The $k$-core decomposition provides insights into the hierarchical structure of the graph, revealing densely connected regions and identifying influential nodes based on their participation in these cores. It is particularly useful in applications such as community detection, and network robustness \citep{peng2014accelerating,dey2020network}.

\subsection{PageRank}  The \emph{PageRank} algorithm, originally developed for ranking webpages, evaluates the importance of nodes within a connected web graph by assigning a numerical weight to each node. This weight reflects the node's relative significance, determined by the structure of incoming and outgoing links \citep{brin1998anatomy}. The concept has since been generalized to apply to any graph structure. A node's PageRank score represents the probability of a random walk visiting that node, making it a fundamental metric for measuring node importance in various networks, particularly in web search algorithms. Mathematically, the PageRank algorithm operates by modeling a random walk over the graph. If the graph $\G$  contains nodes with no outgoing links, the random walk is no longer well-defined. To address this, a natural approach involves allowing the random walk to jump uniformly at random to any node in $\V$ with a probability of $\frac{1}{\n}$. The transition matrix $P_{i,j}$ is then defined as,
\begin{equation}
    P_{ij} =
\begin{cases}
\frac{a_{ij}}{\degr^+(j)}, & \text{if } \degr^+(j) > 0, \\
\frac{1}{\n}, & \text{otherwise}.
\end{cases}
\end{equation}

Here, $\degr^+(j)$ is the out-degree of node $j$, \ie, the  total number of edges directed away from node $j$. While this ensures that the random walk is well defined, the process may still lack irreducibility, as certain nodes may remain inaccessible from others.

To enforce irreducibility, a damping factor $\alpha \in [0, 1]$ is introduced. With probability $\alpha$, the random walk follows the transition matrix $P$, and with probability $1 - \alpha$, it restarts at a random node. The modified transition matrix $\mathbf{P}(\alpha)$ becomes,
\begin{equation}
    \mathbf{P}(\alpha) = \alpha \mathbf{P} + (1 - \alpha) \frac{1}{n} \mathbb{1}\mathbb{1}^T,
\end{equation}

where $\mathbb{1}$ is the column vector of ones. The stationary distribution $\pi^{(\alpha)}$, known as the PageRank vector, satisfies the equation,
 \begin{equation}
     \pi^{(\alpha)} = \alpha \pi^{(\alpha)} \mathbf{P} + (1 - \alpha) \frac{1}{n} \mathbb{1}^T.
 \end{equation}

This stationary distribution $\pi^{(\alpha)}$ gives the PageRank score for each node, indicating the steady-state probability of visiting that node during a random walk \citep{hollocou2016improving}.

\subsection{Walk Count} 
The \emph{Walk Count} centrality quantifies the importance of a node based on the number of walks of a given length $\ell$ starting from that node. This measure captures higher-order connectivity patterns within the graph, going beyond direct neighbors to account for paths that traverse intermediate nodes. Formally, let $\adj \in \mathbb{R}^{\n \times \n}$ denote the adjacency matrix of the graph $\G$, and $\mathbb{1} \in \mathbb{R}^{\n}$ represent the vector of ones. The vector $\mathbf{P}_{\ell\text{-paths}} \in \mathbb{R}^{\n}$ records the number of paths of length $\ell$ originating from each node $i$, defined as,
\begin{equation}
    \mathbf{P}_{\ell\text{-paths}} =  \mathbf{A}^\ell \mathbb{1} ,
\end{equation}
where $\mathbf{A}^\ell$ represents the $\ell$-th power of the adjacency matrix, corresponding to the number of walks of length $\ell$ between nodes. For the special case of $\ell = 2$, $\mathbf{P}_{\ell\text{-paths}}$ can be further expressed as, $ diag(\mathbf{P}_{2\text{-paths}}) = \mathbf{W}_{M_{13}} \mathbb{1} \mathbf{D},$ where $\mathbf{W}_{M_{13}}$ is the motif adjacency matrix introduced by \citep{benson2016higher}, counting the number of open bidirectional wedges (motif $M_{13}$), $\mathbf{D}$ is the degree matrix of the graph, and $diag(\cdot)$ is an operator that converts a vector into a diagonal matrix, placing the vector's elements on the diagonal while setting all off-diagonal elements to zero. This motif network captures higher-order structures and gives new insights into the organization of complex systems.

\section{Graph Neural Networks}

Before the advent of GNNs, machine learning on graphs relied on methods such as handcrafted embedding techniques. Approaches like Node2Vec \citep{grover2016node2vec}, DeepWalk \citep{perozzi2014deepwalk}, and LINE \citep{tang2015line} used random walks or other strategies to generate low-dimensional embeddings of nodes. While these methods achieved reasonable success, they lacked the ability to generalize to unseen graphs or directly leverage node and edge attributes during training. Furthermore, their reliance on static embeddings made them suboptimal for capturing dynamic and multi-hop dependencies within a graph.

GNNs address these limitations by extending the deep learning paradigm to graph-structured data. Unlike traditional approaches, GNNs can dynamically learn representations of nodes, edges, and even entire graphs by iteratively aggregating and transforming information from local neighborhoods. This capability enables GNNs to capture both the structural and semantic properties of graphs, making them highly effective for a wide range of tasks such as node classification, link prediction, and graph classification.

In the following sections, we delve deeper into the key architectures of GNNs, focusing on their design principles and mathematical foundations. We start with the basic framework of the \emph{message-passing scheme}, a fundamental concept in most recent GNN models.

\subsection{Key GNN Architectures}

\noindent A Graph Neural Network (GNN) consists of multiple neighborhood aggregation layers, where each GNN layer generates new node representations relying on the graph structure $\G=(\V,\E)$ and the nodes' feature vectors from the previous layer. Suppose we have a GNN model that contains $\numlayers $ neighborhood aggregation layers.
Let also $h_i^{(0)}$ denote the initial feature vector of node $i$, \ie~ $\mathbf{h}_i = \mathbf{x}_i$.
At each layer ($\llayer > 0$), the hidden state $h_j^{(\llayer)}$ of a node $j$ is updated as follows:
\begin{align}
&a_j^{(\llayer)} = \aggregate^{(\llayer)} \Big( \big\{ h_i^{(\llayer-1)} \colon i \in \neigb (j) \big\} \Big), \\
&h_j^{(\llayer)} = \combine^{(\llayer)} \Big(h_j^{(\llayer-1)}, \mathbf{a}_j^{(\llayer)} \Big),
\end{align}
where $\aggregate(\cdot)$ is a permutation invariant function that maps the feature vectors of the neighbors of a node $v$ to an aggregated vector.
This aggregated vector is passed along with the previous representation of $j$, \ie~$\mathbf{h}_j^{(\llayer-1)}$, to the $\combine(\cdot)$ function which combines those two vectors and produces the new representation of $v$. In the Graph Convolutional Network (GCN) \citep{Kipf:2017tc},  the $\aggregate$ function aggregates features from the neighbors of each node using a normalized sum, and the $\combine$ function is a linear transformation followed by a non-linear activation as follows:
\begin{align}
    a_j^{(\llayer)} & = \sum_{i \in \neigb (j) \cup \{j\}} \frac{1}{ \sqrt{\degr (i) \degr (j)}} h_i^{(\llayer-1)} \\
   h_j^{(\llayer)} & = \sigma \left( W^{(\llayer} a_j^{(\llayer)} \right),
\end{align}
where $W^{( \llayer )} \in \mathbb{R}^{d_{\llayer -1}, d_{ \llayer }}$ are learnable weight matrix, $d_\llayer$ is the dimension of the hidden representation at the $\llayer$-th layer, and $\sigma$ is a non-linear activation function, often chosen as ReLU.

In the Graph Attention Network (GAT) \citep{veličković2018graph}, the $\aggregate$ function incorporates a learnable attention mechanism, which assigns different importance scores to neighbors. The attention coefficients are computed as follows:  

\begin{equation}
    \alpha_{ij}^{(\llayer)} = \frac{\exp \left( \text{LeakyReLU} \Big( \mathbf{a}^\top [W^{(\llayer)} h_i^{(\llayer-1)} \parallel W^{(\llayer)} h_j^{(\llayer-1)}] \Big) \right)}{\sum_{k \in \neigb (j) \cup \{j\}} \exp \left( \text{LeakyReLU} \Big( \mathbf{a}^\top [W^{(\llayer)} h_k^{(\llayer-1)} \parallel W^{(\llayer)} h_j^{(\llayer-1)}] \Big) \right)},
\end{equation}

where $a\in \mathbb{R}^{ 2d^{'}_{\llayer}}$ is a learnable vector, $W^{(\llayer)}  \in \mathbb{R}^{d_{\llayer} \times d^{'}_{\llayer}}$ is a learnable weight matrix, and $\parallel$ denotes the vector concatenation. The new nodes representation are expressed as follows:

\begin{align}
    a_j^{(\llayer)} & = \sum_{i \in \neigb (j)} \alpha_{ij}^{(\llayer)} W^{(\llayer)} h_i^{(\llayer-1)},\\
    h_j^{(\llayer)} & = \sigma \left( a_j^{(\llayer)} \right).
\end{align}

GATv2 \citep{brody2022how} improves upon GAT by introducing a more expressive attention mechanism. Instead of applying the linear attention coefficients directly, GATv2 computes attention scores as:

\begin{equation}
    e_{ij}^{(\llayer)} = \mathbf{a}^\top \sigma \Big( W^{(\llayer)} \big[ h_i^{(\llayer-1)} \parallel h_j^{(\llayer-1)} \big] \Big),
\end{equation}
where $e_{ij}^{(\llayer)}$ is the unnormalized attention score. These scores are normalized using a softmax function:

\begin{equation}
    \alpha_{ij}^{(\llayer)} = \frac{\exp(e_{ij}^{(\llayer)})}{\sum_{k \in \neigb(j) \cup \{j\}} \exp(e_{kj}^{(\llayer)})}.
\end{equation}

The rest of the aggregation and combination steps are similar to GAT.

Graph Isomorphism Network (GIN) \citep{xu2018how} takes a different approach by focusing on maximizing the expressive power of the aggregation function. The $\aggregate$ function in GIN uses a sum, which is theoretically proven to be as powerful as the Weisfeiler-Lehman graph isomorphism test. It is defined as:

\begin{equation}
    a_j^{(\llayer)} = \sum_{i \in \neigb(j) \cup \{j\}} h_i^{(\llayer-1)},
\end{equation}

and the combination function is:

\begin{equation}
    h_j^{(\llayer)} = \mlp^{(\llayer)} \Big( (1 + \epsilon^{(\llayer)}) h_j^{(\llayer-1)} + a_j^{(\llayer)} \Big),
\end{equation}
where $\epsilon^{(\llayer)}$ is a learnable parameter, and $\mlp^{(\llayer)}$ is a multi-layer perceptron that applies non-linear transformations.

These architectures highlight the diverse strategies in designing $\aggregate$ and $\combine$ functions to achieve expressive and effective graph learning. After $\numlayers$ layers of message passing, each node’s representation incorporates information from its -hop neighborhood. Formally, an $\numlayers$-hop neighborhood of a node  includes all nodes that are at most  edges away from . This property allows GNNs to capture local and higher-order structural information from the graph. However, the effective receptive field of a GNN is inherently limited by the number of layers, which can impact its ability to model long-range dependencies in graphs. Furthermore, as  increases, challenges such as over-smoothing, where node representations become indistinguishable, may arise. Understanding this aggregation behavior is critical for designing effective GNN architectures tailored to specific tasks and datasets. 

After $\numlayers$ iterations of neighborhood aggregation, to produce a graph-level representation, GNNs apply a permutation invariant readout function, \eg~the sum operator, to nodes feature as follows,
\begin{equation}
    h_\G = \readout  \Big( \big\{ h_j^{( \numlayers )} \colon j \in \V \big\} \Big).
\end{equation}

\subsection{Challenges in GNNs}
Despite their growing popularity and success across various applications, GNNs encounter several challenges that limit their performance. Specifically, we delve into issues such as oversmoothing, oversquashing, robustness, generalization, and scalability. 

\subsubsection{Oversmoothing}
mai
In graph neural networks (GNNs), \emph{oversmoothing} refers to the phenomenon where, as the number of layers increases, the node representations become indistinguishable from each other. This typically happens because repeated message passing leads to information from different nodes blending together excessively, making it difficult to distinguish nodes based on their features. This phenomenon can be theoretically understood through the graph Laplacian and quantified using the \emph{Dirichlet Energy} \citep{zhou2021dirichlet}, defined as $$\mathcal{E}_{dir}\left( \left\{ h_i^{(\llayer)}  \mid i \in \V \right\} \right)  =  \frac{1}{2} \sum_{i,j} a_{ij} \left\| \frac{h_i^{(\llayer)}}{\sqrt{1 + \degr(i)}} - \frac{h_j^{(\llayer)}}{\sqrt{1 + \degr(j)}} \right\|_2^2,$$
where  $\left\{ h_i^{(\llayer)}  \mid i \in \V \right\}$ represents the node representations at the $\llayer$-th GNN layer. If the Dirichlet Energy, it indicates that representations of connected nodes become close to one another. Consequently, all node features collapse to an almost constant signal on the graph, making it impossible to distinguish between individual nodes based on their learned embeddings.

\subsubsection{Oversquashing}
Oversquashing refers to the compression of information from a large neighborhood into a fixed-size vector, making it difficult for message-passing neural networks (MPNNs) to propagate and preserve all relevant information across the graph. This problem becomes especially severe in tasks involving long-range dependencies, where distant nodes must exchange signals through multiple hops \citep{dwivedi2022long}. Addressing these challenges requires careful architectural design, such as limiting the number of layers, incorporating skip connections, or using alternative aggregation mechanisms to better preserve and propagate information across the graph.
\subsubsection{Robustness}
Robustness in GNNs refers to their ability to maintain performance under various perturbations, such as noise, adversarial attacks, or incomplete data. In the context of adversarial attacks, referred to as adversarial robustness, the focus is on reducing the effect of changes  to degrade model performance. These attacks typically involve perturbing node features or graph structure and can be categorized into evasion attacks, which occur during testing, and poisoning attacks, which target the training process \citep{zugner2018adversarial}. Enhancing robustness against such attacks can be achieved through methods like adversarial training, which incorporates adversarial examples during training to bolster resistance, or certifiable robustness, which involves developing models with formal guarantees against specific types of perturbations \citep{bojchevski2019certifiable}.

\subsection{Generalization}

GNNs have demonstrated significant success in tasks such as node and graph classification, yet they often face difficulties in generalizing, particularly to unseen or out-of-distribution (OOD) \citep{tang2023towards,li2022ood}.  These challenges are intensified when training data is limited in size or diversity. Various theoretical tools, including Vapnik-Chervonenkis dimension \citep{pfaff2020learning}, Rademacher complexity \citep{yin2019rademacher}, and algorithm stability \citep{pfaff2020learning}, have been employed to provide generalization bounds for GNNs. Notably, \citep{liao2020pac} was the first to establish generalization bounds for GCNs and message-passing neural networks using the PAC-Bayesian approach, while Neural Tangent Kernels have also been utilized to analyze the generalization of infinitely wide GNNs trained via gradient descent \citep{jacot2018neural,du2019graph}.

\subsection{Expressivity}
Expressivity in GNNs measures their capacity to model and distinguish complex graph structures and capture rich node or edge-level dependencies. This is influenced by both the underlying architecture and theoretical constraints. 
The expressivity of GNNs is often analyzed through the \emph{Weisfeiler-Lehman (WL) test} of graph isomorphism, which iteratively updates node labels based on their neighbors and assesses the ability to distinguish non-isomorphic graphs \citep{xu2018how}. Standard message-passing GNNs are constrained by the expressive power of the 1-WL test \citep{li2022expressive}. To address this, researchers have proposed enhancements such as higher-order GNNs, which leverage higher-order WL tests to model interactions among larger substructures \citep{morris2019weisfeiler}. Additional methods include incorporating subgraph information to capture higher-order features \citep{bevilacqua2022equivariant} and using attention mechanisms to enable more nuanced feature aggregation by assigning different levels of importance to neighbors \citep{joshi2023expressive}.

\subsubsection{Scalability}
Scalability is a critical challenge for GNNs, particularly when applied to large-scale graphs with millions or even billions of nodes and edges \citep{hu2020open}. Training a GNN on the large graphs is computationally expensive, necessitating methods like sampling or clustering to handle such scale while maintaining performance. Sampling-based methods  reduce the computational cost by sampling a fixed number of neighbors at each layer rather than processing the full neighborhood \citep{hamilton2017inductive}. Cluster-based methods partition the graph into smaller subgraphs or clusters, allowing GNNs to process these independently and then combine the results. Additionally, efficient architectures, such as scalable attention mechanisms and sparse matrix operations, further optimize performance \citep{gao2022partition}.

\section{Graph Learning Tasks}
Now, we introduce a set of important tasks performed on graph data,
\subsection{Semi-Supervised Node classification}

In semi-supervised node classification, the goal is to predict the labels of nodes within a graph when only a subset of nodes has known labels. Formally, we are given a set of labels $\mathcal{Y}_{train} = \{ y_i, i \in \V_{train} \}$ for a fixed subset of nodes $\V_{train}\subset \V$. The task is to leverage both the labeled nodes in $\V_{\text{train}}$ and the graph structure $\G =(\V,\E)$ to learn a model that predicts the labels $\mathcal{Y}_{test}$ of the remaining test nodes $\V_{\text{test}}$, the unlabeled set.

Semi-supervised node classification is commonly encountered in applications like social networks (\eg, predicting user attributes), citation networks (\eg, predicting article categories), and molecular graphs (\eg, predicting functional groups) \citep{shchur2018pitfalls,wu2018moleculenet}. The GNN architecture is well-suited for this task since it uses the structure and feature information of neighboring nodes, effectively propagating label information from labeled nodes through the graph to inform predictions on unlabeled nodes.
 
\subsection{Graph Classification} 
In the Graph Classification task, the goal is to predict a label for each graph in a given set $\Gset$. Formally, we are provided with a dataset $\Gset = \{\G_1, \G_2, \dots, \G_p\}$, where each graph $\G_i$ is defined by its structure $(\V_i, \E_i)$, the set of nodes and edges, and its node attributes $X_{\G_i}$. Each graph $\G_i$ is associated with a label $y_i \in \Y$, where $\Y$ is the label space, such as binary or multi-class categories. The task is to learn a function $f: \Gset \to \Y$, \eg, a GNN,  that maps graphs to their respective labels. During training, the model learns from a labeled training set $\Gset_{\text{train}}$ to extract meaningful patterns. The learned model is then evaluated on a separate test set $\Gset_{\text{test}}$, to assess its ability to generalize and correctly classify new graphs. 

Graph classification is widely used in many real-word applications. For example, GNNs can predict molecular properties where molecules are represented as graphs with atoms as nodes and chemical bonds as edges \citep{godwin2022simple}. This enables efficient screening of potential drug candidates and understanding molecular interactions. Similarly, in Bioinformatics, graph classification plays a critical role in analyzing biological networks, such as protein-protein interaction networks or gene regulatory networks, to identify disease related patterns or classify functional behaviors of biological entities \citep{hsu2022learning}.

\section{Graph Distance Metrics}
Graphs are widely used to model complex systems across various domains, including social networks, biological systems, and transportation networks. Comparing graphs is a fundamental task that enables the analysis of structural and functional similarities or differences. Graph distance metrics provide a mathematical framework to quantify these similarities and differences by evaluating the adjacency structures and node attributes. These metrics are particularly useful in applications such as graph classification, clustering, and anomaly detection. This section introduces key graph distance measures, focusing on both structural and feature-based differences, along with methods to account for node alignment issues.

There are several approaches to compute graph distance metrics, such as \emph{Edit Distance}, which measures the number of edit operations (additions, deletions, or substitutions) required to transform one graph into another \citep{gao2010survey}; \emph{Graph Kernels}, which map graphs into a feature space where distances can be computed using inner products (e.g., the Weisfeiler-Lehman kernel and Random Walk kernels) \citep{shervashidze2011weisfeiler,nikolentzos2020random}; \emph{Spectral Distances}, which compare eigenvalues and eigenvectors of adjacency or Laplacian matrices to capture structural differences \citep{jovanovic2012spectral}; \emph{Optimal Transport based Distances}, which measures distances by solving optimization problems for node alignments and weights—particularly useful when graphs differ in node ordering \citep{chen2020graph}; and \emph{Embedding-Based Methods}, which represent graphs in vector spaces using graph embeddings and compute Euclidean distances between these representations. By focusing on both structural and feature-based differences and accounting for node alignment issues, these measures provide a robust foundation for numerous graph analysis tasks. 

Building upon the overview of key graph distance measures, we now formalize these metrics within a mathematical framework. This involves defining graph distances that capture structural and feature-based changes. Let us consider the graph space $(\mathbb{A}, \lVert \cdot \rVert_{\mathbb{A}})$ and the feature space $(\mathbb{X}, \lVert \cdot \rVert_{\mathbb{X}})$, where $\lVert \cdot \rVert_{\mathbb{A}}$ and $\lVert \cdot \rVert_{\mathbb{X}}$ denote the norms applied to the graph structure and features, respectively. When considering only structural changes, with fixed node features, the distance between two graphs $\mathcal{G}_1,\mathcal{G}_2$ is defined as

\begin{equation}\label{eq:norm_1}
    \left \| \mathcal{G}_1 - \mathcal{G}_2 \right \|= \lVert \adj_1 - \adj_2 \rVert_{\mathbb{A}},
\end{equation}
where $A_1,A_2$ are respectively the adjacency matrices of $\mathcal{G}_1,\mathcal{G}_2,$. The norm $\lVert\cdot \rVert_{\mathbb{G}}$ can be expressed using different metrics. The Hamming Distance is defined as:
\begin{equation}
    \lVert \adj_1 - \adj_2 \rVert_H = \sum_{i,j} \mathbb{1}(\adj_1(i, j) \neq \adj_2(i, j)),
\end{equation}
which counts the number of differing entries between adjacency matrices. The Frobenius Norm is given by:
\begin{equation}
    \lVert \adj_1 - \adj_2 \rVert_F = \sqrt{\sum_{i,j} (\adj_1(i,j) - \adj_2(i,j))^2},
\end{equation}
measuring the element-wise differences. Finally, the Spectral Norm is expressed as:
\begin{equation}
    \lVert \adj_1 - \adj_2 \rVert_2 = \sigma_{\text{max}}(\adj_1 - \adj_2),
\end{equation}
where $\sigma_{\text{max}}$ denotes the largest singular value of the matrix difference, capturing global structural deviations. These norms provide different perspectives on measuring graph similarity depending on the application. If both structural and feature changes are considered, the distance extends to:

\begin{equation}\label{eq:norm_2}
    \left \| \mathcal{G}_1 - \mathcal{G}_2 \right \| = \alpha \lVert \adj_1 - \adj_2 \rVert_{\mathbb{G}} +\beta \lVert X_1 - X_2 \rVert_{\mathbb{X}},
\end{equation}
where $X_1, X_2$ are the node feature matrices of $\mathcal{G}_1,\mathcal{G}_2$ respectively, and $\alpha, \beta$ are positive hyperparameters controlling the contribution of structural and feature differences.

In cases where the node alignment between the two graphs is unknown, we must take into account node permutations. The distance between the two graphs is then defined as

\begin{equation} \label{eq:norm_3}
    \left \| \mathcal{G}_1 - \mathcal{G}_2 \right \|  = \min_{\mathbf{P} \in \Pi} \left( \alpha \lVert \adj_1 - \mathbf{P} \adj_2 \mathbf{P}^T \rVert_{\mathbb{G}}  + \beta \lVert X_1 - \mathbf{P} X_2 \rVert_{\mathbb{X}} \right),
\end{equation}
where $\Pi$ is the set of permutation matrices. A permutation matrix $\mathbf{P}$ is a binary matrix where each row and column contains exactly one entry equal to 1, and all other entries are 0. Mathematically, it satisfies $\mathbf{P} \mathbf{P}^T = \mathbf{P}^T \mathbf{P}= \identity$,
where $\identity$ is the identity matrix. The set $\Pi$ includes all such matrices of a given size, representing all possible node reorderings.

Permutation matrices are necessary because graphs often do not have a fixed node ordering. For instance, two structurally identical graphs may have different node indices. Direct comparison without alignment may lead to misleading results. The permutation matrix $P$ reorders nodes in $\mathcal{G}_2$ to align with $\mathcal{G}_1$, ensuring structural and feature comparisons are valid.

Using Optimal Transport, we find the minimum distance over the set of permutation matrices, corresponding to the optimal matching between nodes in the two graphs. This approach generalizes graph distance computation to handle cases where node correspondences are unknown.

\section{Synthetic Graphs}
Synthetic graphs play a crucial role in the study of graph-based algorithms  models \citep{tsitsulin2022synthetic}, and machine learning approaches. These graphs are artificially generated using predefined rules and statistical models, enabling researchers to evaluate the scalability, robustness, and effectiveness of various algorithms under controlled conditions. Unlike real-world graphs, synthetic graphs provide flexibility in tuning parameters such as size, density, degree distribution, and clustering, making them ideal for hypothesis testing and benchmarking. In this section, we will introduce two well-known types of synthetic graphs, the Stochastic Block Model (SBM) and the Barabási–Albert (BA) model,which are employed in this thesis analyze graph properties.

\subsection{Stochastic Block Model (SBM)}
The Stochastic Block Model (SBM) is a generative model designed to create graphs with community structures. It partitions nodes into $K$ distinct blocks or communities and specifies intra- and inter-community edge probabilities.

Formally, let $\G = (\V, \E)$ represent a graph with $n$ nodes. Nodes are divided into $K$ communities, and each node $i$ belongs to a community $c_i \in \{1, \dots, K\}$. The probability of an edge between two nodes $i$ and $j$ depends on their community memberships:
\begin{equation}
    \mathbb{P}(a_{ij} = 1) = p_{c_i, c_j},
\end{equation}
where $p_{c_i, c_j}$ is the probability of an edge existing between communities $c_i$ and $c_j$. In a common special case, edges within the same community share a fixed probability values $p$ and edges between different communities share a fixed probability values $q$.  When $p \gg q$, the network exhibits pronounced homophily, \ie, nodes within the same community form densely connected subgroups that are relatively sparse between groups. Conversely, if $ p\approx q$, the graph becomes more uniform, \ie, the model becomes similar to an Erdős–Rényi random graph, making it more challenging to detect distinct communities. SBM is widely used for studying community detection algorithms and modeling social and biological networks \citep{lee2019review,wilkinson2018stochastic}.

\subsection{Barabási–Albert (BA) Model}
The Barabási–Albert (BA) model generates scale-free networks characterized by a power-law degree distribution, mimicking real-world graphs such as social and biological systems. This model builds a graph through a preferential attachment mechanism, which simulates the \emph{rich-get-richer} phenomenon, \ie, nodes with higher degrees are more likely to attract new connections. The construction process involves the following steps,
\begin{enumerate}
    \item Start with a small set of $m_0$ nodes.
    \item At each step, add a new node with $m \leq m_0$ edges.
    \item Connect the new node to existing nodes with a probability proportional to their degree:
    \begin{equation}
        \mathbb{P}(i) = \frac{\degr(i)}{\sum_j \degr(j)},
    \end{equation}
    where $deg(i)$ is the degree of node $i$.
\end{enumerate}

Despite its simplicity, the BA model captures fundamental features of complex graphs, offering a foundation for understanding networks that evolve over time. For example, in social networks, new users tend to connect preferentially with individuals who already have a large number of connections, leading to the formation of hubs that represent highly influential users. 

\section{Benchmarks and Datasets}

In this section, we present the datasets and benchmarks used to evaluate the models in this thesis. The experiments focus on two key tasks, node and graph classifications. For each task, we describe the datasets, their key characteristics Tables \ref{tab:data_statistics_node} and \ref{tab:data_statistics_graph} provide detailed statistical summaries of each dataset.

\subsection{Datasets for the Node Classification Task}
In this thesis, we run experiments on the node classification task using the citation networks Cora, CiteSeer, and PubMed \citep{dataset_node_classification}, the co-authorship networks CS and Physiscs \citep{cs_data}, the citation network between Computer Science arXiv papers OGBN-Arxiv  \citep{hu2020open}, the Amazon Computers and Amazon Photo networks \citep{cs_data}, the non-homophilous datasets Penn94 \citep{traud2012social}, genius \citep{lim2021expertise}, deezer-europe \citep{rozemberczki2020characteristic} and arxiv-year \citep{hu2020open}, and the disassortative datasets Chameleon, Squirrel \citep{rozemberczki2021multi},  and Cornell, Texas, Wisconsin from the WebKB dataset \citep{lim2021large}.  For the Cora, CiteSeer, and Pubmed datasets, we used the provided train/validation/test splits. For the remaining datasets, we followed the framework in \citep{lim2021large,rozemberczki2021multi}. Characteristics and information about the datasets utilized in the node classification part of the study are presented in Table \ref{tab:data_statistics_node}. 

\begin{table}[h]

\caption{Statistics of the node classification datasets.}
\label{tab:data_statistics_node}
\vskip 0.15in
\begin{center}
\begin{tabular}{lrrrrr}
\hline
Dataset & \#Features & \#Nodes & \#Edges & \#Classes & Edge Homophily \\
\hline
Cora    & 1,433 & 2,708   & 5,208    & 7 & 0.809 \\
CiteSeer   & 3,703 & 3,327 & 4,552 & 6 & 0.735\\
PubMed    & 500 & 19,717 & 44,338 & 3 & 0.802\\
CS    & 6,805 & 18,333 & 81,894 & 15 & 0.808\\
arxiv-year    & 128 & 169,343 & 1,157,799 & 5 &  0.218\\
chameleon  &   2,325  &   2,277  &   62,792  &   5  &   0.231\\
Cornell  &   1,703  &   183  &   557  &   5  &   0.132 \\
deezer-europe  &   31,241  &   28,281  &   185,504  &   2  &   0.525 \\
squirrel  &   2,089  &   5,201  &   396,846  &   5  &   0.222 \\
Wisconsin  &   1,703  &   251  &   916  &   5  &   0.206  \\
Texas  &   1,703  &   183  &   574  &   5  &   0.111 \\
Photo  &   745  &   7,650  &   238,162  &   8  &   0.827 \\
ogbn-arxiv  &   128  &   169,343  &   2,315,598  &   40  &   0.654  \\
Computers  &   767  &   13752  &   491,722  &   10  &   0.777 \\
Physics  &   8,415  &   34,493  &   495,924  &   5  &   0.931 \\
Penn94  &   4,814  &   41,554  &   2,724,458  &   3  &   0.470  \\
\hline
\end{tabular}

\end{center}
\vskip -0.1in

\end{table}

\subsection{Datasets for the Graph Classification Task}

For the graph classification task, we evaluate our models on widely used datasets from the GNN literature, specifically IMDB-BINARY, IMDB-MULTI, PROTEINS, MUTAG, and DD, all sourced from the TUD Benchmark \citep{ivanov2019understanding}. These datasets consist of either molecular or social graphs. Detailed statistics for each dataset are provided in Table \ref{tab:data_statistics_graph}.

\begin{table}[h]
\caption{Statistics of the graph classification datasets.}
\label{tab:data_statistics_graph}
\begin{center}
\begin{small}
\begin{tabular}{lcccc}
\toprule
Dataset & \#Graphs & Avg. Nodes & Avg. Edges & \#Classes \\
\midrule
IMDB-BINARY    & 1,000 & 19.77 & 96.53  & 2\\
IMDB-MULTI    & 1,500 & 13.00 & 65.94 & 3\\
MUTAG    & 188  &17.93  & 19.79 & 2\\
PROTEINS    & 1,113 & 39.06 & 72.82 & 2 \\
DD    & 1,178 & 284.32 & 715.66 & 2 \\

\bottomrule
\end{tabular}
\end{small}
\end{center}
\end{table}

\section{Evaluation Metrics}
In this section, we describe the evaluation metrics used to assess the performance of the proposed models for both node and graph classification tasks.

\subsection{Node Classification}
Node classification aims to predict labels for nodes in a graph based on their features and connectivity. Given a graph $\G = (\V, \E)$ with $|\V|$ nodes and a set of true labels $\Y = \{y_i \mid y_i \in \{1, \dots, C\}\}$ for each node $i \in V$, where $C$ is the number of classes, we define the predicted labels as $\widehat{\Y} = \{\widehat{y}_i\}$.

The classification accuracy for node classification is computed as:
\begin{equation}
    \text{Accuracy} = \frac{1}{|\V_{test}|} \sum_{i \in V_{test}} \mathbb{1}(y_i = \widehat{y}_i),
\end{equation}
where $V_{test}$ denotes the set of test nodes, and $\mathbb{1}(\cdot)$ is the indicator function that equals 1 if its argument is true and 0 otherwise.

\subsection{Graph Classification}
For graph classification, the task is to assign a label $y_i \in \{1, \dots, C\}$ to each graph $G_i$ in a set of graphs $G = \{\G_1, \G_2, \dots, \G_N\}$, where $N$ is the number of graphs.

Let $\widehat{y}_i$ denote the predicted label for graph $G_i$. The classification accuracy for graph classification is then defined as:
\begin{equation}
    \text{Accuracy} = \frac{1}{N_{test}} \sum_{i=1}^{N_{test}} \mathbb{1}(y_i = \widehat{y}_i),
\end{equation}
where $N_{test}$ is the number of test graphs, and $\mathbb{1}(\cdot)$ is the indicator function.

\section{Software and Tools}

In this section, we outline the software libraries, frameworks, and tools used to conduct the experiments and analyses presented in this thesis. The selected tools were chosen based on their efficiency, scalability, and support for graph-based computations, making them particularly suitable for graph neural network (GNN) research.
\subsection{Programming Languages and Frameworks}

\begin{itemize}
    \item \textbf{Python:} The primary programming language used for implementing models, algorithms, and experiments. Its extensive ecosystem supports scientific computing and machine learning tasks.
    \item \textbf{PyTorch: \citep{paszke2019pytorch}} A flexible and efficient deep learning framework utilized for implementing and training neural network architectures, including GNN models.
\end{itemize}
\subsection{Graph Processing Libraries}
\begin{itemize}
    \item \textbf{NetworkX: \citep{hagberg2008exploring}} Employed for graph creation, manipulation, and analysis. It provides a comprehensive collection of graph algorithms and visualization tools, aiding in initial data exploration and preprocessing.
    \item \textbf{PyTorch Geometric (PyG): \citep{Fey/Lenssen/2019}} A specialized library for deep learning on graphs, enabling the implementation of graph neural networks. It supports tasks such as node classification, graph classification, and link prediction through optimized message-passing operations.
\end{itemize}
\part{Representation Learning in Graph Neural Networks}
\chapter[{Rethinking GSOs in GNNs for Graph Representation Learning}]{Rethinking Graph Shift Operators in GNNs for Graph Representation Learning} \label{ch:Representation}
\lettrine[lines=3]{G}{\small{raph}} Shift Operators (GSOs), such as the adjacency and graph Laplacian matrices, play a fundamental role in graph theory and graph representation learning. Traditional GSOs are typically constructed by normalizing the adjacency matrix by the degree matrix, a local centrality metric. In this work, we instead propose and study Centrality GSOs (CGSOs), which normalize adjacency matrices by global centrality metrics such as the PageRank, $k$-core or count of fixed length walks. We study spectral properties of the CGSOs, allowing us to get an understanding of their action on graph signals. We confirm this understanding by defining and running the spectral clustering algorithm based on different CGSOs on several synthetic and real-world datasets. We furthermore outline how our CGSO can act as the message passing operator in any Graph Neural Network and in particular demonstrate strong performance of a variant of the Graph Convolutional Network and Graph Attention Network using our CGSOs on several real-world benchmark datasets. 

\section{Introduction}\label{cgsO:introduction}
We propose and study a new family of operators defined on graphs that we call Centrality Graph Shift Operators (CGSOs).  
To insert these into the rich history of matrices representing graphs and centrality metrics, the two concepts married in CGSOs, we begin by recalling major advances in these two topics in turn (readers interested purely in recent developments in Graph Representation Learning and Graph Neural Networks are recommended to begin reading in Paragraph 3 of this section). 
The study of graph theory and with it the use of matrices to represent graphs have a long-standing history.  
Graph theory is often said to have its origins in 1736 when Leonard Euler posed and solved the K{\"o}nigsberg bridge problem \citep{euler1736solutio}. His solution did not involve any matrix calculus. In fact, it seems that the first matrix defined to represent graph structures is the \textit{incidence matrix} defined by Henri Poincaré in 1900 \citep{poincare1900second}. It is difficult to pinpoint the first definition of \textit{adjacency matrices}, but by 1936 when the first book on the topic of graph theory was published by D{\'e}nes K{\"o}nig adjacency matrices had certainly been defined and began to be used to solve graph theoretic problems \citep{konig1936theorie}. Two seemingly concurrent works in 1973 defined an additional matrix structure to represent graphs that later became known as the \textit{unnormalized graph Laplacian} \citep{donath1973lower,fiedler1973algebraic}. Then, it was Fan Chung in her book ``Spectral Graph Theory'' published in 1997 who extensively characterized the spectral properties of \textit{normalized Laplacians} \citep{Chung:1997}. In the emerging field of Graph Signal Processing (GSP) \citep{sandryhaila2013discrete,ortega2018graph} these different graph representation matrices were all defined to belong to a more general family of operators defined on graphs, the \textit{Graph Shift Operators (GSOs)}. GSOs currently play a crucial role in graph representation learning research, since the choice of GSO, used to represent a graph structure, corresponds to the choice of message passing function in the currently much-used Graph Neural Network (GNN) models. 

In parallel to advances in graph representation via matrices, centrality metrics have proved to be insightful in the study of graphs. Chief among them is the success of the PageRank centrality criterion revealing the significance of certain webpages \citep{brin1998anatomy} and playing a role in the formation of what is now one of the largest companies worldwide. But also an even older metric, the $k$-core centrality \citep{seidman1983network,kcore-vldbj20}, as well as the degree centrality, closeness centrality, and betweenness centrality, have proven to be impactful in revealing key structural properties of graphs \citep{freeman1977set,zhang2017degree}. 

A commonality of the most frequently used GSOs is their property to encode purely local information in the graph, with the adjacency matrix encoding neighborhoods in the graph and the graph Laplacians relying on the node degree, a local centrality metric, to normalize the adjacency matrix. In this work, we study a novel class of GSOs, the Centrality GSOs (CGSOs) that arise from the normalization of the adjacency matrix by centrality metrics such as the PageRank, $k$-core and the count of fixed length walks emanating from a given node. Our CGSOs introduce global information into the graph representation without altering the connectivity pattern encoded in the original GSO and therefore, maintain the sparsity of the adjacency matrix. We provide several theorems characterizing the spectral properties of our CGSOs. We confirm the intuition gained from our theoretical study by running the spectral clustering algorithm on the basis of our CGSOs on 1) synthetic graphs that are generated from a stochastic blockmodel in which each block is sampled from the Barrabasi-Albert model and 2) the real-world Cora graph in which we aim to recover the partition provided by the $k$-core number of each node. We will furthermore describe how our CGSOs can be inserted as the message passing operator into any GNN and observe strong performance of the resulting GNNs on real-world benchmark datasets. 


In particular, our contributions can be summarized as follows. 

\begin{enumerate}
    \item[\textbf{(i)}] We define Centrality GSOs, a novel class of GSOs based on the normalization of the adjacency matrix with different centrality metrics, such as the degree, PageRank score, $k$-core number, and the count of walks of a fixed length,
    \item[\textbf{(ii)}] We conduct a comprehensive spectral analysis to unveil the fundamental properties of the CGSOs. Our gained understanding of the benefits of CGSOs is confirmed by running the spectral clustering algorithm using our CGSOs on synthetic and real-world graphs. 
    \item[\textbf{(iii)}] We incorporate the proposed CGSOs within GNNs and evaluate performance of a Graph Convolutional Network and Graph Attention Network v2 with a CGSO message passing operator on several real-world datasets.
\end{enumerate}

\section{Related Work}\label{cgsO:background}

This section builds upon the earlier introduction to GNNs by providing a rigorous introduction of their matrix-based formulation.

\textbf{Matrix Formulations of Graph Neural Networks.} Graph Neural Networks (GNNs) are neural networks that operate on graph-structured data that is defined as the combination of a graph $\G = (\V, \E)$, and a node feature matrix $\mathbf{X}\in\mathbb{R}^{n\times K},$ containing the node feature vector of node $i$ in its $i^{\mathrm{th}}$ row.  GNNs are formed by stacking several computational layers, each of which produces a hidden representation for each node in the graph, denoted by $\mathbf{H}^{(\ell)} = [h^{(\ell)}_v]_{v\in \mathcal{V}}$. A GNN layer $\ell$ updates node representations relying on the structure of the graph and the output of the previous layer $\mathbf{H}^{(\ell-1)}$. Conventionally, the node features are used as input to the first layer $\mathbf{H}^{0} =\mathbf{X}$. The most popular framework of GNNs is that of Message Passing Neural Networks \citep{hamilton2020graph}, where the computations are split into two main steps:

    \textbf{Message Passing:} Given a node $v$, this step applies a permutation-invariant function to its neighbors, denoted by $\mathcal{N}(v),$ to generate the aggregated representation, 
    \begin{equation}
        \mathbf{M}^{(\ell+1)} = 
        \Phi(\mathbf{A})   \mathbf{H}^{(\ell)}, \label{cgsO:eq:message_passing}
    \end{equation}
    where $\Phi(\mathbf{A}): \mathbb{R}^{N\times N}\rightarrow \mathbb{R}^{N\times N}$, a function of the adjacency matrix, is the chosen GSO. 
    
    \textbf{Update:} In this step, we combine the aggregated hidden states with the previous hidden representation of the central node $v,$ usually by making use of a learnable function,
        \begin{equation}
        \mathbf{H}^{(\ell+1)} = \sigma (\mathbf{M}^{(\ell+1)}  W^{(\ell)}), \label{cgsO:eq:update}
        \end{equation}
where $W^{(\ell)} \in \mathbb{R}^{d_{\ell-1}, d_{\ell}}$ are learnable weight matrices and $d_\ell$ is the dimension of the hidden representation at the $\ell$-th layer.  

With the emergence and increasing popularity of GNNs, the importance of GSOs has significantly increased. Numerous GNN architectures, such as notably Graph Convolutional Networks (GCNs), rely on these operators in their message passing step.  
In the context of GCNs \citep{kipf2016semi}, the used message passing operator, i.e., the chosen GSO, corresponds to $\Phi(\mathbf{A}) = \mathbf{D}_1^{-1/2}\mathbf{A}\mathbf{D}_1^{-1/2}$, where $\mathbf{D}_1 = \mathbf{D}+\mathbf{I}$ is the degree matrix of the graph corresponding to the adjacency matrix $\mathbf{A}_1= \mathbf{A}+ \mathbf{I}$. 
For Graph Attention Networks v2 (GATv2) \citep{brody2022how}, \eqnref{cgsO:eq:message_passing} becomes $\mathbf{M}^{(\ell+1)} = \Phi(\mathbf{A}_{\text{GATv2}}^{(\ell)})\mathbf{H}^{(\ell)},$ where, in this setting, $\Phi$ corresponds to the identity function and the rows of $\mathbf{A}_{\text{GATv2}}^{(\ell)}$  contain the edge-wise attention coefficients.

As we will present shortly, in this work, we generalize the concept of GSOs to encompass global structural information beyond node degree. The proposed CGSO framework encapsulates several global centrality criteria, demonstrating intriguing spectral properties. We further leverage CGSOs to formulate a new class of message passing operators for GNNs, enhancing model flexibility.

\textbf{Global Information in GNNs.} Besides our GNNs, which leverage the CGSO to make global information accessible to any given GNN layer, there exists a plethora of other approaches to achieve this goal. These include for example the PPNP and APPNP \citep{gasteiger2019predict}, as well as the PPRGo \citep{bojchevski2020scaling} models that use the PageRank centrality to define a completely new graph over which to perform message passing in GNNs. The work of \citet{ramos2022improving} extends these models to consider both the PageRank and $k$-core centrality.  In addition, there is the AdaGCN \citep{sun2019adagcn} and the VPN model \citep{jin2021power} which propose to message pass using powers of the adjacency matrix to incorporate global information and increase the robustness of GNNs, respectively. \citet{lee2019graph} propose the Motif Convolutional Networks, that define motif adjacency matrices and then use these in the message passing scheme. Also the $k$-hop GNNs of \cite{nikolentzos2020k} consider neighbors several hops away from a given central node in the message passing scheme of a single GNN layer to consider more global information in a GNN. Additionally there exists a rich and long-standing literature on spectral GNNs that facilitate global information exchange by explicitly or approximately making use of the spectral decomposition of the GSO chosen to be the GNN's message passing operator \citep{bruna2013spectral, defferrard2016convolutional, koke2023holonets}. Finally, there is an arm of research investigating graph transformers, where usually the graph structure is only used to provide structural encodings of nodes and the optimal message passing operator is learning using an attention mechanism \citep{kreuzer2021rethinking, rampášek2023recipe, ma2023graph}.  
All these approaches increase the computational complexity of the GNN, whereas our CGSO based GNNs maintain the complexity of the underlying GNN model by preserving the sparsity of the original adjacency matrix.

\section{CGSO: Centrality Graph Shift Operators}\label{cgsO:topo_gso}

In this section, we introduce the Centrality GSOs (CGSO), a family of shift operators that incorporate the global position of nodes in a graph. We discuss different instances of CGSOs corresponding to widely used centrality criteria. We further conduct a comprehensive spectral analysis to unveil the fundamental properties of CGSOs, including the eigenvalue structure and the expansion properties, examining how these operators influence information spread across the graph. Then, we leverage CGSOs in the design of flexible GNN architectures.

\subsection{Mathematical Formulation} \label{cgsO:sec:formulation}
For a given node $i \in \mathcal{V}$, let $v(i)$ denote a centrality metric associated with $i$, such as the node degree, $k$-core number, PageRank, or the count of walks of specific length starting from node~$i$. The Hilbert space $L^2(G)$ is characterized by the set of functions $\varphi$ defined on $\mathcal{V}$ such that $\sum_{i\in \mathcal{V}} v(i) \vert \varphi(i)\vert $ converges, equipped with the inner product:
$\langle\varphi_1, \varphi_2\rangle_{G} = \sum_{i\in  \mathcal{V} } v(i) \varphi_1(i)  \bar{\varphi}_2(i).$
The \textit{Markov Averaging Operator} on $L^2(G)$ is defined as the linear map $\mathbf{M}_{G}: \varphi \mapsto \mathbf{M}_{G} \varphi$ such that
\begin{equation*}
\left (\mathbf{M}_{G} \varphi \right )(i)= \left ( \mathbf{V}^{-1} \mathbf{A} \varphi \right )(i) = \frac{1}{v(i)} \sum_{j \in \mathcal{N}_i} \varphi(j),
\end{equation*}
where $\mathbf{V} = \text{\textit{diag}}(v(1),\ldots, v(N) )$ and $\mathcal{N}_i$ is the neighborhood set of node $i$. The form of this Markov Averaging Operator gives rise to the simplest formulation of our CGSOs, which is a left normalization of the adjacency matrix by a diagonal matrix containing node centralities on the diagonal, i.e., $ \mathbf{V}^{-1} \mathbf{A}.$ Note that the \textit{mean aggregation} operator, as discussed in \citet{xu2018how}, represents a specific instance of these CGSOs where the degree corresponds to the chosen centrality metric, namely $\mathbf{V}=\mathbf{D}$. We will further extend the concept of CGSOs in \eqnref{cgsO:eqn:PCGSO} where we extend and parameterize these CGSOs. In this paper, we focus on three global centrality metrics, in addition to the local node degree. We recall the definitions of these global centrality metrics now. 

\textbf{$\boldsymbol{k}$-core.} The $k$-core number of a node can be determined in the process of the $k$-core decomposition of a graph, which captures how well-connected nodes are within their neighborhood \citep{kcore-vldbj20}. The process of $k$-core decomposition involves iteratively removing vertices with degree less than $k$ until no such vertices remain. The core number $k$ of a node is then equal to the largest $k$ for which the considered nodde is still present in the graph's $k$-core decomposition. We define $\mathbf{V}_{\text{\textit{core}}}\in \mathbb{R}^{N\times N}$ to be the diagonal matrix indicating the core number of each node, \ie
$ \forall i \in \mathcal{V}, ~\mathbf{V}_{\text{\textit{core}}}[i,i] = core(i).$  

\textbf{PageRank.} We choose $\mathbf{V}_{\text{\textit{PR}}} \in \mathbb{R}^{N\times N}$ such that, $ \forall i \in \mathcal{V}, ~\mathbf{V}_{\text{\textit{PR}}}[i,i] = (1-\text{\textit{PR}}(i))^{-1},$
where $\text{\textit{PR}}(i)$ corresponds to the PageRank score \citep{brin1998anatomy}. The PageRank score quantifies the likelihood of a random walk visiting a particular node, serving as a fundamental metric for evaluating node significance in various networks.

\textbf{Walk Count.} Here, we consider $\mathbf{V}_{\ell\text{\textit{-walks}}}\in \mathbb{R}^{N\times N}$, the diagonal matrix indicating the number of walks of length $\ell$ starting from each node $i$, i.e., $ \forall i \in \mathcal{V}, ~\mathbf{V}_{\ell\text{\textit{-walks}}}[i,i] = \left (\mathbf{A}^\ell \mathbb{1}\right )[i], $ where $\mathbb{1} \in \mathbb{R}^N$ is the vector of ones. When $\ell=2$, $\mathbf{V}_{\ell\text{\textit{-walks}}}$ corresponds to $\mathbf{W_{M_{13}}} \mathbb{1} - \mathbf{D} $, where $\mathbf{W_{M_{13}}}$ the graph operator presented by \citet{benson2016higher}, which corresponds to the count of open bidirectional wedges, i.e., the motif $M_{13}$. This motif network captures higher-order structures and gives new insights into the organization of complex systems.

In what follows, we delve into the theoretical properties of Markov Averaging Operators, since all three CGSOs $\mathbf{V}_{\text{\textit{core}}},\mathbf{V}_{\text{\textit{PR}}}$ and $\mathbf{V}_{\ell\text{\textit{-walks}}}$ are instances of Markov Averaging Operators.   

\begin{proposition}\label{cgsO:prop:spectral_properties} The following properties of operator $\mathbf{M}_{G}$ hold.
  \begin{enumerate}[label={(\arabic*)}]
    \item $\mathbf{M}_{G}$ is self-adjoint.

    \item $\mathbf{M}_{G}$ is diagonalizable in an orthonormal basis, its eigenvalues are real numbers, and all eigenvalues have absolute values at most $\gamma = \min_{i \in \mathcal{V} } \left ( \frac{v(i)}{deg(i)} \right ) $.
    \end{enumerate}  
\end{proposition}

The proof of Proposition \ref{cgsO:prop:spectral_properties} and all subsequent theoretical results in this section can be found in Appendix \ref{cgsO:app_proof_spectral}. 
Hence, we have shown in Proposition \ref{cgsO:prop:spectral_properties} that all CGSOs have a real set of eigenvalues, which is of real use in practice. 

In the now following Proposition \ref{cgsO:prop:eigenvalues} we provide the mean and standard deviation of the spectrum of $M_G$, i.e., the set of $M_G$'s eigenvalues.
\begin{proposition}\label{cgsO:prop:eigenvalues}
The following properties hold for the spectrum of $M_G$.
\begin{enumerate}[label={(\arabic*)}]
    \item In a graph $G=(\mathcal{V},\mathcal{E})$ with multiple connected components $\mathcal{C} \subset \mathcal{V},$ where each connected component $\mathcal{C}$ induces a subgraph of $G$ denoted by $G_\mathcal{C},$ a complete set of eigenvectors of $\mathbf{M}_G$ can be constructed from the eigenvectors of the different $\mathbf{M}_{G_\mathcal{C}},$ where eigenvectors of $\mathbf{M}_{G_\mathcal{C}}$ are extended to have dimension $N$ via the addition of zero entries in all entries corresponding to nodes not in the currently considered component $\mathcal{C}.$
    
    \item The mean $\mu(\mathbf{M}_{G})$ and standard deviation $\sigma(\mathbf{M}_{G})$ of $\mathbf{M}_{G}$'s spectrum have the following analytic form  
    $$         
    \begin{array}{ll}
        \mu\left (  {\mathbf{M}_{G}}  \right ) = \frac{1}{n} \sum_{i=1}^n \frac{1}{v(i)} ,\\ 
        \sigma\left (  {\mathbf{M}_{G}} \right ) =  \left [ \left ( \frac{1}{n} \sum_{(i,j) \in \mathcal{E}} \frac{1}{v(i)  v(j) }    \right ) -  \mu\left (  sp_\phi  \right )^2   \right ]^{1/2}.
        \end{array}
 $$

\end{enumerate}
\end{proposition}



We define the \textit{normalized spectral gap} $\lambda_1(G)$ as the smallest non-zero eigenvalue of $\mathbf{I}-\mathbf{M}_{G}$. In Proposition \ref{cgsO:prop:cheeger}, we link $\lambda_1(G)$ to the expansion properties of the graph.  In the literature, we characterize graph expansion via the \textit{expansion} or \textit{Cheeger constant} \citep{Chung:1997}, which measures the minimum ratio between the size of a vertex set and the minimum degree of its vertices, reflecting the graph's connectivity. In our work, we generalize this definition to any centrality metric. 

\begin{definition}\label{cgsO:defn:ExtendedCheeger}
For a graph  $G=(\mathcal{V},\mathcal{E})$  we define the \textit{centrality-based Cheeger constant} $h_v(G) $ as follows
\begin{equation}
h_v(\G) = \min \left \{\frac{|\partial U|}{|U|_v} \mid U \subset V, |U|_v\leq \frac{1}{2} |\mathcal{V}|_v \right\},
\end{equation} 
where $|\partial U|$ equals the number of vertices that are connected to a vertex in $U$ but are not in $U$, and $|\cdot|_v : U \subset \mathcal{V} \mapsto \sum_{i\in \mathcal{V} }v(i)$. When the chosen centrality is the degree, $h_v(G)$ corresponds to the classical Cheeger constant.
\end{definition}
Definition \ref{cgsO:defn:ExtendedCheeger} allows us to establish a link between the spectrum of our considered Markov operators, i.e., CGSOs, and the centrality-based Cheeger constant in Proposition \ref{cgsO:prop:cheeger}. 
    \begin{proposition}  \label{cgsO:prop:cheeger}
        Let $G$ be a connected, non-empty, finite graph without isolated vertices. We have,
        $$\lambda_1(G) \leq \left (  2 N \frac{ v_{+}^2}{v_{-}} \right )h_v(\G)  ,$$
            where we denote 
            $v_{-} = \min_{i\in \mathcal{V}} v(i) $ and $v_{+} = \max_{i\in \mathcal{V}} v(i).$
    \end{proposition}

\subsection{CGNN: Centrality Graph Neural Network} \label{cgsO:gnn_experiments}

CGSOs, as defined above, normalize the adjacency matrix based on the centrality of the nodes, thereby providing a refined representation of graph connectivity. Here, we leverage CGSOs to design flexible message passing operators in GNNs. Incorporating CGSOs within GNNs aims to harness structural information, enhancing the model's ability to discern subtle topological patterns for prediction tasks. To achieve this, we integrate these operators, without loss of generality, in Graph Convolutional Networks (GCNs)  \citep{kipf2016semi} and Graph Attention Networks v2 (GATv2)\citep{brody2022how}. We replace the initial shift operator $\Phi(\mathbf{A})$ in  \eqnref{cgsO:eq:message_passing}, with the proposed CGSOs $\Phi(\mathbf{A}, \mathbf{V})$, incorporating different types centrality operators $\mathbf{V}$ defined in Section \ref{cgsO:sec:formulation}.

It has been shown that the maximum PageRank score converges to zero when the total number of nodes is very high \citep{cai2021rankings}, which is the case in many real-world dense graph data \citep{leskovec2010kronecker,leskovec2012learning}. Also, the number of walks is high when the expansion of the graph is high. Thus, training a GNN with the proposed CGSOs can lead to numerical instabilities such as vanishing and exploding gradients. To avoid such issues, we can control the range of the eigenvalues of CGSOs. We particularly consider a learnable parameterized CGSO framework which is a generalization of the work of \cite{dasoulas2021learning}. This has the further advantage that the CGSOs are fit to the given datasets and learning tasks, which leads to more accurate and higher performing graph representation. The exact formula of the new parametrized CGSO is
\begin{equation} \label{cgsO:eqn:PCGSO}
\Phi(\mathbf{A}, \mathbf{V}) =  m_1 \mathbf{V}^{e_1} + m_2 \mathbf{V}^{e_2} \mathbf{A}_a \mathbf{V}^{e_3} + m_3 \mathbf{I}_N,
\end{equation}

where $\mathbf{A}_a = \mathbf{A}+a\mathbf{I}_N$, and $(m_1,m_2,m_3,e_1,e_2,e_3,a)$ are  scalar parameters that are learnable via backpropagation. Here $m_1$ controls the additive centrality normalization of the adjacency matrix. The parameter  $e_1$ controls whether the additive centrality normalization is performed with an emphasis on large centrality values (for large positive values of $e_1$) or with an emphasis on small centrality values (for large negative values of $e_1$). Similarly, we have $e_2$ and $e_3$ controlling the emphasis on large or small centralities, as well as whether the multiplicative centrality normalization of the adjacency matrix is performed symmetrically or predominantly as a column or row normalization. The parameter $m_2$ controls the magnitude and sign of the adjacency matrix term; in particular, a negative $m_2$ corresponds to a more Laplacian-like CGSO, while a positive $m_2$ gives rise to a more adjacency-like CGSO. Finally, $a$ determines the weight of the self-loops that are added to the adjacency matrix, and $m_3$ controls a further diagonal regularization term of the CGSO.  More details on the experimental setup are provided in Section \ref{cgsO:experim_setup}.  

In our experiments, we notice the best centrality to vary across datasets, although the walk-based centrality CGSO appears to be frequently outperformed  by the $k$-core and PageRank CGSO. More particularly, in some cases e.g. PubMed, it is desirable to use local centrality metrics such as the degree, while for other datasets e.g., Cornell, it's preferable to normalize the adjacency with global centrality metrics. In light of this uncertainty, we can opt for a dynamic, trainable choice of centrality by including both local and global centrality-based CGSO in our CGNN; this can be done by summing the CGSO of the degree matrix with the CGSO of a global centrality metric, e.g., $$\mathbf{ \Phi} = \mathbf{ \Phi} (\mathbf{A}, \mathbf{D})+\mathbf{ \Phi}(\mathbf{A}, \mathbf{V_{\textit{core}}}) $$. 

The parameters $m_1,m_2,m_3$ controlling the magnitude of both the local and global CGSOs are then able to learn the relative importance of the local and the global CGSO. In Section \ref{cgsO:experim_setup}, we provide experimental results for GNNs with such combined CGSOs.

\textbf{Time Complexity.} We recall that the main complexity of our CGCN model is concentrated around the pre-computation of each centrality score. Computing the degree of all nodes in a graph has a time complexity of $\mathcal{O}(|\mathcal{V}| + |\mathcal{E}|)$, where $|\mathcal{V}|$ is the number of nodes and $|\mathcal{E} |$ is the number of edges in the graph \citep{cormen2022introduction}. For the PageRank algorithm, each iteration requires one vector-matrix multiplication, which on average requires $\mathcal{O}(|\mathcal{V}|^2)$ time complexity. To compute the core numbers of nodes, we iteratively remove nodes with a degree less than a specified value until all remaining nodes have a degree greater than or equal to that value. This operation can be done with a complexity of $\mathcal{O}(|\mathcal{V}| + |\mathcal{E}|)$. Finally, counting the number of walks of length $\ell$ for all the nodes can be done via matrix multiplication $\mathbf{A}^\ell  \mathbb{1}$ where $\mathbb{1} \in \mathbb{R}^N$ is the vector of ones. Since our CGSOs preserve the sparsity pattern of the original adjacency matrix, the complexity of the GNNs in which the CGSOs are inserted is unaltered.

\section{A Spectral Clustering Perspective of CGSOs}\label{cgsO:synth_exps}
In this section, we analyze CGSOs through the lens of spectral clustering \citep{von2007tutorial,ng2001spectral}. Spectral clustering is a powerful technique that relies on the spectrum of GSOs to reveal underlying structures within graphs, providing insights into their connectivity properties.

\subsection{Spectral Clustering on Stochastic Block Barabási–Albert Models} \label{cgsO:sec:SBBAM}
Here, we investigate the behavior of CGSOs in the spectral clustering task on synthetic data. Specifically, we propose a new graph generator that is a trivial combination of the well-known Stochastic Block Models (SBM) \citep{holland1983stochastic} and Barabási–Albert (BA) models \citep{albert:02statistical}, we call this generator the Stochastic Block Barabási–Albert Models~(SBBAM). We will now discuss the properties and parameterizations of these two graph generators in turn to then discuss their combination in the SBBAMs. 

\textbf{SBMs.} Firstly, in SBMs the node set of the graph is partitioned into a set of $K$ disjoint blocks $\mathcal{B}_1, \dots, \mathcal{B}_K$, where both the number and size of these blocks is a parameter of the model. In SBMs edges are drawn uniformly at random with probability $p_{ij}$ for $i,j\in \{1, \ldots, K\}$ between nodes in blocks $\mathcal{B}_i$ and $\mathcal{B}_j.$ Note that this parameterization is often simplified by the following constraints $p_{ij} = q$ if $i=j$ and $p_{ij} = p$ if $i\neq j.$ SBMs produce graphs which exhibit cluster structure if $p\neq q,$ which makes them a common benchmark for clustering algorithms and subject to extensive theoretical study \citep{abbe2018community}. Note that SBMs can produce both homophilic graphs if $p<q$ and heterophilic graphs if $q>p$ \citep[Figure 1.2]{Lutzeyer2020}. 

\textbf{BA.} The second ingredient of our SBBAMs are the Barabási–Albert (BA) models \citep{albert:02statistical}. This model generates random scale-free networks using a preferential attachment mechanism, which is why these models are also sometimes referred to as preferential attachment (PA) models. In this PA mechanism we start out with a seed graph and then add nodes to it one-by-one at successive time steps. For each added node $r$ edges are sampled between the added node and nodes existing in the graph, where the probability of connecting to existing nodes is proportional to their degree in the graph. Hence, high degree nodes are more likely to have their degree rise even further than low degree nodes in future time steps of the generation process (an effect, that is some time referred to as `the rich get richer'). BA models characterize several real-world networks \citep{barabasi1999emergence}. A key characteristic of a BA model is their degree distribution. In Lemma \ref{cgsO:lem:BA_average_degree}, we prove that the density and connectivity of a BA model strongly depend on and positively correlate with the hyperparameter $r$. Thus, we can generate structurally different BA models by choosing different values of $r$. Lemma \ref{cgsO:lem:BA_average_degree} is proved in Appendix \ref{cgsO:appendix:proof_lemma_BA}.

\begin{lemma}
\label{cgsO:lem:BA_average_degree}
Let $G^{\text{BA}}$ be a Barabási–Albert graph of $N$ nodes generated with the hyperparameters $N_0<N$ the initial number of nodes, $r_0\leq N_0^2$ the initial number of random edges and $r$ the number of added edges at each time step. Then, the average degree in the network is,
\begin{equation*}
\overline{\text{\textit{deg}}}(\G^{\text{BA}}) = 2r + 2\frac{r_0}{N} - 2N_0 \frac{r }{N},    
\end{equation*}
and thus, as the number of nodes grows, i.e., $N \rightarrow  \infty$, the average degree becomes $\overline{\text{\textit{deg}}}(\G^{\text{BA}}) \sim 2r. $
\end{lemma}

\textbf{SBBAMs.} In our SBBAMs we combine SBMs and BA models, by sampling $K$ BA graphs each of size $|\mathcal{B}_1|, \dots, |\mathcal{B}_K|$ and with parameters $r_1, \ldots, r_K.$ We then randomly draw edges between nodes in different BA graphs, $\mathcal{B}_i$ and $\mathcal{B}_j,$ uniformly at random with probability $p_{ij}$ for $i,j\in \{1, \ldots, K\}.$ In other words, SBBAMs trivally extend SBMs to graph in which each block is generated using a BA model. This allows us to generate graphs with cluster structure, in which the different clusters exhibit potentially interesting centrality distributions, which will serve as an interesting testbed to explore the clustering obtained from the eigenvectors of our CGSOs.

\textbf{Experimental Setting.} To better understand the information contained in the spectral decomposition of our CGSOs we will now generate graphs from our SBBAMs and use the spectral clustering algorithm defined on the basis of our CGSOs to attempt to cluster our generated graphs. 
In our experimental setting, each block or BA graph has 100 nodes and an individual parameter $r,$ specifically, $r_1=5,$ $r_2=10$ and $r_3=15.$ In addition we set $p_{ij}=0.1$ for all $i\neq j$ with $i,j\in \{1, 2, 3\}.$ Figure \ref{cgsO:fig:adj_matrix_combined_graph} in Appendix \ref{cgsO:app:_combined_BA} gives an example of an adjacency matrix sampled from this model. We observe variations in edge density across different blocks and in particular observe homophilic cluster structure in the third block, while the first block appears to be predominantly heterophilic, a rather challenging and interesting structure.

Figure \ref{cgsO:fig:sbm-pam} in Appendix \ref{cgsO:appendix_k_core_distrib} illustrates the $k$-core distribution of the three individual BA blocks and the combined SBBAM. Notably, the $k$-core distribution distinguishes the three BA graphs, while the nodes in the combined graph exhibit less discernibility by $k$-core. 

Following the graph generation, we perform spectral clustering  (see Algorithm \ref{cgsO:algo:spectral_clustering} in Appendix~\ref{cgsO:app:spectral_clustering_algo}) using our CGSOs to asses their ability to recover the blocks in our generated SBBAM. Specifically, we utilize the three eigenvectors of $\mathbf{ \Phi} = \mathbf{V}^{e_2} \mathbf{A} \mathbf{V}^{e_3}$ corresponding to the three largest eigenvalues of different CGSOs defined in Section \ref{cgsO:sec:formulation}. Working with this particular parametrized form our CGSOs further allows us to study the effect of different centrality normalizations with $e_2, e_3 \in [-1.5, 1.5].$ We repeated each experiment 200 times, and then reported the mean and standard deviation of Adjusted Mutual Information (AMI) and Adjusted Rand Index (ARI) values. For consistency, we used the same 200 generated graphs for all the GSOs and the baselines.

\begin{figure*}[t]
\centering 

\includegraphics[width=\textwidth]{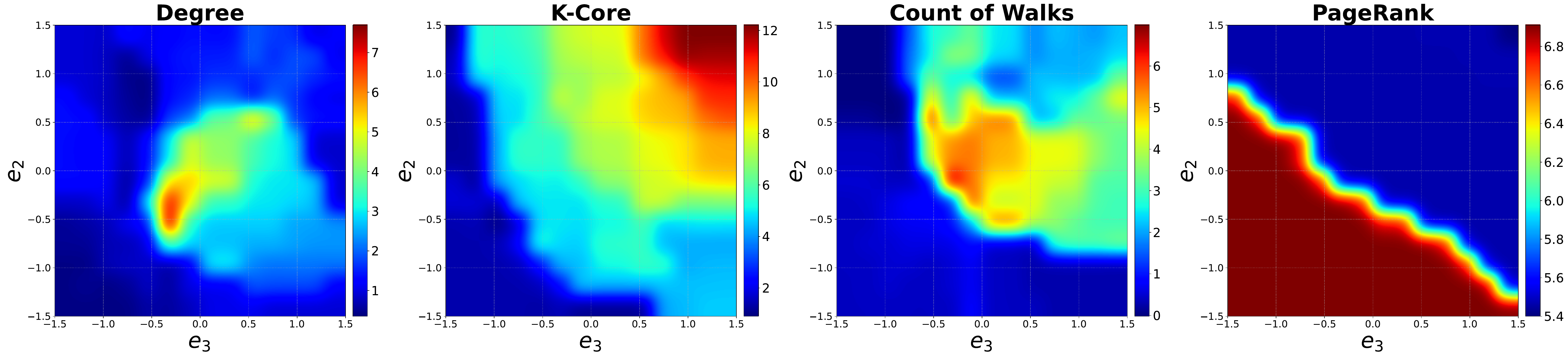}
\caption[Result for the spectral clustering task on the Cora graph]{Result for the spectral clustering task on the Cora graph \citep{dataset_node_classification} with core numbers considered as clusters. We report the values of the Adjusted Mutual Information (AMI) in percentage for different combinations of the exponents $(e_2, e_3)$ in $\mathbf{V}^{e_2} \mathbf{A} \mathbf{V}^{e_3}.$} 
\label{cgsO:fig:ami_cora} 
\end{figure*}

In Figure \ref{cgsO:fig:ami_cora}, we report the AMI values using the four centralities. As noticed, while having competitive results between the degree centrality, the PageRank score and the count of walks, we reach the highest AMI values by using the $k$-core centrality metrics. Using the degree centrality, we reach the highest AMI value when both exponent $e_2$ and $e_3$ are negative, while for the $k$-core and the number of walks, we notice a different behavior as the AMI increase when both the exponents $e_2$ and $e_3$ are positive. Thus, we conclude that nodes with higher $k$-core and count of walks are important for this setup, i.e., when the node labels are positively correlated with global centrality metrics such as the $k$-core. We report the ARI values of the same experiment in Appendix~\ref{cgsO:ari_results_appendix}.  


\subsection{Centrality Recovery in Spectral Clustering}

\begin{table}[h] 

\caption[The result of the spectral clustering task on synthetic graphs]{The result of the spectral clustering task on the synthetic graph data.  We present the mean and standard values of AMI and ARI in percentage. \textcircled{1} Spectral clustering using the centrality based GSOs, \textcircled{2} Other baselines.}
\centering
\resizebox{0.5\textwidth}{!}{%
\begin{tabular}{clrr}\toprule
   & \textbf{Method} & \textbf{AMI in \%}      & \textbf{ARI in \%}            \\
    \toprule
\multirow{4}{*}{\textcircled{2}}& Fast Greedy & $17.27 {\scriptstyle \pm 4.28}$ &  $19.98 {\scriptstyle \pm 5.03}$ \\
& Louvain &$14.37  {\scriptstyle \pm 3.34}$ & $14.82  {\scriptstyle \pm 3.92}$  \\
& Node2Vec & $1.11 {\scriptstyle \pm 0.92}$ & $1.17 {\scriptstyle \pm 0.96}$  \\
& Walktrap & $1.39 {\scriptstyle \pm 1.16}$ & $1.14 {\scriptstyle \pm 0.97}$  \\
\midrule
\multirow{4}{*}{\textcircled{1}} & CGSO  w/ $\mathbf{D}$  & $23.26 {\scriptstyle \pm 3.36}$ & $22.95 {\scriptstyle \pm 3.86}$ \\
& CGSO w/ $\mathbf{V}_{\text{\textit{core}}}$      & $\mathbf{35.78 {\scriptstyle \pm 4.67}} $ &  $\mathbf{33.76 {\scriptstyle \pm 5.83}} $\\
& CGSO w/ $\mathbf{V}_{\ell\text{\textit{-walks}}}$ & $23.85 {\scriptstyle \pm 3.64} $ &   $25.00 {\scriptstyle \pm 4.18} $\\
& CGSO  w/ $\mathbf{V}_{\text{\textit{PR}}}$ &  $35.62 {\scriptstyle \pm 4.90}$ &  $33.19 {\scriptstyle \pm 6.03}$ \\ 
\bottomrule
\end{tabular}
}
\label{cgsO:tab:SBM_PAM}
\end{table}
In this experiment, we aim to discern the CGSOs' effectiveness in recovering clusters based on centrality within a real-world graph. Using the Cora dataset,  we chose core numbers to indicate centrality-based clusters.  We aim to assess the capacity of various CGSOs to effectively recover clusters reflective of core numbers. This investigation aims to shed light on their potential utility in capturing centralities and hierarchical structures within intricate graphs.

\textbf{Spectral Clustering on Cora.}  
In this experiment, we consider only the largest connected component of the Cora graph. We use the spectral clustering algorithm on the different CGSOs to recover $K$ clusters, where $K$ is the number of possible core numbers in the graph. We repeat each experiment 10 times, and report the average AMI and ARI values. We also compared our CGSOs with the popular Louvain community detection method \citep{blondel2008fast}, the node2vec node embedding methods \citep{grover2016node2vec} combined with the $k$-means algorithm, the Walktrap algorithm \citep{pons2005computing}, and the Fast Greedy Algorithm which also optimizes modularity by greedily adding nodes to communities \citep{clauset2004finding}. For the walk count node centrality matrix, we used $\ell=2$ in all our experiments. 
We consider the CGSO $\mathbf{ \Phi} = \mathbf{V}^{e_2} \mathbf{A} \mathbf{V}^{e_3}$, where we normalize the adjacency matrix with the topological diagonal matrix $\mathbf{V}$ using different exponents $(e_2, e_3)$. 

The results of the spectral clustering on this synthetic graph are presented in Table \ref{cgsO:tab:SBM_PAM}. As expected, normalizing the adjacency matrix with $k$-core yields higher AMI and ARI values. This observation indicates an improved discernment of each node's membership in its respective cluster, achieved through the incorporation of global centrality metrics. Our CGSO outperforms well-known community detection techniques, such as the Louvain algorithm,  which optimizes the modularity, measuring the density of links inside communities compared to links between communities. However, in our setting, some blocks have fewer inter-edges than intra-edges with other blocks, thus making it difficult for the Louvain algorithm to cluster these nodes using the edge density. This experiment further reinforces the intuition that if different clusters exhibit different centrality distributions then our CGSOs are able to capture  this difference better than other clustering alternatives which leads to better clustering performance.

\section{Experimental Evaluation}\label{cgsO:experim_setup}
We begin by discussing our experimental setup. Further details on the datasets we evaluate on and the training set-up can be found in Appendix \ref{cgsO:app_data_impl}.

\textbf{Baselines.} We experiment with two particular instances of our proposed CGNN model, using a GCN and GATv2 as the backbone models, we refer to this instance as CGCN and CGATv2, respectively. We compared the proposed CGCN to GCN with classical GSOs: the adjacency matrix $\mathbf{A}$,  Unormalised Laplacian $\mathbf{L} = \mathbf{D}-\mathbf{A}$,  Singless Laplacian $\mathbf{Q} = \mathbf{D}+\mathbf{A}$ \citep{cvetkovic2010towards},  Random-walk Normalised Laplacian $\mathbf{L_{rw}} = \mathbf{I}-\mathbf{D}^{-1}\mathbf{A}$,  Symmetric Normalised Laplacian $\mathbf{L_{sym}} = \mathbf{I}-\mathbf{D}^{-1/2}\mathbf{A}\mathbf{D}^{-1/2}$,  Normalised Adjacency $\mathbf{\hat{A}} = \mathbf{D}^{-1/2}\mathbf{A}\mathbf{D}^{-1/2}$ \citep{kipf2016semi} and  Mean Aggregation $\mathbf{H} = \mathbf{D}^{-1}\mathbf{A}$ \citep{xu2018how}. We also compare to other standard GNN baselines: Graph Attention Network (GAT) \citep{veličković2018graph}, Graph Attention Network v2 (GATv2) \citep{brody2022how}, Graph Isomorphism Network (GIN) \citep{xu2018how}, and Principal Neighbourhood Aggregation (PNA) \citep{corso2020principal}.

\begin{table*}[t] 
\centering
\caption[Performance of CGCN, CGATv2 and other vanilla models]{Classification accuracy ($\pm$ standard deviation) of the models on different benchmark node classification datasets. The higher the accuracy (in \%) the better the model. \textcircled{1} GCN-based models \textcircled{2} Other vanilla GNN baselines \textcircled{3} CGCN \textcircled{4} CGATv2. Highlighted are the \textbf{first}, \underline{second} best results. OOM means \textit{Out of memory}.}
\label{cgsO:tab:topological_GCN}
\resizebox{\textwidth}{!}{%
\begin{tabular}{cl|llllllll}\toprule
\multicolumn{1}{l}{} & Model & CiteSeer  & PubMed & arxiv-year  & chamelon & Cornell  & deezer-europe  & squirrel & Wisconsin  \\ \midrule
\multirow{7}{*}{\textcircled{1}}   
    & GCN w/ $\mathbf{A}$ & $64.95 {\scriptstyle \pm 0.58}$ & $77.12 {\scriptstyle \pm 0.61}$ & $38.55 {\scriptstyle \pm 0.71}$ & $61.03 {\scriptstyle \pm 1.31}$ & $57.03 {\scriptstyle \pm 3.91}$ & $57.65 {\scriptstyle \pm 0.84}$ & $22.38 {\scriptstyle \pm 6.06}$ & $54.51 {\scriptstyle \pm 1.47}$ \\

    & GCN w/ $\mathbf{L}$ & $28.11 {\scriptstyle \pm 0.54}$ & $43.65 {\scriptstyle \pm 0.71}$ & $32.81 {\scriptstyle \pm 0.29}$ & $56.97 {\scriptstyle \pm 0.75}$ & $54.32 {\scriptstyle \pm 0.81}$ & $53.92 {\scriptstyle \pm 0.59}$ & $36.20 {\scriptstyle \pm 0.84}$ & $60.00 {\scriptstyle \pm 2.00}$ \\

    & GCN w/ $\mathbf{Q}$ & $63.28 {\scriptstyle \pm 0.80}$ & $76.57 {\scriptstyle \pm 0.59}$ & $33.76 {\scriptstyle \pm 2.36}$ & $53.88 {\scriptstyle \pm 2.35}$ & $35.41 {\scriptstyle \pm 2.55}$ & $56.79 {\scriptstyle \pm 1.79}$ & $27.69 {\scriptstyle \pm 2.21}$ & $53.33 {\scriptstyle \pm 0.78}$ \\

    & GCN w/ $\mathbf{L_{rw}}$ & $30.18 {\scriptstyle \pm 0.74}$ & $59.68 {\scriptstyle \pm 1.03}$ & $36.36 {\scriptstyle \pm 0.24}$ & $48.77 {\scriptstyle \pm 0.54}$ & $61.62 {\scriptstyle \pm 1.08}$ & $54.04 {\scriptstyle \pm 0.44}$ & $34.27 {\scriptstyle \pm 0.35}$ & $65.10 {\scriptstyle \pm 0.78}$ \\

    & GCN w/ $\mathbf{L_{sym}}$ & $29.90 {\scriptstyle \pm 0.66}$ & $57.68 {\scriptstyle \pm 0.45}$ & $36.49 {\scriptstyle \pm 0.14}$ & $50.81 {\scriptstyle \pm 0.24}$ & $60.27 {\scriptstyle \pm 1.24}$ & $53.30 {\scriptstyle \pm 0.45}$ & $35.96 {\scriptstyle \pm 0.28}$ & $66.08 {\scriptstyle \pm 2.16}$ \\

    & GCN w/ $\mathbf{\hat{A}}$ & $68.74 {\scriptstyle \pm 0.82}$ & $78.45 {\scriptstyle \pm 0.22}$ & $42.23 {\scriptstyle \pm 0.25}$ & $58.44 {\scriptstyle \pm 0.26}$ & $56.22 {\scriptstyle \pm 1.62}$ & $60.68 {\scriptstyle \pm 0.45}$ & $37.73 {\scriptstyle \pm 0.33}$ & $57.45 {\scriptstyle \pm 0.90}$ \\

    & GCN w/ $\mathbf{H}$ & $66.15 {\scriptstyle \pm 0.55}$ & $76.45 {\scriptstyle \pm 0.48}$ & $41.27 {\scriptstyle \pm 0.21}$ & $56.51 {\scriptstyle \pm 0.47}$ & $54.86 {\scriptstyle \pm 1.24}$ & $59.45 {\scriptstyle \pm 0.50}$ & $38.23 {\scriptstyle \pm 0.47}$ & $54.31 {\scriptstyle \pm 0.90}$ \\ 
\midrule

\multirow{5}{*}{\textcircled{2}}   
    & GIN & $66.62 {\scriptstyle \pm 0.44}$ & $78.22 {\scriptstyle \pm 0.52}$ & $38.27 {\scriptstyle \pm 3.43}$ & $61.60 {\scriptstyle \pm 1.05}$ & $45.95 {\scriptstyle \pm 3.42}$ & OOM & $25.78 {\scriptstyle \pm 5.12}$ & $58.82 {\scriptstyle \pm 1.75}$ \\

    & GAT & $59.84 {\scriptstyle \pm 3.14}$ & $71.55 {\scriptstyle \pm 4.69}$ & $41.26 {\scriptstyle \pm 0.30}$ & $63.60 {\scriptstyle \pm 1.70}$ & $49.46 {\scriptstyle \pm 8.11}$ & $57.67 {\scriptstyle \pm 0.74}$ & $40.37 {\scriptstyle \pm 2.89}$ & $55.88 {\scriptstyle \pm 2.81}$ \\

    & GATv2 & $63.01 {\scriptstyle \pm 2.97}$ & $73.96 {\scriptstyle \pm 2.22}$ & $41.16 {\scriptstyle \pm 0.25}$ & $64.14 {\scriptstyle \pm 1.53}$ & $43.78 {\scriptstyle \pm 4.80}$ & $56.77 {\scriptstyle \pm 1.19}$ & $\mathbf{42.63 {\scriptstyle \pm 2.61}}$ & $53.53 {\scriptstyle \pm 4.12}$ \\

    & PNA & $48.89 {\scriptstyle \pm 11.15}$ & $70.83 {\scriptstyle \pm 6.51}$ & $32.45 {\scriptstyle \pm 2.34}$ & $22.89 {\scriptstyle \pm 1.09}$ & $40.54 {\scriptstyle \pm 0.00}$ & OOM & OOM & $53.14 {\scriptstyle \pm 2.55}$ \\  
\midrule  

\multirow{4}{*}{\textcircled{3}}   
    & CGCN w/ $\mathbf{D}$ & $68.35 {\scriptstyle \pm 0.45}$ & $\mathbf{78.70 {\scriptstyle \pm 1.10}}$ & $45.39 {\scriptstyle \pm 0.45}$ & $\underline{64.17 {\scriptstyle \pm 8.10}}$ & $72.43 {\scriptstyle \pm 13.09}$ & $58.04 {\scriptstyle \pm 1.06}$ & $\underline{42.30 {\scriptstyle \pm 1.34}}$ & $76.86 {\scriptstyle \pm 7.70}$ \\

    & CGCN w/ $\mathbf{V}_{\text{\textit{core}}}$ & $68.40 {\scriptstyle \pm 0.75}$ & $77.91 {\scriptstyle \pm 0.41}$ & $\mathbf{47.27 {\scriptstyle \pm 0.31}}$ & $63.68 {\scriptstyle \pm 5.00}$ & $73.78 {\scriptstyle \pm 12.16}$ & $\underline{60.90 {\scriptstyle \pm 2.28}}$ & $40.59 {\scriptstyle \pm 2.21}$ & $74.90 {\scriptstyle \pm 6.52}$ \\

    & CGCN w/ $\mathbf{V}_{\ell\text{\textit{-walks}}}$ & $67.31 {\scriptstyle \pm 0.75}$ & $77.57 {\scriptstyle \pm 0.37}$ & $39.35 {\scriptstyle \pm 0.49}$ & $\mathbf{66.21 {\scriptstyle \pm 2.49}}$ & $72.70 {\scriptstyle \pm 3.24}$ & $59.15 {\scriptstyle \pm 1.24}$ & $36.03 {\scriptstyle \pm 5.81}$ & $74.90 {\scriptstyle \pm 4.19}$ \\

    & CGCN w/ $\mathbf{V}_{\text{\textit{PR}}}$ & $67.11 {\scriptstyle \pm 0.56}$ & $78.17 {\scriptstyle \pm 4.27}$ & $\underline{47.14 {\scriptstyle \pm 0.31}}$ & $60.94 {\scriptstyle \pm 7.00}$ & $\underline{76.22 {\scriptstyle \pm 16.3}}$ & $\mathbf{63.41 {\scriptstyle \pm 0.77}}$ & $32.17 {\scriptstyle \pm 3.94}$ & $80.78 {\scriptstyle \pm 11.7}$ \\  
\midrule 

\multirow{4}{*}{\textcircled{4}}   
    & CGATv2 w/ $\mathbf{D}$ & $68.60 {\scriptstyle \pm 0.60}$ & $77.46 {\scriptstyle \pm 0.51}$ & $45.09 {\scriptstyle \pm 0.17}$ & $58.22 {\scriptstyle \pm 2.74}$ & $\mathbf{76.49 {\scriptstyle \pm 4.37}}$ & OOM & $35.30 {\scriptstyle \pm 2.32}$ & $\mathbf{85.69 {\scriptstyle \pm 3.17}}$ \\

    & CGATv2 w/ $\mathbf{V}_{\text{\textit{core}}}$ & $\underline{68.83 {\scriptstyle \pm 0.66}}$ & $77.99 {\scriptstyle \pm 0.43}$ & $44.38 {\scriptstyle \pm 0.25}$ & $55.83 {\scriptstyle \pm 2.28}$ & $75.95 {\scriptstyle \pm 3.72}$ & OOM & $34.17 {\scriptstyle \pm 1.45}$ & $\underline{85.10 {\scriptstyle \pm 2.80}}$ \\

    & CGATv2 w/ $\mathbf{V}_{\ell\text{\textit{-walks}}}$ & $68.11 {\scriptstyle \pm 0.91}$ & $75.43 {\scriptstyle \pm 0.89}$ & $46.70 {\scriptstyle \pm 0.21}$ & $55.59 {\scriptstyle \pm 2.57}$ & $74.32 {\scriptstyle \pm 5.70}$ & OOM & $34.25 {\scriptstyle \pm 2.15}$ & $83.53 {\scriptstyle \pm 2.66}$ \\

    & CGATv2 w/ $\mathbf{V}_{\text{\textit{PR}}}$ & $\mathbf{68.97 {\scriptstyle \pm 0.65}}$ & $\underline{78.46 {\scriptstyle \pm 0.23}}$ & $41.64 {\scriptstyle \pm 0.18}$ & $58.82 {\scriptstyle \pm 1.68}$ & $74.05 {\scriptstyle \pm 4.55}$ & OOM & $38.41 {\scriptstyle \pm 1.66}$ & $80.78 {\scriptstyle \pm 2.45}$ \\  
\bottomrule
\end{tabular}
}
\end{table*}

\subsection{Experimental Results} 

We present the performance of our CGCN and CGATv2 in Table \ref{cgsO:tab:topological_GCN}. The performance of CSGC, i.e. centrality based Simple Graph Convolutional Networks \citep{wu2019simplifying}, in Appendix \ref{cgsO:app:sgc}. We also incorporated our learnable CGSOs into H2GCN \citep{zhu2020beyond} resulting CH2GCN, that go beyond the message passing scheme and which is designed for heterophilic graphs, we detailed the experiment and the results in Appendix \ref{cgsO:app:heterop_gnn}. The results of CGCN, CGATv2, CSGC and the other baselines on additional datasets can be found in Table \ref{cgsO:tab:additional_CA_GCN} of Appendix \ref{cgsO:node_classification_results_appendix}, and Tables \ref{cgsO:tab:sgc_1} and \ref{cgsO:tab:sgc_2} of Appendix \ref{cgsO:app:sgc}. It has been observed that, across numerous datasets, CGCN and CGATv2 outperform classical GSOs and vanilla GNNs. Moreover, it is noteworthy that the optimal choice of centrality for CGCN varies depending on the specific dataset. To better understand the choice of each centrality, we displayed the learned weights of CGCN  together with some statistics of each dataset in Tables \ref{cgsO:tab:degree_hyper}, \ref{cgsO:tab:kcore_hyper}, \ref{cgsO:tab:pagerank_hyper} and  \ref{cgsO:tab:count_walks_hyper}. Several trends are clear: \textit{i)} For all the centrality metrics, the exponent $e_1$ is usually positive for most of the datasets, which indicates that an additive normalization of the GSO with our centralities in-style of the unnormalized Laplacian often leads to optimal graph representation. However, the exponent values $e_2$ and $e_3$ have different behaviors across centrality metrics, e.g., when using the PageRank centrality, the exponents $e_2$ and $e_3$ are almost null for the graph datasets that are strongly homophilous indicating that an unnormalized sum over neighborhoods is optimal. \textit{ii)} When using the PageRank and Count of walks centrality metrics, we notice that the parameter $a$ is always negative for non-homophilous datasets. This is a very interesting finding indicating that a representation with negatively weighted self-loops is advantageous for non-homophilous datasets (an observation that we have not previously seen in the literature). \textit{iii)} For the datasets where the $k$-core centrality performs well (i.e. Cornell, arxiv-year, Penn94, and deezer-europe), we notice that the parameter $m_3$ is very close to zero, i.e, the regularization by adding an identity matrix to the CGSO turns out to be best-ignored in these settings. These findings suggest that the optimal GSO components vary depending on the graph type, highlighting the need for adaptable CGSO approaches rather than relying solely on classical GSOs.

General intuition on the choice of centrality that we can provide relates to the fact that the node degree is a local centrality metric, while the remaining three centralities we consider correspond to global metrics. Therefore, it is apparent that if the learning task only requires local information a degree-based normalization of the GSO is likely beneficial, while global centrality metrics are appropriate if more global information is required. Beyond this statement it seems to be difficult to provide general guidance on the choice of the global centrality metrics. Therefore, including both local and global centrality-based CGSO in the CGNN  might be optimal to dynamically distinguish the best type of centrality. We present the results of this experiment in Tables \ref{cgsO:tab:comb_centralities_1} and \ref{cgsO:tab:comb_centralities_2}. By combining local and global centralities in the CGNNs, we usually increase their performance.

\begin{table}[t] 
\centering
\caption[Performance of the Local-Global based CGSO]{Classification accuracy ($\pm$ standard deviation) of the models on different benchmark node classification datasets. The higher the accuracy (in \%) the better the model.}
\label{cgsO:tab:comb_centralities_1}
\resizebox{\textwidth}{!}{%
\begin{tabular}{l|llllllll}\toprule
 Model & CiteSeer  & PubMed & arxiv-year  & chamelon & Cornell  & deezer-europe  & squirrel & Wisconsin  \\ 
\hline

CGCN w/ $\mathbf{D}$   
    & $68.35 {\scriptstyle \pm 0.45}$ 
    & $78.70 {\scriptstyle \pm 1.10}$ 
    & $45.39 {\scriptstyle \pm 0.45}$ 
    & $64.17 {\scriptstyle \pm 8.10}$ 
    & $72.43 {\scriptstyle \pm 13.09}$  
    & $58.04 {\scriptstyle \pm 1.06}$ 
    & $42.30 {\scriptstyle \pm 1.34}$ 
    & $76.86 {\scriptstyle \pm 7.70}$ \\

CGCN w/ $\mathbf{V}_{\text{\textit{core}}}$    
    & $68.40 {\scriptstyle \pm 0.75}$ 
    & $77.91 {\scriptstyle \pm 0.41}$ 
    & $47.27 {\scriptstyle \pm 0.31}$ 
    & $63.68 {\scriptstyle \pm 5.00}$ 
    & $73.78 {\scriptstyle \pm 12.16}$        
    & $\mathbf{60.90 {\scriptstyle \pm 2.28}}$ 
    & $40.59 {\scriptstyle \pm 2.21}$ 
    & $74.90 {\scriptstyle \pm 6.52}$  \\

CGCN w/ $\mathbf{V}_{\ell\text{\textit{-walks}}}$  
    & $67.31 {\scriptstyle \pm 0.75}$ 
    & $77.57 {\scriptstyle \pm 0.37}$ 
    & $39.35 {\scriptstyle \pm 0.49}$ 
    & $\mathbf{66.21 {\scriptstyle \pm 2.49}}$ 
    & $72.70 {\scriptstyle \pm 3.24}$ 
    & $59.15 {\scriptstyle \pm 1.24}$ 
    & $36.03 {\scriptstyle \pm 5.81}$ 
    & $74.90 {\scriptstyle \pm 4.19}$ \\

CGCN w/ $\mathbf{V}_{\text{\textit{PR}}}$  
    & $67.11 {\scriptstyle \pm 0.56}$ 
    & $78.17 {\scriptstyle \pm 4.27}$ 
    & $47.14 {\scriptstyle \pm 0.31}$ 
    & $60.94 {\scriptstyle \pm 7.00}$ 
    & $\mathbf{76.22 {\scriptstyle \pm 16.3}}$ 
    & $63.41 {\scriptstyle \pm 0.77}$ 
    & $32.17 {\scriptstyle \pm 3.94}$ 
    & $80.78 {\scriptstyle \pm 11.7}$  \\ 
\midrule 

CGCN w/ $\mathbf{D} - \mathbf{V}_{\text{\textit{core}}}$      
    & $\mathbf{69.0 {\scriptstyle \pm 0.64}}$ 
    & $\mathbf{78.77 {\scriptstyle \pm 0.34}}$  
    & $48.37 {\scriptstyle \pm 0.15}$ 
    & $65.04 {\scriptstyle \pm 4.37}$ 
    & $73.24 {\scriptstyle \pm 6.56}$ 
    & $59.81 {\scriptstyle \pm 0.51}$ 
    & $40.74 {\scriptstyle \pm 4.77}$ 
    & $74.51 {\scriptstyle \pm 3.62}$   \\

CGCN w/ $\mathbf{D} - \mathbf{V}_{\ell\text{\textit{-walks}}}$ 
    & $67.99 {\scriptstyle \pm 0.55}$ 
    & $78.53 {\scriptstyle \pm 0.39}$ 
    & $\mathbf{49.12 {\scriptstyle \pm 0.41}}$ 
    & $58.09 {\scriptstyle \pm 3.78}$ 
    & $74.32 {\scriptstyle \pm 2.77}$ 
    & $59.30 {\scriptstyle \pm 0.70}$ 
    & $34.49 {\scriptstyle \pm 2.66}$ 
    & $\mathbf{81.37 {\scriptstyle \pm 3.64}}$ \\

CGCN w/ $\mathbf{D} - \mathbf{V}_{\text{\textit{PR}}}$   
    & $68.45 {\scriptstyle \pm 0.6}$ 
    & $77.75 {\scriptstyle \pm 0.55}$ 
    & $39.63 {\scriptstyle \pm 1.27}$ 
    & $64.32 {\scriptstyle \pm 3.13}$ 
    & $72.97 {\scriptstyle \pm 4.98}$ 
    & $59.28 {\scriptstyle \pm 0.75}$ 
    & $\mathbf{42.80 {\scriptstyle \pm 6.58}}$ 
    & $74.31 {\scriptstyle \pm 3.97}$ \\  
\bottomrule
\end{tabular}
}

\end{table}

\section{Conclusion}\label{cgsO:sec:conclusion}
In this work, we have proposed CGSOs, a novel class of Graph Shift Operators (GSOs) that can leverage different centrality metrics, such as node degree, PageRank score, core number, and the count of walks of a fixed length. Furthermore, we have modified the message-passing steps of Graph Neural Networks (GNNs) to integrate these CGSOs, giving rise to a novel model class the CGNNs. Experimental results comparing our CGNN models to existing vanilla GNNs show the superior performance of CGNN on many real-world datasets. These experiments furthermore allowed us to analyse the optimal parameters of our CGSO, which led to new and interesting insight such as for example an apparent benefit of negatively weighted self-loops for non-homophilous graphs. 
To further understand the cases where each centrality is beneficial, we conducted additional experiments focused on spectral clustering using two distinct types of synthetic graphs. Through these experiments, we identified instances where CGSOs outperformed conventional GSOs.

\chapter{Adaptive Depth Message Passing GNN} \label{ch:Dynamic}
\lettrine[lines=3]{G}{\small{r}}aph Neural Networks (GNNs) have proven to be highly effective in various graph learning tasks. A key characteristic of GNNs is their use of a fixed number of message-passing steps for all nodes in the graph, regardless of each node's diverse computational needs and characteristics.  Through empirical real-world data analysis, we demonstrate that the optimal number of message-passing layers varies for nodes with different characteristics. This finding is further supported by experiments conducted on synthetic datasets. To address this, we propose \emph{Adaptive Depth Message Passing  GNN (ADMP-GNN)}, a novel framework that dynamically adjusts the number of message passing layers for each node, resulting in improved performance. This approach applies to any model that follows the message passing scheme. We evaluate ADMP-GNN on the node classification task and observe performance improvements over baseline GNN models.

\section{Introduction}
A plethora of structured data comes in the form of graphs \citep{bornholdt2001handbook,cao2020spectral}. This has driven the need to develop neural network models, known as Graph Neural Networks (GNNs), that can effectively process and analyze graph-structured data. GNNs have garnered significant attention for their ability to learn complex node and graph representations, achieving remarkable success in several practical applications \citep{rampavsek2022recipe,corso2022diffdock,pmlr-v202-duval23a,castro-correa-tnnls24}. Many of these models are instances of Message Passing Neural Networks (MPNNs) \citep{xu2019powerful,Kipf:2017tc}. A common characteristic of GNNs is that they typically employ a fixed number of message passing steps for all nodes, determined by the number of layers in the GNN. This static framework raises an intriguing question: \emph{Should the number of message passing steps be adapted individually for each node to better capture their unique characteristics and computational needs?}

Determining the optimal number of message passing layers for each node in a GNN presents a significant challenge due to the intricate and diverse nature of graph structures, node features, and learning tasks. While deeper GNNs can capture long-range dependencies \citep{liu2021eignn}, they can also encounter issues like oversmoothing, where nodes become indistinguishably similar \citep{luan2022revisiting,giraldo2023sjlr}.  In dense graphs, where information can propagate quickly, even shallow GNNs can effectively capture local information \citep{zeng2020deep}. Conversely, sparse graphs, particularly those with isolated nodes or limited connectivity, may require additional layers to facilitate effective information sharing \citep{zhang2021evaluating,Zhao2020PairNorm}. This underscores the importance of selecting the appropriate number of layers for a GNN to capture the necessary graph information effectively. An even more compelling idea is to adjust the GNN depth for each node based on its local structural properties. This adaptive approach could be especially beneficial for graphs with varied local structures, ensuring that each node is processed according to its unique requirements.

This need for per-node customization naturally aligns with the concept of \emph{Dynamic Neural Networks}, also referred to as \emph{Adaptive Neural Networks} \citep{bolukbasi2017adaptive}. Dynamic Neural Networks represent a class of models that can adjust their architecture or parameters depending on the input. Dynamic Neural Networks have gained significant popularity, especially in the field of computer vision. Examples of adaptation include varying the number of layers and implementing skip connections  \citep{li2017pruning,huang2016deep,sabour2017dynamic}. Beyond computer vision, other types of adaptive neural networks have been explored in various domains. In natural language processing, adaptive computation time models allow recurrent neural networks (RNNs) to determine the required number of recurrent steps dynamically based on the complexity of the input sequence \citep{graves2016adaptive}. However, applying these adaptations to graph learning tasks presents unique challenges. Unlike the structured and homogeneous data often encountered in computer vision, graph data involves overcoming nuisances related to the inherent complex structure of graphs. Graphs may have varying node degrees, non-uniform connectivity, and feature Heterogeneity. While some dynamic approaches may extend effectively to graph classification tasks, applying these methods to node classification presents additional complexities. In node classification, each input sample (node) is part of a larger interconnected system where information propagates through edges, i.e., the prediction for one node often depends on the features of the other nodes and the structure of the graph. As a result, specific techniques are needed to account for the dependencies and relational information encoded in the graph structure.

In this work, we focus on the task of node classification by proposing ADMP-GNN, a novel approach that dynamically adapts the number of layers for each node within a GNN. Our main contributions are as follows:
\begin{itemize}
    \item \textbf{Node-specific depth analysis in GNNs.} We demonstrate through empirical analysis that different nodes within the same graph may require varying numbers of message passing steps to accurately predict their labels. This finding underscores the importance of node-specific depth in GNNs.
    \item \textbf{Adaptive message passing layer integration.} We present ADMP-GNN, a novel approach that enables any GNN to make predictions for each node at every layer. Training the GNN to predict labels across all layers is a multi-task setting, which often suffers from gradient conflicts, leading to suboptimal performance. To address this, we propose a sequential training methodology where layers are progressively trained, and their gradients are subsequently frozen, thereby mitigating conflicts and improving overall performance.
    \item \textbf{Adaptive layer policy learning for node classification.} We introduce a heuristic method to learn a layer selection policy using a set of validation nodes. This policy is then applied to select the optimal layer for predicting the labels of test nodes, ensuring that each node exits the GNN at the most appropriate layer for its specific classification task.
    \item \textbf{Model-agnostic flexibility.} Our approach is model-agnostic and can be integrated with any GNN architecture that employs a message passing scheme. This flexibility enhances the GNN's performance on node classification tasks, providing a significant improvement over traditional fixed-layer approaches.
    
\end{itemize}

\section{Related Work}
In this section, we review key developments in GNNs and dynamical neural networks, which form the foundation of our work.

\paragraph{Dynamical Neural Networks.}
Dynamic neural networks are gaining significance in the field of deep learning. Unlike static models with fixed computational graphs and parameters during inference, dynamic networks adapt their structures or parameters based on varying inputs. This dynamic flexibility gives models significant benefits, such as improved accuracy, enhanced computational efficiency, and superior adaptability \citep{DBLP:journals/corr/abs-2010-05300,DBLP:journals/corr/abs-2006-04152}. A popular type of dynamic neural networks includes those that dynamically adjust network depth based on each input. For instance, in natural language processing, some adaptive large language models employ adaptive depth to optimize both inference speed and computational memory usage of the Transformer architecture \citep{Elbayad2020Depth-Adaptive,schuster2022confident,DBLP:journals/corr/VaswaniSPUJGKP17}. In the field of computer vision, there are studies that dynamically generate filters conditioned on each input, enhancing flexibility without significantly increasing the number of model parameters \citep{jia2016dynamic}.

In Graph Machine Learning, dynamic adaptations have been proposed for GNN message passing. These methods include dynamically determining which neighbors to consider at each layer, enabling more flexible and adaptive message passing \citep{finkelshtein2023cooperative}, or allowing nodes to react to individual messages at varying times rather than processing aggregated neighborhood information synchronously \citep{faber2024gwac}. Another line of work focuses on adapting normalization layers for each node to enhance expressive power, generating representations that reflect local neighborhood structures \citep{eliasof2024granola}. Additionally, other related approaches use residual connections to mitigate issues like oversmoothing \citep{errica2023adaptive,DBLP:journals/corr/abs-2007-02133}. To the best of our knowledge, in the field of GNNs, there has been no prior work proposing adaptive depth for each node. However, several studies have focused on combining all GNN layers. These works typically aim to adapt GNN architectures for  heterophilic graphs \citep{chien2020adaptive} and leverage information from higher-order neighbors \citep{DBLP:journals/corr/abs-1806-03536}. While combining GNN layers can be viewed as a form of depth-adaptive strategy, where the final node representation is guided by the optimal intermediate hidden states, this approach remains \emph{static} because the same inference policy is applied uniformly across all nodes and learned layer aggregators stay fixed after training.

\section{Adaptive Depth Message Passing GNN} \label{dynam:method}
\begin{figure}[t]
    \centering
    \includegraphics[width=.7\linewidth]{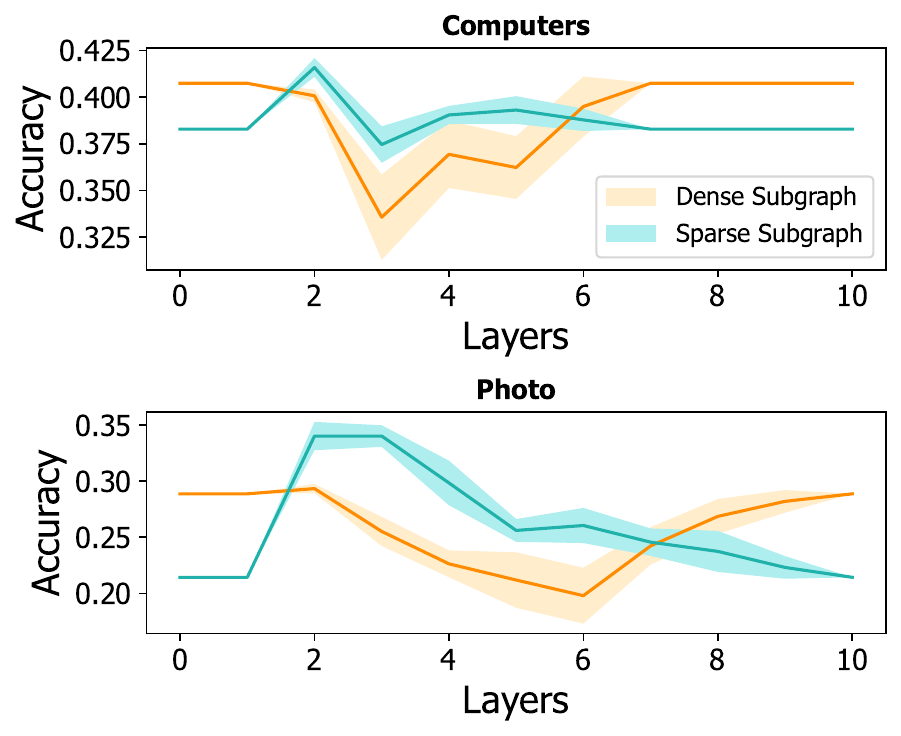}
    \caption[Effect of GCN's depth on sparse and dense subgraphs]{Effect of GCN's depth on sparse and dense subgraphs. The figure shows the performance of GCNs when varying layer depths, and comparing its effectiveness on both sparse and dense subgraphs.}
    \label{dynam:fig:photo_synthetic}
\end{figure}

In this section, we first present an empirical analysis highlighting the necessity for node-specific depth in GNNs. Then, we introduce our ADMP-GNN. Our study includes experiments on both synthetic and real-world datasets to illustrate the importance and potential benefits of this methodology. 

\subsection{Depth Analysis on Synthetic Graphs}
This analysis aims to highlight the importance of employing a varying number of message passing steps based on the specific characteristics of individual nodes. As outlined in the introduction, this approach is particularly relevant for graphs where nodes exhibit diverse properties, such as varying local structures. In this experiment, we investigate the effect of the number of message-passing layers on nodes with varied levels of local neighborhood sparsity. To achieve this, we construct a graph by merging two subgraphs extracted from real-world datasets, namely Computers and Photo \citep{cs_data}. Both subgraphs contain the same number of nodes, exhibit nearly identical homophily, and have equally distributed node labels. The main difference between these subgraphs lies in their structure, as one subgraph is sparse while the other is dense. Consequently, nodes within each subgraph share comparable structural characteristics. Details on the construction of these synthetic datasets and visualizations of their adjacency matrices are provided in Appendix \ref{dynam:app:motivation}, including Fig. \ref{dynam:fig:motivation_hybid}.

We have trained $L+1$ different GCN models, with a varying number of layers $\ell\in \{ 0,\ldots,L\}$, where $L$ represents the maximum depth, set to $L=10$. Each GCN was trained on the entire synthetic graph, composed of both sparse and dense subgraphs, but the performance was evaluated separately on the individual subgraphs. This allows us to assess the impact of GNN depth on different types of local subgraph structures. The results, presented in Fig. \ref{dynam:fig:photo_synthetic}, reveal notable differences in behavior between the subgraph types.
As observed, in dense subgraphs, the test accuracy decreases at a faster rate, while in sparse subgraphs, the drop in accuracy occurs later, typically around layers 2 or 3. Moreover, the optimal number of layers differs between sparse and dense subgraphs. For instance, in the Computers dataset, the highest accuracy is achieved at layer 2 for the sparse subgraph, while for the dense subgraph, the optimal performance is reached at layer 0. Additionally, in the Photo dataset, we observe a distinct behavior starting from layer 6, where the impact of GNN depth diverges between sparse and dense subgraphs. This highlights the need to adapt the number of layers per node based on its characteristics.

\subsection{Adaptive Message Passing Layer Integration}
Building on the insights from the previous analysis, it is essential to develop a framework that can efficiently adjust the GNN depth. For a maximum GNN depth $L$, a traditional approach would involve training $L+1$ different GNNs, each with a distinct number of layers $\ell$, where $\ell$ ranges from 0 and $L$. Subsequently, a policy must be established to determine the optimal GNN with the appropriate number of layers for each individual node. However, training $L+1$ GNNs separately can be computationally expensive. A more efficient approach involves designing a single GNN with $L+1$ layers that provide predictions at each intermediate layer. To ensure equivalence to the previous approach, i.e., training $L+1$ GNNs separately, the computational graph for predictions at layer $\ell$ must match that of a standard GNN with $\ell$ layers, and the classification performance at layer $\ell$ should yield results comparable to those of a conventional GNN with $\ell$ layers. 
\begin{figure*}
    \centering
    \includegraphics[width=\textwidth]{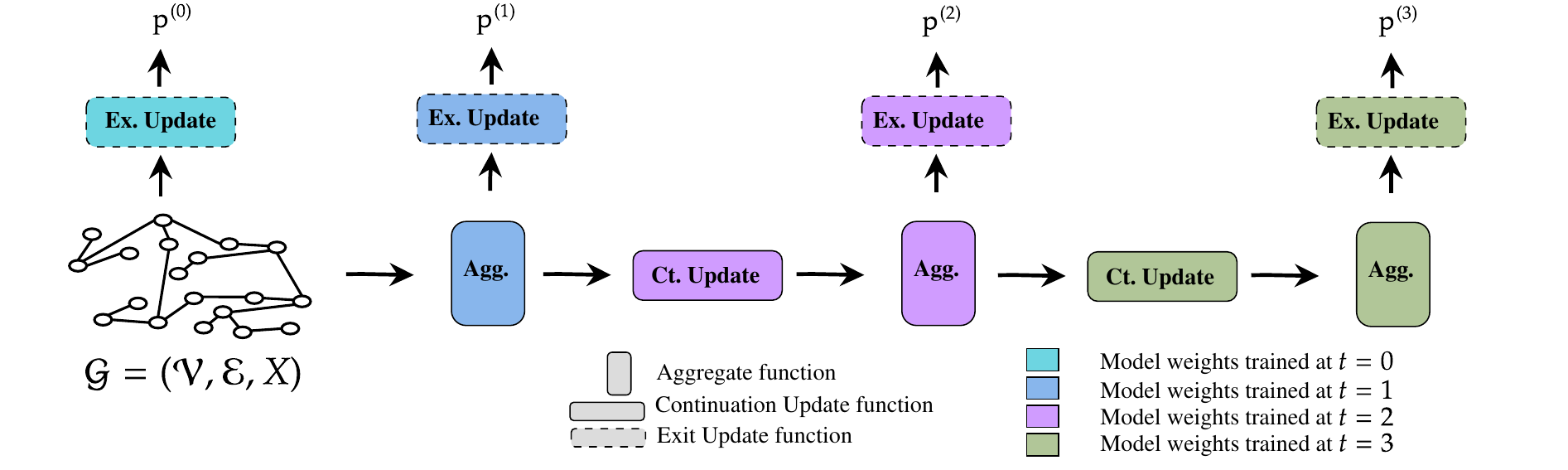}
    \caption[Illustration of ADMP-GNN]{Illustration of ADMP-GNN, when the maximum GNN depth is $L=3$.}
    \label{dynam:fig:AMP_architerchture}
\end{figure*}

To address these challenges, we introduce ADMP-GNN, an adaptation of a Message Passing Neural Network with a maximum depth of $L$ layers.  The goal is to ensure that the computational graph and the performance of ADMP-GNN at a certain layer match that of traditional GNNs when trained and tested on the same number of layers. To achieve this, we incorporate an additional \textit{Update} function, denoted as $\phi^{(\ell)}_{\mathbf{Ex}}$ ($\mathbf{Ex}$ stands for `Exit'), to directly predict node labels at a given layer $\ell$, i.e., $p^{(\ell)}$. The function $\phi^{(\ell)}_{\mathbf{Ex}}$ is defined as follows,
\begin{equation*}
    p^{(\ell)}_v = \phi^{(\ell)}_{\mathbf{Ex}} \left (   h^{(\ell-1)}_v , m^{(\ell)}_v \right ) \\
             = \text{Softmax} \left (   \widetilde{W}^{(\ell)} m^{(\ell)}_v \right ),
\end{equation*}
where $\widetilde{W}^{(\ell)} \in \mathbb{R}^{d^{(\ell)}\times c}$ is a learnable weight matrix, $d^{(\ell)}$ is the dimension of the hidden representation at the $\ell$-th layer, and $c$ is the number of classes. To obtain predictions at a deeper layer $\ell^\prime  \geq \ell $, we continue the message passing using another \textit{Update} function $\phi^{(\ell)}_{\mathbf{Ct}}$ ($\mathbf{Ct}$ stands for `Continuation'),
    \begin{align*}
        p^{(\ell)}_v  &= \phi^{(\ell)}_{\mathbf{Ex}} \left (   h^{(\ell-1)}_v , m^{(\ell)}_v \right ),\\
        h^{(\ell+1)}_v &= \phi^{(\ell)}_{\mathbf{Ct}}\left (  h^{(\ell-1)}_v, m^{(\ell)}_v\right ).
    \end{align*}
For $\ell=0$, we directly use the \textit{Exit Update} function on the node features, i.e.,  $m^{(0)}_v = x_v$,
$$\forall v \in \mathcal{V}, \quad p_v^{(0)} =  \phi^{(0)}_{\mathbf{Ex}} \left ( m^{(0)}_v \right ) \\
              = \text{Softmax} \left (   \widetilde{W}^{(0)} x_v \right ).$$
In Fig. \ref{dynam:fig:AMP_architerchture}, we illustrate the architecture of the proposed ADMP-GNN.
\subsection{Training Scheme of ADMP-GNN}
Our next objective is to train ADMP-GNN to predict node labels across all layers $ \ell \in \{ 0,\ldots , L\}$ simultaneously. For each layer $\ell$, let $\theta_\ell$ denote the weights of the function $\psi ^{(\ell)}\circ\phi_{\mathbf{Ct}}^{(\ell)}(\cdot)$. Two distinct strategies for training ADMP-GNN are explored:

\paragraph{1. Aggregate Loss Minimization (ALM).} This straightforward approach minimizes the aggregate loss over all layers. The total loss is formulated as,
\begin{align}
\mathcal{L}_{\text{ALM}} & := 
    \argmin_{\theta} \mathbb{E}_{v\in \mathcal{V}}\left [ S_L(v) \right ] \label{dynam:eq:multi_obj} \\ 
    & =\argmin_{\theta_0, \ldots, \theta_{L}} \mathbb{E}_{v\in \mathcal{V}}\left [ \sum_{\ell=0}^{L}\mathcal{L}\left ( p_v^{(\ell)}(m^{(0)}_v, \theta_0, \ldots, \theta_{\ell}) , y_v\right ) \right ]
\end{align}
where  $p_v^{(\ell)}(m^{(0)}_v, \theta_0, \ldots, \theta_{\ell}) = \phi^{(\ell)}_{\mathbf{Ex}}(m^{(\ell)}_v)$ is the prediction for the node $v$ at  layer $\ell$, and $\mathcal{L}$ is the Cross Entropy Loss. This approach may encounter gradient conflicts, particularly for early layers involved in both computation and back-propagation across upper layers. 

\paragraph{2. Sequential Training (ST).} We have studied an alternative training setup where we progressively train one GNN layer at a time, subsequently freezing each layer after training. More formally, the problem in \eqref{dynam:eq:multi_obj} can be tackled using dynamic programming as follows,
\begin{align*}
    \forall v\in \mathcal{V}, S_{\ell+1}(v) &= \mathcal{L}\left ( p_v^{(\ell)}(m^{(0)}_v,\theta_0^\star, \ldots, \theta_{\ell}^\star, \theta_{\ell+1}) , y_v\right ) \\
    & ~~~~~~~~~~~~~~~~~~~~~~~~~~~~~~~~~~~~~~~~+  S_\ell (v)\\
    \theta_{\ell}^\star&  = \argmin_{\theta_{\ell}} \mathbb{E}_{v\in V}\left [S_\ell(v)\right ],
\end{align*}
where $\forall v \in \mathcal{V}, ~ S_0(v) = \mathcal{L}(\phi_{\mathbf{Ex}}^{(0)}, y_v)$. For each intermediate layer $\ell<L$, by training this layer on the node classification task, we obtain high-quality node representations $\{h^{(\ell)}_v:v \in \mathcal{V}\}$. These representations are directly employed for predictions and serve as a robust foundation for the label predictions of the subsequent layer $\ell+1$. Algorithm \ref{dynam:algo:seq_training} offers a summary of the approach. 
 
\smallskip
\noindent  \textbf{Comparison of ADMP-GNN Training Paradigms.} To identify the optimal multi-task training configuration, we evaluate how much of a performance drop we lose at each layer compared to the single-task setting in GNNs. The comparative analysis of the three strategies for both GCN and GIN is detailed in Tables \ref{dynam:tab:multi_task_updated}, \ref{dynam:tab:multi_task_2}, \ref{dynam:tab:multi_task_GIN}, and \ref{dynam:tab:multi_task_GIN_2}. Our findings indicate that ADMP-GNN ST outperforms ADMP-GNN ALM. Notably, the performance of ADMP-GNN ST is comparable to, or even exceeds, that of GNN when trained under the single-task setting. Furthermore, ADMP-GNN ST exhibits a smaller standard deviation, suggesting more consistent performance. These results underscore the effectiveness of the ST setting in addressing gradient conflicts inherent in aggregate loss minimization (ALM). Furthermore, ADMP-GNN ST effectively mimics the results of training $L+1$ separate GNNs, each with a different number of layers, while requiring only a single unified model. The next step, outlined in Section \ref{dynam:subsec:policy}, is to learn a policy that selects the optimal prediction layer for each node, completing the framework of ADMP-GNN.

\begin{table}[t]
\caption[Comparison of ADMP-GCN training paradigms ALM and ST]{Comparison of ADMP-GCN training paradigms ALM and ST. These paradigms are also compared to the single-task training setting to evaluate which approach most closely mimics the classical GCN under single-task training.  The best results for each dataset are \textbf{bolded}.}
\resizebox{\textwidth}{!}{%
\begin{tabular}{c|clcccccc}
\toprule
\# Layers & Training Paradigm & Model & Cora & CiteSeer & CS & PubMed & Genius & Ogbn-arxiv \\
\midrule
\multirow{3}{*}{\circled{0}} & \multirow{1}{*}{Single-task} & GCN & $56.38 {\scriptstyle \pm 0.04}$ & $57.18 {\scriptstyle \pm 0.12}$ & $88.04 {\scriptstyle \pm 0.49}$ & $72.50 {\scriptstyle \pm 0.09}$ & $80.82 {\scriptstyle \pm 1.00}$ & $48.88 {\scriptstyle \pm 0.06}$ \\
\cmidrule(lr){2-9}
& \multirow{2}{*}{Multi-task} & ADMP-GCN (ALM) & $\mathbf{56.96 {\scriptstyle \pm 0.20}}$ & $\mathbf{58.44 {\scriptstyle \pm 0.21}}$ & $87.06 {\scriptstyle \pm 1.06}$ & $72.11 {\scriptstyle \pm 0.18}$ & $80.03 {\scriptstyle \pm 0.37}$ & $36.50 {\scriptstyle \pm 0.12}$ \\
& & ADMP-GCN (ST) & $56.38 {\scriptstyle \pm 0.06}$ & $57.17 {\scriptstyle \pm 0.09}$ & $\mathbf{87.27 {\scriptstyle \pm 1.29}}$ & $\mathbf{72.48 {\scriptstyle \pm 0.14}}$ & $\mathbf{80.17 {\scriptstyle \pm 0.79}}$ & $\mathbf{48.86 {\scriptstyle \pm 0.03}}$ \\
\midrule
\multirow{3}{*}{\circled{1}} & \multirow{1}{*}{Single-task} & GCN & $76.90 {\scriptstyle \pm 0.14}$ & $69.68 {\scriptstyle \pm 0.06}$ & $91.74 {\scriptstyle \pm 0.80}$ & $76.63 {\scriptstyle \pm 0.13}$ & $80.23 {\scriptstyle \pm 0.37}$ & $55.21 {\scriptstyle \pm 0.50}$ \\
\cmidrule(lr){2-9}
& \multirow{2}{*}{Multi-task} & ADMP-GCN (ALM) & $75.67 {\scriptstyle \pm 0.18}$ & $\mathbf{70.12 {\scriptstyle \pm 0.04}}$ & $90.55 {\scriptstyle \pm 0.74}$ & $73.74 {\scriptstyle \pm 0.16}$ & $\mathbf{80.13 {\scriptstyle \pm 0.29}}$ & $39.54 {\scriptstyle \pm 1.44}$ \\
& & ADMP-GCN (ST) & $\mathbf{76.90 {\scriptstyle \pm 0.00}}$ & $69.70 {\scriptstyle \pm 0.00}$ & $\mathbf{90.89 {\scriptstyle \pm 0.81}}$ & $\mathbf{76.60 {\scriptstyle \pm 0.00}}$ & $79.93 {\scriptstyle \pm 0.00}$ & $\mathbf{55.15 {\scriptstyle \pm 0.00}}$ \\
\midrule
\multirow{3}{*}{\circled{2}} & \multirow{1}{*}{Single-task} & GCN & $81.06 {\scriptstyle \pm 0.50}$ & $71.05 {\scriptstyle \pm 0.48}$ & $91.67 {\scriptstyle \pm 0.94}$ & $79.46 {\scriptstyle \pm 0.31}$ & $79.88 {\scriptstyle \pm 0.51}$ & $66.92 {\scriptstyle \pm 0.67}$ \\
\cmidrule(lr){2-9}
& \multirow{2}{*}{Multi-task} & ADMP-GCN (ALM) & $70.82 {\scriptstyle \pm 4.05}$ & $60.95 {\scriptstyle \pm 2.39}$ & $31.20 {\scriptstyle \pm 12.85}$ & $75.10 {\scriptstyle \pm 2.41}$ & $79.64 {\scriptstyle \pm 0.60}$ & $55.25 {\scriptstyle \pm 0.92}$ \\
& & ADMP-GCN (ST) & $\mathbf{80.73 {\scriptstyle \pm 0.33}}$ & $\mathbf{71.33 {\scriptstyle \pm 0.40}}$ & $\mathbf{91.49 {\scriptstyle \pm 0.66}}$ & $\mathbf{79.02 {\scriptstyle \pm 0.21}}$ & $\mathbf{80.06 {\scriptstyle \pm 0.11}}$ & $\mathbf{66.51 {\scriptstyle \pm 0.65}}$ \\
\midrule
\multirow{3}{*}{\circled{3}} & \multirow{1}{*}{Single-task} & GCN & $79.14 {\scriptstyle \pm 1.58}$ & $66.33 {\scriptstyle \pm 1.35}$ & $89.80 {\scriptstyle \pm 0.87}$ & $78.50 {\scriptstyle \pm 0.68}$ & $80.00 {\scriptstyle \pm 0.04}$ & $67.33 {\scriptstyle \pm 0.55}$ \\
\cmidrule(lr){2-9}
& \multirow{2}{*}{Multi-task} & ADMP-GCN (ALM) & $68.64 {\scriptstyle \pm 5.14}$ & $51.30 {\scriptstyle \pm 6.74}$ & $47.82 {\scriptstyle \pm 12.98}$ & $74.17 {\scriptstyle \pm 1.98}$ & $\mathbf{80.04 {\scriptstyle \pm 0.10}}$ & $56.22 {\scriptstyle \pm 0.50}$ \\
& & ADMP-GCN (ST) & $\mathbf{80.21 {\scriptstyle \pm 0.52}}$ & $\mathbf{70.08 {\scriptstyle \pm 0.90}}$ & $\mathbf{89.83 {\scriptstyle \pm 1.02}}$ & $\mathbf{78.25 {\scriptstyle \pm 0.51}}$ & $79.93 {\scriptstyle \pm 0.00}$ & $\mathbf{68.31 {\scriptstyle \pm 0.48}}$ \\
\midrule
\multirow{3}{*}{\circled{4}} & \multirow{1}{*}{Single-task} & GCN & $75.96 {\scriptstyle \pm 1.93}$ & $60.33 {\scriptstyle \pm 2.38}$ & $78.90 {\scriptstyle \pm 22.33}$ & $76.59 {\scriptstyle \pm 0.98}$ & $80.01 {\scriptstyle \pm 0.04}$ & $65.49 {\scriptstyle \pm 0.99}$ \\
\cmidrule(lr){2-9}
& \multirow{2}{*}{Multi-task} & ADMP-GCN (ALM) & $67.88 {\scriptstyle \pm 6.05}$ & $49.29 {\scriptstyle \pm 7.87}$ & $52.76 {\scriptstyle \pm 11.79}$ & $73.40 {\scriptstyle \pm 1.95}$ & $\mathbf{80.04 {\scriptstyle \pm 0.10}}$ & $56.43 {\scriptstyle \pm 0.31}$ \\
& & ADMP-GCN (ST) & $\mathbf{81.05 {\scriptstyle \pm 0.49}}$ & $67.99 {\scriptstyle \pm 0.74}$ & $\mathbf{88.86 {\scriptstyle \pm 0.70}}$ & $\mathbf{75.27 {\scriptstyle \pm 1.02}}$ & $79.93 {\scriptstyle \pm 0.00}$ & $\mathbf{69.29 {\scriptstyle \pm 0.74}}$ \\
\midrule
\multirow{3}{*}{\circled{5}} & \multirow{1}{*}{Single-task} & GCN & $70.09 {\scriptstyle \pm 4.01}$ & $57.40 {\scriptstyle \pm 3.43}$ & $77.96 {\scriptstyle \pm 15.24}$ & $74.32 {\scriptstyle \pm 3.66}$ & $80.01 {\scriptstyle \pm 0.04}$ & $63.04 {\scriptstyle \pm 1.33}$ \\
\cmidrule(lr){2-9}
& \multirow{2}{*}{Multi-task} & ADMP-GCN (ALM) & $67.68 {\scriptstyle \pm 7.04}$ & $50.14 {\scriptstyle \pm 7.65}$ & $54.53 {\scriptstyle \pm 10.61}$ & $73.27 {\scriptstyle \pm 1.51}$ & $80.04 {\scriptstyle \pm 0.10}$ & $56.66 {\scriptstyle \pm 0.31}$ \\
& & ADMP-GCN (ST) & $\mathbf{81.22 {\scriptstyle \pm 0.36}}$ & $\mathbf{67.42 {\scriptstyle \pm 0.75}}$ & $\mathbf{86.65 {\scriptstyle \pm 1.52}}$ & $\mathbf{75.08 {\scriptstyle \pm 0.94}}$ & $\mathbf{79.93 {\scriptstyle \pm 0.00}}$ & $\mathbf{68.92 {\scriptstyle \pm 1.00}}$ \\
\bottomrule
\end{tabular}}
\label{dynam:tab:multi_task_updated}
\end{table}

\paragraph{Time Complexity.} The training setup ST, where we sequentially train the deep ADMP-GNN, incurs relatively higher time costs due to the need for $L+1$ training iterations. However, in each iteration, backpropagation is performed on a limited number of parameters, approximately equivalent to those in a single message passing layer. Consequently, only a small number of epochs are required for each training iteration. We report the training time of each approach in Table \ref{dynam:tab:time_complexity} in Appendix \ref{dynam:app:time_comp}.

\subsection{Empirical Insights into Node Specific Depth}
We define \textit{Oracle Accuracy} as the maximum achievable test accuracy under an optimal policy for selecting layers. This is formally expressed as,
$$\mathcal{A}_{\text{oracle}} = \frac{1}{|\mathcal{V}_{\text{test}}|} \sum_{v \in \mathcal{V}_{\text{test}}} \mathbf{1}\left( y_v \in  \{\widehat{y}_v^{(\ell)}: ~~\ell = 0, \ldots , L\}  \right),$$
where $\mathcal{V}_{\text{test}}$ is the set of test nodes, $\{\widehat{y}_v^{(\ell)}: \ell = 0, \ldots , L\}$ represents the predictions for node $v$ at each layer, and $\mathbf{1}(\cdot)$ is the indicator function. Specifically, $\mathcal{A}_{\text{oracle}}$ corresponds to the accuracy obtained if, for each test node, we could perfectly choose the layer that provides the correct prediction. Assuming such an optimal selection for each node, it represents the upper bound of accuracy achievable when employing adaptive strategies to determine the best layer for each node.

Based on the findings presented in Table \ref{dynam:tab:oracle} (and Table \ref{dynam:tab:oracle_2} in Appendix \ref{dynam:app:GCN}) for GCN, as well as Table \ref{dynam:tab:oracle_GIN} (and Table \ref{dynam:tab:oracle_GIN_2} in Appendix \ref{dynam:app:GIN}), it is clear that $\mathcal{A}_{\text{oracle}}$ outperforms the highest accuracies achieved by both GCN and ADMP-GCN ALM. This empirical evidence strongly indicates that adopting a distinct exit layer for each node is highly beneficial in an optimal configuration.

\begin{table}[ht]
\centering
\caption[Comparison of highest accuracy for GCN and ADMP-GCN ST]{Comparison of highest accuracy ($\pm$ standard deviation) for GCN and ADMP-GCN ST, with the layer achieving the best accuracy indicated in brackets. The final row shows the \textit{Oracle Accuracy} for ADMP-GCN ST. Best results for each dataset are \textbf{bolded}.}
\resizebox{0.95\textwidth}{!}{%
\begin{tabular}{lllllll}
\toprule
 Model & Cora & CiteSeer & CS & PubMed & Genius & Ogbn-arxiv \\  \midrule
 GCN & $81.06 {\scriptstyle \pm 0.50} ~[2]$  &  $71.05 {\scriptstyle \pm 0.48} ~[2]$ & $91.67 {\scriptstyle \pm 0.94} ~[2]$  &  $79.46 {\scriptstyle \pm 0.31} ~[2]$ & $80.82 {\scriptstyle \pm 1.00} ~[0]$ &  $67.33 {\scriptstyle \pm 0.55} ~[3]$ \\

ADMP-GCN (ST) & $81.22 {\scriptstyle \pm 0.36} ~[5]$ & $71.33 {\scriptstyle \pm 0.40} ~[2]$ & $91.49 {\scriptstyle \pm 0.66} ~[2]$ & $79.02 {\scriptstyle \pm 0.21} ~[2]$ & $80.17 {\scriptstyle \pm 0.79} ~[0]$ & $69.29 {\scriptstyle \pm 0.74} ~[4]$ \\
ADMP-GCN (ST) - Oracle & $\mathbf{89.43 {\scriptstyle \pm 0.19}}$ & $\mathbf{81.96 {\scriptstyle \pm 0.49}}$ & $\mathbf{97.24 {\scriptstyle \pm 0.52}}$ & $\mathbf{90.13 {\scriptstyle \pm 0.36}}$ & $\mathbf{85.97 {\scriptstyle \pm 7.59}}$ & $\mathbf{79.64 {\scriptstyle \pm 0.27}}$ \\ \bottomrule

\end{tabular}
}

\label{dynam:tab:oracle}
\end{table}

\subsection{Generalization to Test Nodes}\label{dynam:subsec:policy}
In this section, we introduce a heuristic approach to predict the optimal exit layer for test nodes based on the assumption that nodes exhibiting \textit{structural similarity} should share the same exit layer. The notion of structural similarity can be assessed using various metrics. In our work, we define the structural similarity based on node centrality metrics, i.e., nodes are considered structurally similar if they exhibit closely aligned centrality values within the graph. We measured node centralities using both local metrics like degree and global metrics such as $k$-core \citep{kcore-vldbj20}, PageRank scores \citep{brin1998anatomy}, and Walk Count, which are detailed below.

\paragraph{$\boldsymbol{k}$-core.} The $k$-core decomposition of a graph is a method that involves systematically pruning nodes. In each iteration, a subgraph $\mathcal{G}_k \subset \mathcal{G}$ is constructed by removing all vertices whose degree is less than $k$. The core number of a vertex $u$, denoted as $\text{core}(u)$, represents the largest value of $k$ for which $i$ is included in the $k$-core. Formally, this is expressed as $\text{core}(i) = \max \{ k : i \in \mathcal{G}_k \}$. Nodes with higher core numbers are typically found in denser subgraphs, reflecting strong interconnections with their neighbors and indicating a higher level of centrality.

\paragraph{PageRank.} A node’s PageRank score represents the probability of a random walk visiting that node, making it a fundamental metric for measuring node importance in social network analysis and the web \citep{brin1998anatomy}.

\paragraph{Walk Count.} The Walk Count centrality quantifies the importance of a node based on the number of walks of a given length $\ell$ starting from that node. This measure captures higher-order connectivity patterns within the graph, going beyond direct neighbors to account for paths that traverse intermediate nodes. In this work, we consider the Walk Count centrality for $\ell=2$, as it effectively captures higher-order structures and provides valuable insights into the organization of complex systems \citep{benson2016higher}.

 \smallskip
Using this assumption, we employ node clustering to partition the node set into $C$ clusters, denoted by $\mathcal{P} = \cup_{1\leq c \leq C}  \mathcal{P}_c $. Each cluster $c\in \{1,\ldots,C\}$ is assigned a common exit layer $\ell_c\in \{0,\ldots, L\}$, determined using nodes excluded from the test set. Validation nodes are utilized for this purpose. (i) Centrality scores are computed for all nodes, considering metrics. (ii) Nodes are ranked based on their centrality scores. (iii)  These centrality scores are discretized into $C$ equal-sized buckets to facilitate the clustering process. (iv) The optimal exit layer for each cluster is determined by evaluating the classification accuracy on validation nodes within that cluster, i.e.,
$$ \forall c\in \{1,\ldots,C\}, \quad \ell_c = \argmax_{\ell \in \{0,\ldots, L\}} \text{Acc} \left \{ p^{(\ell)}_v \in \mathcal{V}_{val} \cap \mathcal{P}_c\right \}.$$

For the cluster-based layer selection policy, we considered alternative deep learning approaches to predict the optimal exit layer for each node. This high accuracy leads to a significant distribution shift between the predictions for train nodes and those for test nodes. Conversely, validation nodes, which were not utilized during the training of the ADMP-GNN, present a more suitable option for learning the policy due to their unbiased predictions.  For the choice of the cluster-based layer selection policy, we could use a variety of deep learning mechanisms to predict the best exit layer for each node. This included generalizing the policy by training neural networks to identify layers that accurately predict node outcomes or by framing the problem as an optimal stopping problem following the framework of \citep{hure2021deep}. However, these approaches require learning the policy on a set of nodes larger than the test set, which is impractical for node classification tasks where the training and validation node sets available for policy learning are typically smaller than the test set.

\section{Experimental Results}
\begin{table}[t]
\centering
\caption[Classification accuracy for the baselines based on the GCN backbone]{Classification accuracy ($\pm$ standard deviation) on different node classification datasets for the baselines based on the GCN backbone. The higher the accuracy (in \%), the better the model. Highlighted are the \textbf{first}, \underline{second} best results. OOM means \textit{Out of memory}.}
\resizebox{\textwidth}{!}{%
\begin{tabular}{llllllll}
\toprule
Model & Cora & CiteSeer & CS & PubMed & Genuis &  Ogbn-arxiv\\ \midrule

 JKNET-CAT  & $79.52 {\scriptstyle \pm 1.16} ~[2]$ & $69.69 {\scriptstyle \pm 0.05~[1]} $ & $91.23 {\scriptstyle \pm 1.26~[1]} $ & $77.63 {\scriptstyle \pm 0.59} ~[2]$ & $\mathbf{81.46 {\scriptstyle \pm 0.10} ~[2]}$ & $68.54 {\scriptstyle \pm 0.57} ~[5]$ \\

JKNET-MAX  & $75.67 {\scriptstyle \pm 0.18} ~[1]$ & $70.12 {\scriptstyle \pm 0.04} ~[1]$ & $90.55 {\scriptstyle \pm 0.74} ~[1]$ & $75.10 {\scriptstyle \pm 2.41} ~[2]$ & $80.13 {\scriptstyle \pm 0.29} ~[1]$ & $56.66 {\scriptstyle \pm 0.31} ~[5]$ \\

JKNET-LSTM  & $78.95 {\scriptstyle \pm 0.62} ~[0]$ & $65.83 {\scriptstyle \pm 1.27} ~[0]$ & $90.17 {\scriptstyle \pm 1.41} ~[2]$ & $77.73 {\scriptstyle \pm 0.67} ~[0]$ & OOM & OOM \\

Residuals - GCNII  & $76.84 {\scriptstyle \pm 0.20} ~[1]$ & $69.72 {\scriptstyle \pm 0.06} ~[1]$ & $90.84 {\scriptstyle \pm 1.35} ~[1]$ & $77.82 {\scriptstyle \pm 0.44} ~[2]$ & \underline{$81.36 {\scriptstyle \pm 1.13} ~[2]$} & $61.48 {\scriptstyle \pm 3.10} ~[4]$ \\

AdaGCN  & $75.08 {\scriptstyle \pm 0.27}~[1]$  & $69.58 {\scriptstyle \pm 0.19}~[1]$  & $89.62 {\scriptstyle \pm 0.51}~[0] $ & $76.40 {\scriptstyle \pm 0.12}~[5]$ & $79.85 {\scriptstyle \pm 0.00}~[0]$ & $22.06 {\scriptstyle \pm 1.67}~[5]$ \\

GPR-GNN  & $79.91 {\scriptstyle \pm 0.43} ~[2]$ & $69.21 {\scriptstyle \pm 0.81} ~[2]$ & $91.42 {\scriptstyle \pm 1.12} ~[2]$ & $79.00 {\scriptstyle \pm 0.39} ~[2]$ & $81.04 {\scriptstyle \pm 0.41} ~[2]$ & $68.03 {\scriptstyle \pm 0.23} ~[3]$ \\ \midrule

GCN & $80.78 {\scriptstyle \pm 0.72} ~[2]$ & $71.25 {\scriptstyle \pm 0.72} ~[2]$ & $\mathbf{92.20 {\scriptstyle \pm 0.00} ~[1]}$ & $\mathbf{79.32 {\scriptstyle \pm 0.41} ~[2]}$ & $80.76 {\scriptstyle \pm 1.05} ~[0]$ & $64.37 {\scriptstyle \pm 0.43} ~[2]$ \\
 ADMP-GCN & \underline{$81.22 {\scriptstyle \pm 0.36} ~[5]$} & $\mathbf{71.33 {\scriptstyle \pm 0.40} ~[2]}$ & \underline{$91.49 {\scriptstyle \pm 0.66} ~[2]$} & \underline{$79.02 {\scriptstyle \pm 0.21} ~[2]$} & $80.17 {\scriptstyle \pm 0.79} ~[0]$ & $69.29 {\scriptstyle \pm 0.74} ~[4]$ \\ \midrule

ADMP-GCN w/ \textit{Degree} & $81.03 {\scriptstyle \pm 0.53}$ & $71.10 {\scriptstyle \pm 0.50}$ & $91.26 {\scriptstyle \pm 0.59}$ & $78.71 {\scriptstyle \pm 0.39}$ & $80.73 {\scriptstyle \pm 1.00}$ & \underline{$69.59 {\scriptstyle \pm 0.28}$} \\

ADMP-GCN w/ \textit{$k$-core} & $\mathbf{81.19 {\scriptstyle \pm 0.40}}$ & \underline{$71.27 {\scriptstyle \pm 0.53}$} & $91.29 {\scriptstyle \pm 0.68}$ & $78.73 {\scriptstyle \pm 0.41}$ & $80.73 {\scriptstyle \pm 1.00}$ & $69.55 {\scriptstyle \pm 0.33}$ \\

ADMP-GCN w/ \textit{Walk Count} & $81.14 {\scriptstyle \pm 0.41}$ & $71.14 {\scriptstyle \pm 0.50}$ & $91.19 {\scriptstyle \pm 0.64}$ & $78.64 {\scriptstyle \pm 0.60}$ & $80.68 {\scriptstyle \pm 0.97}$ & $69.55 {\scriptstyle \pm 0.34}$ \\

ADMP-GCN w/ \textit{PageRank} & $81.05 {\scriptstyle \pm 0.46}$ & $71.00 {\scriptstyle \pm 0.29}$ & $91.09 {\scriptstyle \pm 0.99}$ & $78.69 {\scriptstyle \pm 0.56}$ & $81.12 {\scriptstyle \pm 1.41}$ & $\mathbf{69.60 {\scriptstyle \pm 0.29}}$ \\
 \bottomrule
\end{tabular}
}

\label{dynam:tab:results_with_baselines}
\end{table}

\begin{table}[h]
\centering
\caption[Classification accuracy for the baselines based on the GIN backbone]{Classification accuracy ($\pm$ standard deviation) on different node classification datasets for the baselines based on the GIN backbone. The higher the accuracy (in \%), the better the model. Highlighted are the \textbf{first}, \underline{second} best results. OOM means \textit{Out of memory}.}
\resizebox{\textwidth}{!}{%
\begin{tabular}{llllllll}
\toprule
Model & Cora & CiteSeer & CS & PubMed & Genuis &  Ogbn-arxiv\\ \midrule

 JKNET-CAT  & $77.94 {\scriptstyle \pm 0.67} ~[2]$ & $64.82 {\scriptstyle \pm 0.04} ~[1]$ & $89.26 {\scriptstyle \pm 1.18} ~[1]$ & $75.89 {\scriptstyle \pm 2.50} ~[2]$ & OOM & $60.23 {\scriptstyle \pm 0.37} ~[1]$ \\

  JKNET-MAX  & $77.47 {\scriptstyle \pm 0.81} ~[1]$ & $64.96 {\scriptstyle \pm 2.08} ~[2]$ & $87.40 {\scriptstyle \pm 2.11} ~[0]$ & $76.21 {\scriptstyle \pm 1.73} ~[0]$ & OOM & $51.84 {\scriptstyle \pm 4.01} ~[0]$ \\

JKNET-LSTM & $77.39 {\scriptstyle \pm 1.39} ~[2]$ & $64.52 {\scriptstyle \pm 2.02} ~[1]$ & $87.27 {\scriptstyle \pm 3.66} ~[3]$ & $75.90 {\scriptstyle \pm 1.63} ~[0]$ & OOM & OOM \\

GPR-GIN & $76.83 {\scriptstyle \pm 1.22} ~[2]$ & \underline{$66.43 {\scriptstyle \pm 1.15} ~[2]$} & $88.15 {\scriptstyle \pm 1.53} ~[5]$ & $\mathbf{77.27 {\scriptstyle \pm 0.87} ~[2]}$ & $80.82 {\scriptstyle \pm 0.42} ~[3]$ & $\mathbf{63.05 {\scriptstyle \pm 0.44} ~[2]}$ \\  \midrule

GIN & $77.73 {\scriptstyle \pm 0.99} ~[2]$ & $65.23 {\scriptstyle \pm 1.45} ~[2]$ & $90.29 {\scriptstyle \pm 0.99} ~[1]$ & $76.05 {\scriptstyle \pm 1.14} ~[2]$ & $80.78 {\scriptstyle \pm 1.03} ~[0]$ & $60.70 {\scriptstyle \pm 0.15} ~[1]$ \\

 ADMP-GIN & $78.07 {\scriptstyle \pm 0.68} ~[2]$ & $65.41 {\scriptstyle \pm 1.91} ~[2]$ & $90.82 {\scriptstyle \pm 1.15} ~[1]$ & $76.46 {\scriptstyle \pm 1.04} ~[4]$ & $80.47 {\scriptstyle \pm 0.91} ~[0]$ & $60.85 {\scriptstyle \pm 0.01} ~[1]$ \\ \midrule

 ADMP-GIN w/ \textit{Degree} & \underline{$78.12 {\scriptstyle \pm 0.70}$} & $\mathbf{66.82 {\scriptstyle \pm 0.89}}$ & $90.70 {\scriptstyle \pm 0.79}$ & $76.07 {\scriptstyle \pm 1.27}$ & \underline{$81.68 {\scriptstyle \pm 0.70}$} & $60.85 {\scriptstyle \pm 0.01}$ \\

ADMP-GIN w/ \textit{$k$-core} & $78.10 {\scriptstyle \pm 0.64}$ & $66.36 {\scriptstyle \pm 1.03}$ & $\mathbf{90.85 {\scriptstyle \pm 0.93}}$ & $75.93 {\scriptstyle \pm 1.13}$ & $81.48 {\scriptstyle \pm 0.64}$ & \underline{$60.85 {\scriptstyle \pm 0.01}$} \\

ADMP-GIN w/ \textit{Walk Count} & $\mathbf{78.19 {\scriptstyle \pm 0.68}}$ & $65.79 {\scriptstyle \pm 0.81}$ & \underline{$90.77 {\scriptstyle \pm 0.91}$} & \underline{$76.72 {\scriptstyle \pm 0.76}$} & $81.21 {\scriptstyle \pm 0.58}$ & $60.85 {\scriptstyle \pm 0.01}$ \\

ADMP-GIN w/ \textit{PageRank} & $77.78 {\scriptstyle \pm 0.83}$ & $67.08 {\scriptstyle \pm 0.51}$ & $90.72 {\scriptstyle \pm 0.91}$ & $76.63 {\scriptstyle \pm 0.87}$ & $\mathbf{81.89 {\scriptstyle \pm 1.04}}$ & $60.85 {\scriptstyle \pm 0.01}$ \\

 \bottomrule

\end{tabular}
}
\label{dynam:tab:results_with_baselines_GIN}
\end{table}

\subsection{Experimental Setup}
\paragraph{Datasets.} We use thirteen widely used datasets in the GNN literature. We particularly used the citation networks Cora, CiteSeer, and PubMed \citep{dataset_node_classification}, the co-authorship networks CS \citep{cs_data}, the citation network between Computer Science arXiv papers Ogbn-arxiv  \citep{hu2020open}, the Amazon Computers and Amazon Photo networks \citep{cs_data}, the non-homophilous dataset genius \citep{lim2021expertise}, and the disassortative datasets Chameleon, Squirrel \citep{rozemberczki2021multi}, 
 and Cornell, Texas, Wisconsin from the WebKB dataset \citep{lim2021large}. More details and statistics about the used datasets can be found in Appendix \ref{dynam:app:data_stats}. For the Cora, CiteSeer, and Pubmed datasets, we used the provided train/validation/test splits. For the remaining datasets, we followed the framework in \citep{lim2021large,rozemberczki2021multi}.
\begin{algorithm}[h]
\small
\caption{Sequential Training for ADMP-GNN}
\label{dynam:algo:seq_training}

\KwIn{Graph $\mathcal{G}=(\mathcal{V},\mathcal{E},X)$, number of layers $L$, node classification loss function $\mathcal{L}$}

\ForEach{$t \in \{0,\ldots,L\}$}{
    \eIf{$t=0$}{
        Set $ h^{(0)}_v \gets m^{(0)}_v = x_v$ for all $ v \in \mathcal{V}.$\;
        
        Compute predictions at layer $\ell=0$, i.e., 
        $\forall v \in \mathcal{V}, \quad p_v^{(0)} =  \phi^{(0)}_{\mathbf{Ex}} ( h^{(0)}_v  ).$\;
        
        Train the weights of $\phi^{(0)}_{\mathbf{Ex}}$ to minimize $\mathcal{L}(p_v^{(0)}).$\;
        
        Freeze the gradients of $\phi^{(0)}_{\mathbf{Ex}}.$
    }{
        Use $\phi^{(t-1)}_{\mathbf{Ct}}$ to update node representations:
        $\forall v \in \mathcal{V}, \quad \tilde{h}^{(t-1)}_v =  \phi^{(t-1)}_{\mathbf{Ct}}(h^{(t-1)}_v). $\;
        
        Aggregate the information for neighbor nodes:
        $\forall v \in \mathcal{V}, \quad m^{(t)}_v = \psi^{(t)}(\{ \tilde{h}^{(t-1)}_u: u \in \mathcal{N}(v)\}). $\;
        
        Compute predictions at layer $\ell=t$, i.e., 
        $\forall v \in \mathcal{V}, \quad p_v^{(t)} = \phi^{(t-1)}_{\mathbf{Ct}}(  \tilde{h}^{(t-1)}_v, m^{(t)}_v).$\;
        
        Train $\phi^{(t-1)}_{\mathbf{Ct}}, \psi^{(t)}, \text{ and } \phi^{(t)}_{\mathbf{Ex}}$ to minimize $\mathcal{L}(p_v^{(t)}).$\;
        
        Freeze the gradients of $\phi^{(t-1)}_{\mathbf{Ct}}, \psi^{(t)}, \text{ and } \phi^{(t)}_{\mathbf{Ex}}.$ 
    }
}
\end{algorithm}

\paragraph{Baselines.} We compare our approach with architectures that combine all the hidden representations of nodes to form a final node representation used for prediction. For each baseline model, we vary the number of layers from 0 to 5, and we report in Table \ref{dynam:tab:results_with_baselines} the performance of the best number of layers with respect to the test set. (i) This includes \textit{Jumping knowledge}, which combines the nodes representation of all layers using an aggregation layer, e.g., MaxPooling (JKMaxPool), Concatenation (JK-Concat), or LSTM-attention (JK-LSTM) \citep{pmlr-v80-xu18c}. (ii) Residuals-GCNII uses an initial residual connection and an identity mapping for each layer.  The initial residual connection ensures that the final representation of each node retains at least a fraction of $\alpha$ from the input layer \citep{DBLP:journals/corr/abs-2007-02133}. (iii)  GPR-GCN combines adaptive generalized PageRank (GPR) scheme with GNNs \citep{chien2020adaptive}. 
(iv)  Ada-GCN, which proposes an RNN-like deep GNN architecture by incorporating AdaBoost to combine the layers \citep{sun2019adagcn}.  In contrast to the baselines, which aggregate the layers and determine the best performance by performing a grid search over the number of layers $L$, we fix the maximum number of layers to $L=5$ for ADMP-GNN. This comparison gives a strong advantage to the baselines, as their results are optimized for each dataset, while ADMP-GNN's performance is evaluated under a fixed and consistent setup.

\paragraph{Implementation Details.} We train all the models using the Adam optimizer \citep{kingma_adam} and the same hyperparameters. The GNN hyperparameters in each dataset were performed using a Grid search on the classical GCN; we detail the values of these hyperparameters in Table \ref{dynam:tab:gcn_tuned_hyperparams} of Appendix \ref{dynam:app:hyper}. To account for the impact of random initialization, each experiment was repeated 10 times, and the mean and standard deviation of the results were reported. The experiments have been run on both an NVIDIA A100 GPU.

\subsection{Experimental Results}

Through extensive experiments on multiple datasets, we can better understand the scenarios in which ADMP-GNN proves to be effective. As observed in Tables \ref{dynam:tab:results_with_baselines}, \ref{dynam:tab:results_with_baselines_GIN}, \ref{dynam:tab:results_with_baselines_2}, and \ref{dynam:tab:results_with_baselines_GIN_2}, a comparison between ADMP-GCN and ADMP-GIN against their respective baselines, GCN and GIN, demonstrates consistently higher accuracy for most datasets. Regarding the centrality-based layer selection policy, it becomes clear that this policy is particularly efficient when graphs exhibit a wide range of local density and centrality among nodes. Most importantly, no impactful drop in accuracy was observed with ADMP-GNN. It is important to note that the suitability of a given centrality metric can vary significantly across datasets. For instance, while PageRank is widely used, its distribution often forms a peak near zero due to the normalization constraint where the sum of PageRank scores across nodes equals 1, making it challenging to form meaningful clusters. In contrast, other centrality metrics like $k$-core and Walk Count are more adaptable in datasets where clear clustering patterns emerge. For example, in datasets like Cora, Ogbn, and Photo, the $k$-core centrality effectively highlights clusters, enabling an adaptive depth policy. For datasets like Texas and Wisconsin, Walk Count effectively captures the structural diversity necessary for cluster-based layer selection policy. These observations emphasize the importance of selecting appropriate centrality metrics tailored to the specific properties of a dataset to maximize the effectiveness of ADMP-GNN.

\section{Conclusion}
In this work, we propose ADMP-GNN, a novel adaption of message passing neural networks that enables to make predictions for each node at every layer. Additionally, we propose a sequential training approach designed to achieve performance comparable to training multiple GNNs independently in a single-task setting.  Our empirical analysis highlights the importance of node-specific depth in GNNs to effectively capture the unique characteristics and computational requirements of each node. Determining the optimal number of message-passing layers for each node is a challenging task, influenced by factors such as the complexity and connectivity of graph structures and the variability introduced by node features. Through our experiments, we identified that node centrality can serve as a useful indicator for determining the optimal layer for each node. To this end, we heuristically learn a layer selection policy using a set of validation nodes, which is then generalized to test nodes. Extensive experiments across multiple datasets, particularly those characterized by a diversity in local structural properties, demonstrate that ADMP-GNN significantly enhances the prediction accuracy of GNNs, offering an effective solution to address the challenges of layer selection and node-specific learning.

\part{Robustness of Graph Neural Networks}
\chapter{Adversarial Robustness of GNNs: Theory, Bounds, and Defense} \label{ch:Robustness}
\lettrine[lines=3]{G}{\small{raph}} Neural Networks (GNNs) have demonstrated state-of-the-art performance in various graph representation learning tasks. Recently, studies revealed their vulnerability to adversarial attacks. In this work, we theoretically define the concept of expected robustness in the context of attributed graphs and relate it to the classical definition of adversarial robustness in the graph representation learning literature. Our definition allows us to derive an upper bound of the expected robustness of Graph Convolutional Networks (GCNs) and Graph Isomorphism Networks subject to node feature attacks. Building on these findings, we connect the expected robustness of GNNs to the orthonormality of their weight matrices and consequently propose an attack-independent, more robust variant of the GCN, called the Graph Convolutional Orthonormal Robust Networks (GCORNs). We further introduce a probabilistic method to estimate the expected robustness, which allows us to evaluate the effectiveness of GCORN on several real-world datasets. Experimental experiments showed that GCORN outperforms available defense methods.

\section{Introduction}
Graph-structured data is prevalent in a wide range of domains, motivating therefore the development of neural network models that can operate on graphs, known as Graph Neural Networks (GNNs). 
GNNs have emerged as a powerful tool for learning node and graph representations. Many GNNs are instances of Message Passing Neural Networks (MPNNs) \citep{gilmer2017} such as Graph Isomorphism Networks (GIN)\citep{xu2019powerful} and Graph Convolutional Networks (GCN)\citep{Kipf:2017tc}. These models have been successfully applied in real-world applications such as molecular design~\citep{kearnes2016molecular}. In parallel to their success, it has been shown, particularly in the field of computer vision, that deep learning architectures can be susceptible to adversarial attacks \citep{denfeses_cv}. These attacks, which are based on injecting small perturbations into the input, lead to unreliable predictions, limiting therefore the applicability of these models to real-world problems. 
Similar to other deep learning architectures, GNNs are also vulnerable to adversarial attacks. Recent studies ~\citep{dai2018adversarial, zugner2018adversarial, gunnemann2022graph} have shown that a GNN can be attacked by applying small structural perturbations to the input graphs. These attacks pose a threat to the reliability of GNNs, particularly in safety-critical applications such as finance and healthcare. Consequently, different attacks have been proposed to explore the robustness of GNNs. Concurrently, several studies focus on developing methods to mitigate possible perturbation effects and improve the robustness of MPNNs. The proposed methods include augmenting training data with adversarial examples and retraining the model~\citep{graph_adversarial_training}, pre-processing methods such as edge pruning~\citep{gnn_guard}, and more recently robustness certificates~\citep{schuchardt2021collective}. While the majority of these proposed defenses focus on structural perturbations, only limited advances have been made for feature-based adversarial attacks on graphs. Moreover, despite the large amount of research on the robustness of these methods through empirical exploration, there has been limited progress in understanding the theoretical robustness of GNNs. 


In this work, we conduct a theoretical examination of the robustness of MPNNs subject to adversarial attacks. 
First, we formally define the concept of ``Expected Adversarial Robustness'' for structure and feature-based perturbations in the context of graphs defined in a metric space. We furthermore establish a formal relation between our introduced expected robustness and the conventional adversarial robustness formulation.
Further, by analyzing the input sensitivity of the iterative message-passing schemes, we derive an upper bound on the robustness of these models for both structural and feature-based attacks. Motivated by our theoretical results, we propose a refined learning scheme, called \emph{Graph Convolutional Orthonormal Robust Network (GCORN)}, for the GCN, to improve its robustness to feature-based perturbations while maintaining its expressive power. In addition to our theoretical findings, we empirically evaluate the effectiveness of our GCORN in both node and graph classification on commonly used real-world benchmark datasets and compare GCORN to existing methods to defend against feature-based adversarial attacks.
Most empirical evaluations of the effectiveness of defense methods consider worst-case scenarios, hence not taking into account the variability of attacks and their likelihood of occurrence. To overcome this limitation, we propose a novel probabilistic method for evaluating the expected robustness of GNNs, which is based on our introduced robustness definition. The method is model-agnostic and can hence be applied to any architecture to estimate the local robustness. More realistic and comprehensive evaluation of the effectiveness of defense approaches can hence be conducted.

Our main contributions are: \textbf{(i)} We define and theoretically analyze the expected robustness of MPNNs, producing an upper-bound on their expected robustness, \textbf{(ii)} a novel approach (GCORN) for improving the expected robustness of GCNs to feature-based  attacks while maintaining their performance in terms of accuracy, \textbf{(iii)} a theoretically well-founded, probabilistic and model-agnostic evaluation method, and \textbf{(iv)} an empirical evaluation of our GCORN on benchmark datasets, demonstrating its superior ability compared to existing methods to defend against feature-based attacks.


\section{Related Work}\label{sec:related_work}

While most studies on \textit{attacking machine learning models} focus on images~\citep{denfeses_cv}, work on discrete spaces such as graphs has also emerged. Analogous to images, most existing graph-based attack methods frame the task as a search/optimization problem. For instance, the targeted attack Nettack~\citep{zugner2018adversarial} utilized a greedy optimization scheme while \citet{zugner2019adversarial} formulate the problem as a bi-level optimization task and leverage meta-gradients to solve it. \citet{zhan2021black} expanded this work through a black-box gradient algorithm overcoming several limitations. Furthermore, \citet{dai2018adversarial} proposed to use Reinforcement Learning to solve the search problem and generate adversarial attacks. Node injection attacks \citep{Zou_2021, Tao_2021, chen2022understanding, ju2022let} have also proven effective, with attackers introducing malicious nodes instead of modifying existing nodes or edges, affecting therefore the model's performance. In parallel, the field of \textit{defending against adversarial attacks} on GNNs is still relatively under-explored compared to that of image-based models. The majority of methods are primarily focused on heuristic strategies. Similar to computer vision, robust training \citep{zugner_2019} and noise injection \citep{ennadir2024simple} have been used to improve the robustness of GNNs. Additionally, low-rank approximation with graph anomaly detection \citep{Ma_2021} has been used to defend against adversarial attacks. The GNN-Jaccard method~\citep{gnn_jaccard} pre-processes the adjacency matrix to detect potential manipulation of edges. In the same context, GNN-SVD~\citep{gnn_svd} uses a low-rank approximation of the adjacency matrix to filter out noise. Other methods such as edge pruning \citep{gnn_guard} and transfer learning \citep{Tang_2020} have also been proposed. Finally, \citet{lipschitz_bound_and_robust_training} proposed a low-pass adaptation of the message passing to enhance robustness while providing theoretical guarantees.
Although these defense strategies have had some success, for the majority of them, their heuristic nature results in defenses against specific types of attacks without any guarantees on the model's underlying robustness. As a result, these defenses may be susceptible to being circumvented by future new advanced attacks. Consequently, the \textit{investigation of robustness certificates} \citep{zugner_2019, bojchevski_2019, gosch2023revisiting} has gained attention by providing attack-independent guarantees on the stability of the model's predictions such as randomized smoothing \citep{bojchevski_certificate_2020}. 


While the majority of the existing work on defense strategies for GNNs has focused on structural perturbations, there is far less work on addressing feature-based attacks. This represents a significant gap in the literature as feature-based attacks on GNNs can be very effective. \citet{seddik_rgcn} propose to add a node feature kernel to the message passing to enhance the robustness of GCNs. Additionally, RobustGCN \citep{robustGCN} proposes to use Gaussian distributions as the hidden representations, enabling the absorption of the impact of both structural and feature-based attacks. Finally, \citet{airgnn} edited the message passing module using adaptive residual connections and feature aggregation, which have been shown experimentally to enhance the model's robustness against abnormal node features. Robustness certificates have also been proposed for node feature-based attacks \citep{scholten_graybox_certificate, bojchevski_certificate_2020}.

\section{Expected Adversarial Robustness}\label{sec:adversarial_robustness}
In this section we mathematically define the concept of the expected robustness for a graph-based function, such as a GNN. 
Let us consider three metric spaces with defined norms over the graph space ($\mathbb{A}, \lVert  \cdot \rVert_{\mathbb{A}}$), the feature space ($\mathbb{X}$, $\lVert  \cdot \rVert_{\mathbb{X}}$) and the label space ($\mathbb{Y}, \lVert  \cdot \rVert_{\mathbb{Y}}$). \noindent Let $\mathcal{D}$ be the underlying probability distribution defined on $(\mathbb{A}, \mathbb{X}, \mathbb{Y})$.
Given a graph-based function $f: (\mathbb{A}, \mathbb{X}) \rightarrow \mathbb{Y}$, and some input $\G = (\adj, X) \in (\mathbb{G}, \mathbb{X})$ with its corresponding label $y \in \mathcal{Y}$ where $f(\adj, X) = y$, the goal of an adversarial attack is to produce a perturbed graph $\widetilde{G}$ and its corresponding features $\widetilde{X}$ which are \emph{slightly} different from the original input $(G, X)$ such that the predicted class of $(\widetilde{G}, \widetilde{X})$ is different from the predicted class of $(G, X)$. The adversarial task is contingent on defining a similarity measure between the input graph and the adversarially generated graph. Note that for un-attributed graphs, the distance in \ref{eq:norm_3} aligns with the commonly used edit distance on graphs which is a measure of similarity between two graphs quantifying the minimal number of edges that need to be edited to convert one graph into another while taking into account graph isomorphism.
Based on this distance, we can mathematically formulate the adversarial task as finding a perturbed attributed graph $(\widetilde{\G}, \widetilde{\X})$ within a specified budget $\epsilon$ such that $f(\widetilde{\G}, \widetilde{\X}) = \widetilde{y} \neq y$ with $d^{\alpha, \beta}([\G, \X], [\widetilde{\G}, \widetilde{\X}]) < \epsilon$. Moreover, given that in practice the attacker does not have access to the ground-truth labels, we define an adversarial graph attack to be valid when $f(\widetilde{\G}, \widetilde{\X})  \neq f(\G, \X)$. We can now define the expected vulnerability of a graph function as its likelihood to suffer from such attacks in the input's neighborhood defined by $\epsilon.$ Upper bounding this vulnerability allows us to quantify the model's expected robustness, we start by formulating the expected vulnerability of a graph function $f$ as
\begin{equation}\label{equation:robustness_definition_1}
        Adv^{\alpha, \beta}_{\epsilon}[f] = \mathbb{P}_{(\G, \X) \sim \mathcal{D}_{\mathcal{\G}, \mathcal{\X}}} [ (\widetilde{\G}, \widetilde{\X}) \in B^{\alpha, \beta}(\G, \X, \epsilon):   d_{\mathcal{Y}}(f(\widetilde{\G}, \widetilde{\X}), f(\G, \X)) > \sigma],
\end{equation}
with $B^{\alpha, \beta}(\G,\X ,\epsilon) = \{(\widetilde{\G}, \widetilde{\X}): d^{\alpha, \beta}([\G, \X], [\widetilde{\G}, \widetilde{\X}])<\epsilon\}$ for any budget $\epsilon \geq 0$. Additionally, $d_{\mathcal{Y}}$ can be any defined distance in the output space $\mathcal{Y}$ and $\sigma > 0$. Here, we focus on real-valued output mappings and consider the distance metric $d_{\mathcal{Y}}(f(\widetilde{\G}, \widetilde{\X}), f(\G, \X)) = \lVert f(\widetilde{\G}, \widetilde{\X}) - f(\G, \X) \rVert_{\mathcal{Y}}$. This formulation is applicable to both graph and node classification tasks. In node classification, the parameter $\sigma$ determines the minimal number of nodes that need to be successfully attacked to consider the attack to be adversarially successful at the graph-level. This allows for flexibility in different scenarios, where in some cases, even a single node's label flip may be considered a threat, while in others, a limited number of changes can be tolerated. In graph classification, the parameter $\sigma $ acts as a threshold to the continuous softmax output above which an attack is considered effective. We can now introduce the concept of expected robustness of a function defined on graphs, such as a GNN.

\begin{definition}[Expected Adversarial Robustness]
\label{def:robustness} Let $d^{\alpha, \beta}$ be a graph distance on the spaces $(\mathcal{\G}, \mathcal{\X})$ and $d_\mathcal{Y}$ be a distance on the space $\mathcal{Y}$. The graph function $f: (\mathcal{\G}, \mathcal{\X}) \rightarrow \mathcal{Y}$ is $((d^{\alpha, \beta}, \epsilon),( d_{\mathcal{Y}}, \gamma))$--robust if its vulnerability as defined in \ref{equation:robustness_definition_1} can be upper-bounded by $\gamma,$ \ie $Adv^{\alpha, \beta}_{\epsilon}[f] \leq \gamma$.
\end{definition}
In Appendix \ref{sec:proof_proposition} (Proposition \ref{prop:equivalence}) we show how, via the equivalence of metrics, expected robustness in a given metric implies expected robustness in several other metrics.

\textbf{Relating Expected Adversarial Robustness to Worst-Case Adversarial Robustness.} Our introduced formulation represents a broader perspective of the classical ``worst-case'' adversarial robustness, where the attacker aims to identify a single adversarial attack representing ``worst-case'' losses within a predefined budget and neighborhood. 
Our Expected Adversarial Robustness focuses on understanding the overall behavior of the underlying graph-based function within the specified input neighborhood. This approach offers a more comprehensive assessment of the model's robustness. Nevertheless, we note that our formulation encompasses the adversarial robustness as a special case since by definition these examples are included in our considered neighborhood. In fact, by adjusting the hyper-parameter $\sigma$, we can isolate these worst-case examples. Lemma \ref{lemma:worst_case} directly relates Definition \ref{def:robustness} to the classical ``worst-case'' adversarial robustness.

\begin{lemma}
\label{lemma:worst_case}
Let $d^{\alpha, \beta}$ be a defined graph metric on the metric spaces $\mathcal{\G}, \mathcal{\X}$. Let $f: (\mathcal{\G}, \mathcal{\X}) \rightarrow \mathcal{Y}$ be a graph-based function, we have the following result: If $f$ is $((d^{\alpha, \beta}, \epsilon),( d_{\mathcal{Y}}, \gamma))$--robust, then $f$ is also $((d^{\alpha, \beta}, \epsilon),( d_{\mathcal{Y}}, \gamma))$--``worst-case'' robust. 
\end{lemma}

The proof of Lemma \ref{lemma:worst_case} is provided in Appendix \ref{sec:proof_lemma}. As a result, our forthcoming theoretical analysis, which considers the general case, is equally applicable to worst-case adversarial examples, which will be also validated experimentally in Section \ref{sec:experiments}. We finally note that the advantages and pitfalls of the generalization of ``worst-case'' to average robustness have also been studied by \citet{average_and_worst_case_neurips21}.

\section{The Expected Robustness of Message Passing Neural Networks}\label{sec:result_robustness}
We now use our Expected Adversarial Robustness Definition \ref{def:robustness} to derive an upper bound on the expected robustness of the GCN, based on which we introduce our more robust GCN adaptation.

\subsection{On the Expected Robustness of Graph Convolutional Networks}

Our work primarily focuses on the theoretical analysis of GCNs within the broader context of MPNNs. 
The computations of one GCN layer are composed of the aggregation of node hidden states over neighborhoods in the graph and subsequent node-wise updates of the hidden states via a weight matrix and non-linear activation function.  

The updated hidden states are then passed to the next layer for further aggregation and updates. An iteration of this process can be expressed as follows
\begin{equation} \label{equation:gcn}
    h^{(\ell)} = \phi^{(\ell)}(\widetilde{\adj}h^{(\ell-1)}W^{(\ell)}),
\end{equation}
where $h^{(\ell)}$ represents the hidden state in the $\ell$-th GCN layer and $h^{(0)}$ is the initial node features $X\in \mathbb{R}^{n \times K}$, $W^{(\ell)} \in \mathbb{R}^{p \times e}$ is the weight matrix in the $\ell$-th layer, $e$ is the embedding dimension and $\phi^{(\ell)}$ is a 1-Lipschitz continuous non-linear activation function. Moreover, $\widetilde{\adj} \in \mathbb{R}^{n \times n}$, with $n$ being the number the nodes, denotes the normalized adjacency matrix  $\widetilde{\adj} = \D^{-1/2} \adj \D^{-1/2}.$ 

Determining the exact expected adversarial robustness of a graph-based function, as outlined in Section \ref{sec:adversarial_robustness}, poses a significant challenge. To overcome this, we provide an upper bound, referred to as~$\gamma$ in Definition \ref{def:robustness}.
Our definition of adversarial attacks is closely related to the concept of sensitivity analysis. In both cases, the goal is to understand how small input changes can affect the model's output. We hence tackled the adversarial task by adopting an input perturbation perspective. 
Similar approaches based on sensitivity analysis have had some success for deep neural networks (DNNs). While extending to other domains such as Convolutional Neural Networks is direct, generalizing to graphs presents new challenges. Notably, model dynamics differ due to the Message Passing process involving the adjacency and node features, complicating the task of providing an upper-bound. Since, the architecture itself involves the adjacency matrix, any perturbation on the input produces a different dynamic in the model itself, unlike the classical DNNs architecture, which remains static when subject to perturbations. Consequently, the theoretical analysis and results have to reflect the underlying propagation architecture, i.e.,  the graph structure. Theorem \ref{theo:main_result} provides theoretical insight into the expected robustness of GCNs by establishing an upper bound on the amount of perturbation that a GCN can tolerate before its predictions become unreliable.

\begin{theorem}
\label{theo:main_result}
Let $f: \Gset \rightarrow \Y$ denote a graph-based function composed of $L$ GCN layers, with $W^{(i)}$ denoting the weight matrix of the $i$-th layer. Further, let $d^{0,1}$ be a feature distance. For attacks targeting node features of the input graph, with a budget $\epsilon$, with respect to Definition \ref{def:robustness}: 
\begin{itemize}
    \item $f$ is $((d^{0, 1}, \epsilon),( d_1, \gamma))$--robust  with $\gamma  = \prod_{\ell=1}^{L} \lVert W^{(\ell)} \rVert_1  \epsilon (\sum_{u \in \mathcal{V}} \hat{w}_u ) /\sigma,$ with $\hat{w}_u$ denoting the sum of normalized walks of length $(L-1)$ starting from node $u$ and $\mathcal{V}$ is the node set.
    \item $f$ is  $((d^{0, 1}, \epsilon),( d_\infty, \gamma))$--robust   with $\gamma  = \prod_{\ell=1}^{L} \lVert W^{(\ell)} \rVert_\infty  \epsilon \hat{w}_G /\sigma$, with $\hat{w}_G=\underset{u \in \mathcal{V}}{\max} ~ \hat{w}_u.$
\end{itemize}

\end{theorem}

As previously mentioned, the provided upper-bound in Theorem \ref{theo:main_result} is directly dependent on both the graph structure (in terms of walks from the graph's nodes) and the propagation scheme (where the length of the considered walks depends on the number of message-passing iterations). 
The derived upper-bound is intuitive: effectively using feature-based attacks is increasingly difficult with increasing graph sparsity; or conversely, in dense graphs node feature attacks can have greater effect since the message passing scheme propagates them along a greater number of walks. To make this precise, the expected sum of normalized walks in a sparser graph tends to be lower than in a denser one, resulting in a reduced bound, indicating a more robust model within the considered neighborhood. We finally note that this bound applies to both targeted and untargeted feature modifications, whether limited to a specific subset or all the nodes. 
While our main focus is on node feature-based attacks, our analysis provided in Theorem \ref{theo:main_result} can be extended to structural attacks; Theorem \ref{theo:structural_perturbtation} sheds light on this latter case. 

\begin{theorem}
\label{theo:structural_perturbtation}

Let $f: \Gset \rightarrow \Y$ denote a graph function composed of $L$ GCN layers, where $W^{(i)}$ denotes the weight matrix of the $i$-th layer. Further, let $d^{1,0}$ be a graph distance. For structural attacks with a budget $\epsilon$, the function $f$ is $((d^{1, 0}, \epsilon),( d_2, \gamma))$--robust with
\begin{equation}\label{eqn:StructuralAttacks}
\gamma = \prod_{\ell=1}^{L} \lVert W^{(\ell)} \rVert_2 \lVert \X \rVert_2 \epsilon (1 + L  \prod_{\ell=1}^{L} \lVert W^{(\ell)} \lVert_2 )/\sigma.    
\end{equation}

\end{theorem}

We observe the upper bound in \ref{eqn:StructuralAttacks} to functionally depend on the size of the graph via $\lVert X \rVert_2,$ yielding the intuitive result that larger graphs have more potential targets to attack and thereby give rise to less robust models. The proofs of Theorems \ref{theo:main_result} and \ref{theo:structural_perturbtation} are provided in Appendix \ref{sec:proof_theorem}.

\subsection{On the Generalization to Other Graph Neural Networks}

While our work focuses on GCNs, our analysis can be extended to any GNN. Our theoretical analysis is contingent on assuming the input node feature space to be bounded, which is a realistic assumption for real word data. For illustration, Theorem \ref{theo:gin_results} derives the upper-bound of the specific case of GINs. 

\begin{theorem}
\label{theo:gin_results}
Let $f: (\mathcal{G}, \mathcal{X}) \rightarrow \Y$ be composed of $L$ GIN-layers (with parameter $\zeta = 0,$ that is usually denoted by $\epsilon$ in the literature) and $W^{(i)}$ denote the weight matrix of the $i$-th MLP layer. We consider the input node feature space to be bounded, i.e.,  $\lVert X \rVert_2 < B$ for some $B\in \mathbb{R},$  and graphs of maximum degree $\Delta_G.$ For node feature-based attacks, with a budget $\epsilon$, the function $f$ is $((d^{0, 1}, \epsilon),( d_\infty, \gamma))$--robust with
    $$\gamma  = \prod_{l=1}^L \lVert W^{(l)}\lVert_{\infty}  (B ~L ~\Delta_G  + \epsilon)  /\sigma.$$
\end{theorem}

Theorem \ref{theo:gin_results} is proved in Appendix \ref{sec:proof_generalization}. 
In addition, following the same assumption, it is possible to derive an upper bound on the GCN's robustness when subject to both structural and feature-based attacks simultaneously. The complete study and additional information are provided in Appendix \ref{appendix:simultaneous_attacks}.

\subsection{Enhancing the Robustness of Graph Convolutional Networks}

Leveraging the established upper-bound on the expected robustness in Theorem \ref{theo:main_result}, we now introduce a novel approach, called Graph Convolutional Orthonormal Robust Networks (GCORNs), which enhances the robustness of a GCN to node feature perturbations while maintaining its ability to learn accurate node and graph representations. 
To this end, we aim to design a GCN architecture for which the upper bound $\gamma$ as stated in Theorem~\ref{theo:main_result} is low. As this quantity is dependent on the norm of the weight matrices in each layer of the GCN, we propose to control the norms of these matrices. 

Enhancing the robustness through enforcing matrix norm constraints during training has previously been studied in the context of DNNs through methods such as Parseval regularization \citep{parseval_cisse} or optimizing over the orthogonal manifold \citep{orthogonal_manifold_ablin}. However, as observed in our experiments, the added constraints of these methods can negatively impact the performance of the graph function and additionally the hyper-parameter tuning can be tricky especially when dealing with large datasets. In our work, we choose to tackle the task by modifying the mathematical formulation of the GCN in \ref{equation:gcn} to encourage the orthonormality of the weights. 
We have therefore chosen to use an iterative algorithm \citep{bjork_paper}, that mainly was used in the literature for studying Lipschitz approximation \citep{liptschitz_function_approx}, to ensure a fair trade-off between clean and attacked accuracy. We note that, based on the introduced Theorem \ref{theo:main_result}, any orthonormalization method can theoretically enhance the underlying model's robustness.
Given our weight matrix $W$, the iterative process which is computed using Taylor expansion, consists of finding the closest orthonormal matrix $\hat{W}$ to our weight matrix $W$. By considering $\hat{W}_0 = W$, we recursively compute $\hat{W}_k$ from 
\begin{equation}
    \hat{W}_{k+1} = \hat{W}_k\left(I + \tfrac{1}{2} Q_k   + \ldots + (-1)^p \tbinom{-1/2}{p} Q_k^p\right),
    \label{equation:bjork}
\end{equation}
with $k\geq0$, $Q_k = I - \hat{W}_k^T \hat{W}_k$ and $p\geq1$ is the chosen order. A key advantage of this orthonormalization approach is its differentiability, hence its compatibility with the back-propagation process of training GNNs. 
By incorporating this projection into each forward pass during the model's training, we encourage the orthonormality of the weights, consequently enhancing its expected robustness.

\textbf{Training and Convergence of Our Approach.} Encouraging the orthonormality of the weight matrices for each layer in our framework mitigates the problems of vanishing and exploding gradients. Our proposed method preserves the gradient norm, leading to enhanced convergence and improved learning \citep{guo2021orthogonal}. It is important to note that the training of our framework relies on the convergence of the iterative orthonormalization process.
From the original work \citep{bjork_paper} which examined this aspect, convergence is contingent on the condition $\lVert W^T W - I \lVert \leq 1$ being satisfied, which can be guaranteed by applying a scaling factor, based on the spectral norm, to the weight matrices before the iterative process. We noticed that this approach not only guarantees that the weight matrices satisfy the necessary condition for convergence but also helps to speed up the convergence and the training process as analyzed by previous work \citep{salimans2016weight}.

\textbf{Complexity of Our Approach.} Equation \ref{equation:bjork} entails a trade-off between convergence speed and the approximation's precision. A higher order $p$ yields closer projections in each iteration, resulting in an increased computational complexity. This trade-off must be carefully considered to strike a balance between complexity and approximation accuracy. The main complexity of the method results from the matrix products which are $\mathcal{O}(e^3)$ where $e$ represents the embedding dimension. Our experiments indicate that a low order and a small number of iterations often yield a satisfactory approximation in practice. We note that our method's complexity does not increase with growing input graph size, distinguishing it from other defenses such as GNNGuard which has a complexity of $\mathcal{O}(e \times |E|)$, where $|E|$ represents the number of edges or GNN-SVD which computes a low-rank approximation through SVD with a corresponding complexity of $\mathcal{O}(n^3)$, with $n$ being the number of nodes. We point out that we conducted an ablation study (reported in Appendix \ref{sec:time_complexity_analysis}) on the effect of changing the order $p$ and number of iterations $k$ on both the precision of the orthonormal projection (represented by both the clean accuracy and the adversarial accuracy) and the time complexity.

\section{Estimation of Our Robustness Measure} \label{sec:probabilistic_evaluation}

Based on our introduced definition of expected robustness in Section \ref{sec:adversarial_robustness}, we introduce an algorithm to empirically estimate the quantity $Adv^{\alpha, \beta}_{\epsilon}[f]$, serving therefore as a new metric to evaluate GNN robustness. While most defense methods on graphs are evaluated using the worst-case accuracy of a GNN when exposed to specific attack schemes \citep{bojchevski_certificate_2020,pgd_paper,zugner2019adversarial}, $Adv^{\alpha, \beta}_{\epsilon}[f]$ is an attack-independent and model-agnostic robustness metric based on uniform sampling. It can therefore be used as a \textit{security checkpoint}, in addition to existing robustness metrics, to evaluate the GNN robustness against unknown attack distributions.

Numerous model-agnostic robustness metrics \citep{weng2018clever,DBLP:journals/corr/abs-2102-03716} have emerged primarily in computer vision. These metrics mainly assess robustness through measuring the \textit{distortion} between input and output manifolds. While these methods can be extended to the graph classification task with appropriate distortion definitions, their application in the node classification context proves challenging. To our knowledge, there exists no specific or analogous adaptation within the context of graph representation learning. Thus, our proposed evaluation aims to address this gap.

Given that the expected value of an indicator random variable for an event  $E$ is the probability of that event, i.e., $\mathbb{E}[ \mathbf{1} \{ E \} ] = \mathbb{P}(E)$, we use \ref{equation:robustness_expected_value} as an equivalent formulation of $Adv^{\alpha, \beta}_{\epsilon}[f]$ in \ref{equation:robustness_definition_1}.
\begin{equation}\label{equation:robustness_expected_value}
           Adv^{\alpha, \beta}_{\epsilon}[f]   = \mathbb{E}_{\substack{(\G, \X) \sim \mathcal{D}_{\mathcal{G}, \mathcal{X}} ,\\(\widetilde{\G}, \widetilde{\X}) \in B^{\alpha, \beta}((\G,\X) ,\epsilon)}}
        \left [ \mathbf{1} \{d_{\mathcal{Y}}(f(\widetilde{\G}, \widetilde{\X}), f(\G, \X)) > \sigma  \}  \right ]. 
        \end{equation}

From the law of large numbers \citep{bernoulli1713ars}, the mean sampling is an unbiased estimator of the vulnerability quantity $Adv^{\alpha, \beta}_{\epsilon}[f].$ Consequently, we estimate a GNN's expected robustness by generating multiple graph pairs $([\G,\X],[\widetilde{\G} , \widetilde{\X}])$, where each input graph $[\G,\X]$ is sampled from the underlying distribution $D_{\mathcal{\G}, \mathcal{\X}}$ and $[\widetilde{\G},\widetilde{\X}]$ is uniformly sampled from the ball of radius $\epsilon$ centered around $[\G,\X],$ \ie~ $d^{\alpha, \beta}([\G, \X], [\widetilde{\G} , \widetilde{\X} ]) \leq \epsilon$. 
This approach can be used for both structural and feature attacks. Since we focus on the latter, we propose a sampling strategy based on the distance
\begin{equation}
\label{eq:prob_eval_distance}
    d^{0,1}([\G, \X], [\widetilde{\G} , \widetilde{\X}]) = \lVert \X -\widetilde{\X} \lVert_{\mathcal{X}}  = \max_{i \in \{1,\ldots, n\}} \lVert \X_{i}-\widetilde{\X}_{i}\lVert_p ,
\end{equation}
where $\lVert\cdot\lVert_p$ is the $L_p$-norm ($p>0$) and $\X_{i}, \widetilde{\X}_{i}$ are the $i$-th rows of the matrices $X,\widetilde{\X} \in \mathbb{R}^{n\times K}.$

Since the original dataset consists of samples $[\G,\X]$ from the  underlying distribution $D_{\mathcal{G}, \mathcal{X}}$, we only need to sample uniformly graphs $[\widetilde{\G},\widetilde{\X}]$ from the neighborhood of the dataset graphs $[G,X]$ such that $\lVert \X - \widetilde{\X}\lVert_{\mathcal{X}} \leq \epsilon$ and $\widetilde{\G} = \G$. Sampling such $\widetilde{\X}$ is equivalent to first sample $\mathbf{Z} \in  \mathbb{R}^{n \times K}$ from $ \mathcal{B}_{\epsilon} =\left \{ \mathbf{Z}\in \mathbb{R}^{n \times K}: \lVert \mathbf{Z} \lVert_{\mathcal{X}} \leq \epsilon\right \}$ and then set $\widetilde{\X} = \X + \mathbf{Z}$. Sampling from the ball $\mathcal{B}_{\epsilon}$ could be performed by partitioning the ball with respect to the possible values of $\lVert Z \lVert_{\mathcal{X}}$.
\begin{equation*}
    \centering
    S_r =  \{  \mathbf{Z}\in \mathbb{R}^{n\times K}: \lVert \mathbf{Z} \lVert_{\mathcal{X}}  = r \}, \qquad
    \mathcal{B}_{\epsilon}  = \cup_{r \leq \epsilon} S_r;  \qquad
    \forall r\neq r'~~~~S_r \cap S_{r'} = \emptyset.
\end{equation*}

We therefore propose to use the following \textit{Stratified Sampling} approach, which is based on two main steps: \textbf{(i)} Sample a distance $r$ from $[0, \epsilon]$  using the distribution $p_\epsilon$ defined in Lemma \ref{lem:sampling_radius}; \textbf{(ii)} Sample $\mathbf{Z} $ from  $S_r$. Additional details on the prior distribution $r$ and the sampling from $S_r$ are in Appendix~\ref{appendix:prob_eval}.

\begin{lemma}
\label{lem:sampling_radius}
Let $\mathbb{R}^K$ be the real finite-dimensional space and $\epsilon$ a positive real number. If  $R^{(p)}$ is the random variable indicating the maximum of the $L_p$ norm's values inside the ball of radius $\epsilon$, i.e., 
$\mathcal{B}_\epsilon = \left \{  \mathbf{Z} \in \mathbb{R}^{n\times K}: \max_{i \in \{1,\ldots, n\}} \lVert \mathbf{Z}_i \lVert_p \leq \epsilon  \right \} .$
Then, for every $p>0$, the density distribution of $R^{(p)}$ does not depends on $p$ and is defined as follows, $ p_\epsilon(r)   =  K \tfrac{1}{\epsilon} \left ( \tfrac{r}{\epsilon}\right )^{K-1}\mathbf{1}\{ 0 \leq r \leq \epsilon \}. $

\end{lemma}
The proof is available in Appendix \ref{appendix:proof_lemma}. We note that the proposed framework is a practical and theory-based robustness evaluation approach applicable to any GNN, since no assumption has been made on the underlying architecture. Algorithm \ref{algo:prob_eval} in Appendix \ref{appendix:prob_eval} offers a summary of the approach.

\section{Empirical Investigation}\label{sec:experiments}
This section examines the practical impact of our theoretical findings on real-world datasets where we aim to investigate the robustness of GCORN in comparison to other defense benchmarks.

\begin{table}[t]

\tiny
\centering
\caption[Attacked classification accuracy of the models after the feature attacks]{Attacked classification accuracy ($\pm$ standard deviation) of the models on different benchmark node classification dataset after the feature attacks. }
\label{tab:results_node_classification}
\resizebox{\textwidth}{!}{%
\begin{tabular}{@{}llcccccc@{}}
\toprule
Attack                                                                            & Dataset    & \textbf{GCN}                       & \textbf{GCN-k} & \textbf{AirGNN}         & \textbf{RGCN}           & \textbf{ParsevalR} & \textbf{GCORN}           \\ \midrule
\multirow{5}{*}{\begin{tabular}[c]{@{}l@{}}Random \\ ($\psi = 0.5$)\end{tabular}} & Cora       & \multicolumn{1}{l}{68.4 $\pm$ 1.9} & 69.2 $\pm$ 2.6 & 73.5 $\pm$ 1.9          & 71.6 $\pm$ 0.3          & 72.9 $\pm$ 0.9     & \textbf{77.1 $\pm$ 1.8}  \\
                                                                                  & CiteSeer   & 57.8 $\pm$ 1.5                     & 62.3 $\pm$ 1.2 & 64.6 $\pm$ 1.6          & 63.7 $\pm$ 0.6          & 65.1 $\pm$ 0.8     & \textbf{67.8 $\pm$ 1.4}  \\
                                                                                  & PubMed     & \multicolumn{1}{l}{68.3 $\pm$ 1.2} & 71.2 $\pm$ 1.1 & 70.9 $\pm$ 1.3          & 71.4 $\pm$ 0.5          & 71.8 $\pm$ 0.8     & \textbf{73.1 $\pm$ 1.1}  \\
                                                                                  & CS         & 85.3 $\pm$ 1.1                     & 86.7 $\pm$ 1.1 & 87.5 $\pm$ 1.6          & 88.2 $\pm$ 0.9          & 87.6 $\pm$ 0.6     & \textbf{89.8 $\pm$ 1.2}  \\
                                                                                  & OGBN-Arxiv & 68.2 $\pm$ 1.5                     & 52.8 $\pm$ 0.5 & 66.5 $\pm$ 1.3          & 63.8 $\pm$ 1.9          & 68.3 $\pm$ 1.9     & \textbf{69.1 $\pm$ 1.8}  \\ \midrule
\multirow{5}{*}{\begin{tabular}[c]{@{}l@{}}Random \\ ($\psi = 1.0$)\end{tabular}} & Cora       & \multicolumn{1}{l}{41.7 $\pm$ 2.1} & 46.3 $\pm$ 2.8 & 53.7 $\pm$ 2.2          & 52.8 $\pm$ 1.6          & 55.3 $\pm$ 1.2     & \textbf{57.6 $\pm$ 1.9}  \\
                                                                                  & CiteSeer   & 38.2 $\pm$ 1.3                     & 45.3 $\pm$ 1.4 & 49.8 $\pm$ 2.1          & 43.7 $\pm$ 2.2          & 51.2 $\pm$ 1.2     & \textbf{57.3 $\pm$ 1.7}  \\
                                                                                  & PubMed     & \multicolumn{1}{l}{60.1 $\pm$ 1.7} & 62.3 $\pm$ 1.3 & 62.4 $\pm$ 1.2          & 61.9 $\pm$ 1.2          & 61.3 $\pm$ 1.7     & \textbf{65.8 $\pm$ 1.4}  \\
                                                                                  & CS         & 69.9 $\pm$ 1.3                     & 73.2 $\pm$ 0.9 & 76.7 $\pm$ 2.8          & 76.2 $\pm$ 1.4          & 78.7 $\pm$ 1.2     & \textbf{81.3 $\pm$ 1.6}  \\
                                                                                  & OGBN-Arxiv & 66.4 $\pm$ 1.9                     & 46.6 $\pm$ 0.6 & 62.7 $\pm$ 1.6          & 63.0 $\pm$ 2.4          & 66.1  $\pm$ 0.7    & \textbf{67.3  $\pm$ 2.1} \\ \midrule
\multirow{5}{*}{PGD}                                                              & Cora       & \multicolumn{1}{l}{54.1 $\pm$ 2.4} & 58.3 $\pm$ 1.6 & 68.2 $\pm$ 1.8          & 62.5 $\pm$ 1.2          & 68.6 $\pm$ 1.7     & \textbf{71.1 $\pm$ 1.4}  \\
                                                                                  & CiteSeer   & 52.3 $\pm$ 1.1                     & 59.6 $\pm$ 1.6 & 59.3 $\pm$ 2.1          & 61.9 $\pm$ 1.1          & 62.1 $\pm$ 1.5     & \textbf{65.6 $\pm$ 1.4}  \\
                                                                                  & PubMed     & \multicolumn{1}{l}{66.1 $\pm$ 2.1} & 67.3 $\pm$ 1.3 & 70.8 $\pm$ 1.7          & 69.5 $\pm$ 0.9          & 68.9 $\pm$ 2.1     & \textbf{72.3 $\pm$ 1.3}  \\
                                                                                  & CS         & 71.3 $\pm$ 1.1                     & 74.1 $\pm$ 0.8 & 76.3 $\pm$ 2.1          & 76.6 $\pm$ 1.2          & 77.3 $\pm$ 0.6     & \textbf{79.6 $\pm$ 1.2}  \\
                                                                                  & OGBN-Arxiv & 67.5 $\pm$ 0.9                     & 49.9 $\pm$ 0.7 & 55.7 $\pm$ 0.9          & 63.6 $\pm$ 0.7          & 67.6 $\pm$ 1.2     & \textbf{68.1 $\pm$ 1.1}  \\ \midrule
\multirow{5}{*}{Nettack}                                                          & Cora       & \multicolumn{1}{l}{60.9 $\pm$ 2.5} & 64.2 $\pm$ 5.2 & 66.7 $\pm$ 3.8          & 63.4 $\pm$ 3.8          & 67.5 $\pm$ 2.5     & \textbf{68.3 $\pm$ 1.4}  \\
                                                                                  & CiteSeer   & 55.8 $\pm$ 1.4                     & 71.7 $\pm$ 1.4 & 67.5 $\pm$ 2.5          & 70.8 $\pm$ 3.8          & 69.2 $\pm$ 3.8     & \textbf{77.5 $\pm$ 2.5}  \\
                                                                                  & PubMed     & \multicolumn{1}{l}{60.0 $\pm$ 2.5} & 65.8 $\pm$ 2.9 & 69.2 $\pm$ 1.4          & \textbf{71.7 $\pm$ 3.8} & 68.3 $\pm$ 1.4     & 70.8 $\pm$ 1.4           \\
                                                                                  & CS         & 55.8 $\pm$ 1.4                     & 71.6 $\pm$ 1.4 & 76.7 $\pm$ 1.4          & 71.7 $\pm$ 2.9          & 75.8 $\pm$ 2.8     & \textbf{78.3 $\pm$ 1.4}  \\
                                                                                  & OGBN-Arxiv & 49.2 $\pm$ 2.9                     & 53.3 $\pm$ 1.4 & \textbf{56.7 $\pm$ 1.4} & 52.6 $\pm$ 2.5          & 55.8 $\pm$ 1.4     & 55.8 $\pm$ 1.4           \\ \bottomrule
\end{tabular}%
}
\end{table}

\subsection{Experimental Setup}

The necessary code to reproduce all our experiments is available on github \href{https://github.com/Sennadir/GCORN}{https://github.com/Sennadir/GCORN}. In this section, we focus on node classification while we report results on graph classification \citep{morris2020tudataset} in Appendix \ref{appendix:additional_results}. We use the citations networks Cora, CiteSeer, and PubMed \citep{dataset_node_classification}, the Co-author network CS \citep{cs_data}, and the citation network between Computer Science arXiv papers OGBN-Arxiv \citep{hu2021open}. Further information about the datasets and implementation details are provided in Appendix~\ref{sec:dataset_implementation_details}. 

To reduce the impact of initialization, we repeated each experiment 10 times and used the train/validation/test splits provided with the datasets and for the CS dataset, we followed the framework of \citet{yang_2016}. We note that for the node feature-based attacks, we normalized the input features to work in a continuous space allowing more flexibility in terms of available attacks.  

\textbf{Attacks.} We use two evasion and one poisoning feature-based attacks: 
\textbf{(i)} The baseline random attack injecting Gaussian noise $\mathcal{N}(0, \mathbf{I})$ to the features with a scaling parameter $\psi$ controlling the attack budget;  
\textbf{(ii)} The white-box Proximal Gradient Descent \citep{pgd_paper}, which is a gradient-based approach to the adversarial optimization task. We set the perturbation rate to 15\%. While this attack has limited success for structural perturbations due to the discrete space, it is known to be powerful in continuous spaces, such as the node feature space;

\textbf{(iii)} The targeted-attack Nettack~\citep{zugner2018adversarial}, where similar to the original study, we selected 40 correctly classified target nodes, comprising 10 with the largest classification margin, 20 randomly chosen, and 10 with the smallest margin.

\textbf{Baseline Models.} 
Currently, as presented in Section~\ref{sec:related_work}, there are only few methods for defending against feature-based adversarial perturbations. We compare against \textbf{(i)} GCN-k \citep{seddik_rgcn}; \textbf{(ii)} RobustGCN (RGCN) \citep{robustGCN}; \textbf{(iii)} AIRGNN \citep{airgnn} and \textbf{(iv)} we finally compared to the Parseval Regularization (ParsevalR) \citep{parseval_cisse}, another orthonormalization method, with great success in the computer vision. The aim is mainly to motivate the merits of our proposed orthonormalization method based on the iterative weight projection.

\subsection{Experimental Results}

\begin{figure}[t]
    \centering
    \includegraphics[width=\textwidth]{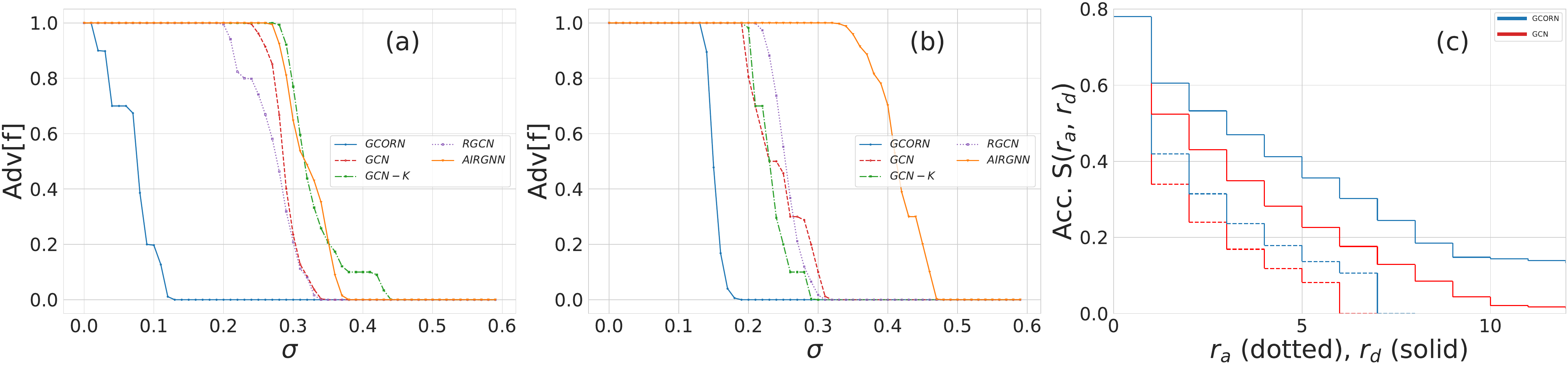}
    \caption[Robustness guarantees on Cora]{(a) and (b) display $Adv^{\alpha, \beta}_{\epsilon}[f]$ for Cora and OGBN-Arxiv. (c) Robustness guarantees on Cora, where $r_a,r_d$ are respectively the maximum number of adversarial additions and deletions.}
    \label{fig:all_plots}
\end{figure}

\textbf{Empirical Estimation of Our Expected Robustness.} We start by investigating the robustness of the different considered benchmarks through the empirical estimation of $Adv^{\alpha, \beta}_{\epsilon}[f]$ as introduced in Section \ref{sec:probabilistic_evaluation} (see Appendix \ref{sec:dataset_implementation_details}).  Figure \ref{fig:all_plots} (a) and (b) report the estimated values of the vulnerability $Adv^{\alpha, \beta}_{\epsilon}[f]$ for our GCORN and the considered defense benchmarks for the Cora and OGBN-Arxiv datasets, respectively. Additional results on further datasets are provided in Appendix \ref{appendix:additional_results}. The figures show that our GCORN yields the lowest adversarial vulnerability indicating therefore that our proposed approach is exhibiting greater expected robustness within the considered neighborhood.

\textbf{Worst-Case Adversarial Evaluation.} Table~\ref{tab:results_node_classification} reports the attacked node classification accuracies for our GCORN and the other considered benchmarks. The results show that, when subject to features-based adversarial attacks with a varying budget, the performance of GCN is severely impacted. Instead, the proposed GCORN demonstrates a significant improvement in defending against these attacks in comparison to other methods. For instance, GCORN is outperforming the GCN-k by an average of $\sim12$\% in accuracy. In certain situations, GCORN demonstrates the ability to recover the GCN's performance to a level equivalent to no adversarial attack.
Furthermore, the baselines are not as effective in defending against gradient-based attacks. In contrast, GCORN, which is based on an attack-agnostic upper bound (Theorem~\ref{theo:main_result}), demonstrates its \textit{ability to defend against gradient-based attacks}. We, additionally study the trade-off between robustness and input sensitivity in Appendix \ref{appendix:additional_results}.

\textbf{Certified Robustness.} In addition to our probabilistic and empirical evaluations, we assessed the certified robustness of a GCN and our proposed GCORN on the Cora dataset using the sparse randomized smoothing approach \citep{bojchevski_certificate_2020}. Specifically, we plotted their certified accuracy $S(r_a, r_d)$ for varying addition $r_a$ and deletion $r_d$ radii. Figure~\ref{fig:all_plots}(c) showcases our theoretical robustness; GCORN improves the certified accuracy with respect to feature perturbations for all radii.


\begin{table}[t] 
\centering
\caption[Attacked classification accuracy of the models after the structural attacks]{Attacked classification accuracy ($\pm$ standard deviation) of the models on different benchmark node classification datasets after the structural attacks.}
\label{tab:structural_perturbations}

\resizebox{\textwidth}{!}{%
\begin{tabular}{@{}llccccccc@{}}
\toprule
Attack                   & Dataset  & GCN              & GCN-Jaccard    & RGCN                    & GNN-SVD        & \multicolumn{1}{l}{GNN-Guard} & ParsevalR               & GCORN                   \\ \midrule
\multirow{4}{*}{Mettack} & Cora     & 73.0   $\pm$ 0.7 & 75.4 $\pm$ 1.8 & 69.2 $\pm$ 0.3          & 73.6 $\pm$ 0.9 & 74.4 $\pm$ 0.8                & 71.9 $\pm$ 0.7          & \textbf{77.3 $\pm$ 0.5} \\
                         & CiteSeer & 63.2 $\pm$ 0.9   & 69.5 $\pm$ 1.9 & 68.9 $\pm$ 0.6          & 65.8 $\pm$ 0.6 & 68.8 $\pm$ 1.5                & 68.3 $\pm$ 0.8          & \textbf{73.7 $\pm$ 0.3} \\
                         & PubMed   & 60.7 $\pm$ 0.7   & 62.9 $\pm$ 1.8 & 65.1 $\pm$ 0.4          & 82.1 $\pm$ 0.8 & \textbf{84.8 $\pm$ 0.3}       & 69.5 $\pm$ 1.1          & 71.8 $\pm$ 0.4          \\
                         & CoraML   & 73.1 $\pm$ 0.6   & 75.4 $\pm$ 0.4 & 77.1 $\pm$ 1.1          & 71.3 $\pm$ 1.0 & 76.5 $\pm$ 0.7                & 76.9 $\pm$ 1.3          & \textbf{79.2 $\pm$ 0.6} \\ \midrule
\multirow{4}{*}{PGD}     & Cora     & 76.7   $\pm$ 0.9 & 78.3 $\pm$ 1.1 & 72.0 $\pm$ 0.3          & 71.6 $\pm$ 0.4 & 75.0 $\pm$ 2.0                & 78.4 $\pm$ 1.2          & \textbf{79.9 $\pm$ 0.4} \\
                         & CiteSeer & 67.8 $\pm$ 0.8   & 70.9 $\pm$ 1.0 & 62.2 $\pm$ 1.8          & 60.3 $\pm$ 2.4 & 68.9 $\pm$ 2.2                & 70.6 $\pm$ 1.0          & \textbf{73.1 $\pm$ 0.5} \\
                         & PubMed   & 75.3 $\pm$ 1.6   & 73.8 $\pm$ 1.3 & 78.6 $\pm$ 0.4          & 81.9 $\pm$ 0.4 & \textbf{84.3 $\pm$ 0.4}       & 77.3 $\pm$ 0.7          & 77.4 $\pm$ 0.4          \\
                         & CoraML   & 76.9 $\pm$ 1.2   & 75.0 $\pm$ 2.4 & 77.5 $\pm$ 0.3          & 73.1 $\pm$ 0.5 & 75.5 $\pm$ 0.8                & 81.3 $\pm$ 0.4          & \textbf{84.1 $\pm$ 0.2} \\ \midrule
\multirow{4}{*}{DICE}    & Cora     & 74.9   $\pm$ 0.8 & 76.9 $\pm$ 0.9 & 79.6 $\pm$ 0.3          & 72.2 $\pm$ 1.4 & 75.6 $\pm$ 1.1                & \textbf{79.7 $\pm$ 0.8} & 78.9 $\pm$ 0.4          \\
                         & CiteSeer & 64.1 $\pm$ 0.5   & 66.0 $\pm$ 0.6 & 68.7 $\pm$ 0.5          & 62.6 $\pm$ 1.2 & 65.5 $\pm$ 1.1                & 68.9 $\pm$ 0.4          & \textbf{74.6 $\pm$ 0.4} \\
                         & PubMed   & 79.4 $\pm$ 0.4   & 78.3 $\pm$ 0.2 & \textbf{79.8 $\pm$ 0.4} & 76.6 $\pm$ 0.5 & 77.8 $\pm$ 0.7                & 79.2 $\pm$ 0.3          & 78.1 $\pm$ 0.6          \\
                         & CoraML   & 78.3 $\pm$ 0.6   & 77.5 $\pm$ 0.3 & 80.1 $\pm$ 0.4          & 58.7 $\pm$ 0.4 & 77.5 $\pm$ 0.2                & 80.5 $\pm$ 1.3          & \textbf{81.1 $\pm$ 0.8} \\ \bottomrule
\end{tabular}%
}
\end{table}

\textbf{Structural Perturbations.} Although our focus is on node feature-based adversarial attacks, we also provided a theoretical analysis for structural perturbations (Theorem \ref{theo:structural_perturbtation}). To evaluate the effectiveness of our approach in handling structural attacks, we conducted an empirical comparison to five baseline defense methods that focus on structural perturbations, GNN-Jaccard \citep{gnn_jaccard}, RobustGCN \citep{robustGCN}, GNN-SVD \citep{gnn_svd}, GNNGuard \citep{gnn_guard} and the Parseval Regularization(ParsevalR) \citep{parseval_cisse}. We considered three structural adversarial attacks with a perturbation budget $ 0.1 |\mathcal{E}|$: ``Mettack'' (with the ‘Meta-Self’ strategy) \citep{zugner2019adversarial}, ``Dice'' \citep{zugner2019adversarial} and ``PGD'' \citep{pgd_paper}. 
The average node classification accuracies of the considered models under attack are presented in Table~\ref{tab:structural_perturbations}. Our GCORN shows remarkable defense capabilities against these attacks, outperforming other methods in $8$ of $12$ experiments.
This result highlights the effectiveness of our proposed defense mechanism in enhancing the robustness of the underlying GCNs against structural perturbations while providing theoretical guarantees which are absent for the considered baselines.

\section{Conclusion}\label{sec:conclusion}
We have proposed GCORN, an adaptation of the GCN, which is robust against feature-based adversarial attacks. The approach utilizes the orthonormalization of weight matrices to control the robustness of GCNs via an upper bound we derived. 
Additionally, we have proposed a probabilistic evaluation method for the robustness of GNNs based on the introduced definition of \emph{Expected Adversarial Robustness}. Experimental results comparing our GCORN model to the standard GCN and existing defense methods show the superior performance of GCORN on different real-world datasets. 

\chapter[A Post-hoc Approach With Conditional Random Fields]{Rethinking Robustness in Graph Neural Networks: A Post-hoc Approach With Conditional Random Fields} \label{ch:CRF}
\lettrine[lines=3]{G}{\small{raph}} Neural Networks (GNNs), which are nowadays the benchmark approach in graph representation learning, have been shown to be vulnerable to adversarial attacks, raising concerns about their real-world applicability. While existing defense techniques primarily concentrate on the training phase of GNNs, involving adjustments to message passing architectures or pre-processing methods, there is a noticeable gap in methods focusing on increasing robustness during inference. In this context, this study introduces RobustCRF, a post-hoc approach aiming to enhance the robustness of GNNs at the inference stage. Our proposed method, founded on statistical relational learning using a Conditional Random Field, is model-agnostic and does not require prior knowledge about the underlying model architecture. We validate the efficacy of this approach across various models, leveraging benchmark node classification datasets. 

\section{Introduction}
Deep Neural Networks (DNNs) have demonstrated exceptional performance across various domains, including image recognition, language modeling, and speech recognition \citep{chen2019looks,minaee2024large,shim2021understanding}. The growing interest in handling irregular and unstructured data, particularly in domains like bioinformatics, has drawn significant attention to graph-based representations. Graphs have emerged as the preferred format for representing such irregular data due to their ability to capture interactions between elements, whether individuals in a social network or interactions between atoms. In response to this need, Graph Neural Networks (GNNs) \citep{Kipf:2017tc,velivckovic2017graph,xu2019powerful} have been proposed as an extension of DNNs tailored to graph-structured data. GNNs excel in the generation of meaningful representations for individual nodes by leveraging both a graph's structural information and its associated features. This approach has demonstrated significant success in addressing challenging applications in protein function prediction \citep{gilmer2017}, materials modeling \citep{pmlr-v202-duval23a},  time-varying data reconstruction, and recommendation systems \citep{wu2019session}.

Alongside their achievements, these deep learning-based approaches have exhibited vulnerability to various data alterations, including noisy, incomplete, or out-of-distribution examples \citep{gunnemann2022graph}. While such alterations may naturally occur in data collection, adversaries can deliberately craft and introduce them, resulting in adversarial attacks. These attacks manifest as imperceptible modifications to the input that can deceive and disrupt the classifier. In these attacks, the adversary's objective is to introduce subtle noise in the features or manipulate some edges in the graph structure to alter the initial prediction made on the input graph. Depending on the attacker's goals and level of knowledge, different attack settings can be considered. For instance, poisoning attacks \citep{zugner2019adversarial} involve manipulating the training data to introduce malicious data points, while evasion attacks \citep{dai2018adversarial} focus on attacking the model during the inference phase without further model adaptation.

Given the susceptibility of GNNs to adversarial attacks, their practical utility is constrained. Therefore, it becomes imperative to study and enhance their robustness. Various defense strategies have been proposed to mitigate this vulnerability, including pre-processing the input graph \citep{gnn_svd, gnn_jaccard}, edge pruning \citep{gnn_guard}, and adapting the message passing scheme \citep{robustGCN, airgnn, abbahaddou2024bounding}. Most of these defense methods operate by modifying the underlying model architecture or the training procedure, \ie~focusing on the training stage of GNNs and, therefore, limiting their applicability. 
Additionally, modifying the model architecture during training poses the risk of degrading accuracy on clean, non-attacked graphs, and usually, these modifications are not adapted for all possible architectures. 

In light of these challenges, our work introduces a post-hoc defense mechanism, denoted as RobustCRF, aimed at bolstering the robustness of GNNs during the \textit{inference phase} using statistical relational learning. RobustCRF is model-agnostic, requiring no prior knowledge of the underlying model, and is adaptable to various architectural designs, providing flexibility and applicability across diverse domains. Central to our approach is the assumption that neighboring points in the input manifold, accounting for graph isomorphism, should yield similar predictions in the output manifold. 
Based on our robustness assumption, we employ a Conditional Random Field (CRF) \citep{lafferty2001conditional} to adapt and edit the model's output to preserve the similarity relationship between the input and output manifold.

We start by introducing our CRF-based post-hoc robustness enhancement model. Recognizing the potential complexity associated with this task and aiming to deliver a robustness technique that remains computationally affordable, we study a sampling strategy for both the discrete space of the graph structure and the continuous space of node features. We finally proceed with an empirical analysis to assess the effectiveness of our proposed technique across various models, conducting also a comprehensive examination of the parameters involved.  In summary, our contributions can be outlined as follows. (1)~\textit{Model-agnostic robustness enhancement}: We present RobustCRF, a post-hoc approach designed to enhance the robustness of underlying GNNs, without any assumptions about the model's architecture.  (2)~\textit{Theoretical underpinnings and complexity reduction}: We conduct a comprehensive theoretical analysis of our proposed approach and enhance our general architecture through the incorporation of sampling techniques, thereby mitigating the complexity associated with the underlying model. (3)~\textit{Empirical Evaluation and Analysis:} We evaluate and compare our RobustCRF model to a wide array of baseline models on several node classification benchmark datasets and observe convincing performance of our RobustCRF approach. 

\section{Related Work}
\textbf{Attacking GNNs.} A multitude of poisoning and evasion adversarial attacks targeting GNNs models has surged lately \citep{gunnemann2022graph, zugner2018adversarial}. Gradient-based techniques \citep{pgd_paper}, such as Proximal Gradient Descent (PGD), have been employed to tackle the adversarial aim, framing it as an optimization task that seeks the closest adversarial example to the input while satisfying the adversarial objective. Building upon this foundation, Mettack \citep{zugner2019adversarial} extends the approach by expressing the problem as a bi-level optimization task and harnessing meta-gradients for its solution. Taking a different perspective, Nettack~\citep{zugner2018adversarial} introduces a targeted poisoning attack strategy, encompassing both structural and node feature perturbations, employing a greedy optimization algorithm to minimize an attack loss with respect to a surrogate model. Diverging from these classical search problems, \citet{dai2018adversarial} approach the adversarial search task using Reinforcement Learning techniques.

\textbf{Defending GNNs.} Different defense strategies have been proposed to counter the previously presented attacks on GNNs. Some of these defense methods employ low-rank approximation of the adjacency matrix to filter out noise \citep{gnn_svd,alchihabi2023efficient}, while similar pre-processing techniques are used to identify potential edge manipulations \citep{gnn_jaccard,deng2022garnet}. Additionally, methods like edge pruning \citep{gnn_guard} and transfer learning \citep{Tang_2020,yang2022not,tang2020transferring} have been used to mitigate the impact of poisoning attacks.
Notably, most research efforts have predominantly focused on addressing structural perturbations, with fewer strategies developed to counter attacks targeting node features. For instance, in \citet{seddik_rgcn}, the inclusion of node feature kernels in the message passing operator was proposed to increase GCN robustness. RobustGCN \citep{robustGCN} uses Gaussian distributions as hidden node representations in each convolutional layer, effectively mitigating the influence of both structural and feature-based adversarial attacks. Finally, in GCORN \citep{abbahaddou2024bounding}, an adaptation of the message passing scheme has been proposed by using orthonormal weight matrices to counter the effect of node feature-based adversarial attacks. Recent advances have further expanded the landscape of GNN defenses. For instance, self-supervised learning has been explored as a robust defense mechanism against adversarial perturbations \citep{yang2023selfsupervised}. These approaches continue to push the boundaries of GNN security by addressing both structural and feature-based vulnerabilities.

The majority of the previously discussed methods intervene during the model's training phase, necessitating modifications to the underlying architecture. However, this strategy exhibits certain limitations; it may not be universally applicable across diverse architectural designs, and it can potentially result in a loss of accuracy when applied to the clean graph. Furthermore, these methods may not be suitable for scenarios where users prefer to employ pre-trained models, as they necessitate model retraining.
Consequently, the need for proposing and crafting post-hoc robustness enhancements becomes increasingly apparent. Addressing this gap in the literature, our work aims to contribute to this essential area. One commonly employed post-hoc approach is randomized smoothing \citep{DBLP:journals/corr/abs-2008-12952,Carmon2019UnlabeledDI}, which involves injecting noise into the inputs at various stages and subsequently utilizing majority voting to determine the final prediction. Note that randomized smoothing has actually been initially borrowed from the optimization community \citep{duchi2012randomized}. Despite its popularity, randomized smoothing exhibits several limitations, including suffering from the ``shrinkage phenomenon'' where decision regions shrink or drift as the variance of the smoothing distribution increases \citep{mohapatra2021hidden}. Other works also identified that the smoothed classifier is \textit{more-constant} than the original model, \ie~it forces the classification to remain invariant over a large input space, resulting in a drop in the accuracy \citep{anderson2022certified,krishnan2020lipschitz,wangpretrain}.

\textbf{Conditional Random Fields.} A probabilistic graphical model (PGM) is a graph framework that compactly models the joint probability distributions $P$ and dependence relations over a set of random variables $ \widetilde{Y} = \{\widetilde{Y}_1, \ldots, \widetilde{Y}_m\}$ represented in a graph. The two most common classes of PGMs are Bayesian Networks (BNs) and Markov Random Fields (MRFs) \citep{heckerman2008tutorial,clifford1990markov}.  The core of the BN representation is a directed acyclic graph (DAG). In the DAG representation of a BN, each node in the DAG corresponds to a random variable and directed edges capture dependency relationships between these random variables. The direction of the edges determines the influence of one random variable on another. Similarly, MRFs are also used to describe dependencies between random variables in a graph. However, MRFs use undirected instead of directed edges and permit cycles.  An important assumption of MRFs is the \textit{Markov property}, i.e., for each pair of nodes $(a,b)$ that are not directly connected, \ie~ $e_{ab} \notin  E^{\text{\textit{MRF}}}$,  node $a$  is independent of  node $b$ conditioned on $a$'s neighbors, $$\forall a \in V^{\text{\textit{MRF}}} , ~~ \widetilde{Y}_a \perp  \widetilde{Y}_{V \setminus \mathcal{N}(a)} \mid \mathcal{N}(a) .$$

An important special case of MRFs arises when they are applied to model a conditional probability distribution $P (\widetilde{Y} \mid Y, V^{\text{\textit{MRF}}}, E^{\text{\textit{MRF}}})$, where $Y = \{Y_1, \ldots, Y_m\}$ are additional observed node features. These types of graphs are called Conditioned Random Fields (CRFs) \citep{sutton2012introduction,wallach2004conditional}. Formally,  a CRF is Markov network over random variables $\widetilde{Y}$ and observation $Y$, with the conditional distribution defined as follows:
\begin{multline*} 
P (\widetilde{Y} \mid Y, V^{\text{\textit{CRF}}}, E^{\text{\textit{CRF}}}) = \frac{1}{Z(Y, V^{\text{\textit{CRF}}}, E^{\text{\textit{CRF}}})}\times  \\ \exp\left \{ -\sum_{a\in V^{\text{\textit{CRF}}}} \phi_a(\widetilde{Y}_a)   - \sum_{(a,b) \in E^{\text{\textit{CRF}}}} \phi_{ab}(\widetilde{Y}_a, \widetilde{Y}_b) \right \},
\end{multline*}

where $Z(Y, V^{\text{\textit{CRF}}}, E^{\text{\textit{CRF}}})$ is the partition function, and $\phi_a(\widetilde{Y}_a)$, $\phi_{ab}(\widetilde{Y}_a, \widetilde{Y}_b)$ are potential functions contributed by each node $a$ and edge $(a,b)$. They are usually defined as simple linear functions or learned by simple regression on the features, e.g., with logistic regression.


\section{Proposed Method: RobustCRF} \label{crf:proposed_method}

In this section, we first formally define adversarial attacks on models processing attributed graphs and outline the common rationale followed in different robustness definitions. We then demonstrate how this rationale can be represented via a CRF. We further illustrate how we can use the mean field approximation and a sampling approach to fit a CRF to draw more robust inference from GNNs.

\begin{figure}[t]
    \centering 
\includegraphics[width=\textwidth]{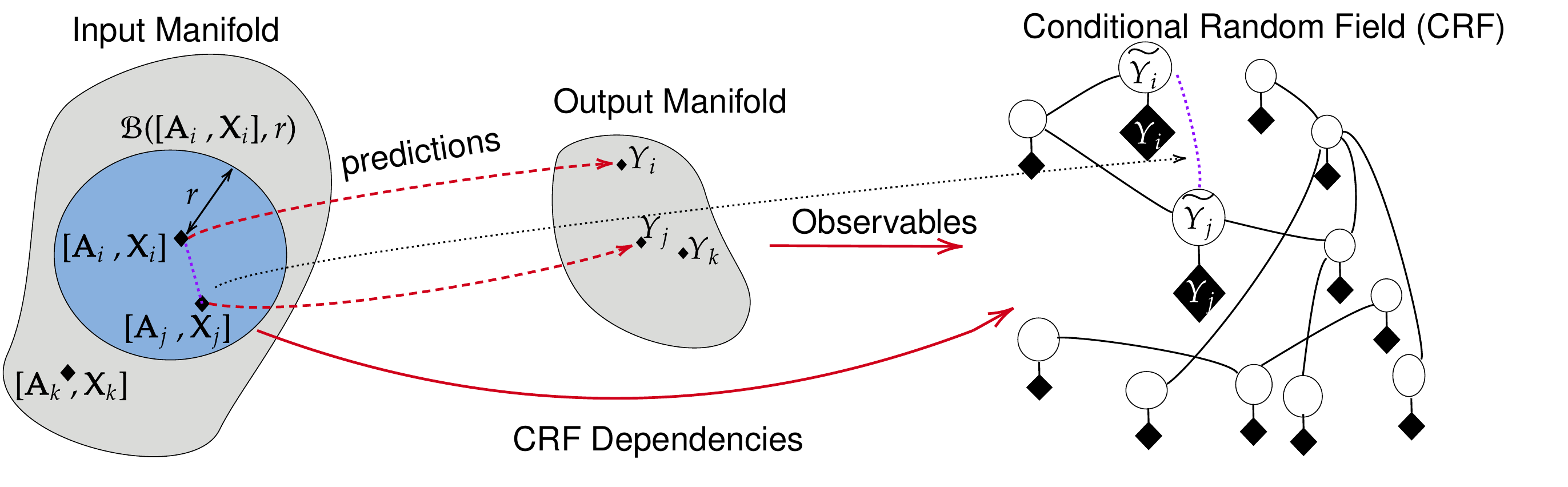}
    \caption[Illustration of our RobustCRF approach]{Illustration of our RobustCRF approach. We use the input graph manifold to generate the structure of the CRF, i.e., $V^{\text{\textit{CRF}}}, E^{\text{\textit{CRF}}}$. We use the GNN's predictions to generate the observables  $\left \{ Y_a:a \in V^{\text{\textit{CRF}}}  \right \}$, we then run the CRF inference to generate the new GNN's predictions $\{ \widetilde{Y}_a: a \in V^{\text{\textit{CRF}}} \}$.}
    \label{crf:fig:crf_construction}
\end{figure}

\subsection{Adversarial Attacks}

Let us consider our graph space ($\mathbb{A}, \lVert  \cdot \rVert_{\mathbb{A}}$), feature space ($\mathbb{X}$, $\lVert  \cdot \rVert_{\mathbb{X}}$), and the label space ($\mathbb{Y}, \lVert  \cdot \rVert_{\mathbb{Y}}$) to be measurable spaces. Given a GNN $f: (\mathbb{A},\mathbb{X}) \rightarrow \mathbb{Y}$, an input data point $[\mathbf{A},\mathbf{X} ]\in \mathbb{A}\times \mathbb{X}$ and its corresponding prediction $y \in \mathbb{Y},$ i.e., $f(\mathbf{A},\mathbf{X}) = y$, the goal of an adversarial attack is to produce a perturbed graph $[\widetilde{\mathbf{A}}, \widetilde{\mathbf{X}}]$ slightly different from the original graph with its predicted class being different from the predicted class of $[\mathbf{A},\mathbf{X} ]$. This could be formulated as finding a $ [\widetilde{\mathbf{A}}, \widetilde{\mathbf{X}}]$ with $f(\widetilde{\mathbf{A}}, \widetilde{\mathbf{X}})\neq y$ subject to $d([\mathbf{A},\mathbf{X} ], [\widetilde{\mathbf{A}}, \widetilde{\mathbf{X}}]) < \epsilon$, with $d$ being some distance function between the original and perturbed graphs. This could be a distance taking into account both the graph structure, in terms of the adjacency matrix, and the corresponding node features, as defined in Equation \ref{eq:norm_3}.

\subsection{Motivation}

There are different theoretical definitions of robustness \citep{DBLP:journals/corr/abs-2102-03716,weng2018evaluating} and the great majority, if not all, rely on one assumption:  \textit{If two inputs are adjacent in the input space, their predictions should be adjacent in the output space}. Many adversarial attack methods use this principle by attempting to find small perturbations that cause significant changes in the model's output, highlighting its vulnerability  \citep{wu2021performance,Goodfellow2014ExplainingAH}. Additionally, there are also works that use the neural networks distortion as a robustness metric \citep{DBLP:journals/corr/abs-2102-03716,weng2018evaluating,Carlini2016TowardsET}. Intuitively, a large distortion implies potentially poor adversarial robustness since a small perturbation applied to these inputs will lead to significant changes in the output. Consequently, most robustness metrics measure the extent to which the network's output is changed when perturbing the input data, indicating the network's vulnerability to adversarial attacks. To respect this common assumption, we construct a CRF, where its node set $V^{\text{\textit{CRF}}}$ represents the set of possible GNN inputs $\mathbb{A} \times \mathbb{X}$, while its edge set $E^{\text{\textit{CRF}}}$ represents the set of pair inputs  $ ( [\mathbf{A},\mathbf{X}],[\widetilde{\mathbf{A}}, \widetilde{\mathbf{X}}]  )$ such that $[\widetilde{\mathbf{A}}, \widetilde{\mathbf{X}}] $ belongs to the ball $\mathcal{B}$  of radius $r>0$ surrounding $[\mathbf{A}, \mathbf{X}]$, as follows,
$$\mathcal{B}\left ( \left [ \mathbf{A}, \mathbf{X} \right ],r\right ) = \left \{ \left [ \widetilde{\mathbf{A}}, \widetilde{\mathbf{X}} \right ]  :   d^{\alpha, \beta}([\mathbf{A}, \mathbf{X}], [\widetilde{\mathbf{A}}, \widetilde{\mathbf{X}}] ) \leq r \right \}. $$

\subsection{Modeling the Robustness Constraint with a CRF}

Let $Y_a$ be the output prediction of the trained GNN $f$ on the input $a= [\mathbf{A},\mathbf{X}] \in V^{\text{\textit{CRF}}}$. The main goal of using a CRF is to update the predictions $Y = \{ Y_a : a\in V^{\text{\textit{CRF}}}\}$ into new predictions $\widetilde{Y}=\{ \widetilde{Y}_a: a\in V^{\text{CRF}}\} $ that respect the \textit{robustness assumption}. To do so, we model the relation between the two predictions $Y$ and $\widetilde{Y}$ using a CRF, maximizing the following conditional probability
\begin{multline}\label{crf:eq:def_crf}
         P (\widetilde{Y} \mid Y, V^{\text{\textit{CRF}}}, E^{\text{\textit{CRF}}}) = \frac{1}{Z} \exp\bigg\{ - \sum_{a\in V^{\text{\textit{CRF}}}} \phi_a(\widetilde{Y}_a, Y_a)  \\ - \sum_{(a,b) \in E^{\text{\textit{CRF}}}} \phi_{ab}(\widetilde{Y}_a, \widetilde{Y}_b) \bigg\}, 
\end{multline}
where $Z$ is the partition function, $\phi_a(\widetilde{Y}_a, Y_a) $ and $\phi_{ab}(\widetilde{Y}_a, \widetilde{Y}_b)$ are potential functions contributed by each CRF node $a$ and each CRF edge $(a,b)$ and usually defined as simple linear functions or learned by simple regression on the features (e.g., logistic regression). The potential functions $\phi_a$ and $\phi_{ab}$ can be either fixed or trainable to optimize a specific objective. In this paper, we define the potential functions as follows:
\begin{align*}
   \phi_a(\widetilde{Y}_a, Y_a) & = \sigma \lVert \widetilde{Y}_a- Y_a\rVert_2^2, \\ 
    \phi_{ab}(\widetilde{Y}_a, \widetilde{Y}_b)  &= (1-\sigma) g_{ab} \lVert \widetilde{Y}_a- \widetilde{Y}_b\rVert_2^2,
\end{align*}
where $g_{ab}$ denotes a chosen similarity term of two inputs $a=[\mathbf{A},\mathbf{X}]$ and $b=[\widetilde{\mathbf{A}}, \widetilde{\mathbf{X}}]$, e.g., the cosine similarity between their feature matrices. The parameter $\sigma$ is used to adjust the importance of the two potential functions. 
Representing GNN inputs in the CRF can be seen as a constrained problem to enhance the robustness: Finding a new prediction $\widetilde{Y}_a$ that satisfies our robustness assumption and, at the same time, stays as close as possible to the original prediction of the GNN, $Y_a = f(a)$ where $a=[\mathbf{A},\mathbf{X}]$ is a GNN input.  In Figure \ref{crf:fig:crf_construction}, we illustrate the main idea of the proposed RobustCRF model.

Now that we have defined the CRF and its potential functions, generating the new prediction $\{ \widetilde{Y}_p ~~:~~ p\in V^{\text{\textit{CRF}}}\} $ is intractable for two reasons. First, the partition function $Z$ is usually intractable. Second, the CRF distribution $ P$ represents a potentially infinite collection of CRF edges $E^{\text{\textit{CRF}}}$. We need, therefore, to derive a tractable algorithm to generate the smoothed prediction $\widetilde{Y}$. In what follows, we show how to overcome these two challenges.

\subsection{Mean Field Approximation}
We aim to derive the most likely $\widetilde{Y}$ from the initial distribution $P$ defined in  \eqref{crf:eq:def_crf}, as,
\begin{equation} \label{crf:eq:initial_goal}
  \widetilde{Y}^* = \argmax\limits_{\widetilde{Y} \in \mathbb{Y} }   P (\widetilde{Y} \mid Y, V^{\text{\textit{CRF}}}, E^{\text{\textit{CRF}}}).
\end{equation}
Since the inference is intractable, we use a variational inference method where we propose a family of densities $\mathcal{D}$ and find a member $Q^* \in \mathcal{D} $ which is close to the posterior $P (\widetilde{Y} \mid Y, V^{\text{\textit{CRF}}}, E^{\text{\textit{CRF}}})$. Thus, we approximate the initial task in \eqref{crf:eq:initial_goal} with a new objective,
\begin{equation}\label{crf:eq:new_objective}
    \centering
    \left\{
    \begin{array}{ll}
     \widetilde{Y}^* = \argmax\limits_{\widetilde{Y}\in \mathbb{Y} }   Q^* (\widetilde{Y} ), \\ 
     Q^* = \argmin\limits_{Q\in\mathcal{D}}\mathcal{KL}\left ( Q \mid P  \right ),
        \end{array}
    \right.
\end{equation}
where $\mathcal{KL}\left ( \cdot \mid \cdot  \right )$
denotes the Kullback-Leibler divergence. The goal is to find the distribution $Q^* $ within the family $\mathcal{D}$, which is the closest to the initial distribution $P$.  In this work, we used the \textit{mean-field approximation} that enforces full independence among all latent variables \citep{wang2013variational}. Mean field approximation is a powerful technique for simplifying complex probabilistic models, widely used in various fields of machine learning and statistics \citep{andrews2017bayesian}. Thus,  the variational distribution over the latent variables factorizes as,
\begin{equation} 
    \forall Q\in\mathcal{D}, ~~ Q(\widetilde{Y} ) = \prod_{a\in V^{\text{\textit{CRF}}}} Q_a(\widetilde{Y}_a ).
\end{equation}
The exact formula of the optimal surrogate distribution $Q$ can be obtained using Lemma \ref{crf:lem:mean_field_expected}.  
\begin{lemma}
\label{crf:lem:mean_field_expected}
By solving the system of \eqref{crf:eq:new_objective}, we can get the optimal distribution $Q^*$ for each $a \in V^{\text{\textit{CRF}}}$ as follows,
 \begin{equation}\label{crf:eq:lema_eq}
Q_a(\widetilde{Y}_a) \propto \exp \left \{  \mathbb{E}_{-a}  \left [ \log P\left ( \widetilde{Y} \mid Y, V^{\text{\textit{CRF}}}, E^{\text{\textit{CRF}}}  \right ) \right ]  \right \},
 \end{equation} 
 where $\mathbb{E}_{-a}$  denotes the expectation taken over the random variables $\widetilde{Y}_{-a}$, corresponding to all nodes except $a$,
\end{lemma}
The proof of Lemma \ref{crf:lem:mean_field_expected} is provided in Appendix \ref{crf:appendix:crf_neighbors}. 
We will use Coordinate Ascent Inference (CAI), iteratively optimizing each variational distribution and holding the others fixed.
The CAI iteratively updates each $Q_a(\widetilde{Y}_a)$. The evidence lower bound (ELBO) converges to a local minimum. Using \eqref{crf:eq:def_crf} and \eqref{crf:eq:lema_eq}, we get the optimal surrogate distribution as follows,
\begin{equation*}
  Q_a(\widetilde{Y}_a) \propto \exp \Bigg \{  \sigma \lVert \widetilde{Y}_a- Y_a\rVert_2^2 + (1-\sigma) \sum_{b \text{ s.t. }(a,b)\in E^{\text{\textit{CRF}}}}  g_{ab} \lVert \widetilde{Y}_a- \widetilde{Y}_b\rVert_2^2 \Bigg \}. 
\end{equation*}
Thus, for each GNN input $a\in V^{\text{\textit{CRF}}}$, $Q_a(\widetilde{Y}_a)$ is a Gaussian distribution that reaches the highest probability at its expectation,
\begin{equation*}
\argmax_{\widetilde{Y}_a\in \mathbb{Y}} Q_a(\widetilde{Y}_a) = \frac{   \sigma Y_a +   (1-\sigma) \sum\limits_{b \text{ s.t. }(a,b)\in E^{\text{\textit{CRF}}}}  g_{ab} \widetilde{Y}_b}{ \sigma  +   (1-\sigma) \sum\limits_{b \text{ s.t. }(a,b)\in E^{\text{\textit{CRF}}}}  g_{ab} }.
\end{equation*}
Using the CAI algorithm, we can thus utilize the following update rule: 
\begin{equation} \label{crf:eq:update_rule}
 \widetilde{Y}_a^{k+1} = \frac{ \sigma Y_a + (1-\sigma)   \sum\limits_{b \text{ s.t. }(a,b)\in E^{\text{\textit{CRF}}}} g_{ab} \widetilde{Y}_b^{k} }{\sigma  + (1-\sigma)  \sum\limits_{b \text{ s.t. }(a,b)\in E^{\text{\textit{CRF}}}}  g_{ab} }.
\end{equation}

\subsection{Reducing the Size of the CRF} 
The number of possible inputs is usually very large, or infinite, for discrete data and infinite for continuous data. This could make the inference intractable due to the potential large size of $E^{\text{\textit{CRF}}}$. Therefore, instead of considering all the possible CRF neighbors of an input $a$, i.e., all inputs $b\in V^{\text{\textit{CRF}}}$ such that $d^{\alpha, \beta}(a,b) \leq r$, we can consider only a subset of $L$ neighbors by randomly sampling from the CRF neighbors of $a$. The update rule in \eqref{crf:eq:update_rule} can then be expressed as,
\begin{equation} \label{crf:eqn:UpdateRule}
    \widetilde{Y}_a^{k+1} = \frac{ \sigma Y_a + (1-\sigma)   \sum\limits_{b \in \mathcal{U}^{L}(a)} g_{ab} \widetilde{Y}_b^{k} }{\sigma  + (1-\sigma)   \sum\limits_{b \in \mathcal{U}^{L}(a)} g_{ab} },
\end{equation}
where $\mathcal{U}^{L}(a)$ denotes a set of $L$ randomly sampled CRF neighbors of a graph $[\mathbf{A},\mathbf{X}]$. 

\begin{remark}
If in \eqref{crf:eqn:UpdateRule} we set the value of $\sigma$ to $0$, the number of iterations to $1$, and all the similarity coefficients $g_{ab}$ to $1$, this scheme corresponds to the \textit{standard randomized smoothing} with uniform distribution. In this setting, we do not take into account the original classification task, which causes a huge drop in the clean accuracy. Therefore, RobustCRF is a generalization of the randomized smoothing that gives a better trade-off between accuracy and robustness.
\end{remark}

Below, we elaborate on how to uniformly sample a neighbor graph $b$ surrounding $a$ in the structural distances, i.e., $(\alpha, \beta) = (1,0)$.

Let $\mathbb{A} = \{0,1\}^{n^2}$ be the adjacency matrix space, where $n$ is the number of nodes. $\mathbb{A}$ is a finite-dimensional compact normed vector space, so all the norms are equivalent. Thus, all the induced $L^p$ distances are equivalent. Without loss of generality, we can consider the $L^1$ loss, which exactly corresponds to the \textit{Hamming} distance,
$$d^{1}([\mathbf{A},\mathbf{X}],[\widetilde{\mathbf{A}},\widetilde{\mathbf{X}}])  = \sum_{i\leq j} | \mathbf{A}^{i,j} - \widetilde{\mathbf{A}}^{i,j}|,$$
where $\mathbf{A},\widetilde{\mathbf{A}}$ correspond to the adjacency matrices of graphs $G, \widetilde{G},$ respectively.
Since we consider undirected graphs, the Hamming distance takes only discrete values in $\{0,\ldots, \frac{n(n+1)}{2}\}$. In Lemma \ref{crf:lem:size_CRF}, we provide a lower bound for the size of CRF neighbors $\mathcal{N}^{\text{\textit{CRF}}}(a) =\{b\in V^{\text{\textit{CRF}}}~~:~~ (a,b) \in E^{\text{\textit{CRF}}}\}$ for any GNN input $a \in V^{\text{\textit{CRF}}}$.
\begin{lemma}
\label{crf:lem:size_CRF}
For any integer $r$ in  $\{0,\ldots, \frac{n(n+1)}{2}\}$, the number of CRF neighboors  $\left | \mathcal{N}^{\text{\textit{CRF}}}(a)   \right |$ for any $a\in V^{\text{\textit{CRF}}}$, i.e., the set of graphs $[\widetilde{\mathbf{A}}, \widetilde{\mathbf{X}}]$ with a Hamming distance smaller or equal than $r$, for each $a \in V^{\text{\textit{CRF}}}$, we have the following lower bound: 
\begin{equation}
 \frac{2^{H(\epsilon)n(n+1)/2}}{\sqrt{4n(n+1)\epsilon (1-\epsilon)}} \leq \left | \mathcal{N}^{\text{\textit{CRF}}}(a)   \right |,
\end{equation}
where $0\leq \epsilon=\frac{2r}{n(n+1)}\leq 1$ and $H(\cdot)$ is the binary entropy function, i.e., 
$H(\epsilon) = -\epsilon \log_2(\epsilon) - (1-\epsilon) \log_2(1-\epsilon).$
\end{lemma}
In Appendix \ref{crf:appendix:proof_lema}, we present the proof of Lemma \ref{crf:lem:size_CRF} and we empirically analyze the asymptotic behavior of this lower bound, motivating, thus, the need for sampling strategies to reduce the size of the CRF.

\begin{table*}[t]
\caption[Attacked classification accuracy of RobustCRF]{Attacked classification accuracy ($\pm$ standard deviation) of the GCN, the baselines and the proposed RobustCRF on different benchmark node classification datasets after the features-based attack. \textcircled{1}~Test accuracy on the original features; \textcircled{2}~Test accuracy on the perturbed features.}
\centering
\resizebox{\textwidth}{!}{
\begin{tabular}{l|llccccc}
\toprule
& Attack & Model  & Cora & CiteSeer  & PubMed   &  CS & Texas \\ 
\midrule
\multirow{5}{*}{\textcircled{1}} 
& \multirow{4}{*}{Clean} 
  & GCN       
    & $80.66 {\scriptstyle \pm 0.41}$ 
    & $70.37 {\scriptstyle \pm 0.53}$ 
    & $78.16 {\scriptstyle \pm 0.67}$ 
    & $89.15 {\scriptstyle \pm 2.06}$ 
    & $51.35 {\scriptstyle \pm 18.53}$ \\

& 
  & RGCN      
    & $77.64 {\scriptstyle \pm 0.52}$ 
    & $69.88 {\scriptstyle \pm 0.47}$ 
    & $75.58 {\scriptstyle \pm 0.65}$ 
    & $\mathbf{92.05 {\scriptstyle \pm 0.72}}$ 
    & $51.62 {\scriptstyle \pm 13.91}$ \\

& 
  & GCORN     
    & $77.83 {\scriptstyle \pm 2.33}$ 
    & $\mathbf{71.68 {\scriptstyle \pm 1.54}}$ 
    & $76.03 {\scriptstyle \pm 1.29}$ 
    & $88.94 {\scriptstyle \pm 1.86}$ 
    & $59.73 {\scriptstyle \pm 4.90}$ \\

& 
  & NoisyGCN  
    & $\mathbf{81.04 {\scriptstyle \pm 0.74}}$ 
    & $70.36 {\scriptstyle \pm 0.79}$ 
    & $78.13 {\scriptstyle \pm 0.53}$ 
    & $91.47 {\scriptstyle \pm 0.92}$ 
    & $48.91 {\scriptstyle \pm 20.02}$ \\

& 
  & RobustCRF 
    & $80.63 {\scriptstyle \pm 0.38}$ 
    & $70.30 {\scriptstyle \pm 0.43}$ 
    & $\mathbf{78.20 {\scriptstyle \pm 0.24}}$ 
    & $88.16 {\scriptstyle \pm 3.41}$ 
    & $\mathbf{61.08 {\scriptstyle \pm 5.01}}$ \\ 

\midrule

\multirow{10}{*}{\textcircled{2}} 
& \multirow{5}{*}{\begin{tabular}[c]{@{}l@{}}Random \\ ($\psi=0.5$)\end{tabular}} 
  & GCN       
    & $77.88 {\scriptstyle \pm 0.90}$ 
    & $66.65 {\scriptstyle \pm 1.00}$ 
    & $73.60 {\scriptstyle \pm 0.75}$ 
    & $88.92 {\scriptstyle \pm 2.04}$ 
    & $46.49 {\scriptstyle \pm 15.75}$ \\

& 
  & RGCN      
    & $67.61 {\scriptstyle \pm 0.80}$ 
    & $59.76 {\scriptstyle \pm 1.01}$ 
    & $61.93 {\scriptstyle \pm 1.18}$ 
    & $90.74 {\scriptstyle \pm 1.08}$ 
    & $43.51 {\scriptstyle \pm 10.22}$ \\

& 
  & GCORN     
    & $76.28 {\scriptstyle \pm 1.96}$ 
    & $67.82 {\scriptstyle \pm 2.18}$ 
    & $72.35 {\scriptstyle \pm 1.43}$ 
    & $88.31 {\scriptstyle \pm 2.10}$ 
    & $60.00 {\scriptstyle \pm 4.95}$ \\

& 
  & NoisyGCN  
    & $78.59 {\scriptstyle \pm 1.09}$ 
    & $66.83 {\scriptstyle \pm 0.98}$ 
    & $73.60 {\scriptstyle \pm 0.58}$ 
    & $\mathbf{91.04 {\scriptstyle \pm 0.85}}$ 
    & $48.91 {\scriptstyle \pm 19.29}$ \\

& 
  & RobustCRF 
    & $\mathbf{78.28 {\scriptstyle \pm 0.68}}$ 
    & $\mathbf{68.23 {\scriptstyle \pm 0.57}}$ 
    & $\mathbf{74.37 {\scriptstyle \pm 0.47}}$ 
    & $88.30 {\scriptstyle \pm 3.25}$ 
    & $\mathbf{57.30 {\scriptstyle \pm 4.32}}$ \\

\cline{2-8}

& \multirow{5}{*}{PGD} 
  & GCN       
    & $76.38 {\scriptstyle \pm 0.72}$ 
    & $67.57 {\scriptstyle \pm 0.77}$ 
    & $74.86 {\scriptstyle \pm 0.65}$ 
    & $86.90 {\scriptstyle \pm 1.91}$ 
    & $52.97 {\scriptstyle \pm 19.19}$ \\

& 
  & RGCN      
    & $68.45 {\scriptstyle \pm 0.97}$ 
    & $64.63 {\scriptstyle \pm 0.82}$ 
    & $73.35 {\scriptstyle \pm 0.81}$ 
    & $\mathbf{90.76 {\scriptstyle \pm 0.68}}$ 
    & $60.81 {\scriptstyle \pm 10.27}$ \\

& 
  & GCORN     
    & $73.32 {\scriptstyle \pm 2.19}$ 
    & $\mathbf{69.05 {\scriptstyle \pm 2.50}}$ 
    & $74.49 {\scriptstyle \pm 1.13}$ 
    & $87.07 {\scriptstyle \pm 2.96}$ 
    & $59.73 {\scriptstyle \pm 4.90}$ \\

& 
  & NoisyGCN  
    & $76.29 {\scriptstyle \pm 1.69}$ 
    & $67.09 {\scriptstyle \pm 1.50}$ 
    & $75.04 {\scriptstyle \pm 0.53}$ 
    & $88.79 {\scriptstyle \pm 0.85}$ 
    & $52.97 {\scriptstyle \pm 19.11}$ \\

& 
  & RobustCRF 
    & $\mathbf{76.41 {\scriptstyle \pm 0.70}}$ 
    & $67.90 {\scriptstyle \pm 0.63}$ 
    & $\mathbf{75.17 {\scriptstyle \pm 0.91}}$ 
    & $85.60 {\scriptstyle \pm 2.75}$ 
    & $\mathbf{62.16 {\scriptstyle \pm 4.83}}$ \\

\bottomrule
\end{tabular}
}
\label{crf:tab:features_results}
\end{table*}

\begin{table*}[t]

\caption[Attacked classification accuracy of RobustCRF when combined with other approaches]{Attacked classification accuracy ($\pm$ standard deviation) of the baselines when combined with the proposed RobustCRF on different benchmark node classification datasets after the features-based attack application.}
\centering
\resizebox{\textwidth}{!}{
\begin{tabular}{l|lcccc}
\toprule
Attack & Model  & Cora & CiteSeer  & PubMed    & Texas \\ 
\midrule

\multirow{6}{*}{Clean} 
  & RGCN 
    & $77.64 {\scriptstyle \pm 0.52}$ 
    & $\mathbf{69.88 {\scriptstyle \pm 0.47}}$ 
    & $\mathbf{75.58 {\scriptstyle \pm 0.65}}$ 
    & $51.62 {\scriptstyle \pm 13.91}$ \\

& RGCN w/ RobustCRF  
    & $\mathbf{77.70 {\scriptstyle \pm 0.46}}$  
    & $69.84 {\scriptstyle \pm 0.39}$ 
    & $75.50 {\scriptstyle \pm 0.60}$ 
    & $\mathbf{52.16 {\scriptstyle \pm 14.20}}$ \\  
    \cdashline{2-6}
& GCORN    
    & $77.83 {\scriptstyle \pm 2.33}$ 
    & $71.68 {\scriptstyle \pm 1.54}$ 
    & $76.03 {\scriptstyle \pm 1.29}$ 
    & $59.73 {\scriptstyle \pm 4.90}$ \\

& GCORN w/ RobustCRF  
    & $\mathbf{78.50 {\scriptstyle \pm 1.17}}$ 
    & $\mathbf{71.72 {\scriptstyle \pm 1.46}}$ 
    & $\mathbf{76.13 {\scriptstyle \pm 1.08}}$ 
    & $\mathbf{59.77 {\scriptstyle \pm 4.68}}$ \\  
    \cdashline{2-6}
& NoisyGCN    
    & $81.04 {\scriptstyle \pm 0.74}$ 
    & $\mathbf{70.36 {\scriptstyle \pm 0.79}}$ 
    & $78.13 {\scriptstyle \pm 0.53}$ 
    & $48.91 {\scriptstyle \pm 20.02}$ \\  

& NoisyGCN w/ RobustCRF 
    & $\mathbf{81.07 {\scriptstyle \pm 0.70}}$ 
    & $70.20 {\scriptstyle \pm 0.85}$ 
    & $\mathbf{78.90 {\scriptstyle \pm 0.46}}$ 
    & $\mathbf{49.18 {\scriptstyle \pm 19.59}}$ \\  
\midrule

\multirow{6}{*}{\begin{tabular}[c]{@{}l@{}}Random \\ ($\psi =0.5$)\end{tabular}} 
  & RGCN 
    & $67.61 {\scriptstyle \pm 0.80}$ 
    & $59.76 {\scriptstyle \pm 1.01}$ 
    & $61.93 {\scriptstyle \pm 1.18}$ 
    & $43.51 {\scriptstyle \pm 10.22}$ \\  

& RGCN w/ RobustCRF    
    & $\mathbf{69.08 {\scriptstyle \pm 0.73}}$ 
    & $\mathbf{60.04 {\scriptstyle \pm 1.01}}$ 
    & $\mathbf{63.05 {\scriptstyle \pm 0.88}}$ 
    & $\mathbf{45.05 {\scriptstyle \pm 9.99}}$ \\  
    \cdashline{2-6}
& GCORN 
    & $76.28 {\scriptstyle \pm 1.96}$ 
    & $67.82 {\scriptstyle \pm 2.18}$ 
    & $\mathbf{72.35 {\scriptstyle \pm 1.43}}$ 
    & $\mathbf{60.00 {\scriptstyle \pm 4.95}}$ \\  

& GCORN w/ RobustCRF    
    & $\mathbf{77.21 {\scriptstyle \pm 1.17}}$ 
    & $\mathbf{68.94 {\scriptstyle \pm 3.00}}$  
    & $72.29 {\scriptstyle \pm 1.29}$  
    & $59.18 {\scriptstyle \pm 3.71}$ \\  
    \cdashline{2-6}
& NoisyGCN    
    & $78.59 {\scriptstyle \pm 1.09}$ 
    & $66.83 {\scriptstyle \pm 0.98}$ 
    & $73.60 {\scriptstyle \pm 0.58}$ 
    & $\mathbf{48.91 {\scriptstyle \pm 19.29}}$ \\  

& NoisyGCN w/ RobustCRF    
    & $\mathbf{81.07 {\scriptstyle \pm 0.99}}$  
    & $\mathbf{67.04 {\scriptstyle \pm 1.38}}$ 
    & $\mathbf{74.18 {\scriptstyle \pm 0.98}}$ 
    & $45.94 {\scriptstyle \pm 14.54}$ \\  
\midrule

\multirow{6}{*}{PGD}                                                             
  & RGCN 
    & $68.45 {\scriptstyle \pm 0.97}$ 
    & $\mathbf{64.63 {\scriptstyle \pm 0.82}}$ 
    & $73.35 {\scriptstyle \pm 0.81}$  
    & $\mathbf{60.81 {\scriptstyle \pm 10.27}}$ \\  

& RGCN w/ RobustCRF    
    & $\mathbf{68.46 {\scriptstyle \pm 0.93}}$ 
    & $64.59 {\scriptstyle \pm 0.84}$ 
    & $\mathbf{73.47 {\scriptstyle \pm 0.72}}$ 
    & $58.10 {\scriptstyle \pm 11.09}$ \\  
    \cdashline{2-6}

& GCORN 
    & $73.32 {\scriptstyle \pm 2.19}$ 
    & $69.05 {\scriptstyle \pm 2.50}$ 
    & $74.49 {\scriptstyle \pm 1.13}$ 
    & $59.73 {\scriptstyle \pm 4.90}$ \\  

& GCORN w/ RobustCRF    
    & $\mathbf{73.65 {\scriptstyle \pm 1.55}}$ 
    & $\mathbf{69.09 {\scriptstyle \pm 2.57}}$ 
    & $\mathbf{74.59 {\scriptstyle \pm 0.90}}$ 
    & $\mathbf{60.27 {\scriptstyle \pm 4.68}}$ \\  
    \cdashline{2-6}
& NoisyGCN    
    & $76.29 {\scriptstyle \pm 1.69}$ 
    & $67.09 {\scriptstyle \pm 1.50}$ 
    & $75.04 {\scriptstyle \pm 0.53}$ 
    & $52.97 {\scriptstyle \pm 19.11}$ \\  

& NoisyGCN w/ RobustCRF  
    & $\mathbf{76.48 {\scriptstyle \pm 1.65}}$ 
    & $\mathbf{67.21 {\scriptstyle \pm 1.35}}$ 
    & $\mathbf{75.39 {\scriptstyle \pm 0.45}}$ 
    & $\mathbf{53.51 {\scriptstyle \pm 19.07}}$ \\  

\bottomrule
\end{tabular}
}
\label{crf:tab:combinations}
\end{table*}

To sample from $\mathcal{B}\left ( \left [ \mathbf{A}, \mathbf{X} \right ],r\right )$, we use a  \textit{stratified sampling} strategy. First, we sample a distance value $d$ from  $\{ 0,1,\ldots,n^2\}$. To do so, we partition $\mathcal{B}\left ( \left [ \mathbf{A}, X \right ],r\right )$ with respect to their distance to the original adjacency matrix $\mathbf{A}$, 
    \begin{equation*}
        \left\{\begin{array}{l}
    S_d= \{\widetilde{\mathbf{A}} \in \mathbb{A} :  d^1(\mathbf{A},\widetilde{\mathbf{A}}) = d \},\\ 
    \mathcal{B}( [ \mathbf{A}, \mathbf{X} ],r ) = \bigcup_{d\leq r}S_d, \\ 
    \forall d\neq d' , S_d\cap S_{d'} = \varnothing  . 
    \end{array} \right.
    \end{equation*}
    To each distance value $d$, we assign the portion of graphs covered by $S_d\left ( \mathbf{A} \right ) $ in $\mathbb{A}$ as
    $$\forall d\in \{0,\ldots,r\}, p(d)=\binom{r}{d}\frac{1}{R},$$ 
    where $R = \sum_{d=0}^{r} \binom{r}{d} = 2^r.$ Second, we uniformly select $d$ positions in the adjacency matrix to be modified and change the element of the $d$ positions by changing the value $\mathbf{A}^{i,j}$ to $1 - \mathbf{A}^{i,j}$. To uniformly sample neighbor graphs when dealing with feature-based distances for graphs, we used the sampling strategy of \citet{abbahaddou2024bounding}. 

RobustCRF is an attack-independent and model-agnostic robustness approach based on uniform sampling without requiring any training. Therefore, it can be used to enhance GNNs' robustness against unknown attack distributions. In Section \ref{crf:sec:empric}, we will experimentally validate this theoretical insight for the node classification task, demonstrating that RobustCRF achieves a good trade-off between robustness and clean accuracy, \ie~the model's initial performance on a clean un-attacked dataset. Moreover, the approach of our work and these baselines are fundamentally different, since our RobustCRF is post-hoc, we can also use the baselines in combination with our proposed RobustCRF approach to make even more robust predictions. We report the results of this experiment in Table \ref{crf:tab:combinations}.

An advantage of RobustCRF is that it allows flexibility in defining the CRF structure based on different criteria. For instance, we can construct the CRF by considering the worst-case scenario, where the CRF neighbors correspond to the most adversarial perturbations within a given threat model. However,  instead of evaluating the model’s behavior only under specific adversarial attacks, our method examines its performance more generally within a defined neighborhood. This perspective leans towards a concept of \textit{average} robustness, expanding on the conventional worst-case based adversarial robustness that is often emphasized in adversarial studies \citep{abbahaddou2024bounding}. A similar average robustness concept was studied and showed to be appropriate for computer vision \citep{rice2021robustness}.

\section{Experimental Results}\label{crf:sec:empric}

In this section, we shift from theoretical exploration to practical validation by evaluating the effectiveness of RobustCRF on real-world benchmark datasets. Our primary experimental objective is to assess how well the proposed method enhances the robustness of a trained GNN in a node classification task. Details about the datasets used in our experiments and implementation details can be found in Appendix \ref{crf:app:empric}.


\textbf{Attacks.} We evaluate RobustCRF via the \textit{Attack Success Rate} (ASR), the percentage of attack attempts that produce successful adversarial examples. For the feature-based attacks, we consider two main types: (1) we first consider a random attack which consists of injecting Gaussian noise $\mathcal{N}(0, \psi \mathbf{I})$ to the features with scaling parameter $\psi=0.5$; (2) we have additionally used the white-box Proximal Gradient Descent \citep{pgd_paper}, which is a gradient-based approach to the adversarial optimization task for which we set the perturbation rate to 15\%. For the structural perturbations, we evaluated RobustCRF using the ``Dice'' adversarial attack in a black-box setting, where we consider a surrogate model \citep{zugner2019adversarial}. For this setting, we used an attack budget of $10\%$ (the ratio of perturbed edges).

\begin{table*}[t]
\caption[Attacked classification accuracy of RobustCRF with respect to structural attacks]{Attacked classification accuracy ($\pm$ standard deviation) of the GCN, the baselines and the proposed RobustCRF on different benchmark node classification datasets after the structural attack application. \textcircled{1}~Test accuracy on the original structure, \textcircled{2}~Test accuracy on the perturbed structure.}
\centering
\resizebox{0.75\textwidth}{!}{
\begin{tabular}{l|llcccc}
\toprule
& Attack & Model  & Cora & CoraML  & CiteSeer   &  PolBlogs  \\ 
\midrule
\multirow{7}{*}{\textcircled{1}} 
& \multirow{6}{*}{Clean} 
  & GCN       
    & $83.42 {\scriptstyle \pm 1.00}$ 
    & $85.60 {\scriptstyle \pm 0.40}$ 
    & $70.66 {\scriptstyle \pm 1.18}$ 
    & $95.16 {\scriptstyle \pm 0.64}$ \\

& 
  & RGCN      
    & $83.46 {\scriptstyle \pm 0.53}$ 
    & $85.61 {\scriptstyle \pm 0.61}$ 
    & $72.18 {\scriptstyle \pm 0.97}$ 
    & $95.32 {\scriptstyle \pm 0.76}$ \\

& 
  & GCNGuard  
    & $\mathbf{83.72 {\scriptstyle \pm 0.67}}$ 
    & $85.54 {\scriptstyle \pm 0.42}$ 
    & $73.18 {\scriptstyle \pm 2.36}$ 
    & $95.07 {\scriptstyle \pm 0.51}$ \\

& 
  & GCNSVD    
    & $77.96 {\scriptstyle \pm 0.61}$ 
    & $81.29 {\scriptstyle \pm 0.51}$ 
    & $68.16 {\scriptstyle \pm 1.15}$ 
    & $93.80 {\scriptstyle \pm 0.73}$ \\

& 
  & GCNJaccard 
    & $82.20 {\scriptstyle \pm 0.67}$ 
    & $84.85 {\scriptstyle \pm 0.39}$ 
    & $\mathbf{73.57 {\scriptstyle \pm 1.21}}$ 
    & $51.81 {\scriptstyle \pm 1.49}$ \\

& 
  & GOOD-AT  
    & $83.43 {\scriptstyle \pm 0.11}$ 
    & $84.87 {\scriptstyle \pm 0.15}$ 
    & $72.80 {\scriptstyle \pm 0.45}$ 
    & $94.85 {\scriptstyle \pm 0.52}$ \\

& 
  & RobustCRF 
    & $83.52 {\scriptstyle \pm 0.04}$ 
    & $\mathbf{85.69 {\scriptstyle \pm 0.09}}$ 
    & $72.16 {\scriptstyle \pm 0.30}$ 
    & $\mathbf{95.40 {\scriptstyle \pm 0.24}}$ \\ 

\midrule

\multirow{7}{*}{\textcircled{2}} 
& \multirow{6}{*}{Dice} 
  & GCN       
    & $81.87 {\scriptstyle \pm 0.73}$ 
    & $83.34 {\scriptstyle \pm 0.60}$ 
    & $71.76 {\scriptstyle \pm 1.06}$ 
    & $87.14 {\scriptstyle \pm 0.86}$ \\

& 
  & RGCN      
    & $81.27 {\scriptstyle \pm 0.71}$ 
    & $83.89 {\scriptstyle \pm 0.51}$ 
    & $69.45 {\scriptstyle \pm 0.92}$ 
    & $87.35 {\scriptstyle \pm 0.76}$ \\

& 
  & GCNGuard  
    & $81.63 {\scriptstyle \pm 0.74}$ 
    & $83.72 {\scriptstyle \pm 0.49}$ 
    & $71.87 {\scriptstyle \pm 1.19}$ 
    & $86.98 {\scriptstyle \pm 1.26}$ \\

& 
  & GCNSVD    
    & $75.62 {\scriptstyle \pm 0.61}$ 
    & $79.13 {\scriptstyle \pm 1.01}$ 
    & $66.10 {\scriptstyle \pm 1.29}$ 
    & $88.50 {\scriptstyle \pm 0.97}$ \\

& 
  & GCNJaccard 
    & $80.67 {\scriptstyle \pm 0.66}$ 
    & $82.88 {\scriptstyle \pm 0.58}$ 
    & $\mathbf{72.32 {\scriptstyle \pm 1.10}}$ 
    & $51.81 {\scriptstyle \pm 1.49}$ \\

& 
  & GOOD-AT  
    & $82.21 {\scriptstyle \pm 0.56}$ 
    & $84.16 {\scriptstyle \pm 0.36}$ 
    & $71.43 {\scriptstyle \pm 0.28}$ 
    & $90.93 {\scriptstyle \pm 0.38}$ \\

& 
  & RobustCRF 
    & $\mathbf{82.44 {\scriptstyle \pm 0.41}}$ 
    & $\mathbf{84.71 {\scriptstyle \pm 0.32}}$ 
    & $71.48 {\scriptstyle \pm 0.15}$ 
    & $\mathbf{90.46 {\scriptstyle \pm 0.25}}$ \\ 

\bottomrule
\end{tabular}
}
\label{crf:tab:structure_results}
\end{table*}

\textbf{Baseline Models.}  When dealing with feature-based attacks, we compare RobustCRF with the vanilla GCN \citep{Kipf:2017tc}, the feature-based defense method RobustGCN (RGCN) \citep{robustGCN}, NoisyGNN \citep{ennadir2024simple}, and GCORN \citep{abbahaddou2024bounding}. For the structural attacks, we again included the RGCN and other baselines such as  GNN-Jaccard \citep{gnn_jaccard}, GNN-SVD \citep{gnn_svd}, GNNGuard \citep{gnn_guard}, and GOOD-AT \citep{li2024boosting}. For all the models, we used the same number of layers $T=2$, with a hidden dimension of $16$. The models were trained using the cross-entropy loss function with the Adam optimizer \citep{Kingma2014AdamAM}, the number of epochs $N_{\text{\textit{epochs}}}=300$, and learning rate $0.01$ were kept similar for the different approaches across all experiments.  To reduce the impact of random initialization, we repeated each experiment 10 times and used the train/validation/test splits provided with the datasets when evaluating against the feature-based attacks, c.f. Table~\ref{crf:tab:features_results}. When evaluating against the structural attacks, c.f. Table~\ref{crf:tab:structure_results}, we used the split strategy of \citep{zugner2018adversarial}, i.e., we select the largest connected components of the graph and use 10\%/10\%/80\% nodes for training/validation/test.

\textbf{Worst-Case Adversarial Evaluation.} We now analyze the results of RobustCRF for the node classification task. Additionally, we have compared our approach to the baseline methods on OGBN-Arxiv, a large dataset, as detailed in Appendix \ref{crf:app:ogbn}. We report all the results for the feature and structure-based adversarial attacks, respectively, in Tables \ref{crf:tab:features_results}, \ref{crf:tab:structure_results}, and \ref{crf:tab:ogbn}. The results demonstrate that the performance of the GCN model is significantly impacted when subject to adversarial attacks of varying strategies. In contrast, the proposed RobustCRF approach shows robust performance against these attacks compared to other baseline models. Furthermore, in contrast to some other benchmarks, RobustCRF offers a good balance between robustness and clean accuracy. Specifically, RobustCRF effectively enhances the robustness against adversarial attacks while maintaining high accuracy on non-attacked, clean datasets. This latter point makes RobustCRF particularly advantageous, as it enhances the model's defenses without compromising its performance on downstream tasks.

 \textbf{Enhancing Baseline Defenses with RobustCRF.}  One key advantage of RobustCRF is its ability to integrate with other defense methods, enhancing their overall robustness.  In Table \ref{crf:tab:combinations}, we report the performance of the baselines when combined with RobustCRF. As noticed, for most of the cases, we further enhance the robustness of the baselines when using RobustCRF.  This demonstrates its adaptability to different defense strategies, particularly those operating at the architectural level.

\textbf{Time and Complexity.} We analyze the computational complexity of RobustCRF, focusing on its inference phase, which depends primarily on the number of iterations $K$ and the number of sampled neighbors 
$L$. Since RobustCRF applies iterative updates following a coordinate ascent inference (CAI) scheme, its complexity is influenced by the size of the CRF and the number of updates required for convergence. In Appendix \ref{crf:appendix:time_robustcrf}, we report the average time needed to compute the RobustCRF inference. The results validate the intuitive fact that the inference time grows by increasing $K$ and $L$. We recall that we need to use the model $L^K$ times in the CRF inference. While increasing $K$ and $L$ can improve robustness, it also increases computational cost. In practical settings, we observe that convergence is often reached within a small number of iterations, making RobustCRF computationally feasible even for large graphs.

\section{Conclusion}
This work addresses the problem of adversarial defense at the inference stage. We propose a model-agnostic, post-hoc approach using Conditional Random Fields (CRFs) to enhance the adversarial robustness of pre-trained models. Our method, RobustCRF, operates without requiring knowledge of the underlying model and necessitates no post-training or architectural modifications. Extensive experiments on multiple datasets demonstrate RobustCRF's effectiveness in improving the robustness of Graph Neural Networks (GNNs) against both structural and node-feature-based adversarial attacks, while maintaining a balance between attacked and clean accuracy, typically preserving their performance on clean, un-attacked datasets, which makes RobustCRF the best trade-off between robustness and clean accuracy.

\part{Generalization of Graph Neural Networks}
\chapter{Improving Generalization in GNNs through Data Augmentation} \label{ch:Generalization}
\lettrine[lines=3]{G}{\small{raph}} Neural Networks (GNNs) have shown great promise in tasks like node and graph classification, but they often struggle to generalize, particularly to unseen or out-of-distribution (OOD) data. These challenges are exacerbated when training data is limited in size or diversity. To address these issues, we introduce a theoretical framework using Rademacher complexity to compute a regret bound on the generalization error and then characterize the effect of data augmentation. This framework informs the design of \method, an efficient graph data augmentation algorithm leveraging the capability of Gaussian Mixture Models (GMMs) to approximate any distribution. Our approach not only outperforms existing augmentation techniques in terms of generalization but also offers improved time complexity, making it highly suitable for real-world applications. 
\section{Introduction}
Graphs are a fundamental and ubiquitous structure for modeling complex relationships and interactions. In biology, graphs are employed to represent complex networks of protein interactions and in drug discovery by modeling molecular relationships \citep{gaudelet2021utilizing, branemf-bioinformatics22}. Similarly,  social networks capture relationships and community interactions \citep{zeng2022toward,MALLIAROS201395,newman2002random}. To address the unique challenges posed by graph-structured data, GNNs have been developed as a specialized class of neural networks designed to operate directly on graphs. Unlike traditional neural networks that are optimized for grid-like data, such as images or sequences, GNNs are engineered to process and learn from the relational information embedded in graph structures. GNNs have demonstrated state-of-the-art performance across a range of graph representation learning tasks such as node and graph classification, proving their effectiveness in various real-world applications \citep{vignac2022digress,corso2022diffdock,pmlr-v202-duval23a,gegenGNN-TNNLS24,chiRLCG22}.

Despite their impressive capabilities, GNNs face significant challenges related to generalization, particularly when handling unseen or out-of-distribution (OOD) data \citep{guo2024investigating,li2022ood}. OOD graphs are those that differ significantly from the training data in terms of graph structure, node features, or edge types, making it difficult for GNNs to adapt and perform well on such data. This challenge is also faced when GNNs are trained on small datasets, where the limited data diversity hampers the model’s ability to generalize effectively. To address these challenges, the community has explored various strategies to improve the robustness and generalization ability of GNNs \citep{abbahaddou2024bounding,yang2022graph}.

Generalization bounds for GNNs have been derived using various theoretical tools, such as the Vapnik-Chervonenkis (VC) dimension \citep{pfaff2020learning, garg2020generalization} and Rademacher complexity \citep{yin2019rademacher, esser2021learning}. Furthermore, \citet{liao2020pac} were among the first to establish generalization bounds for GNNs using the PAC-Bayesian approach. Neural Tangent Kernels have also been employed to study the generalization properties of infinitely wide GNNs trained via gradient descent \citep{jacot2018neural, du2019graph,huang2024enhancing}. While most existing research has focused on the node classification task, enhancing generalization in graph classification presents unique challenges. Techniques to improve generalization in graph classification can be broadly categorized into architectural and dataset-based strategies \citep{tang2023towards,buffelli2022sizeshiftreg}. On the dataset side, techniques like adversarial training and data augmentation play a significant role. More broadly, data augmentation methods create synthetic or modified graph instances to enrich the training set, reducing overfitting and enhancing the model's adaptability to diverse graph structures. Data augmentation has shown its benefits across different types of data structures such as images \citep{krizhevsky12} and time series \citep{aboussalah2023recursive}. For graph data structures, generating augmented versions of the original graphs, such as by adding or removing nodes and edges or perturbing node features \citep{rong2019dropedge,you2020graph}, allows for the creation of a more varied training set. Inspired by the success of the Mixup technique in computer vision \citep{rebuffi2021data,dabouei2021supermix,hong2021stylemix}, additional methods such as $\mathcal{G}$-Mixup and GeoMix have been developed to adapt Mixup for graph data \citep{ling2023graph,han2022g}. These techniques combine different graphs to create new, synthetic training examples, further enriching the dataset and enhancing the GNN’s ability to generalize to new unseen graph structures.

In this work, we introduce \method, a graph augmentation technique based on Gaussian Mixture Models (GMMs), which operates at the level of the final hidden representations. Specifically, guided by our theoretical results, we apply the Expectation-Maximization (EM) algorithm to train a GMM on the graph representations. We then use this GMM to generate new augmented graph representations through sampling, enhancing the diversity of the training data.

The contributions of our work are as follows:

\begin{itemize}
\item \textbf{Theoretical framework for generalization in GNNs.} We introduce a theoretical framework that allows us to rigorously analyze how graph data augmentation impacts the generalization  GNNs. This framework offers new insights into the underlying mechanisms that drive performance improvements through augmentation.

\item \textbf{Efficient graph data augmentation via GMMs.} We propose \method, a fast and efficient graph data augmentation technique leveraging GMMs. This approach enhances the diversity of training data while maintaining computational simplicity, making it scalable for large graph datasets.

\item \textbf{Comprehensive theoretical analysis using influence functions.} We perform an in-depth theoretical analysis of our augmentation strategy through the lens of influence functions, providing a principled understanding of the approach's impact on generalization performance.

\item \textbf{Empirical Validation.} Through experiments on real-world datasets we confirm \method~to be a fast, high-performing graph augmentation scheme in practice.
\end{itemize}

\section{Background and Related Work}\label{gen:background}

Graph data augmentation has become essential to enhance the performance and robustness of GNNs. Classical graph augmentation techniques focus on structural modifications to generate augmented graphs. Key methods here include DropEdge, DropNode, and Subgraph sampling \citep{rong2019dropedge,you2020graph}.  For instance,  DropEdge randomly removes a subset of edges from the graph during training, improving the model’s robustness to missing or noisy connections. Similarly, DropNode removes certain nodes as well as their connections, assuming that the missing part of nodes will not affect the semantic meaning, i.e., the structural and relational information of the original graph. Subgraph sampling, on the other hand, samples a subgraph from the original graph using random walks to use as a training graph.

Beyond classical methods, recent advancements have explored more sophisticated augmentation techniques, focusing on manipulating graph embeddings and leveraging the geometric properties of graphs. Following the effectiveness of the Mixup technique in computer vision \citep{rebuffi2021data,dabouei2021supermix,hong2021stylemix}, several works describe variations of the Mixup for graphs. For example, the Manifold-Mixup model conducts a Mixup operation for graph classification in the embedding space. This technique interpolates between graph-level embeddings after the \texttt{READOUT} function, blending different graphs in the embedding space \citep{wang2021mixup}.  Similarly, $\mathcal{G}$-Mixup \citep{han2022g} uses graphons to model the topological structures of each graph class and then interpolates the graphons of different classes, subsequently generating synthetic graphs by sampling from mixed graphons across different classes. It is important to note that $\mathcal{G}$-Mixup operates under a significant assumption: graphs belonging to the same class can be produced by a single graphon. Other advanced techniques include $S$-Mixup method, which interpolates graphs by first determining node-level correspondences between a pair of graphs \citep{ling2023graph}, and FGW-Mixup, which adopts the Fused Gromov-Wasserstein barycenter to compute mixup graphs but suffers from heavy computation time \citep{ma2024fused}. Finally,  GeoMix  \citep{zeng2024graph} leverages Gromov-Wasserstein geodesics to interpolate graphs more efficiently.  By leveraging these structural augmentation techniques, GNNs can better generalize to unseen graph structures.

\section{\method: Gaussian Mixture Model for Graph Data Augmentation} \label{gen:math_form}
In this section, we introduce the mathematical framework for graph data augmentation and its connection to the generalization of GNNs. Then, we present our proposed model \method, which is based on GMMs for graph augmentation. 
\subsection{Formalism of Graph Data Augmentation}

We focus on the task of graph classification, where the objective is to classify graphs into predefined categories. Let $\mathcal{D}$ denote the distribution of graphs. Given a training set of graphs $\mathcal{D}_{\text{train}} = \{ (\G_n, y_n) \mid n = 1, \ldots, N \}$, $\G_n$ is the $n$-th graph and $y_n$ is its corresponding label belonging to a set $\{0, \ldots,C\}$. Each graph $\G_n$ is represented as a tuple $(\V_n, \E_n, \mathbf{X}_n)$, where $\V_n$ denotes the set of nodes with cardinality $p_n = |\V_n|$, $\E_n \subseteq \V_n \times \V_n$ is the set of edges, and $\mathbf{X}_n \in \mathbb{R}^{p_n \times d}$ is the node feature matrix of dimension $d$. The objective is to train a GNN $f(\cdot, \theta)$ that can accurately predict the class labels for unseen graphs in the test set $\mathcal{D}_{\text{test}} = \{ \G_n^{\text{test}}\mid n = 1, \ldots, N_{\text{test}} \}$. The classical training approach involves minimizing the following loss function,
\begin{equation}\label{gen:eq:standard_loss}
\mathcal{L}  = \ell(f(\G_n, \theta), y_n),
\end{equation}
where $\ell$ denotes the cross-entropy loss function.

To improve the generalization performance of GNNs, we introduce a graph data augmentation strategy. For each training graph  $\G_n$ in the dataset, we generate $M$ augmented graphs, denoted as $\{\widetilde{\G}_{n,m}, \widetilde{y}_{n,m} \mid m = 1, \ldots, M\}$, where $M$ is the number of augmented graphs generated per training graph.  These augmented graphs are obtained using a graph augmentation generator $A_{\lambda}$, parameterized by $\lambda$ as a mapping $A_{\lambda}: \G_n, y_n\rightarrow A_{\lambda}\left(\G_n, y_n\right) \in  \mathbb{G} \times \mathbb{Y},$ where  $\mathbb{G}$ denotes the space of all possible graphs, and $\mathbb{Y}$ is the label space. The generator $A_{\lambda}$ may be either deterministic or stochastic with a dependence on a prior distribution $\mathcal{P}(\lambda)$. Examples of such augmentation strategies can be found in Appendix \ref{gen:app:aug_strat}.

We use the notation $\widetilde{\G}_{n}^{m}\sim A_{\lambda}$ to represent an augmented graph sampled from the augmentation strategy $A_\lambda$, i.e., $(\widetilde{\G}_{n}^{m},\widetilde{y}_{n}^{m})\sim A_{\lambda}(\G_n, y_n)$. With the augmented data, the loss function is modified to account for multiple augmented versions of each graph,
\begin{equation}
\mathcal{L}^{\text{aug}} = \frac{1}{N} \sum_{n=1}^{N} \mathbb{E}_{\widetilde{\G}_{n}^{m}\sim A_{\lambda}} \left [ \ell(f(\widetilde{\G}_{n}^{m}, \theta), \widetilde{y}_{n}^{m}) \right ].
\end{equation}
For simplicity,  we denote the loss for the original graph by $\ell(f(\G_n, \theta), y_n) = \ell(\G_n, \theta )$  and the loss for an augmented graph as $ \mathbb{E}_{\widetilde{\G}_{n}^{m}\sim A_{\lambda}} \left [ \ell(f(\widetilde{\G}_{n}^{m}, \theta), \widetilde{y}_{n}^{m}) \right ]= \ell^{\text{aug}}(\widetilde{\G}_n, \theta )$. Via the law of large numbers,  $\mathcal{L}^{\text{aug}}$ is empirically estimated,
\begin{align*}
\mathcal{L}^{\text{aug}} & = \frac{1}{N} \sum_{n=1}^{N} \ell^{\text{aug}}(\widetilde{\G}_n, \theta )\\
& \simeq \frac{1}{N   M} \sum_{n=1}^N \sum_{m=1}^M \ell(f(\widetilde{\G}_{n}^{m}, \theta),\widetilde{y}_{n}^{m}).
\end{align*}
To understand the impact of data augmentation on the graph classification performance, we analyze the effect of the augmentation strategy on the generalization risk $\mathbb{E}_{\G \sim \mathcal{D}}\left [ \ell(\G, \theta) \right ]$. More specifically, we want to study the generalization error,
$$\eta = \mathbb{E}_{\G \sim \mathcal{D}}\left [ \ell(\G,  \theta_{\text{aug}}) \right ] - \mathbb{E}_{\G \sim \mathcal{D}}\left [ \ell(\G, \theta_{\star}) \right ],$$ 
where $\theta_{\text{aug}}$ and $\theta_{\star}$ are the optimal GNN parameters for the augmented and non-augmented settings,  respectively,
\begin{equation}\theta_{\star} =\argmin_\theta\mathcal{L}_\theta,~~~~\theta_{\text{aug}} = \argmin_\theta\mathcal{L}^{\text{aug}}_\theta ,\end{equation}
which can be estimated empirically as follows, 
\begin{align*}
    \hat{\theta } &  = \argmin_\theta  \frac{1}{N} \sum_{n=1}^N \ell(\G_n, \theta),\\
    \hat{\theta }_{\text{aug}} & = \argmin_\theta  \frac{1}{N  M} \sum_{n=1}^N \sum_{m=1}^M  \ell(\widetilde{\G}_{n}^{m}, \theta).
\end{align*}
By theoretically studying the generalization error $\eta$, we aim to quantify the effect of each augmentation strategy on the overall classification performance, providing insights into the benefits and potential trade-offs of data augmentation in graph-based learning tasks. In Theorem \ref{gen:thm:rademacher}, we present a regret bound of the generalization error using Rademacher complexity defined as follows \cite{yin2019rademacher},
$$\mathcal{R}(\ell )  = \mathbb{E}_{\epsilon_n \sim P_{\epsilon}}\left [  \sup_{\theta \in \Theta} \left | \frac{1}{N} \sum_{n=1}^{N} \epsilon_n \ell(\G_n, \theta)  \right | \right ],$$
where $\epsilon_n$ are independent Rademacher variables, taking values $+1$ or $-1$ with equal probability, $P_{\epsilon}$ is the Rademacher distribution, and $\Theta$ is the hypothesis class. Rademacher complexity is a fundamental concept in statistical learning, which indicates how well a learned function will perform on unseen data \citep{shalev2014understanding}. Intuitively, the Rademacher complexity measures the capacity of a GNN to fit random noise \cite{zhu2009human}, where a lower Rademacher complexity indicates better generalization.

\begin{theorem}\label{gen:thm:rademacher}
Let $\ell$ be a classification loss function with $\text{L}_{\text{Lip}}$ as a Lipschitz constant and $\ell(\cdot,\cdot)\in [0,1].$ Then, with a probability at least $1-\delta$ over the samples $\mathcal{D}_{\text{train}}$, we have, 
\begin{equation} \begin{split}\mathbb{E}_{\G  \sim \mathcal{D}}\left [ \ell(\G,\hat{\theta}_{\text{aug}}) \right ] - \mathbb{E}_{\G  \sim \mathcal{D}}\left [ \ell(\G,\theta_{\star}) \right ] \leq 2 \mathcal{R}(\ell_{\text{aug}} ) + \\ 5\sqrt{\frac{2\log(4/\delta)}{N}} + 2 \text{L}_{\text{Lip}}  \mathbb{E}_{\G \sim \mathcal{D}, \widetilde{\G}\sim A_\lambda} \left [ \left \| \widetilde{\G}- \G \right \| \right ].\end{split} \end{equation}
Moreover, we have,
$$\mathcal{R}(\ell_{\text{aug}} )  \leq \mathcal{R}(\ell ) +  \max_{n}  \text{L}_{\text{Lip}}    \mathbb{E}_{\widetilde{\G}_{n}^{m} \sim A_\lambda} \left [ \left \| \widetilde{\G}_{n}^{m}  -\G_n \right \| \right ].  
$$
\end{theorem}
Theorem \ref{gen:thm:rademacher} relies on the assumption that the loss function is Lipschitz continuous. This assumption is realistic, given that the input node features and graph structures in real-world datasets are typically bounded, i.e., node features are typically normalized or constrained within a fixed range, while graph structures, represented by adjacency matrices or their normalized forms, have bounded spectral properties, ensuring a constrained input space. Additionally, we can ensure that the loss function is bounded within $[0,1]$ by composing any standard classification loss with a strictly increasing function that maps values to the interval $[0,1]$. We provide the proof of this theorem in Appendix~\ref{gen:appendix:proof_lemma}.

A direct implication of Theorem \ref{gen:thm:rademacher} is that if we chose the right data augmentation strategy $A_{\lambda}$ that minimizes the expected distance between original graphs and augmented ones $ \mathbb{E}_{\G \sim \mathcal{D}, \widetilde{\G} \sim A_\lambda} \left [ \left \| \widetilde{\G} - \G \right \| \right ]$, we can guarantee with a high probability that the data augmentation decreases both the Rademacher complexity and the generalization risk. Specifically, we need enough diversity to ensure a non-zero difference in the Rademacher complexity compared to the non-augmented case while controlling the expected distance so that, with high probability, we reduce both the Rademacher complexity and the overall generalization error $\eta$. On the other hand, if the distance is large, we cannot guarantee that data augmentation will outperform the normal training setting.

The findings of Theorem \ref{gen:thm:rademacher} hold for all norms defined on the graph input space. Specifically, let us consider the graph structure space $(\mathbb{A}, \lVert \cdot \rVert_{\mathbb{A}})$ and the feature space $(\mathbb{X}, \lVert \cdot \rVert_{\mathbb{X}})$, where $\lVert \cdot \rVert_{\mathbb{A}}$ and $\lVert \cdot \rVert_{\mathbb{X}}$ denote the norms applied to the graph structure and features, respectively. Assuming a maximum number of nodes per graph, which is a realistic assumption for real-world data, the product space $\mathbb{A} \times \mathbb{X}$ is a finite-dimensional real vector space, and all the norms are equivalent. Thus, the choice of norm does not affect the theorem, as long as the Lipschitz constant is adjusted accordingly. Additional details on graph distance metrics and the comparison between the Lipschitz constants of GCN and GIN are provided in Appendices \ref{gen:app:metrics} and \ref{gen:app:gnn_definitions}.

The variety of distance metrics on the original graph space offers different upper bounds, which leads to distinct criteria for data augmentation based on these metrics to control the upperbound in Theorem \ref{gen:thm:rademacher}. Instead, we propose to shift the focus to the hidden representation space of graphs, where we aim to derive more consistent and meaningful augmentations. If $\text{L}_{\text{Lip}}$ is taken to be the Lipschitz constant of the post-readout function only, then Theorem \ref{gen:thm:rademacher} still applies for the norm on the graph-level embeddings produced by the readout function, i.e., $\|\mathbf{h}_{\widetilde{\G}} - \mathbf{h}_{\G}\|$.

Working at the level of hidden representations of graphs, rather than directly in the graph input space, offers additional advantages. Hidden representations capture both the structural information and node features of each graph, enabling augmentation that enhances the generalization of both aspects simultaneously. Moreover, node alignment is needed to compare the original and augmented graph, which is computationally expensive. By operating on hidden representations instead, node alignment becomes unnecessary. Furthermore, as we will discuss in Section \ref{gen:sub_sec:influence}, the effectiveness of augmented data depends on the specific GNN architecture. By leveraging graph representations learned through a GNN, we ensure that the augmentation process remains architecture-specific, aligning with the inductive biases of the chosen model.

\subsection{Proposed Approach}
Based on the theoretical findings, it is crucial to employ a data augmentation technique that effectively controls the term  $ \mathbb{E}_{ \G\sim \mathcal{D}, \widetilde{\G}\sim A_\lambda} \left [ \|\mathbf{h}_{\widetilde{\G}} - \mathbf{h}_{\G}\| \right ]$ which measures the expected deviation in the hidden representations of graphs under augmentation, to achieve stronger generalization guarantees. To better understand this term, we express it in terms of graph representations rather than graphs themselves,
\begin{align*}
\mathbb{E}_{ \G\sim \mathcal{D},\widetilde{\G}\sim A_\lambda} \left [ \|\mathbf{h}_{\widetilde{\G}} - \mathbf{h}_{\G}\| \right ] & = \mathbb{E}_{ \mathbf{h}\sim \delta_\mathcal{D},\widetilde{\mathbf{h}}\sim Q_\lambda} \left [ \|\widetilde{\mathbf{h}} - \mathbf{h}\| \right ],
\end{align*}
where $Q_\lambda$ represents the augmentation strategy at the level of hidden representations, replacing $A_\lambda$, which operates at the graph level, and  $\delta_\mathcal{D}: \mathbf{h}\mapsto \frac{1}{N} \sum_{n=1}^{N} \delta_{\mathbf{h}_{\G_n}}(\mathbf{h})$ represents the Dirac distribution over the training graph representations, capturing the empirical distribution of graph embeddings. In Proposition \ref{gen:prop:sampling_ineq}, we upper bound this expected perturbation.

\begin{proposition} \label{gen:prop:sampling_ineq}
    Let $\delta_\mathcal{D}$ denote the discrete distribution of the training graph representations. Suppose we sample new augmented graph representations from a distribution $Q_\lambda$  defined on $\mathbb{R}^d$. Then, the following inequality holds,
\begin{equation}\mathbb{E}_{\mathbf{h} \sim \delta_\mathcal{D},\widetilde{\mathbf{h}} \sim Q_\lambda} \left[ \| \mathbf{h} - \widetilde{\mathbf{h}} \| \right]  
\leq   \sqrt{2} \cdot \sup_{\substack{\mathbf{h} \sim \delta_\mathcal{D} \\ \widetilde{\mathbf{h}} \sim Q_\lambda}} \| \mathbf{h} - \widetilde{\mathbf{h}} \| \left( \sqrt{ \mathcal{KL}(\delta_\mathcal{D} \parallel Q_\lambda) } + \sqrt{2} \right),\end{equation} 
where $\mathcal{KL}(\cdot \parallel \cdot)$ denotes the Kullback-Leibler divergence. 
\end{proposition}

We provide a proof of Proposition \ref{gen:prop:sampling_ineq} in Appendix \ref{gen:proposition:KL_proof}. A way to control the left side of this inequality is to choose a generator $Q_\lambda$ that minimizes both the $\mathcal{KL}(\delta_\mathcal{D} \parallel Q_\lambda)$ and the supremum distance $\sup_{\mathbf{h} \sim \delta_\mathcal{D}, \widetilde{\mathbf{h}} \sim Q_\lambda} \| \mathbf{h} - \widetilde{\mathbf{h}} \|$. Various universal approximators can be used to minimize $\mathcal{KL}(\delta_\mathcal{D} \parallel Q_\lambda)$, including generative models like Generative Adversarial Networks (GANs) \cite{yang2012robust}. These models are capable of approximating any probability distribution, making them powerful tools for learning complex augmentations. However, we specifically choose Gaussian Mixture Models (GMMs), which are well-suited for this purpose, and can effectively approximate any data distribution, c.f. Theorem \ref{gen:thm:GMM}.  GMMs are computationally fast compared to other generative approaches, making them suitable for large-scale graph datasets. As shown in Appendix \ref{gen:app:ablation_study}, GMM-based augmentation yields better results compared to alternative generative strategies. Moreover, due to the exponential decay of Gaussian distributions, the supremum distance $\sup_{\mathbf{h} \sim \delta_\mathcal{D}, \widetilde{\mathbf{h}} \sim Q_\lambda} \| \mathbf{h} - \widetilde{\mathbf{h}} \|$ is naturally constrained, ensuring a better control of the expected distance $\mathbb{E}_{\mathbf{h} \sim \delta_\mathcal{D},\widetilde{\mathbf{h}} \sim Q_\lambda} \left[ \| \mathbf{h} - \widetilde{\mathbf{h}} \| \right]$.

\begin{theorem}{\citep[Page 65]{goodfellow2016deep}}\label{gen:thm:GMM} A Gaussian mixture model is a universal approximator of densities, in the sense that any smooth density can be approximated with any speciﬁc nonzero amount of error by a Gaussian mixture model with enough components.
\end{theorem}

To achieve this, we first train a standard GNN on the graph classification task using the training set. Next, we obtain embeddings for all training graphs using the \texttt{READOUT} output, resulting in $\mathcal{H} = \left\{\mathbf{h}_{\G_n} ~\text{s.t.}~\G_n \in \mathcal{D}_{\text{train}}  \right\}$. These embeddings are used as the basis for generating augmented training graphs. We then partition the training set $\mathcal{D}_{\text{train}}$ by classes, such that $\mathcal{D}_{\text{train}} = \bigcup_{c}\mathcal{D}_c$ where $\mathcal{D}_c = \left\{\G_n \in \mathcal{D}_{\text{train}} ~,~ y_n=c \right\}$. The objective is to learn new graph representations from these embeddings, and create augmented data for improved training.

We use the EM algorithm to learn the best-fitting GMM for the embeddings of each partition $\mathcal{D}_c$, denoted as $\mathcal{H}_c = \left\{\mathbf{h}_{\G_n} ~\text{s.t.}~\G_n \in \mathcal{D}_c \right\}$. The EM algorithm finds maximum likelihood estimates for each cluster $\mathcal{H}_c$, following the procedure described in \cite{bishop2006pattern}.
    
\begin{figure*}
    \centering
    \includegraphics[width=\textwidth]{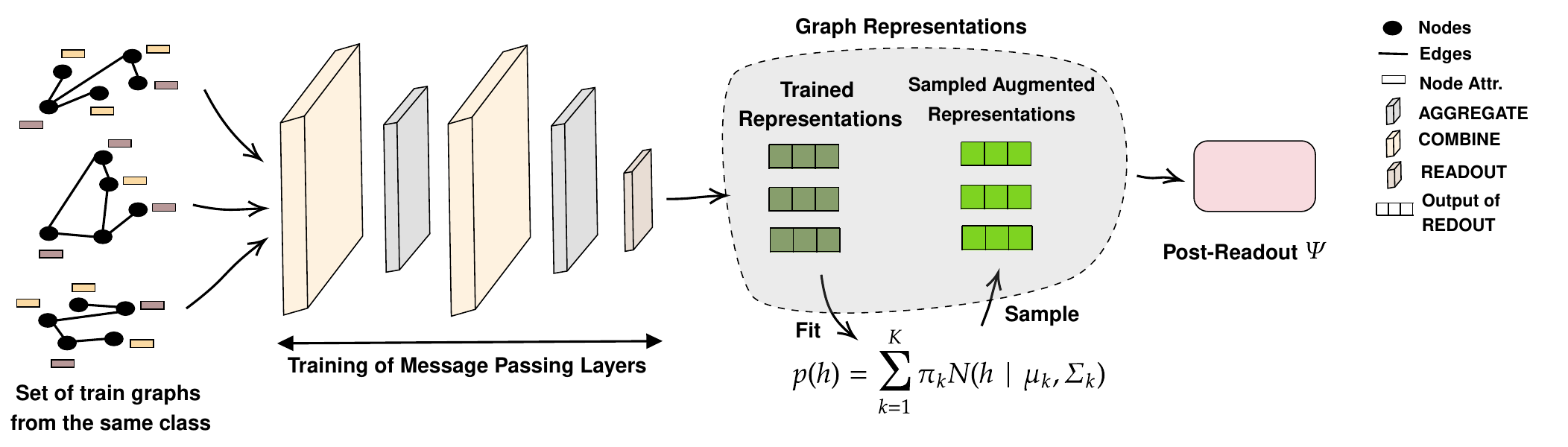}
    \caption[Illustration of \method]{Illustration of \method. \textit{Step 1}. We first train the GNN on the graph classification task using the training graphs. \textit{Step 2}. Next, we utilize the weights from the message passing layers to generate graph representations for the training graphs. \textit{Step 3}. A GMM  is then fit to these graph representations, from which we sample new graph representations. \textit{Step 4}. Finally, we fine-tune the post-readout function for the graph classification task, using both the original training graphs and the augmented graph representations. For inference on the test set, we use the message passing weights trained in \textit{Step 1} and the post-readout function weights trained in \textit{Step 4}.} 
    \label{gen:fig:architechture}
\end{figure*}

Once a GMM distribution $p_c$ is fitted for each partition $\mathcal{D}_c$, we use this GMM to generate new augmented data by sampling hidden representations from $p_c$. Each new sample drawn from $p_c$ is then assigned the corresponding partition label $c$, ensuring that the augmented data inherits the label structure from the original partitions. 
After merging the hidden representations of both the original training data and the augmented graph data, we finetune the post-readout function, i.e., the final part of the GNN, which occurs after the readout function, on the graph classification task. Since the post-readout function consists of a linear layer followed by a Softmax function, the finetuning process is relatively fast. To evaluate our model during inference on test graphs, we input the test graphs into the GNN layers trained in the initial step to compute the hidden graph representations. For the post-readout function, we use the weights obtained from the second stage of training. Algorithm \ref{gen:algo:GMM_GDA} and Figure \ref{gen:fig:architechture} provide a summary of the \method~model.

\begin{algorithm}[t]
\small
\textbf{Inputs: } GNN of $T$ layers $f(\cdot,\theta)=\Psi \circ \texttt{READOUT} \circ g $, where $g$ is the composition of message passing layers,  i.e, $g = \cup_{t=0}^T \{\texttt{AGGREGATE}^{(t)} \circ \texttt{COMBINE}^{(t)}(\cdot)  \} $ and $\Psi$ is the post-readout function, graph classification dataset $\mathcal{D}$, loss function $\mathcal{L}$;\\
\textbf{Steps:}

\textbf{1.} Train GNN $f$  on the training set $\mathcal{D}_{\text{train}}$;

\textbf{2.} Use the trained message passing layers and the readout function to generate graph representations $\mathcal{H} = \left\{\mathbf{h}_{\G_n} ~s.t.~\G_n \in \mathcal{D}_{\text{train}}  \right\}$ for the training set;

\textbf{3.} Partition the training set $\mathcal{D}_{\text{train}}$ by classes, such that $\mathcal{D}_{\text{train}} = \bigcup_{c}\mathcal{D}_c$ where $\mathcal{D}_c = \left\{\G_n \in \mathcal{D}_{\text{train}} ~,~ y_n=c \right\}$;\\
    \ForEach{$c \in \{0,\ldots,C\}$}{
     \textbf{3.1.} Fit a GMM distribution $p_c$ on the graph representations $\mathcal{H}_c = \left\{\mathbf{h}_{\G_n} ~\text{s.t.}~\G_n \in \mathcal{D}_c \right\}$;\\
     \textbf{3.2.} Sample new graph representation $\widetilde{\mathcal{H}_c} = \{\widetilde{\mathbf{h}} ~~\text{s.t.}~~ \widetilde{\mathbf{h}} \sim p_c\}$ from the distribution $p_c$; \\
      \textbf{3.3.} Include the sampled representations $\widetilde{\mathcal{H}_c}$ with trained representations $\mathcal{H}_c  = \mathcal{H}_c  \cup \widetilde{\mathcal{H}_c} $;
    }
\textbf{4.} Finetune the post-readout function $\Psi $ on the graph classification task directly on the new training set $\mathcal{H} = \cup_c \mathcal{H}_c$;
\caption{Graph classification with \method}\label{gen:algo:GMM_GDA}
\end{algorithm}

\subsection{Time Complexity}
One advantage of \method~is its efficiency, as it generates new augmented graph representations with minimal computational time. Unlike baseline methods, which apply augmentation strategies to each individual training graph (or pair of graphs in Mixup-based approaches) separately, our method learns the distribution of graph representations across the entire training dataset simultaneously using the EM algorithm \citep{ng2000cs229}. If $N = \left| \mathcal{D}_{\text{train}} \right|$ is the number of training graphs in the dataset, $d$ is the dimension of graph hidden representations $\left\{  \mathbf{h}_{\G}, ~\G \in \mathcal{D}_{\text{train}}\right\}$, and $K$ is the number of Gaussian components in the GMM, then the complexity to fit a GMM on $T$ iterations is $\mathcal{O}(N \cdot K \cdot T \cdot d^2)$ \citep{yang2012robust}. We compare the data augmentation times of our approach and the baselines in Table \ref{gen:tab:complexity}. Due to our different training scheme, i.e., where we first train the message passing layers and then train the post-readout function after learning the GMM distribution, we have measured the total backpropagation time and compared it with the backpropagation time of the baseline methods. The training time of baseline models varies depending on the augmentation strategy used, specifically whether it involves pairs of graphs or individual graphs. Even in cases where a graph augmentation has a low computational cost for some baselines, training can still be time-consuming as multiple augmented graphs are required to achieve satisfactory test accuracy. In contrast, \method~generates only one augmented graph per training graph, demonstrating effective generalization on the test set. Overall, our data augmentation approach is highly efficient during the sampling of augmented data,  with minimal impact on the training time. A complete analysis of the time complexity of \method and the baselines can be found in Appendix \ref{gen:app:time}.

\subsection{Analyzing the Generalization Ability of the Augmented Graphs via Influence Functions} \label{gen:sub_sec:influence}

We use \textit{influence functions} \citep{law1986robust,koh2017understanding,kong2021resolving}  to understand the impact of augmented data on the model performance on the test set and thus motivate the use of a data augmentation strategy, which is specific to the model architecture and model weights.
In Theorem \ref{gen:thm:closed_formula}, we derive a closed-form formula for the impact of adding an augmented graph $\widetilde{\G}_{n}^{m}$ on the GNN's performance on a test graph $\G_k^{\text{test}}$, where the GNN is trained solely on the training set, without the augmented graph.

\begin{theorem}\label{gen:thm:closed_formula}
Given a test graph $\G_k$, let $\hat{\theta} = \arg\min_\theta \mathcal{L}$ be the GNN parameters that minimize the objective function in \eqref{gen:eq:standard_loss}. The impact of upweighting the objective function $\mathcal{L}$ to $\mathcal{L}_{n,m}^{\text{aug}} = \mathcal{L} + \epsilon_{n,m} \ell(\widetilde{\G}_{n}^{m}, \theta)$, where $\widetilde{\G}_{n}^{m}$ is an augmented graph candidate of the training graph $\G_n$ and $\epsilon_{n,m}$ is a sufficiently small perturbation parameter, on the model performance on the test graph $\G_k^{\text{test}}$ is given by
\begin{equation}
\frac{d \ell(\G_k^{\text{test}}, \hat{\theta}_{\epsilon_{n,m}})}{d \epsilon_{n,m}} = - \nabla_\theta \ell(\G_k^{\text{test}}, \hat{\theta})\mathbf{H}_{\hat{\theta}}^{-1} \nabla_\theta \ell(\widetilde{\G}_{n}^{m}, \hat{\theta} ),
\end{equation}
where $\hat{\theta}_{\epsilon_{n,m}} = \argmin_{\theta} \mathcal{L}_{n,m}^{\text{aug}}$ denotes the parameters that minimize the upweighted objective function $\mathcal{L}_{n,m}^{\text{aug}}$ and $\mathbf{H}_{\hat{\theta}}  =\nabla_{\theta}^2\mathcal{L}(\hat{\theta})$ is the Hessian matrix of the loss w.r.t. the model parameters.
\end{theorem}

We provide the proof of Theorem \ref{gen:thm:closed_formula} in Appendix \ref{gen:app:proof_thm_infl}. The influence scores are useful for evaluating the effectiveness of the augmented data on each test graph. The strength of influence function theory lies in its ability to analyze the effect of adding augmented data to the training set without actually retraining on this data. As noticed, these influence scores depend not only on the augmented graphs themselves but also on the model's weights and architecture. This highlights the need for a graph data augmentation strategy tailored specifically to the GNN backbone in use, as opposed to traditional techniques like DropNode, DropEdge, and $\G$-Mixup, which are general-purpose methods that can be applied with any GNN architecture. 

\begin{figure}
    \centering
    \includegraphics[width=\textwidth]{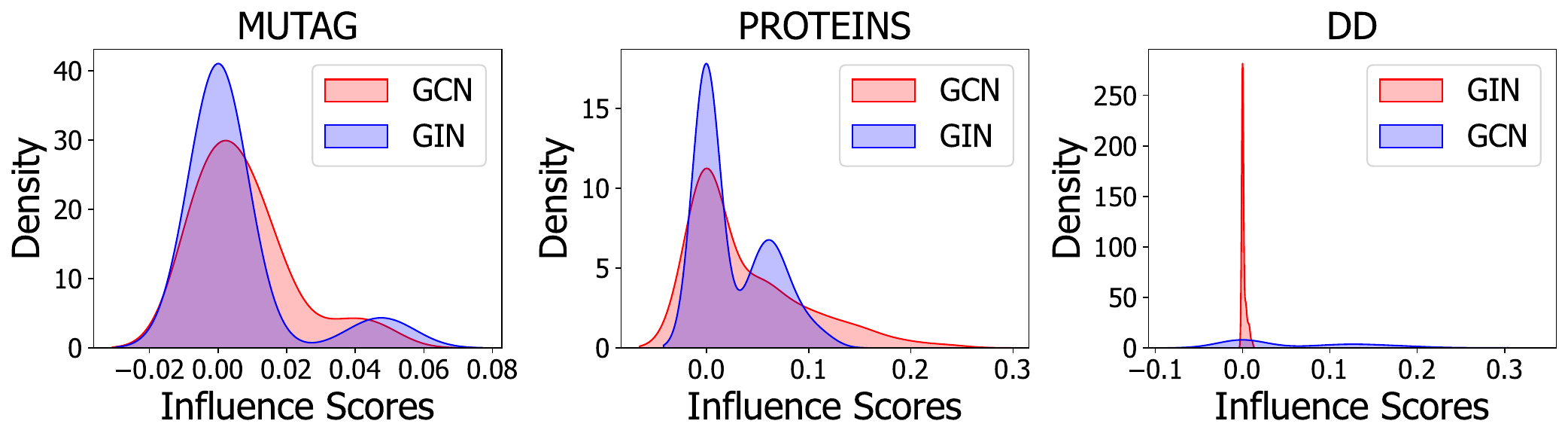}
    \caption[The density of the average influence scores]{The density of the average influence scores of each augmented data on the test set.} 
    \label{gen:fig:influence_scores}
\end{figure}

Theorem \ref{gen:thm:closed_formula} is valid for any differentiable loss function. More specifically, if the chosen loss is the cross entropy or the negative log-likelihood, then the hessian matrix corresponds to the Fisher information matrix \cite{barshan2020relatif,Lee_2022_CVPR}. Consequently, the norm of $\mathbf{H}_{\hat{\theta}}^{-1}$, i.e., the inverse of the hessian matrix,  can be bounded above using the Cramér–Rao inequality \cite{nielsen2013cramer}. Therefore, a trivial case where the norm of influence scores is zero arises when the gradient of the loss function with respect to the input graphs vanishes. This scenario, for instance, can occur in the DD dataset when using GIN. A detailed analysis of this phenomenon is provided in Section \ref{gen:exps}. In these cases, data augmentation becomes ineffective, having minimal impact on the GNN’s ability to generalize. We can measure the average influence $\mathcal{I}(\widetilde{\G}_{n}^{m})$ of a augmented graph $\widetilde{\G}_{n}^{m}$ on the test set by averaging the derivatives as follows, 
$$\mathcal{I}(\widetilde{\G}_{n}^{m}) = \frac{-1}{|\mathcal{D}_{\text{test}}|}\sum_{\G_k^{\text{test}} \in \mathcal{D}_{\text{test}}} \frac{d \ell(\G_k^{\text{test}}, \hat{\theta}_{\epsilon_{n,m}})}{d \epsilon_{n,m}}.$$
A negative value of $\mathcal{I}(\widetilde{\G}_{n}^{m})$ indicates that adding the augmented data to the training set would increase the prediction loss on the test set, negatively affecting the GNN's generalization. In contrast, a good augmented graph is one with a positive $\mathcal{I}(\widetilde{\G}_{n}^{m})$, indicating improved generalization. In Figure \ref{gen:fig:influence_scores}, we present the density of the average influence scores of each augmented data on the test set.

\subsection{Fisher-Guided GMM Augmentation} \label{gen:subsec:fisher_gmm_augmentation}

Using influence scores, we can further improve the generalization of the GNN by filtering candidate augmented representations. The process consists of three key stages. \textit{(i) Primary GNN training:} The GNN model is first trained on the original training set without incorporating any augmented graphs. \textit{(ii) Augmentation and filtering:} A pool of candidate augmented graph representations is generated using a data augmentation strategy based on GMMs. The influence of each augmented representation on the validation set is then computed using influence scores, enabling us to rank the augmented graphs by their impact. \textit{(iii) Filtering:} Finally, we combine a subset of the highest-ranked augmented graphs with the original training set to train the post-readout function. Our experiments in Appendix~\ref{gen:app:fisher_gmm_experiment} demonstrate that this two-phase training paradigm improves generalization across various datasets and GNN architectures.

\section{Experimental Results} \label{gen:exps}

\begin{table}[t!]
\centering
\caption[Classification accuracy of the data augmentation baselines and \method on the GCN backbone]{Classification accuracy ($\pm$ std) on different benchmark graph classification datasets for the data augmentation baselines based on the GCN backbone. The higher the accuracy (in \%) the better the model. Highlighted are the \textbf{first}, \underline{second} best results. }
\resizebox{0.8\columnwidth}{!}{%
\begin{tabular}{lllllll}
\toprule
Model & IMDB-BIN  & IMDB-MUL & MUTAG & PROTEINS  & DD \\ \midrule

No Aug.  & $73.00  {\scriptstyle \pm 4.94}$ & $47.73  {\scriptstyle \pm 2.64}$ & $73.92  {\scriptstyle \pm 5.09}$ &  $69.99  {\scriptstyle \pm 5.35}$ &  $69.69  {\scriptstyle \pm 2.89}$ \\
DropEdge  & $71.70  {\scriptstyle \pm 5.42}$ & $45.67  {\scriptstyle \pm 2.46}$ & $73.39  {\scriptstyle \pm 8.86}$  &   $70.07  {\scriptstyle \pm 3.86}$ &  $69.35  {\scriptstyle \pm 3.37}$ \\
DropNode  & $\mathbf{74.00  {\scriptstyle \pm 3.44}}$ & $43.80  {\scriptstyle \pm 3.54}$ & $73.89  {\scriptstyle \pm 8.53}$ &  $69.81  {\scriptstyle \pm 4.61}$ &  $69.01  {\scriptstyle \pm 3.95}$ \\
SubMix   & $\underline{72.70  {\scriptstyle \pm 5.59}}$ & $46.00  {\scriptstyle \pm 2.44}$ & $\underline{77.13  {\scriptstyle \pm 9.69}}$ & $67.57  {\scriptstyle \pm 4.56}$ & $70.11  {\scriptstyle \pm 4.48}$ \\
$\mathcal{G}$-Mixup & $72.10  {\scriptstyle \pm 3.27}$ & $48.33  {\scriptstyle \pm 3.06}$ & $\mathbf{88.77  {\scriptstyle \pm 5.71}}$ &  $65.68  {\scriptstyle \pm 5.03}$ & $61.20  {\scriptstyle \pm 3.88}$ \\
GeoMix  &  $69.69  {\scriptstyle \pm 3.37}$& $\underline{49.80  {\scriptstyle \pm 4.71}}$ & $74.39  {\scriptstyle \pm 7.37}$ &  $69.63  {\scriptstyle \pm 5.37}$ & $68.50  {\scriptstyle \pm 3.74}$ \\
\method  & $71.00  {\scriptstyle \pm 4.40}$ & $\mathbf{49.82  {\scriptstyle \pm 4.26}}$  & $76.05  {\scriptstyle \pm 6.74}$ & $\mathbf{70.97  {\scriptstyle \pm 5.07}}$ &  $\mathbf{71.90  {\scriptstyle \pm 2.81}}$\\
 \bottomrule

\end{tabular}
}

\label{gen:tab:results_gcn}
\end{table}

\begin{table}[bt]
\centering
\caption[Classification accuracy of the data augmentation baselines and \method on the GIN backbone]{Classification accuracy ($\pm$ std) on different benchmark graph classification datasets for the data augmentation baselines based on the GIN backbone. The higher the accuracy (in \%) the better the model. Highlighted are the \textbf{first}, \underline{second} best results. }
\resizebox{0.8\columnwidth}{!}{%
\begin{tabular}{lllllll}
\toprule 
Model & IMDB-BIN  & IMDB-MUL & MUTAG & PROTEINS  & DD \\ \midrule

No Aug. & $70.30 {\scriptstyle \pm 3.66}$ & $\underline{48.53 {\scriptstyle \pm 4.05}}$ & $83.42 {\scriptstyle \pm 2.12}$ &  $69.54 {\scriptstyle \pm 3.61}$ &   $68.00 {\scriptstyle \pm 3.18}$ \\
DropEdge & $70.40 {\scriptstyle \pm 4.03}$ & $46.80 {\scriptstyle \pm 3.91}$ & $74.88 {\scriptstyle \pm 9.62}$ &  $68.27 {\scriptstyle \pm 5.21}$& $67.82 {\scriptstyle \pm 4.46}$\\
DropNode   & $70.30 {\scriptstyle \pm 3.49} $& $45.20 {\scriptstyle \pm 4.24}$ & $75.53 {\scriptstyle \pm 7.89}$ & $65.40 {\scriptstyle \pm 4.71}$  & $\mathbf{69.01 {\scriptstyle \pm 3.95} }$   \\
SubMix   & $\mathbf{72.50 {\scriptstyle \pm 4.98}}$ & $48.13 {\scriptstyle \pm 2.12}$ & $81.90 {\scriptstyle \pm 9.21}$ & $\underline{70.44 {\scriptstyle \pm 2.58}}$ & $68.59 {\scriptstyle \pm 5.04}$ \\
$\mathcal{G}$-Mixup  & $70.70 {\scriptstyle \pm 3.10}$ & $47.73 {\scriptstyle \pm 4.95}$ & \underline{$87.77 {\scriptstyle \pm 7.48}$}  & $68.82 {\scriptstyle \pm 3.48}$ & $63.91 {\scriptstyle \pm 2.09}$ \\
GeoMix   & $70.60 {\scriptstyle \pm 4.61}$ & $47.20 {\scriptstyle \pm 3.75}$ & $81.90 {\scriptstyle \pm 7.55}$  & $69.80 {\scriptstyle \pm 5.33}$ & $68.34 {\scriptstyle \pm 5.30}$   \\ 
\method  & \underline{$71.70 {\scriptstyle \pm 4.24}$} & $\mathbf{49.20 {\scriptstyle \pm 2.06}}$ & $\mathbf{88.83 {\scriptstyle \pm 5.02}}$ &  $\mathbf{71.33 {\scriptstyle \pm 5.04}}$ &  $\underline{68.61 {\scriptstyle \pm 4.62}}$ \\
 \bottomrule

\end{tabular}
}

\label{gen:tab:results_gin}
\end{table}

\begin{table*}[t]
\centering
\caption[Robustness performance against structure corruption]{Robustness against structure corruption: We present the Classification accuracy ($\pm$ standard deviation). We highlighted the best data augmentation strategy \textbf{bold}. For this experiment, we use the GCN backbone.}
\resizebox{\columnwidth}{!}{%

\begin{tabular}{lllllllll}  \toprule 
Noise Budget & \multicolumn{4}{c}{10\%}             & \multicolumn{4}{c}{20\%}\\
\cmidrule(lr){2-5} \cmidrule(lr){6-9}
  Dataset 
  & IMDB-BIN & IMDB-MUL & PROTEINS & DD & IMDB-BIN & IMDB-MUL & PROTEINS & DD \\ \midrule
DropNode &  $66.40  {\scriptstyle \pm  5.51}$ & $44.46  {\scriptstyle \pm  2.13}$ & $69.18  {\scriptstyle \pm  4.87}$ & $65.79  {\scriptstyle \pm  3.23}$ & $64.80  {\scriptstyle \pm  5.01}$ & $43.06  {\scriptstyle \pm  2.86}$ & $67.73  {\scriptstyle \pm 6.43}$ & $64.35  {\scriptstyle \pm 4.56}$  \\
DropEdge &  $66.70 {\scriptstyle \pm 5.10}$ & $43.80  {\scriptstyle \pm  3.11}$ & $69.36  {\scriptstyle \pm  5.90}$ & $68.42  {\scriptstyle \pm 4.76}$ & $63.20  {\scriptstyle \pm  6.30}$ & $41.80  {\scriptstyle \pm  3.15}$ & $68.10  {\scriptstyle \pm  5.05}$ & $67.06  {\scriptstyle \pm 2.53}$  \\
SubMix   & $69.30 {\scriptstyle \pm 3.76}$ &$46.73  {\scriptstyle \pm 2.67}$ & $69.80  {\scriptstyle \pm 4.73}$ & $68.04  {\scriptstyle \pm 7.64} $   & $63.70 {\scriptstyle \pm 5.64}$ & $43.73  {\scriptstyle \pm 3.60}$ &    $69.09  {\scriptstyle \pm 4.58}$ &  $59.18  {\scriptstyle \pm 6.29}$  \\
GeoMix   &$ 72.20  {\scriptstyle \pm 5.19}$ & $49.20  {\scriptstyle \pm 4.31}$ & $70.25  {\scriptstyle \pm 4.75}$ & $68.00  {\scriptstyle \pm 3.64}$ & $70.90  {\scriptstyle \pm 3.85}$ & $48.86  {\scriptstyle \pm 5.18} $ & $68.36  {\scriptstyle \pm 6.01}$  & $67.31 {\scriptstyle \pm  3.91}$  \\
$\mathcal{G}$-Mixup  & $68.30  {\scriptstyle \pm 5.13}$  & $45.53  {\scriptstyle \pm 4.12}$ & $61.71  {\scriptstyle \pm 5.81}$ & $51.26  {\scriptstyle \pm 8.76}$ & $63.20  {\scriptstyle \pm 5.54}$ & $44.00  {\scriptstyle \pm 4.63}$ & $46.63  {\scriptstyle \pm  5.05}$ & $43.71  {\scriptstyle \pm 7.12}$  \\ 
NoisyGNN & $70.50  {\scriptstyle \pm 4.71}$  & $40.66  {\scriptstyle \pm 3.12}$ & $69.45  {\scriptstyle \pm 4.32}$ & $64.18  {\scriptstyle \pm 5.71}$ & $63.50  {\scriptstyle \pm 5.43}$ & $38.66  {\scriptstyle \pm 4.12}$ & $69.99  {\scriptstyle \pm 3.78}$ & $63.24  {\scriptstyle \pm 5.02}$  \\ 
\method  &  $\mathbf{72.80  {\scriptstyle \pm 2.99}}$ & $\mathbf{49.36  {\scriptstyle \pm 4.53}}$ & $\mathbf{70.61  {\scriptstyle \pm 4.30}}$ & $\mathbf{68.68  {\scriptstyle \pm 3.72}}$ & $\mathbf{73.10  {\scriptstyle \pm 3.04}}$ & $\mathbf{49.53  {\scriptstyle \pm 3.54}}$ & $\mathbf{70.32  {\scriptstyle \pm 4.04}}$ &  $\mathbf{69.01  {\scriptstyle \pm 3.09}}$ \\ \bottomrule
\end{tabular}
}

\label{gen:tab:results_structure_corr}
\end{table*}

In this section, we present our results and analysis. Our experimental setup is described in Appendix \ref{gen:app:dataset_impl}.

\textbf{On the Generalization of GNNs.}  In Tables \ref{gen:tab:results_gcn} and \ref{gen:tab:results_gin},  we compare the test accuracy of our data augmentation strategy against baseline methods. We trained all baseline models using the same train/validation/test splits, GNN architectures, and hyperparameters to ensure a fair comparison. It is worth noting that the baselines exhibit high standard deviations, which is a common characteristic in graph classification tasks. Unlike node classification, graph classification is known to have a larger variance in performance metrics  \citep{errica2019fair,duval2022higherorder}. Overall, our proposed approach consistently achieves the best or highly competitive performance for most of the datasets.

Additionally, we observed that the results of the baseline methods vary depending on the GNN backbone, motivating further investigation using influence functions. As demonstrated in Theorem \ref{gen:thm:closed_formula}, the gradient, and more generally, the model architecture, significantly influence how augmented data impacts the model’s performance on the test set.

\textbf{Robustness to Structure Corruption.} Besides generalization, we assess the robustness of our data augmentation strategy, following the methodology outlined by \citep{zeng2024graph}. Specifically, we test the robustness of data augmentation strategies against graph structure corruption by randomly removing or adding 10\% or 20\% of the edges in the training set. By corrupting only the training graphs, we introduce a distributional shift between the training and testing datasets. This approach allows us to evaluate \method's ability to generalize well and predict the labels of test graphs, which can be considered OOD examples. The results of these experiments are presented in Table \ref{gen:tab:results_structure_corr} for the IMDB-BIN, IMDB-MUL, PROTEINS, and DD datasets. As noted, our data augmentation strategy exhibits the best test accuracy in all cases and improves model robustness against structure corruption.

\textbf{Influence Functions.} In Figure \ref{gen:fig:influence_scores}, we show the density distribution of the average influence of augmented data sampled using \method. These findings are consistent with the empirical results presented in Tables \ref{gen:tab:results_gcn}. For the MUTAG and PROTEINS datasets, we observe that \method's data augmentation has a positive impact on both GCN and GIN models. In contrast, for the DD dataset, \method~shows no effect on GIN, while it generates many augmented samples with positive values of the influence scores on GCN, thereby enhancing its performance. This behavior is consistent with the baselines, as most graph data augmentation strategies tend to enhance test accuracy more significantly for GCN than for GIN when applied to DD. This is an interesting phenomenon and worthy of deeper analysis. For the DD dataset, when using the GIN model at inference, we observe Softmax saturation, where the predicted class probabilities approach extreme values (close to $0$ or $1$), c.f. Appendix \ref{gen:app:softmax_saturation}. We suspect this saturation to happen due to the large average number of nodes in DD and the fact that GIN does not normalize the node representations, which may lead to a graph representation with large norms. This saturation results in the model making predictions with very high confidence. Consequently, the gradient of the loss function with respect to the input graphs eventually converges to $0$. In such cases, the influence scores become negligible, as explained in Section \ref{gen:sub_sec:influence}.

\textbf{Configuration Models.} As part of an ablation study, we propose a simple yet effective graph augmentation strategy inspired by \textit{configuration models} \citep{yang2013networks}. As shown in Theorem \ref{gen:thm:rademacher}, the objective is to control the term $ \mathbb{E}_{\mathcal{G} \sim \mathcal{D}, \widetilde{\mathcal{G}} \sim A_\lambda }\left [  \| \mathbf{h}_{\widetilde{\mathcal{G}}} -\mathbf{h}_{\mathcal{G}}\| \right ]$, which can be achieved by regulating the distance between the original and the sampled graph within the input manifold, i.e., $\mathbb{E}_{\mathcal{G} \sim \mathcal{D}, \widetilde{\mathcal{G}} \sim A_\lambda } \left [  \| \widetilde{\mathcal{G}} -\mathcal{G}  \| \right ]$. The approach involves generating a sampled version of each training graph by randomly breaking existing edges into \textit{half-edges} with probability $r$ and then randomly connecting half-edges until all edges are connected. The strength of this method lies in its simplicity and in preserving the degree distribution. If the distance norm is the $L_1$ distance between adjacency matrices, $|\mathcal{E}|  r^2$ is an upper bound of $\mathbb{E}_{\mathcal{G} \sim \mathcal{D},\widetilde{\mathcal{G}} \sim A_\lambda } \left [  \| \widetilde{\mathcal{G}} -\mathcal{G}  \| \right ]$, where $|\mathcal{E}|$ is the average number of edges. The results of this experiment are available in Appendix \ref{gen:app:config_models}.

\section{Conclusion}

We introduced \method, a novel approach for graph data augmentation that enhances both the generalization and robustness of GNNs. Our method uses Gaussian Mixture Models (GMMs) applied at the output level of the Readout function, an approach motivated by theoretical findings. Using the universal approximation property of GMMs, we can sample new graph representations to effectively control the upper bound of the Rademacher complexity, ensuring improved generalization of GNNs. Through extensive experiments on widely used datasets, we demonstrated that our approach not only exhibits strong generalization ability but also maintains robustness against structural perturbations. An additional advantage of our approach is its efficiency in terms of time complexity. Unlike baselines that generate augmented data for each individual or pair of training graphs, \method~fits the GMM to the entire training dataset at once, allowing for fast graph data augmentation without incurring significant additional backpropagation time.

\part{Conclusion}
\chapter{Conclusion} \label{ch:conclusion}

\section{Summary of contributions}
\subsection{Representation learning}  Chapter \ref{ch:Representation} introduced Centrality Graph Shift Operators (CGSOs), a novel class of graph shift operators. These operators incorporate both local and global centrality measures, such as degree, $k$-core, walk-count, and PageRank centralities, enabling flexible propagation. Spectral analysis demonstrated that CGSOs more accurately capture community structure compared to traditional degree-based GSOs, resulting in consistent performance gains across synthetic clustering tasks and a variety of node classification benchmarks.

In Chapter \ref{ch:Dynamic}, we presented Adaptive-Depth Message Passing GNNs (ADMP-GNNs), which eliminate the need to predefine the number of message-passing steps. Instead, ADMP-GNNs employ centrality-informed policies to dynamically learn the optimal propagation depth for each node. Empirical results highlight that this node-specific flexibility leads to improvements in predictive accuracy.

\subsection{Robustness}  Chapter \ref{ch:Robustness} introduced a new formulation of expected robustness for graphs, providing a framework to evaluate GNNs vulnerability to adversarial  perturbations. We derived theoretical bounds connecting a model’s robustness to the orthonormality of its weight matrices, revealing that orthonormality leads to improved resistance against attacks. Building on this insight, we proposed the Graph Convolutional Orthonormal Robust Network (GCORN), a variant of the classical GCN that enforces strict orthonormality. 

Complementing the training-time defense, Chapter \ref{ch:CRF} introduced RobustCRF, a post-hoc, model-agnostic method based on conditional random fields that make predictions more robust during inference. RobustCRF improves performance on both clean and adversarially perturbed data, and could me combined with other baselines.

\subsection{Generalization}

Finally, Chapter \ref{ch:Generalization} addresses the generalization of GNNs through the introduction of GRATIN, a novel graph data augmentation technique. GRATIN fits a Gaussian Mixture Model to the latent representations produced by a trained GNN, enabling the generation of additional graph representations in that learned space. The method is theoretically supported via an upper bound on the Rademacher complexity. Empirically, GRATIN improves test accuracy on the graph classification task and demonstrates robustness to structural noise, all by maintaining low computational overhead during both augmentation and training.

\section{Broader Impact}

The methods introduced in this thesis aim to enhance the robustness, and generalization of GNNs, with the goal of making them more reliable and adaptable for real-world use.

In terms of representation learning, the development of Centrality Graph Shift Operators (CGSOs) provides more flexibility for GNNs to capture structural information in graphs. This has broad implications for domains such as social network analysis (e.g., Twitter), music recommendation systems, e.g., Deezer \citep{salha2022contributions} and biological network modeling \citep{liu2023deep}, where understanding the importance and roles of nodes within a graph is crucial. By aligning message passing with global or local structural properties, CGSOs enable more interpretable and efficient graph representations.

For robustness, techniques like GCORN and RobustCRF help mitigate the vulnerabilities of GNNs to adversarial manipulations. This is especially significant in safety-critical applications such as navigation systems, e.g., Google Maps \citep{lange2020traffic} or fraud detection. These methods contribute to the broader effort of building trustworthy machine learning systems, independent of specific attack models or retraining procedures.

On the front of generalization, the GRATIN framework addresses a key challenge in real-world GNN deployment: the ability to perform well under limited data or when confronted with distributional shifts. Applications in evolving social networks and drug discovery \citep{liuexplanations,kaur2024molecular}, where molecular graphs often vary, stand to benefit from GRATIN’s ability to augment training data. Importantly, this is achieved without introducing significant computational cost.

Together, the contributions of this thesis open up new directions toward GNNs that are not only more accurate but also more robust, interpretable, and deployable in real-world scenarios. At the same time, they highlight new ethical considerations, such as the potential for structural biases, or the risk of overfitting to learned latent representations. These call for attention to fairness in future GNN research and deployment.

\section{Future directions} 
This thesis opens several avenues for further investigation. Below we highlight a few especially promising directions:

\begin{itemize}
    \item \textbf{Generalization to Node Classification under Non-IID Settings.} Most classical generalization bounds, such as those based on Rademacher complexity or McDiarmid’s inequality, assume that training and test examples are drawn independently and identically distributed. While this assumption is reasonable for graph-level tasks, it breaks down in semi-supervised node classification, i.e.,  nodes are connected by edges, and label information propagates through the same graph on which we test. Developing a theory of generalization for node classification task should accounts for this inherent dependency structure in order to derive new concentration inequalities. Such a framework would enable rigorous bounds on node-level error and guide the design of augmentation or regularization techniques for non-IID node distributions.
    \item \textbf{3D Geometric Deep Learning.}  An exciting direction is to develop E(3) and SE(3) Equivariant Graph Shift Operators tailored to three-dimensional data \cite{duval2023hitchhiker}. Current GNNs achieve permutation equivariance, but modeling 3D structures (e.g., molecular conformations, point clouds, meshes) demands invariance to rotations and translations.  By designing GSOs that explicitly incorporate spatial coordinates, edge angles, and embedding them into the message-passing scheme of Equivariant GNN layers—we can respect the symmetries of the Special Euclidean group and enhance the network’s ability to capture geometric relations.  Such 3D-specific GSOs could update both node features and coordinate embeddings at each layer, enabling more accurate property prediction in physics and chemistry, and more faithful modeling of complex engineering or biological structures.
    \item \textbf{Transformers and Large Language Models for Structured Data.} Recent advances in Large Language Models (LLMs) open the door to in-context learning for discrete, structured data tasks. By prompting an LLM with examples of graph transformations or denoising steps, one could perform tasks like diffusion-based graph generation or adaptive message passing without extensive retraining. Exploring how to leverage LLMs for link prediction, node classification, or discrete diffusion on graphs represents a novel, training-free paradigm for graph machine learning.
\end{itemize}

\defbibheading{bibintoc}[\bibname]{%
  \phantomsection
  \manualmark
  \markboth{\spacedlowsmallcaps{#1}}{\spacedlowsmallcaps{#1}}%
  \addtocontents{toc}{\protect\vspace{\beforebibskip}}%
  \addcontentsline{toc}{chapter}{\tocEntry{#1}}%
  \chapter*{#1}%
}
\printbibliography[heading=bibintoc]

\include{chapter_Representation/Content/reference}
\appendix
\renewcommand{\thechapter}{\alph{chapter}}

\part{Appendices}
\chapter[Appendix: Rethinking GSOs in GNNs for Graph Representation Learning]{Appendix: Rethinking Graph Shift Operators in GNNs for Graph Representation Learning}
\section{Datasets and Implementation Details}\label{cgsO:app_data_impl} 
In this section, we present the benchmark datasets used for our experiments, and the process used for the training.
\subsection{Statistics of the Node Classification Datasets}

We use ten widely used datasets in the GNN literature. In particular, we run experiments on the node classification task using the citation networks Cora, CiteSeer, and PubMed \citep{dataset_node_classification}, the co-authorship networks CS and Physiscs \citep{cs_data}, the citation network between Computer Science arXiv papers OGBN-Arxiv  \citep{hu2020open}, the Amazon Computers and Amazon Photo networks \citep{cs_data}, the non-homophilous datasets Penn94 \citep{traud2012social}, genius \citep{lim2021expertise}, deezer-europe \citep{rozemberczki2020characteristic} and arxiv-year \citep{hu2020open}, and the disassortative datasets Chameleon, Squirrel \citep{rozemberczki2021multi}, 
 and Cornell, Texas, Wisconsin from the WebKB dataset \citep{lim2021large}. 
Characteristics and information about the datasets utilized in the node classification part of the study are presented in Table \ref{cgsO:tab:data_statistics}.

\begin{table}[h]

\caption{Statistics of the node classification datasets used in our experiments.}
\label{cgsO:tab:data_statistics}
\vskip 0.15in
\begin{center}
\begin{tabular}{lrrrrr}
\hline
Dataset & \#Features & \#Nodes & \#Edges & \#Classes & Edge Homophily \\
\hline
Cora    & 1,433 & 2,708   & 5,208    & 7 & 0.809 \\
CiteSeer   & 3,703 & 3,327 & 4,552 & 6 & 0.735\\
PubMed    & 500 & 19,717 & 44,338 & 3 & 0.802\\
CS    & 6,805 & 18,333 & 81,894 & 15 & 0.808\\
arxiv-year    & 128 & 169,343 & 1,157,799 & 5 &  0.218\\
chameleon  &   2,325  &   2,277  &   62,792  &   5  &   0.231\\
Cornell  &   1,703  &   183  &   557  &   5  &   0.132 \\
deezer-europe  &   31,241  &   28,281  &   185,504  &   2  &   0.525 \\
squirrel  &   2,089  &   5,201  &   396,846  &   5  &   0.222 \\
Wisconsin  &   1,703  &   251  &   916  &   5  &   0.206  \\
Texas  &   1,703  &   183  &   574  &   5  &   0.111 \\
Photo  &   745  &   7,650  &   238,162  &   8  &   0.827 \\
ogbn-arxiv  &   128  &   169,343  &   2,315,598  &   40  &   0.654  \\
Computers  &   767  &   13752  &   491,722  &   10  &   0.777 \\
Physics  &   8,415  &   34,493  &   495,924  &   5  &   0.931 \\
Penn94  &   4,814  &   41,554  &   2,724,458  &   3  &   0.470  \\
\hline
\end{tabular}

\end{center}
\vskip -0.1in

\end{table}
\subsection{Implementation Details}

We train all the models  using the Adam optimizer \citep{kingma_adam}. To account for the impact of random initialization, each experiment was repeated 10 times, and the mean and standard deviation of the results were reported. The experiments have been run on both a NVIDIA A100 GPU and a RTX A6000 GPU.

\textbf{Training of our CGNN.}
We train our model using the Adam optimizer \citep{kingma_adam}, with a weight decay on the parameters of $5 \times 10^{-4}$, an initial learning rate of $0.005$ for the exponential parameters and an initial learning rate of $0.01$ for all other model parameters. We repeated the training 10 times to test the stability of the model. We tested 7 initialization of the weights $(m_1,m_2,m_3,e_1,e_2,e_3,a)$. These initializations are reported in Table \ref{cgsO:tab:weights_initializations} in Appendix \ref{cgsO:app_data_impl}, and correspond to classical GSOs when the chosen centrality is the degree. 
For the  Cora, CiteSeer, and Pubmed datasets, we used the provided train/validation/test splits. For the remaining datasets, we followed the
framework of \citet{lim2021large,rozemberczki2021multi}.

\subsection{Weights Initialization}
In this part, we present the different initializations of CGSO. When the chosen centrality is the degree, i.e. $\mathbf{V}=\mathbf{D}$, the  initializations corresponds to popular classical GSO \citep{dasoulas2021learning}.

\begin{table}[h]

\caption[Differenet initialization of the weights in CGSO]{Differenet initialization of the weights $(m_1,m_2,m_3,e_1,e_2,e_3,a)$.}
\label{cgsO:tab:weights_initializations}
\begin{center}
\resizebox{\textwidth}{!}{
\begin{tabular}{lll}
\toprule
Initialization of  $(m_1,m_2,m_3,e_1,e_2,e_3,a)$ & Corresponding GSO & Description when  $\mathcal{V}=\mathcal{D}$   \\
\midrule
$(0, 1, 0, 0, 0, 0, 0)$   &  $\mathbf{A}(\mathbf{V}) = \mathbf{A}$ & Adjacency matrix\\
$(1, -1, 0, 1, 0, 0, 0)$ &  $\mathbf{L}(\mathbf{V}) =\mathbf{V}-\mathbf{A}$& Unnormalised Laplacian matrix \\
$(1, 1, 0, 1, 0, 0, 0) $  & $\mathbf{Q}(\mathbf{V}) =\mathbf{V}+\mathbf{A}$ & Signless Laplacian matrix\\
$(0, -1, 1, 0, -1, 0, 0) $   & $\mathbf{L_{rw}}(\mathbf{V}) =\mathbf{I} - \mathbf{V}^{-1}\mathbf{A}$ & Random-walk Normalised Laplacian\\
$(0, -1, 1, 0, -1/2, -1/2, 0) $    &$\mathbf{L_{sym}}(\mathbf{V}) =\mathbf{I} - \mathbf{V}^{-1/2}\mathbf{A}V^{-1/2}$ & Symmetric Normalised Laplacian\\
$(0, 1, 0, 0, -1/2, -1/2, 1) $     & $\mathbf{\hat{A}}(\mathbf{V}) = \mathbf{V}^{-1/2}\mathbf{A}_1\mathbf{V}^{-1/2}$&Normalised Adjacency matrix \\
$(0, 1, 0, 0, -1, 0, 0)  $     & $ \mathbf{H}(\mathbf{V}) =\mathbf{V}^{-1}\mathbf{A}$& Mean Aggregation Operator\\
\bottomrule
\end{tabular}
}
\end{center}
\end{table}

\subsection{Hyperparameter Configurations}
For a more balanced comparison, however, we use the same training procedure for all the models. The hyperparameters in each dataset where performed using a Grid search on the classical GCN (i.e. with the GSO : Normalised adjacency) over the following search space:

\begin{itemize}
    \item Hidden size : $[16,32,64,128,256,512],$
    \item Learning rate : $[0.1, 0.01, 0.001],$
    \item Dropout probability: $[0.2,0.3,0.4,0.5,0.6,0.7,0.8].$
\end{itemize}

The number of layers was fixed to 2. The optimal hyperparameters can be found in Table \ref{cgsO:tab:gcn_tuned_hyperparams}.
\begin{table*}[h]
\caption{Hyperparameters used in our experiments.}
\label{cgsO:tab:gcn_tuned_hyperparams}
\vskip 0.15in
\begin{center}
\begin{tabular}{lccc}
\toprule
Dataset & Hidden Size & Learning Rate & Dropout Probability   \\
\midrule
Cora    & 64 & 0.01 & 0.8   \\
CiteSeer    &64 & 0.01 & 0.4   \\
PubMed     &64 & 0.01 & 0.2   \\\
CS     & 512 & 0.01 & 0.4 \\
arxiv-year   & 512 & 0.01  & 0.2 \\
chameleon   & 512 & 0.01  & 0.2 \\
cornell   & 512 & 0.01  & 0.2 \\
deezer-europe   & 512 & 0.01  & 0.2 \\
squirrel   & 512 & 0.01  & 0.2 \\
Wisconsin   &  512 & 0.01  & 0.2 \\
Texas   & 512 & 0.01  & 0.2 \\
Photo     & 512 & 0.01 & 0.6 \\
OGBN-Arxiv    & 512 & 0.01 & 0.5  \\
Computers     & 512 & 0.01 & 0.2 \\
Physics   & 512 & 0.01 & 0.4 \\
Penn94     & 64 & 0.01 & 0.2 \\

\bottomrule
\end{tabular}
\end{center}
\vskip -0.1in
\end{table*}

\section{Additional Results for the Node Classification Task}\label{cgsO:node_classification_results_appendix} 
To further evaluate our \textit{CGCN} and \textit{CGATv2}, we compute its performance on additional datasets. The results of this study are presented in Table \ref{cgsO:tab:additional_CA_GCN}. 

\begin{table*}[h] 
\centering
\caption[Performance of CGCN, CGATv2 and other vanilla models on additional datasets]{Classification accuracy ($\pm$ standard deviation) of the models on different benchmark node classification datasets. The higher the accuracy (in \%) the better the model. \textcircled{1} GCN Based models \textcircled{2} Other Vanilla GNN baselines \textcircled{3} CGCN \textcircled{4} CGATv2. Highlighted are the \textbf{first}, \underline{second} best results. OOM means \textit{Out of memory}.}
\label{cgsO:tab:additional_CA_GCN}
\resizebox{\textwidth}{!}{%
\begin{tabular}{c|l|llllllll}
\toprule
\multicolumn{1}{l}{} & Model & Cora  & Texas & Photo  & ogbn-ariv & CS  & Computers  & Physics & Penn94  \\ 
\hline

\multirow{7}{*}{\textcircled{1}}   
    & GCN w/ $\mathbf{A}$ & $78.61 {\scriptstyle \pm 0.51}$ & $63.51 {\scriptstyle \pm 2.18}$ & $82.31 {\scriptstyle \pm 2.61}$ & $13.23 {\scriptstyle \pm 6.44}$ & $87.70 {\scriptstyle \pm 1.25}$ & $69.32 {\scriptstyle \pm 3.64}$ & $88.92 {\scriptstyle \pm 1.93}$ & $52.35 {\scriptstyle \pm 0.36}$ \\

    & GCN w/ $\mathbf{L}$ & $31.57 {\scriptstyle \pm 0.41}$ & $\mathbf{84.32 {\scriptstyle \pm 2.65}}$ & $27.42 {\scriptstyle \pm 6.23}$ & $10.91 {\scriptstyle \pm 1.49}$ & $23.75 {\scriptstyle \pm 3.22}$ & $26.27 {\scriptstyle \pm 3.89}$ & $35.31 {\scriptstyle \pm 3.71}$ & $65.31 {\scriptstyle \pm 0.59}$ \\

    & GCN w/ $\mathbf{Q}$ & $77.32 {\scriptstyle \pm 0.50}$ & $60.54 {\scriptstyle \pm 1.32}$ & $77.06 {\scriptstyle \pm 6.73}$ & $10.50 {\scriptstyle \pm 1.97}$ & $89.42 {\scriptstyle \pm 1.31}$ & $47.72 {\scriptstyle \pm 18.37}$ & $90.69 {\scriptstyle \pm 2.13}$ & $53.46 {\scriptstyle \pm 2.16}$ \\

    & GCN w/ $\mathbf{L_{rw}}$ & $26.59 {\scriptstyle \pm 1.11}$ & $78.38 {\scriptstyle \pm 2.09}$ & $24.60 {\scriptstyle \pm 4.21}$ & $8.07 {\scriptstyle \pm 0.07}$ & $26.34 {\scriptstyle \pm 4.09}$ & $13.76 {\scriptstyle \pm 3.96}$ & $28.19 {\scriptstyle \pm 3.75}$ & $69.82 {\scriptstyle \pm 0.44}$ \\

    & GCN w/ $\mathbf{L_{sym}}$ & $26.79 {\scriptstyle \pm 0.50}$ & $71.35 {\scriptstyle \pm 1.32}$ & $22.82 {\scriptstyle \pm 2.67}$ & $20.18 {\scriptstyle \pm 0.24}$ & $24.39 {\scriptstyle \pm 1.96}$ & $16.06 {\scriptstyle \pm 5.19}$ & $30.94 {\scriptstyle \pm 3.11}$ & $70.57 {\scriptstyle \pm 0.30}$ \\

    & GCN w/ $\mathbf{\hat{A}}$ & $\mathbf{80.84 {\scriptstyle \pm 0.40}}$ & $60.81 {\scriptstyle \pm 1.81}$ & $78.94 {\scriptstyle \pm 1.65}$ & $65.80 {\scriptstyle \pm 0.14}$ & $91.52 {\scriptstyle \pm 0.75}$ & $68.91 {\scriptstyle \pm 3.00}$ & $\mathbf{93.72 {\scriptstyle \pm 0.80}}$ & $74.60 {\scriptstyle \pm 0.42}$ \\

    & GCN w/ $\mathbf{H}$ & $\underline{80.15 {\scriptstyle \pm 0.37}}$ & $59.46 {\scriptstyle \pm 0.00}$ & $73.95 {\scriptstyle \pm 4.75}$ & $63.34 {\scriptstyle \pm 0.15}$ & $90.98 {\scriptstyle \pm 1.84}$ & $62.01 {\scriptstyle \pm 4.36}$ & $92.16 {\scriptstyle \pm 1.12}$ & $71.78 {\scriptstyle \pm 0.47}$ \\  
\midrule
\multirow{5}{*}{\textcircled{2}}   
    & GIN & $79.06 {\scriptstyle \pm 0.47}$ & $57.03 {\scriptstyle \pm 1.89}$ & $83.00 {\scriptstyle \pm 2.52}$ & $9.30 {\scriptstyle \pm 6.42}$ & $89.53 {\scriptstyle \pm 1.20}$ & $55.89 {\scriptstyle \pm 13.45}$ & $89.15 {\scriptstyle \pm 2.44}$ & OOM \\

    & GAT & $77.73 {\scriptstyle \pm 1.83}$ & $52.16 {\scriptstyle \pm 6.74}$ & $71.56 {\scriptstyle \pm 3.48}$ & $67.36 {\scriptstyle \pm 0.13}$ & $67.67 {\scriptstyle \pm 3.96}$ & $59.73 {\scriptstyle \pm 3.59}$ & $80.91 {\scriptstyle \pm 4.48}$ & $73.85 {\scriptstyle \pm 1.38}$ \\

    & GATv2 & $74.53 {\scriptstyle \pm 2.48}$ & $48.11 {\scriptstyle \pm 3.78}$ & $73.49 {\scriptstyle \pm 2.49}$ & $68.14 {\scriptstyle \pm 0.07}$ & $70.13 {\scriptstyle \pm 4.92}$ & $58.18 {\scriptstyle \pm 4.76}$ & $83.28 {\scriptstyle \pm 3.68}$ & $\underline{75.54 {\scriptstyle \pm 2.54}}$ \\

    & PNA & $56.67 {\scriptstyle \pm 10.53}$ & $63.51 {\scriptstyle \pm 4.05}$ & $16.75 {\scriptstyle \pm 5.59}$ & OOM & OOM & $13.62 {\scriptstyle \pm 6.39}$ & OOM & OOM \\  
\hline  

\multirow{4}{*}{\textcircled{3}}   
    & CGCN w/ $\mathbf{D}$ & $79.45 {\scriptstyle \pm 0.58}$ & $81.89 {\scriptstyle \pm 9.38}$ & $\underline{88.78 {\scriptstyle \pm 1.74}}$ & $69.09 {\scriptstyle \pm 0.21}$ & $91.28 {\scriptstyle \pm 1.29}$ & $\mathbf{79.26 {\scriptstyle \pm 1.87}}$ & $92.51 {\scriptstyle \pm 1.16}$ & $73.06 {\scriptstyle \pm 0.34}$ \\

    & CGCN w/ $\mathbf{V}_{\text{\textit{core}}}$ & $79.80 {\scriptstyle \pm 0.43}$ & $77.84 {\scriptstyle \pm 5.51}$ & $88.53 {\scriptstyle \pm 1.40}$ & $65.54 {\scriptstyle \pm 0.57}$ & $91.37 {\scriptstyle \pm 1.18}$ & $77.35 {\scriptstyle \pm 2.67}$ & $91.98 {\scriptstyle \pm 1.49}$ & $\mathbf{78.11 {\scriptstyle \pm 3.74}}$ \\

    & CGCN w/ $\mathbf{V}_{\ell\text{\textit{-walks}}}$ & $79.52 {\scriptstyle \pm 0.35}$ & $78.11 {\scriptstyle \pm 5.82}$ & $83.72 {\scriptstyle \pm 2.03}$ & $22.54 {\scriptstyle \pm 8.22}$ & $89.87 {\scriptstyle \pm 1.20}$ & $68.56 {\scriptstyle \pm 3.39}$ & $89.84 {\scriptstyle \pm 2.74}$ & $68.44 {\scriptstyle \pm 0.37}$ \\

    & CGCN w/ $\mathbf{V}_{\text{\textit{PR}}}$ & $79.51 {\scriptstyle \pm 15.01}$ & $82.70 {\scriptstyle \pm 4.95}$ & $81.28 {\scriptstyle \pm 6.08}$ & $68.56 {\scriptstyle \pm 0.18}$ & $88.76 {\scriptstyle \pm 30.68}$ & $65.54 {\scriptstyle \pm 6.43}$ & $89.64 {\scriptstyle \pm 10.3}$ & $72.59 {\scriptstyle \pm 0.84}$ \\  
\midrule 

\multirow{4}{*}{\textcircled{4}}   
    & CGATv2 w/ $\mathbf{D}$ & $79.07 {\scriptstyle \pm 0.64}$ & $82.70 {\scriptstyle \pm 5.30}$ & $87.97 {\scriptstyle \pm 1.77}$ & $70.09 {\scriptstyle \pm 0.10}$ & $\underline{91.48 {\scriptstyle \pm 1.05}}$ & $78.62 {\scriptstyle \pm 2.35}$ & $91.32 {\scriptstyle \pm 1.18}$ & $72.81 {\scriptstyle \pm 0.36}$ \\

    & CGATv2 w/ $\mathbf{V}_{\text{\textit{core}}}$ & $79.03 {\scriptstyle \pm 0.96}$ & $83.78 {\scriptstyle \pm 6.62}$ & $\mathbf{89.72 {\scriptstyle \pm 1.54}}$ & $\underline{69.93 {\scriptstyle \pm 0.13}}$ & $\mathbf{91.91 {\scriptstyle \pm 1.06}}$ & $77.31 {\scriptstyle \pm 3.33}$ & $91.15 {\scriptstyle \pm 1.07}$ & $72.86 {\scriptstyle \pm 0.41}$ \\

    & CGATv2 w/ $\mathbf{V}_{\ell\text{\textit{-walks}}}$ & $78.58 {\scriptstyle \pm 0.58}$ & $79.73 {\scriptstyle \pm 4.72}$ & $88.11 {\scriptstyle \pm 2.02}$ & $\mathbf{70.51 {\scriptstyle \pm 0.24}}$ & $90.73 {\scriptstyle \pm 1.46}$ & $\underline{79.09 {\scriptstyle \pm 1.66}}$ & $89.98 {\scriptstyle \pm 1.37}$ & $72.79 {\scriptstyle \pm 0.43}$ \\

    & CGATv2 w/ $\mathbf{V}_{\text{\textit{PR}}}$ & $78.60 {\scriptstyle \pm 0.38}$ & $\underline{83.78 {\scriptstyle \pm 4.98}}$ & $88.38 {\scriptstyle \pm 2.09}$ & $69.26 {\scriptstyle \pm 0.12}$ & $91.77 {\scriptstyle \pm 1.00}$ & $74.95 {\scriptstyle \pm 3.05}$ & $\underline{92.73 {\scriptstyle \pm 1.44}}$ & $75.16 {\scriptstyle \pm 0.69}$ \\  
\bottomrule
\end{tabular}
}
\end{table*}

\section{Simple Graph Convolutional Networks}\label{cgsO:app:sgc}
In Tables \ref{cgsO:tab:sgc_1} and \ref{cgsO:tab:sgc_2}, we present the results of our centrality-aware Simple Graph Convolutional Networks \textit{CGSC} of 2 layers. As noticed in most cases, by incorporating our CGSO, we outperform the classical SGC. To also understand the effect of the centrality on the oversmoothing effect, we analyzed the variation of Dirichlet Energy \citep{zhao2024understanding} of \textit{CGSC} across different numbers of layers. As noticed, while the centrality has a lower effect on the oversmoothing in the homophilous dataset Cora, we notice a larger impact on the heterophilious dataset Chameleon.

\begin{table*}[h] 
\centering
\caption[Classification accuracy of CSGC on additional datasets]{Classification accuracy ($\pm$ standard deviation) of the models on different benchmark node classification datasets. The higher the accuracy (in \%) the better the model. \textcircled{1}~CSGC with nodes centrality,  \textcircled{2}~SGC. Highlighted are the \textbf{best results}.}
\label{cgsO:tab:sgc_1}
\resizebox{\textwidth}{!}{%
\begin{tabular}{c|l|llllllll}\toprule
\multicolumn{1}{l}{} & Model & CiteSeer  & PubMed & arxiv-year  & chamelon & Cornell  & deezer-europe  & squirrel & Wisconsin  \\ 
\hline

\multirow{4}{*}{\textcircled{1}}   
    & CSGC w/ $\mathbf{D}$ 
        & $67.70 {\scriptstyle \pm 0.17}$ 
        & $77.37 {\scriptstyle \pm 0.25}$ 
        & $35.01 {\scriptstyle \pm 0.16}$ 
        & $59.10 {\scriptstyle \pm 1.66}$ 
        & $72.70 {\scriptstyle \pm 4.26}$ 
        & $58.22 {\scriptstyle \pm 0.47}$ 
        & $\mathbf{40.12 {\scriptstyle \pm 1.69}}$ 
        & $75.88 {\scriptstyle \pm 4.96}$  \\

    & CSGC w/ $\mathbf{V}_{\text{\textit{core}}}$    
        & $66.85 {\scriptstyle \pm 0.15}$ 
        & $\mathbf{78.19 {\scriptstyle \pm 0.12}}$ 
        & $\mathbf{37.71 {\scriptstyle \pm 0.17}}$ 
        & $\mathbf{63.11 {\scriptstyle \pm 4.56}}$ 
        & $72.16 {\scriptstyle \pm 5.55}$ 
        & $61.29 {\scriptstyle \pm 0.50}$ 
        & $38.66 {\scriptstyle \pm 2.27}$ 
        & $75.10 {\scriptstyle \pm 4.12}$  \\

    & CSGC w/ $\mathbf{V}_{\ell\text{\textit{-walks}}}$  
        & $\mathbf{67.09 {\scriptstyle \pm 0.05}}$ 
        & $77.50 {\scriptstyle \pm 0.18}$ 
        & $36.67 {\scriptstyle \pm 0.22}$ 
        & $45.26 {\scriptstyle \pm 2.51}$ 
        & $\mathbf{74.32 {\scriptstyle \pm 6.07}}$ 
        & $59.69 {\scriptstyle \pm 0.50}$ 
        & $27.85 {\scriptstyle \pm 1.38}$ 
        & $\mathbf{81.76 {\scriptstyle \pm 3.73}}$ \\

    & CSGC w/ $\mathbf{V}_{\text{\textit{PR}}}$   
        & $64.91 {\scriptstyle \pm 0.47}$ 
        & $76.47 {\scriptstyle \pm 0.37}$ 
        & $23.87 {\scriptstyle \pm 0.51}$ 
        & $55.18 {\scriptstyle \pm 3.36}$ 
        & $69.46 {\scriptstyle \pm 5.14}$ 
        & $58.94 {\scriptstyle \pm 0.48}$ 
        & $26.73 {\scriptstyle \pm 2.25}$ 
        & $75.29 {\scriptstyle \pm 4.31}$ \\ 

\midrule 

\multirow{1}{*}{\textcircled{2}}   
    & SGC  
        & $64.96 {\scriptstyle \pm 0.10}$ 
        & $75.72 {\scriptstyle \pm 0.12}$ 
        & $26.61 {\scriptstyle \pm 0.24}$ 
        & $38.44 {\scriptstyle \pm 4.41}$ 
        & $45.41 {\scriptstyle \pm 5.77}$ 
        & $\mathbf{62.66 {\scriptstyle \pm 0.48}}$ 
        & $19.88 {\scriptstyle \pm 0.79}$ 
        & $53.53 {\scriptstyle \pm 8.09}$ \\  

\bottomrule
\end{tabular}
} 
\end{table*}

\begin{table*}[h] 
\centering
\caption[Classification accuracy of CSGC on additional datasets]{Classification accuracy ($\pm$ standard deviation) of the models on different benchmark node classification datasets. The higher the accuracy (in \%) the better the model. \textcircled{1}~CSGC with nodes centrality,  \textcircled{2}~SGC. Highlighted are the \textbf{best results}.}
\label{cgsO:tab:sgc_2}
\resizebox{\textwidth}{!}{%
\begin{tabular}{c|l|llllllll}\toprule
\multicolumn{1}{l}{} & Model & Cora  & Texas & Photo  & ogbn-arxiv & CS  & Computers  & Physics & Penn94  \\
\hline

\multirow{4}{*}{\textcircled{1}}   
    & CSGC w/ $\mathbf{D}$  
        & $\mathbf{80.10 {\scriptstyle \pm 0.11}}$ 
        & $76.76 {\scriptstyle \pm 3.24}$ 
        & $\mathbf{89.38 {\scriptstyle \pm 1.81}}$ 
        & $\mathbf{67.94 {\scriptstyle \pm 0.06}}$ 
        & $\mathbf{92.29 {\scriptstyle \pm 1.04}}$ 
        & $\mathbf{79.04 {\scriptstyle \pm 1.94}}$ 
        & $\mathbf{92.32 {\scriptstyle \pm 1.2}}$ 
        & $\mathbf{78.84 {\scriptstyle \pm 4.15}}$   \\

    & CSGC w/ $\mathbf{V}_{\text{\textit{core}}}$     
        & $78.80 {\scriptstyle \pm 0.17}$ 
        & $77.30 {\scriptstyle \pm 3.86}$ 
        & $88.58 {\scriptstyle \pm 1.68}$ 
        & $62.54 {\scriptstyle \pm 0.16}$ 
        & $91.82 {\scriptstyle \pm 1.10}$ 
        & $76.46 {\scriptstyle \pm 2.29}$ 
        & $91.71 {\scriptstyle \pm 1.63}$ 
        & $76.25 {\scriptstyle \pm 1.21}$   \\

    & CSGC w/ $\mathbf{V}_{\ell\text{\textit{-walks}}}$  
        & $77.32 {\scriptstyle \pm 0.29}$ 
        & $\mathbf{80.27 {\scriptstyle \pm 5.41}}$ 
        & $88.78 {\scriptstyle \pm 2.69}$ 
        & $66.41 {\scriptstyle \pm 0.05}$ 
        & $91.96 {\scriptstyle \pm 0.84}$ 
        & $76.17 {\scriptstyle \pm 4.92}$ 
        & $91.71 {\scriptstyle \pm 1.58}$ 
        & $73.20 {\scriptstyle \pm 0.36}$  \\

    & CSGC w/ $\mathbf{V}_{\text{\textit{PR}}}$  
        & $76.92 {\scriptstyle \pm 0.39}$ 
        & $77.30 {\scriptstyle \pm 4.86}$ 
        & $84.33 {\scriptstyle \pm 3.06}$ 
        & $44.82 {\scriptstyle \pm 1.16}$ 
        & $90.24 {\scriptstyle \pm 0.86}$ 
        & $61.51 {\scriptstyle \pm 2.71}$ 
        & $91.57 {\scriptstyle \pm 1.70}$ 
        & $77.24 {\scriptstyle \pm 0.67}$ \\ 

\midrule 

\multirow{1}{*}{\textcircled{2}}   
    & SGC  
        & $78.79 {\scriptstyle \pm 0.13}$ 
        & $58.65 {\scriptstyle \pm 4.20}$ 
        & $24.0 {\scriptstyle \pm 11.82}$ 
        & $60.48 {\scriptstyle \pm 0.14}$ 
        & $70.78 {\scriptstyle \pm 5.47}$ 
        & $11.34 {\scriptstyle \pm 11.67}$ 
        & $91.69 {\scriptstyle \pm 1.48}$ 
        & $66.63 {\scriptstyle \pm 0.62}$ \\  

\bottomrule
\end{tabular}
} 
\end{table*}

\begin{figure}[h]
    \centering
    \includegraphics[width=\textwidth]{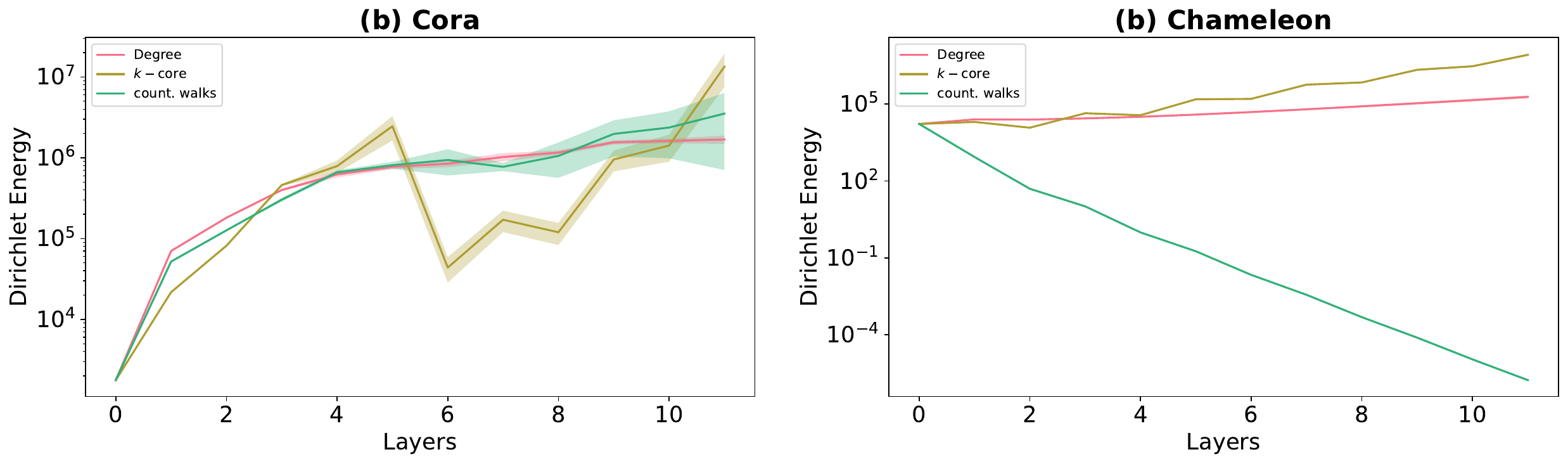}
    \caption[Dirichlet Energy variation]{Dirichlet Energy variation with layers in (a) Cora and (b) Chamelon.}
    \label{cgsO:fig:enter-label}
\end{figure}

\section{Combining Local and Global Centralities}\label{cgsO:combine_centralities}

\begin{table}[h] 
\centering
\caption{Classification accuracy ($\pm$ standard deviation) of the models on different benchmark node classification datasets. The higher the accuracy (in \%) the better the model. }
\label{cgsO:tab:comb_centralities_2}
\resizebox{\textwidth}{!}{%
\begin{tabular}{l|llllllll}\toprule
 Model & Cora  & Texas & Photo  & ogbn-arxiv & CS  & Computers  & Physics & Penn94  \\ 
\hline

CGCN w/ $\mathbf{D}$  
    & $79.45 {\scriptstyle \pm 0.58}$ 
    & $81.89 {\scriptstyle \pm 9.38}$ 
    & $88.78 {\scriptstyle \pm 1.74}$ 
    & $69.09 {\scriptstyle \pm 0.21}$ 
    & $91.28 {\scriptstyle \pm 1.29}$ 
    & $\mathbf{79.26 {\scriptstyle \pm 1.87}}$ 
    & $\mathbf{92.51 {\scriptstyle \pm 1.16}}$ 
    & $73.06 {\scriptstyle \pm 0.34}$ \\

CGCN w/ $\mathbf{V}_{\text{\textit{core}}}$   
    & $79.80 {\scriptstyle \pm 0.43}$ 
    & $77.84 {\scriptstyle \pm 5.51}$ 
    & $88.53 {\scriptstyle \pm 1.40}$ 
    & $65.54 {\scriptstyle \pm 0.57}$ 
    & $91.37 {\scriptstyle \pm 1.18}$ 
    & $77.35 {\scriptstyle \pm 2.67}$ 
    & $91.98 {\scriptstyle \pm 1.49}$ 
    & $78.11 {\scriptstyle \pm 3.74}$ \\

CGCN w/ $\mathbf{V}_{\ell\text{\textit{-walks}}}$  
    & $79.52 {\scriptstyle \pm 0.35}$ 
    & $78.11 {\scriptstyle \pm 5.82}$ 
    & $83.72 {\scriptstyle \pm 2.03}$ 
    & $22.54 {\scriptstyle \pm 8.22}$ 
    & $89.87 {\scriptstyle \pm 1.20}$ 
    & $68.56 {\scriptstyle \pm 3.39}$ 
    & $89.84 {\scriptstyle \pm 2.74}$ 
    & $68.44 {\scriptstyle \pm 0.37}$ \\

CGCN w/ $\mathbf{V}_{\text{\textit{PR}}}$ 
    & $79.51 {\scriptstyle \pm 15.01}$ 
    & $\mathbf{82.70 {\scriptstyle \pm 4.95}}$ 
    & $81.28 {\scriptstyle \pm 6.08}$ 
    & $68.56 {\scriptstyle \pm 0.18}$ 
    & $88.76 {\scriptstyle \pm 30.68}$ 
    & $65.54 {\scriptstyle \pm 6.43}$ 
    & $89.64 {\scriptstyle \pm 10.3}$ 
    & $72.59 {\scriptstyle \pm 0.84}$ \\ 

\midrule 

CGCN w/  $\mathbf{D}~~ \& ~~\mathbf{V}_{\text{\textit{core}}}$     
    & $\mathbf{79.88 {\scriptstyle \pm 0.38}}$ 
    & $78.92 {\scriptstyle \pm 4.32}$ 
    & $\mathbf{89.06 {\scriptstyle \pm 1.28}}$ 
    & $67.67 {\scriptstyle \pm 0.26}$ 
    & $91.63 {\scriptstyle \pm 0.95}$ 
    & $78.41 {\scriptstyle \pm 1.94}$ 
    & $91.28 {\scriptstyle \pm 3.17}$ 
    & $\mathbf{80.28 {\scriptstyle \pm 2.93}}$ \\

CGCN w/ $\mathbf{D}~~ \& ~~\mathbf{V}_{\ell\text{\textit{-walks}}}$ 
    & $79.38 {\scriptstyle \pm 0.72}$ 
    & $81.89 {\scriptstyle \pm 4.69}$ 
    & $86.78 {\scriptstyle \pm 2.75}$ 
    & $\mathbf{69.57 {\scriptstyle \pm 0.24}}$ 
    & $\mathbf{91.78 {\scriptstyle \pm 1.04}}$ 
    & $78.39 {\scriptstyle \pm 2.36}$ 
    & $91.2 {\scriptstyle \pm 1.56}$ 
    & $72.5 {\scriptstyle \pm 0.48}$ \\

CGCN w/ $\mathbf{D}~~ \& ~~\mathbf{V}_{\text{\textit{PR}}}$  
    & $79.84 {\scriptstyle \pm 0.4}$ 
    & $78.11 {\scriptstyle \pm 2.55}$ 
    & $82.76 {\scriptstyle \pm 2.06}$ 
    & $21.28 {\scriptstyle \pm 9.89}$ 
    & $90.04 {\scriptstyle \pm 0.57}$ 
    & $65.66 {\scriptstyle \pm 4.96}$ 
    & $90.24 {\scriptstyle \pm 1.86}$ 
    & $71.03 {\scriptstyle \pm 5.83}$ \\  

\bottomrule
\end{tabular}
}

\end{table}

\section{The Graph Structure of the Stochastic Block Barabási–Albert Models}\label{cgsO:app:_combined_BA} 
\begin{figure}[h]
    \centering
    \includegraphics[width=0.5\textwidth]{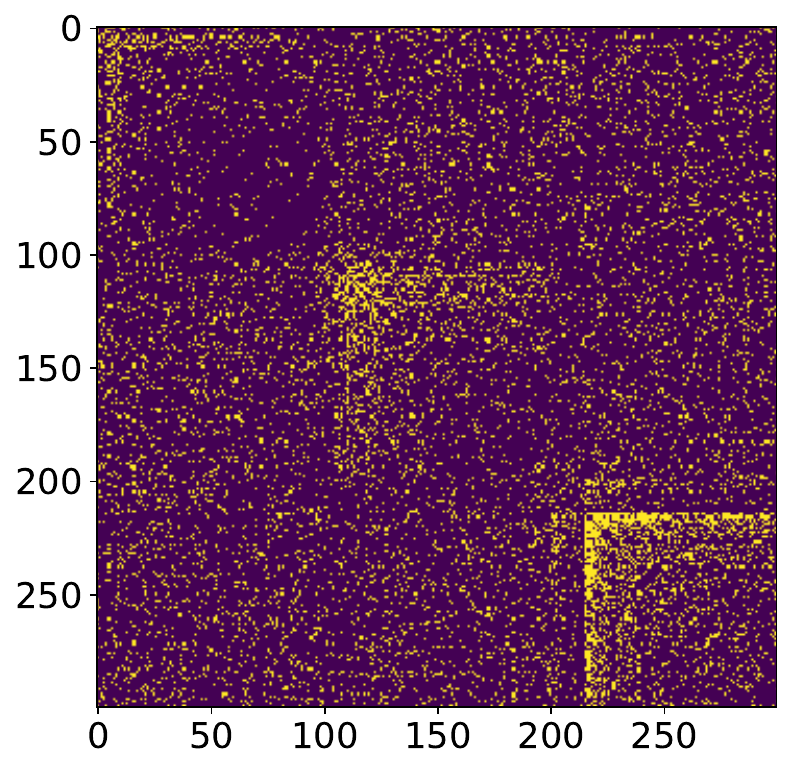}
  \caption[The synthetic graph generated from an SBBA]{The adjacency matrix of the synthetic graph generated from an SBBAM with 3 blocks. }
  \label{cgsO:fig:adj_matrix_combined_graph}
\end{figure}

In Figure \ref{cgsO:fig:adj_matrix_combined_graph}, we give an example of an adjacency matrix sampled from SBBAM model, presented previously in Section \ref{cgsO:sec:SBBAM}. Yellow points indicate edges, while purple points represent non-edges. Notably, there are variations in edge density across different blocks: while the first block appears to be predominantly heterophilic, the third block inherits a homophilic cluster structure.

\section{$k$-core distribution in Stochastic Block Barabási–Albert Models}
\label{cgsO:appendix_k_core_distrib} 
\begin{figure*}[h]
    \centering

    \includegraphics[width=\textwidth]{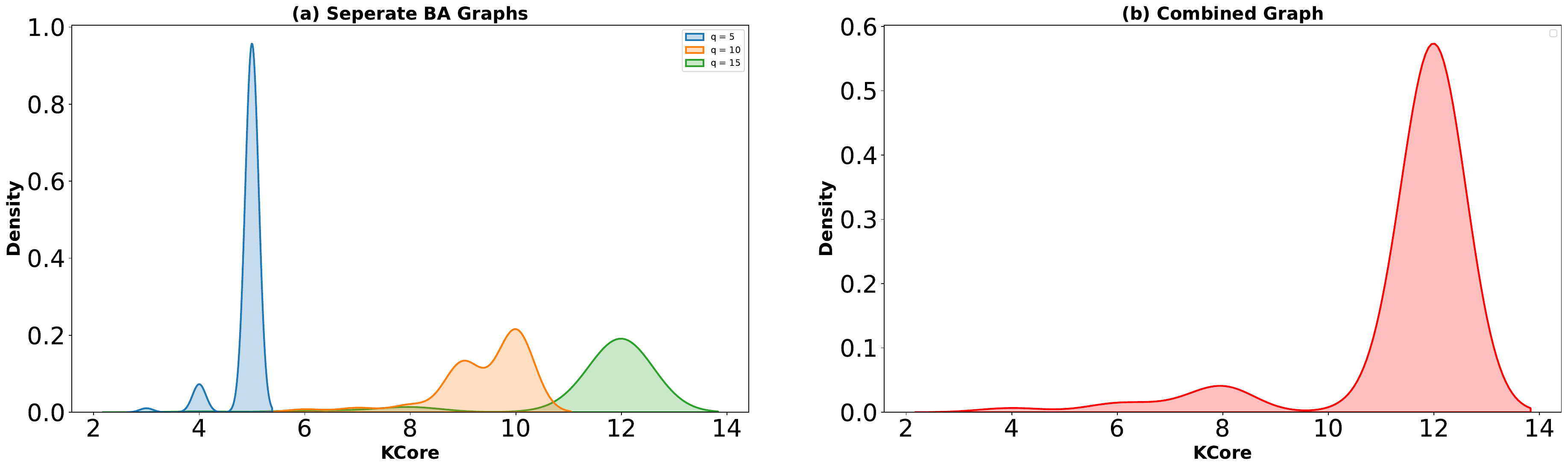} 
    \caption[$k$-core distributions of three different BA models]{The left figure represents the $k$-core distributions of three different BA models with the hyperparameters $r=5,10$ and $15$ serving as blocks of of our SBBAM. The right figure represents the $k$-core distribution of the SBBAM. }
    \label{cgsO:fig:sbm-pam}
\end{figure*}

In Figure \ref{cgsO:fig:sbm-pam}, we illustrate the $k$-core distribution of the three individual blocks, each corresponding to BA graphs, and the combined SBBAM graph.

\section{Spectral Clustering Algorithm}
\label{cgsO:app:spectral_clustering_algo}

 In Algorithm \ref{cgsO:algo:spectral_clustering}, we outline the steps of the Spectral Clustering Algorithm using CGSOs. This algorithm is applied on SBBAMs for cluster recovery and on the Cora dataset for centrality recovery.
\begin{algorithm}[h]

\small
\textbf{Inputs: } Graph $G$, Centrality GSO $\Phi$, Number of clusters to retrieve $C.$\\
\begin{enumerate}
    \item Compute the eigenvalues $\{\lambda \}_{i=1}^n$ and eigenvectors $\{u \}_{i=1}^n$ of $\mathbf{\Phi}$;
    \item Consider only the eigenvectors $U\in \mathbb{R}^{N\times C}$ corresponding to the $C$ largest eigenvalues;
    \item Cluster rows of $U$, corresponding to nodes in the graph,  using the $K$-Means algorithm to retrieve a node partition $P$ with $C$ clusters;
    $$\mathcal{P} = \text{K-Means}(U, C ) $$
\end{enumerate}

\Return $\mathcal{P}$; 
\caption{Spectral Clustering using the Centrality GSOs}\label{cgsO:algo:spectral_clustering}

\end{algorithm}

\section{Additional Results for the Spectral Clustering Task}\label{cgsO:ari_results_appendix} 
In this section, we report the ARI value of the spectral clustering task described in Section \ref{cgsO:synth_exps}.

\begin{figure}[h]
    \centering
    \includegraphics[width=\textwidth]{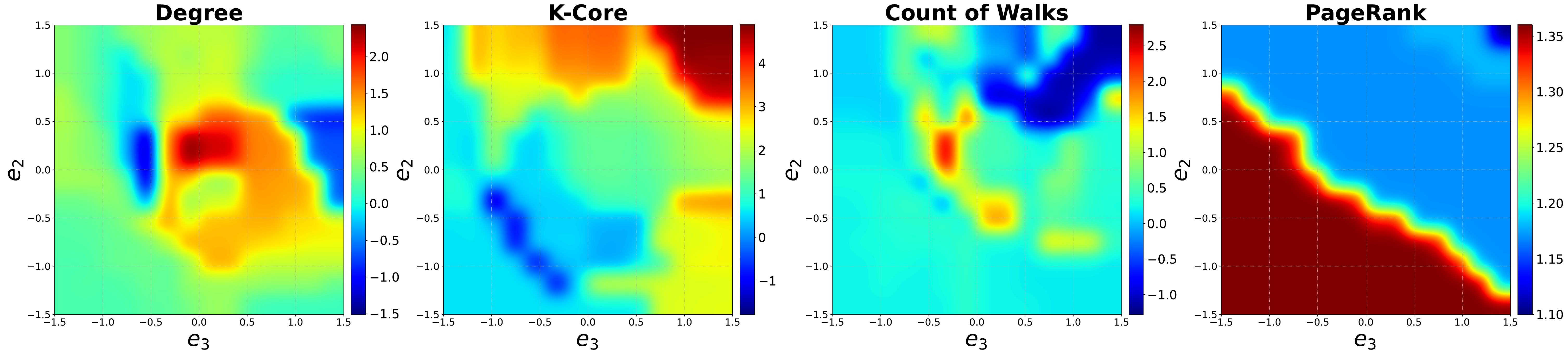}
    \caption[Result for the spectral clustering task on Cora]{Result for the spectral clustering task on Cora graph with core numbers considered as clusters. We report the values of the Adjusted Rand Information (ARI) in \% different combination of the exponents $(e_2,e_3)$ in $\mathbf{V}^{e_2} \mathbf{A} \mathbf{V}^{e_3}.$}    
    \label{cgsO:fig:ari_cora}
\end{figure}

\section{CGNN with Heterophily}\label{cgsO:app:heterop_gnn}
In this section, we incorporate our learnable CGSOs into \textit{H2GCN} \cite{zhu2020beyond}, designed for heterophilic graphs. We compared the results of \textit{CH2GCN} and \textit{H2GCN} on datasets with low homophily. We report the results of this experiment in Table \ref{cgsO:tab:H2GCN}. As noticed, our \textit{CH2GCN} outperforms \textit{H2GCN}.

\begin{table*}[h] 
\centering
\caption[Classification accuracy of CH2GCN]{Classification accuracy ($\pm$ standard deviation) of the models on different benchmark node classification datasets. The higher the accuracy (in \%) the better the model. \textcircled{1}~CH2GCN with nodes centrality,  \textcircled{2}~H2GCN. Highlighted are the \textbf{best results}.}
\label{cgsO:tab:H2GCN}
\resizebox{0.8\textwidth}{!}{%
\begin{tabular}{c|l|llll}\toprule
\multicolumn{1}{l}{} & Model & Texas  & Cornell & Wisconsin   & chameleon   \\
\hline

\multirow{4}{*}{\textcircled{1}}   
    & CH2GCN w/ $\mathbf{D}$  
        & $\mathbf{79.73 {\scriptstyle \pm 5.02}}$ 
        & $68.65 {\scriptstyle \pm 5.16}$ 
        & $79.80 {\scriptstyle \pm 4.02}$ 
        & $\mathbf{67.89 {\scriptstyle \pm 4.23}}$  \\

    & CH2GCN w/ $\mathbf{V}_{\text{\textit{core}}}$    
        & $78.92 {\scriptstyle \pm 5.77}$ 
        & $\mathbf{68.92 {\scriptstyle \pm 7.28}}$ 
        & $\mathbf{79.80 {\scriptstyle \pm 3.40}}$     
        & $60.00 {\scriptstyle \pm 5.63}$  \\

    & CH2GCN w/ $\mathbf{V}_{\ell\text{\textit{-walks}}}$   
        & $78.11 {\scriptstyle \pm 6.10}$ 
        & $68.92 {\scriptstyle \pm 6.19}$ 
        & $82.35 {\scriptstyle \pm 5.04}$ 
        & $44.28 {\scriptstyle \pm 2.32}$ \\

    & CH2GCN w/ $\mathbf{V}_{\text{\textit{PR}}}$  
        & $60.27 {\scriptstyle \pm 5.41}$ 
        & $44.86 {\scriptstyle \pm 7.76}$ 
        & $52.35 {\scriptstyle \pm 7.75}$ 
        & $31.95 {\scriptstyle \pm 5.79}$   \\ 

\midrule 

\multirow{1}{*}{\textcircled{2}}   
    & H2GCN    
        & $56.76 {\scriptstyle \pm 6.73}$ 
        & $51.08 {\scriptstyle \pm 6.89}$ 
        & $55.29 {\scriptstyle \pm 5.10}$ 
        & $63.93 {\scriptstyle \pm 2.07}$    \\   

\bottomrule
\end{tabular}
} 
\end{table*}

\section{Proofs of Propositions}\label{cgsO:app_proof_spectral}
In this section, we details the proofs of the propositions \ref{cgsO:prop:spectral_properties}, \ref{cgsO:prop:eigenvalues} and \ref{cgsO:prop:cheeger}.
\subsection{Proof of Proposition \ref{cgsO:prop:spectral_properties}}


\begin{proof}[Proof of Proposition \ref{cgsO:prop:spectral_properties}]
We first prove that the operator $\mathbf{M}_{G}$ is self-adjoint. \\
For $ \varphi_1, \varphi_2 \in L^2(G)$, we have:
\begin{align*}
    <\mathbf{M}_{G} \varphi_1, \varphi_2>_{G} & = \sum_{i\in  \mathcal{V} } v(i) \left (  \mathbf{M}_{G} \varphi_1 \right )(i)  \bar{\varphi_2}(i) \\
    &  = \sum_{i\in  \mathcal{V} } v(i) \left ( \frac{1}{v(i)} \sum_{j \in \mathcal{N}_i} \varphi_1(j)  \right ) \bar{\varphi_2}(i)  \\
    & = \sum_{i\in  \mathcal{V} } \bar{\varphi_2}(i) \left (  \sum_{j \in \mathcal{N}_i} \varphi_1(j)  \right )  \\
    & =  \sum_{i\in  \mathcal{V} }   \sum_{j \in \mathcal{N}_i} a_{i,j}  \bar{\varphi_2}(i)  \varphi_1(j) \\
    & =  \sum_{i,j\in  \mathcal{V} }   a_{i,j}  \bar{\varphi_2}(i)  \varphi_1(j). \\
\end{align*}

Similarly, we also have that, 
\begin{align*}
    <\varphi_1, \mathbf{M}_{G}  \varphi_2>_{G} & = \sum_{i\in  \mathcal{V} } v(i)   \varphi_1(i) \overline{\left (  \mathbf{M}_{G} \varphi_2 \right )}(i) \\
    &  =  \sum_{i\in  \mathcal{V} } v(i)   \varphi_1(i) \overline{  \left ( \frac{1}{v(i)} \sum_{j \in \mathcal{N}_i} \varphi_2(j)  \right )   }\\
 &  =  \sum_{i\in  \mathcal{V} }  \varphi_1(i) \overline{  \left ( \sum_{j \in \mathcal{N}_i} \varphi_2(j)  \right )   }\\
 & =  \sum_{i,j\in  \mathcal{V} }   a_{i,j}  \bar{\varphi_2}(i)  \varphi_1(j) \\
 & = \sum_{i,j\in  \mathcal{V} }   a_{j,i}  \bar{\varphi_2}(j)  \varphi_1(i) .
\end{align*}
\\

Thus, 

\begin{equation}
\forall \varphi_1, \varphi_2 \in L^2(G), \left\{ \begin{array}{l}
                                 <\mathbf{M}_{G} \varphi_1, \varphi_2>_{G} =  \sum_{i,j\in  \mathcal{V} }   a_{i,j}  \bar{\varphi_2}(i)  \varphi_1(j) , \\ 
                               <\varphi_1, \mathbf{M}_{G}  \varphi_2>_{G} = \sum_{i,j\in  \mathcal{V} }  a_{j,i}  \bar{\varphi_2}(j)  \varphi_1(i). \\ 
                                    \end{array} \right.
    \label{cgsO:eq:inner_prodcut}
\end{equation}
Since $a_{i,j}  = a_{j,i} $, we conclude that  $\mathbf{M}_{G}$ is self-adjoint, i.e. 
$$<\mathbf{M}_{G} \varphi_1, \varphi_2>_{G} = <\varphi_1, \mathbf{M}_{G}  \varphi_2>_{G}. $$

$\mathbf{M}_{G}$ is self-adjoint, the space $L^2(G)$ is finite-dimensional, thus is diagonalizable in an orthonormal basis, and its eigenvalues are real.\\
We define the following norm,  \begin{equation*}
     \lVert \mathbf{M}_{G} \rVert = \sup_{\varphi \neq 0} \frac{\langle \mathbf{M}_{G} \varphi, \varphi \rangle_{G}}{\lVert \varphi \rVert ^2}.   
    \end{equation*}
We will now prove that all eigenvalues have absolute values at most $\gamma = \min_{i \in \mathcal{V} } v(i)/deg(i) $. For that, we will first compute the two inner-products $<\left ( I - \mathbf{M}_{G} \right )  \varphi, \varphi>_{G} $ and $<\left ( I + \mathbf{M}_{G} \right )  \varphi, \varphi>_{G} $. For any $ \varphi \in L^2(G)$, using \eqnref{cgsO:eq:inner_prodcut}, we have that: 

$$\left\{ \begin{array}{l}
< \varphi, \varphi>_{G} =  \sum_{i \in \mathcal{V}} v(i) |\varphi(i)|^2 , \\ 
< \mathbf{M}_{G}  \varphi, \varphi>_{G} = \sum_{i,j\in  \mathcal{V}} a(i,j) \bar{\varphi}(i)  \varphi(j). \\ 
                                    \end{array} \right.$$

Let's first take the simple case, where $\gamma = \min_{i \in \mathcal{V} } \left ( \frac{v(i)}{deg(i)} \right ) \leq 1$, 
then,

\begin{align*}
   2 < \varphi, \varphi>_{G} & =  2\sum_{i \in  \mathcal{V}} v(i) |\varphi(i)|^2 \\
    & \geq 2  \gamma \sum_{i \in  \mathcal{V}}  deg(i) |\varphi(i)|^2   \\
    & \geq 2 \sum_{i \in  \mathcal{V}}  deg(i) |\varphi(i)|^2   \\
    & \geq 2 \sum_{i \in  \mathcal{V}}  \left ( \sum_{j \in  \mathcal{V}} a(i,j)  \right ) |\varphi(i)|^2   \\
    & \geq 2 \sum_{i,j \in  \mathcal{V}}   a(i,j)   |\varphi(i)|^2  . \\
\end{align*}

Therefore, 
\begin{align*}
   2 <(\mathbf{I}-\mathbf{M}_{G}) \varphi, \varphi>_{G} & =   2 <\varphi, \varphi>_{G} -  2 <\mathbf{M}_{G} \varphi, \varphi>_{G} \\
    & \geq 2 \sum_{i,j \in  \mathcal{V}}   a(i,j)   |\varphi(i)|^2   - 2 \sum_{i,j\in  \mathcal{V}} a(i,j) \bar{\varphi}(i)  \varphi(j) \\
    & \geq \sum_{i,j \in  \mathcal{V}}   a(i,j)   |\varphi(i)|^2 + \sum_{i,j \in  \mathcal{V}}   a(i,j)   |\varphi(j)|^2   - 2 \sum_{i,j\in  \mathcal{V}} a(i,j) \bar{\varphi}(i)  \varphi(j) \\
    & \geq \sum_{i,j \in  \mathcal{V}}   a(i,j)   |\varphi(i) - \varphi(j)|^2 .\\
\end{align*}

Similarly, we can prove that, 

$$ 2 <(\mathbf{I}+\mathbf{M}_{G}) \varphi, \varphi>_{G} \geq \sum_{i,j \in  \mathcal{V}}   a(i,j)   |\varphi(i) + \varphi(j)|^2 .$$

Therefore, if $\phi \neq 0$, then, 
\begin{align*}
\left\{ \begin{array}{l}
<(\mathbf{I}-\mathbf{M}_{G}) \varphi, \varphi>_{G} \geq 0  , \\ 
<(\mathbf{I}+\mathbf{M}_{G}) \varphi, \varphi>_{G} \geq 0  \\ 
\end{array} \right. & \Rightarrow  < \varphi, \varphi>_{G} \leq  <\mathbf{M}_{G} \varphi, \varphi>_{G} \leq  < \varphi, \varphi>_{G} .\\ 
& \Rightarrow  \frac{|<\mathbf{M}_{G} \varphi, \varphi>_{G} |}{< \varphi, \varphi>_{G}} \leq 1.
\end{align*}

Thus, $\lVert \mathbf{M}_{G} \rVert \leq 1$, i.e. all the eigenvalues have absolute values at most 1.
Let now consider the general case, where $\gamma$ is not necessarily smaller than 1.
Let's consider $\widetilde{V} = \frac{1}{\gamma} V=diag( \frac{v(1)}{\gamma}, \ldots , \frac{v(N)}{\gamma})$. Since,

\begin{align*}
   \Tilde{\gamma } & = \min_{i \in \mathcal{V} } \left ( \frac{\Tilde{v(i)}}{deg(i)} \right )\\
   & = \frac{1}{\gamma }  \left ( \frac{v(i)}{deg(i)} \right ) \\
   & = \frac{\gamma}{\gamma}\\
   & =1.
\end{align*}
Therefore, all the eigenvalues of $\widetilde{\mathbf{M}}_{G} = \widetilde{V}^{-1}A = \frac{1}{\gamma}\mathbf{V}^{-1}\mathbf{A} =\frac{1}{\gamma} \mathbf{M}_{G} $ have absolute values at most 1. Thus, all the eigenvalues of $ \mathbf{M}_{G}$ have absolute values at most $\gamma$. 
\end{proof}

\subsection{Proof of Proposition \ref{cgsO:prop:eigenvalues}}

    
\begin{proof}[Proof of Proposition \ref{cgsO:prop:eigenvalues}]

We will prove the first property.\\
We consider $P$ as the number of connected components, i.e. $G = \bigcup_{i=1}^{P} \mathcal{C}_i$.

The adjacency matrix of the graph $G$ is,
$$A = \begin{bmatrix}
A_{\mathcal{C}_1}  &     & \text{\LARGE 0}& &  &\text{\LARGE 0} \\
 &  & \ddots  & & & \\
   & \text{\LARGE 0}   &  &    A_{\mathcal{C}_i } &  & \text{\LARGE 0} &  \\
   &   &  &    &  & \ddots&   \\
   &   \text{\LARGE 0}&  &    &\text{\LARGE 0}  & & A_{ \mathcal{C}_P}\\
\end{bmatrix}.$$
And the transformation of $A$ by the Markov Average operator $\mathbf{M}_{G}$ is,
$$\mathbf{M}_{G} = \begin{bmatrix}
M_{\mathcal{C}_1}  &     & \text{\LARGE 0}& &  &\text{\LARGE 0} \\
 &  & \ddots  & & & \\
   & \text{\LARGE 0}   &  &    M_{\mathcal{C}_i } &  & \text{\LARGE 0} &  \\
   &   &  &    &  & \ddots&   \\
   &   \text{\LARGE 0}&  &    &\text{\LARGE 0}  & & M_{ \mathcal{C}_P}\\
\end{bmatrix}.$$

According to Proposition \ref{cgsO:prop:spectral_properties}, for each connected component $\mathcal{C}_i$, the matrix $\mathbf{M}_{\mathcal{C}_i}$ is diagonalizable in an orthonormal basis, and its eigenvalues are real numbers. We denote by $\mathbf{e^{\mathcal{C}_i} = [e_1^{\mathcal{C}_i},\ldots, e_{|\mathcal{C}_i|}^{\mathcal{C}_i} ]}$ the eigenvectors basis of $M_{\mathcal{C}_i}$ corresponding the eigenvalues $\lambda^{\mathcal{C}_i} = [\lambda_1^{\mathcal{C}_i},\ldots, \lambda_{|\mathcal{C}_i|}^{\mathcal{C}_i} ]$.

We consider the set of vectors

$$\mathbf{e}  =\begin{bmatrix}
\mathbf{e^{\mathcal{C}_1}} &     & \text{\LARGE 0}& &  &\text{\LARGE 0} \\
 &  & \ddots  & & & \\
   & \text{\LARGE 0}   &  &   \mathbf{e^{\mathcal{C}_i}} &  & \text{\LARGE 0} &  \\
   &   &  &    &  & \ddots&   \\
   &   \text{\LARGE 0}&  &    &\text{\LARGE 0}  & &  \mathbf{e^{\mathcal{C}_P}}\\
\end{bmatrix}.$$
The column vectors of $\mathbf{e} $ are eigenvectors of the matrix $\mathbf{M}_{G}$, and which achieves the conditions of Property 1.
Let's now prove the formulas of the mean and standard deviation of the $\mathbf{M}_{G}$ spectrum. The matrix $\mathbf{V}^{-1}\mathbf{A}$ is defined as follow, 
$$\forall 1\leq i,j \leq N, ~~~~(\mathbf{D}^{-1}\mathbf{A})_{i,j} = \frac{1}{v(i)} A_{i,j}.$$
Therefore, the diagonal elements of the matrix $ \left ( \mathbf{D}^{-1}\mathbf{A} \right )^2$ is defined as follow,
\begin{align*}
  \forall 1\leq i\leq N, ~   \left ( \left ( \mathbf{D}^{-1}\mathbf{A} \right )^2 \right )_{i,i} & =   \left ( \mathbf{D}^{-1}\mathbf{A}\mathbf{D}^{-1}\mathbf{A}\right )_{i,i} \\ 
  &=    \sum_j (\mathbf{D}^{-1}\mathbf{A})_{i,k} (\mathbf{D}^{-1}\mathbf{A})_{k,i} \\
     &= \sum_j \frac{ A_{i,j} A_{j,i}}{v(i) \times  v(j)}  \\
  &= \sum_j \frac{ A_{i,j}^2 }{v(i)\times v(j)} \\
   &= \sum_{j \in \mathcal{N}_i} \frac{ 1 }{v(i) \times v(j)} .
\end{align*}
Thus, 
\begin{align*}
  \mu\left (  {\mathbf{M}_{G}}  \right )& =    \text{Mean}\left (  Spectrum  \left [  \mathbf{V}^{-1}\mathbf{A}  \right ] \right )  \\
  & = \frac{1}{N} Sum \left (  Spectrum  \left [  \mathbf{V}^{-1}\mathbf{A}  \right ] \right )  \\
  & = \frac{1}{N} \sum_{i=1}^N (\mathbf{D}^{-1}\mathbf{A})_{i,i} \\
  & = \frac{1}{N} \sum_{i=1}^N \frac{1}{v(i)} A_{i,i} \\
  & = \frac{1}{N} \sum_{i=1}^N \frac{1}{v(i)}, \\
\end{align*}
 and,
\begin{align*}
  \sigma\left (  {\mathbf{M}_{G}} \right )   & =   \text{Stdev}\left (  Spectrum  \left [  \mathbf{V}^{-1}\mathbf{A}  \right ] \right )  \\
   & = \sqrt{\frac{1}{N}  \sum_{\lambda \in Spectrum  \left [  \mathbf{V}^{-1}\mathbf{A}  \right ] } \left (  \lambda  - \text{Mean}\left (  sp_\phi  \right )  \right )^2   }  \\ 
& = \sqrt{ \left ( \frac{1}{N}  \sum_{\lambda \in Spectrum  \left [  \mathbf{V}^{-1}\mathbf{A}  \right ] }  \lambda^2 \right )  - \text{Mean}\left (  sp_\phi  \right )^2   }  \\ 
& = \sqrt{ \left ( \frac{1}{N}  Sum(Spectrum\left [  \phi^2\right ]) \right )  - \text{Mean}\left (  sp_\phi  \right )^2   } \\ 
& = \sqrt{ \left ( \frac{1}{N}  Sum(Spectrum\left [  \left ( \mathbf{D}^{-1}\mathbf{A} \right )^2 \right ]) \right )  - \text{Mean}\left (  sp_\phi  \right )^2   } \\ 
& = \sqrt{ \left ( \frac{1}{N} Tr\left [  \left ( \mathbf{D}^{-1}\mathbf{A} \right )^2 \right ] \right )  - \text{Mean}\left (  sp_\phi  \right )^2   } \\ 
& = \sqrt{ \left ( \frac{1}{N} \sum_{i=1}\sum_{j\in \mathcal{N}_i} \frac{1}{v(i) \times v(j) }    \right )  - \text{Mean}\left (  sp_\phi  \right )^2   } \\ 
   & = \sqrt{ \left ( \frac{1}{N} \sum_{(i,j) \in E} \frac{1}{v(i) \times v(j) }    \right )  - \text{Mean}\left (  sp_\phi  \right )^2   }.
\end{align*}
 
\end{proof}

    \subsection{Proof of Proposition \ref{cgsO:prop:cheeger}}

    
    \begin{proof}[Proof of Proposition \ref{cgsO:prop:cheeger}]
        Let $W \subset V$, such that $|W|\leq \frac{1}{2} |\mathcal{V}|.$
    
    For $\varphi = \mathbb{1}_W - \mu_{G}(W)$ where $\mu_{G}(W) = \frac{|W|}{N_v}$ and $N_v= \sum_{i\in  \mathcal{V}} v(i) = |\mathcal{V}|_v.$
    \begin{align*}
       2 <(\mathbf{I}-\mathbf{M}_{G}) \varphi, \varphi>_{G} & =   2 <\varphi, \varphi>_{G} -  2 <\mathbf{M}_{G} \varphi, \varphi>_{G} \\
        & \leq 2 \sum_{i \in  \mathcal{V}}  v(i)   |\varphi(i)|^2   - 2 \sum_{i,j\in  \mathcal{V}} a(i,j) \bar{\varphi}(i)  \varphi(j) \\
        & \leq 2 \sum_{i \in  \mathcal{V}}  \beta \times deg(i)   |\varphi(i)|^2   - 2 \sum_{i,j\in  \mathcal{V}} a(i,j) \bar{\varphi}(i)  \varphi(j) \\
        & \leq 2 \sum_{i \in  \mathcal{V}}   deg(i)   |\varphi(i)|^2   - 2 \sum_{i,j\in  \mathcal{V}} a(i,j) \bar{\varphi}(i)  \varphi(j) \\
        & \leq 2 \sum_{i \in  \mathcal{V}}   \left ( \sum_{j \in  \mathcal{V}} a(i,j)  \right )   |\varphi(i)|^2   - 2 \sum_{i,j\in  \mathcal{V}} a(i,j) \bar{\varphi}(i)  \varphi(j) \\
        & \leq 2 \sum_{i,j \in  \mathcal{V}}   a(i,j)   |\varphi(i)|^2   - 2 \sum_{i,j\in  \mathcal{V}} a(i,j) \bar{\varphi}(i)  \varphi(j) \\
        & \leq \sum_{i,j \in  \mathcal{V}}   a(i,j)   |\varphi(i)|^2 + \sum_{i,j \in  \mathcal{V}}   a(i,j)   |\varphi(j)|^2   - 2 \sum_{i,j\in  \mathcal{V}} a(i,j) \bar{\varphi}(i)  \varphi(j) \\
        & \leq \sum_{i,j \in  \mathcal{V}}   a(i,j)   |\varphi(i) - \varphi(j)|^2 \\
    & \leq \sum_{i,j \in  \mathcal{V}}   a(i,j)   |\mathbb{1}_W (i) - \mathbb{1}_W (j)|^2. \\
    \end{align*}
    
    The non-zero terms in $\sum_{i,j \in  \mathcal{V}}   a(i,j)   |\varphi(i) - \varphi(j)|^2$ are those where $i$ and $j$ are adjacent, but one of them is in $W$ and the other not. 
    
    \begin{align*}
       <(\mathbf{I}-\mathbf{M}_{G}) \varphi, \varphi>_{G} &  \leq \frac{1}{2} \sum_{i,j \in  \mathcal{V}}   a(i,j)   |\mathbb{1}_W (i) - \mathbb{1}_W (j)|^2 \\
        &  = \#\mathcal{E}(W).\\
    \end{align*}
    There $\frac{1}{2}$ was removed because of the symmetry. 
    We also have that, 
    $$\frac{1}{N_v}<\mathbb{1}_W , \mathbb{1}_W >_{G} = \frac{1}{N_v} \sum_{i \in  \mathcal{V}} v(i) = \mu_G(W),$$
    and,
  $$\frac{1}{N_v}<\mathbb{1}_W , \mu_G(W)>_{G} = \frac{1}{N_v} \sum_{i \in  \mathcal{V}} v(i) \mu_G(W)=  \left ( \mu_G(W) \right )^2,$$   
  Therefore, 
  \begin{align*}
    \frac{1}{N_v}<\varphi , \varphi>_{G} & =  \frac{1}{N_v}< \mathbb{1}_W - \mu_{G}(W) ,  \mathbb{1}_W - \mu_{G}(W)>_{G}  \\
    & =  \frac{1}{N_v}< \mathbb{1}_W ,  \mathbb{1}_W - \mu_{G}(W)>_{G} - \frac{1}{N_v}<\mu_{G}(W) ,  \mathbb{1}_W - \mu_{G}(W)>_{G}  \\
    & = \frac{1}{N_v}<\mathbb{1}_W , \mathbb{1}_W >_{G} - \frac{2}{N_v} < \mathbb{1}_W  ,  \mu_{G}(W)>_{G}  +\frac{1}{N_v} < \mu_{G}(W) ,   \mu_{G}(W)>_{G} \\
    & = \mu_G(W) - 2  \left ( \mu_G(W) \right )^2 + \left ( \mu_G(W) \right )^2 \frac{1}{N_v}<1 , 1>_{G} \\
    & =  \mu_G(W) - 2  \left ( \mu_G(W) \right )^2 + \left ( \mu_G(W) \right )^2 \frac{1}{N_v} \sum_{i \in  \mathcal{V}} v(i) \\
    & =  \mu_G(W) - 2  \left ( \mu_G(W) \right )^2 + \left ( \mu_G(W) \right )^2 \frac{N_v}{N_v}  \\
    & = \mu_G(W) -   \left ( \mu_G(W) \right )^2  \\
    & = \mu_G(W)  \left (1- \mu_G(W) \right )  \\
    & = \mu_G(W)\mu_G(W' ),
\end{align*}
where $W' = \mathcal{V} - W.$\\
By definition,
$$\lambda_1(G) = \min_{\Tilde{\varphi}\neq 0} \frac{<(\mathbf{I}-\mathbf{M}_{G})\Tilde{\varphi} , \Tilde{\varphi}>_{G}}{<\Tilde{\varphi} , \Tilde{\varphi}>_{G}}.$$
Therefore,
\begin{align*}
    \lambda_1(G)  & \leq \frac{<(\mathbf{I} - \mathbf{M}_{G})\varphi , \varphi>_{G}}{<\varphi , \varphi>_{G}} \\
    & \leq \frac{\#\mathcal{E}(W)}{N_v} \frac{N_v }{<\varphi , \varphi>_{G}} \\
    & \leq \frac{\#\mathcal{E}(W)}{N_v} \frac{N_v }{  \mu_G(W)\mu_G(W' ) } .
\end{align*}
Since,
$$\frac{v_{-}}{v_{+}} \frac{|W|_v}{|\mathcal{V}|_v} \leq\mu_G(W) \leq \frac{v_{-}}{v_{+}} \frac{|W|_v}{|\mathcal{V}|_v}, $$
then,
\begin{align*}
N_v   \mu_G(W)\mu_G(W' ) & \geq N_v \frac{|W|_v}{|\mathcal{V}|_v} \frac{v_{-}}{v_{+}} \frac{|W'|_v}{|\mathcal{V}|_v} \\
& \geq  \frac{\sum_{i\in\mathcal{V} } v(i)}{|\mathcal{V}|_v} |W|_v \frac{v_{-}}{v_{+}} \frac{|W'|_v}{|\mathcal{V}|_v} \\
& \geq  \frac{\sum_{i\in\mathcal{V} } v(i)}{\sum_{i\in\mathcal{V} } v(i)} |W|_v \frac{v_{-}}{v_{+}} \frac{|W'|_v}{|\mathcal{V}|_v} \\
& \geq  |W|_v \frac{v_{-}}{v_{+}} \frac{|W'|_v}{|\mathcal{V}|_v} \\
& \geq  \frac{v_{-}}{v_{+}}|W|_v\frac{|W'|_v}{|\mathcal{V}|_v}  \\
& \geq \frac{v_{-}}{2v_{+}}|W|_v,
\end{align*}
because 
$$ \left\{\begin{matrix}
|W|_v\leq \frac{1}{2} |\mathcal{V}|_v\\ 
W' = \mathcal{V}-W
\end{matrix}\right. \Rightarrow  |W'|_v\geq \frac{1}{2} |\mathcal{V}|_v.$$
Thus, 
$$\forall W\subset \mathcal{V}, |W|_v\leq \frac{1}{2} |\mathcal{V}|_v \Rightarrow  \lambda_1(G) \leq \frac{2v_{+}}{v_{-}} \frac{\#\mathcal{E}(W)}{|W|_v}.$$

Thus, 
$$ \lambda_1(G) \leq \frac{2v_{+}}{v_{-}}  N_v h(G) \leq 2 N \frac{ v_{+}^2}{v_{-}} h_v(G).$$

    \end{proof}

\section{Average Degree of a Barabasi–Albert Model}\label{cgsO:appendix:proof_lemma_BA}
In this section, we details the proofs of the propositions \ref{cgsO:lem:BA_average_degree}.
\begin{proof}[Proof of Proposition \ref{cgsO:lem:BA_average_degree}]
    We start with a small graph of $N_0$ nodes and $r_0$ edges. At each time step, we increase the number of edges by $r$. Thus, if $N$ is the number of nodes at a certain time step, then there are exactly $r_0 + r(N-N_0)$ edges.

    As each edge contributes to the degree of two nodes, thus, the average degree is twice the number of edges divided by the number of nodes $N$. Therefore,
    \begin{align*}
        \overline{deg}(G^{BA}) & = \frac{2}{N} \left ( r_0 + r(N-r_0) \right ) \\
        & = 2r + 2\frac{r_0}{N} - 2N_0 \frac{r }{N} .
    \end{align*}
\end{proof}

\section{Learned Parameters of Different Centrality Based GSOs} \label{cgsO:appendix:learned_params}
In this section, we present some graph properties of the used dataset. We specifically present the node density, the homophily coefficient as well as the average value of different centrality metrics in Table \ref{cgsO:tab:graph_prop}. We also present the $(m_1,m_2,m_3,e_1,e_2,e_3,a)$ learned by the GNN in Tables \ref{cgsO:tab:degree_hyper}, \ref{cgsO:tab:kcore_hyper}, \ref{cgsO:tab:pagerank_hyper} and \ref{cgsO:tab:count_walks_hyper}.

\begin{table}[h]
\centering
\caption{Detailed graph properties of the used datasets.}
\label{cgsO:tab:graph_prop}
\resizebox{\textwidth}{!}{%
\begin{tabular}{l|rrrrrr}\toprule
Dataset  &  \multicolumn{1}{l}{density}  &  \multicolumn{1}{l}{Avg. Degree} &  \multicolumn{1}{l}{Avg. PageRank} &  \multicolumn{1}{l}{Avg. K-core}  & \multicolumn{1}{l}{Avg. Count. Paths} & \multicolumn{1}{l}{homophily}   \\ \midrule
   Physics & $4.16\times 10^{-4}$ & $14.37$ & $2.89 \times 10^{-5}$ & $7.71$ & $449.22$ & $0.931$ \\
Photo & $4.07\times 10^{-3}$ & $31.13$ & $1.30\times 10^{-4}$ & $16.97$ & $3204.098$ & $0.827$ \\
Cora & $1.43\times 10^{-3}$ & $3.89$ & $3.69\times 10^{-4}$ & $2.31$ & $42.52$ & $0.809$  \\
CS & $4.87\times 10^{-4}$ & $8.93$ & $5.45 \times 10^{-5}$ & $4.94$ & $162.75$ & $0.808$  \\
PubMed & $2.28\times 10^{-4}$ & $4.49$ & $5.07 \times 10^{-5}$ & $2.39$ & $75.43$ & $0.802$  \\
Computers & $2.60\times 10^{-3}$ & $35.75$ & $7.27 \times 10^{-5}$ & $18.84$ & $6221.39$ & $0.777$ \\
CiteSeer & $8.22\times 10^{-4} $& $2.73 $& $3.00\times 10^{-4}$ & $1.73$ & $18.91$ & $0.735$ \\
ogbn-arxiv & $8.07 \times 10^{-5}$ & $13.67$ & $5.90 \times 10^{-6}$ & $7.13$ & $4898.16$ & $0.654$  \\
deezer-europe & $2.31\times 10^{-4}$ & $6.55$ & $3.53 \times 10^{-5}$  & $3.57$ & $106.16$ & $0.525$  \\
Penn94 & $1.57\times 10^{-3}$ & $65.56$ & $2.40 \times 10^{-5}$ & $33.68$ & $10662.08 $& $0.470$ \\
chameleon & $1.21\times 10^{-2}$  &$ 27.57$ & $4.39\times 10^{-4}$ & $16.60 $ & $2913.48$ & $0.231$  \\
squirrel & $1.46\times 10^{-2}$  & $76.30$ &$ 1.92\times 10^{-4}$ &$ 41.55$ & $31888.02$ & $0.222$ \\
arxiv-year & $8.07 \times 10^{-5}$  & $6.88 $&$ 5.90 \times 10^{-6}$ & $7.13$ & $82.85$ & $0.218 $ \\
Wisconsin & $1.48\times 10^{-2}$  & $3.64$ & $3.98\times 10^{-3}$ & $2.05$ & $76.26$ & $0.206$ \\
Cornell & $1.68\times 10^{-2} $ & $3.04 $& $5.46\times 10^{-3}$ & $1.74$ & $58.47$ & $0.132 $ \\
Texas & $1.77\times 10^{-2}$  & $3.13$ & $5.46\times 10^{-3}$ &$ 1.71$ & $70.72$ & $0.111$     
  \\ \bottomrule      
\end{tabular}
}
\end{table}

\newpage
\subsection{Degree Centrality}
\begin{table}[h]
\centering
\caption{Graph Properties of the used datasets and the corresponding learned hyperparameters in GAGCN w/ Degree}
\label{cgsO:tab:degree_hyper}
\resizebox{\textwidth}{!}{%
\begin{tabular}{l|rrr|rrrrrrr}\toprule
Dataset & \multicolumn{3}{c}{Graph Properties} & \multicolumn{7}{c}{Hyperparameters}     \\ \midrule
 &   \multicolumn{1}{l}{Avg. K-core}  & \multicolumn{1}{l}{Avg. Count. Walks} & \multicolumn{1}{l}{homophily}  & \multicolumn{1}{r}{$e_1$} &  \multicolumn{1}{r}{$e_2$}  &  \multicolumn{1}{r}{$e_3$} & \multicolumn{1}{r}{$m_1$}  &  \multicolumn{1}{r}{$m_2$}&\multicolumn{1}{r}{$m_3$} &  \multicolumn{1}{r}{$a$}  \\ \midrule
Physics & $7.71$ & $449.22$ & $0.931 $&$ 0.28~(0.01)$&$ -0.31~(0.00)$& $-0.32~(0.00)$& $0.34~(0.01)$& $1.33~(0.01)$& $0.31~(0.01)$& $1.36~(0.01)$\\
Photo   & $16.97$ & $3204.098$& $0.827$ & $0.39~(0.06)$& $-0.26~(0.01) $& $-0.25~(0.01)$ & $0.59~(0.05)$& $1.51~(0.01)$& $0.53~(0.04)$& $1.70~(0.04)$\\
Cora   &$ 2.31$ & $42.52$  & $0.809$ & $0.31~(0.04)$& $0.02~(0.01)$& $-0.02~(0.01)$ & $0.67~(0.02)$& $1.43~(0.04)$& $0.66~(0.02)$& $0.69~(0.01)$\\
CS & $4.94$ & $162.75$ & $0.808$ & $0.33~(0.00)$& $-0.25~(0.00)$ & $-0.26~(0.00)$ & $0.44~(0.01)$& $1.44~(0.00)$& $0.40~(0.01)$& $1.47~(0.01)$\\
PubMed & $2.39$ & $75.43$  & $0.802$ & $0.28~(0.01)$& $-0.27~(0.00) $& $-0.28~(0.00)$ &$ 0.39~(0.00)$& $1.40~(0.01)$& $0.38~(0.00)$&$ 1.39~(0.01)$\\
Computers  & $18.84$ & $6221.39$ & $0.777$ & $0.40~(0.05)$&$ -0.74~(0.02)$ & $0.24~(0.03)$& $0.74~(0.05)$& $1.60~(0.05)$& $0.66~(0.04)$& $0.86~(0.10)$\\
CiteSeer & $1.73 $& $18.91 $ & $0.735$ & $0.35~(0.00)$& $-0.21~(0.01)$ & $-0.22~(0.01)$ & $ 0.49~(0.01)$& $1.49~(0.01)$& $0.47~(0.01)$& $1.50~(0.01)$\\
ogbn-arxiv  &$ 7.13$ & $4898.16$ & $0.654 $& $-0.08~(0.02)$ & $-0.29~(0.01)$ & $-0.41~(0.00) $& $0.13~(0.01)$& $1.31~(0.04)$&$ 0.13~(0.01)$& $1.00~(0.01)$\\
deezer-europe & $3.57$ &$ 106.16$ &$ 0.525$ & $0.31~(0.04)$& $-0.51~(0.03)$ & $-0.54~(0.02)$ & $0.59~(0.04)$& $-0.96~(0.04)$ & $1.55~(0.03)$& $-0.59~(0.03)$\\
Penn94  & $33.68 $& $10662.08$& $0.470$ & $0.51~(0.01)$& $-1.00~(0.02)$ & $-0.09~(0.02)$ & $0.98~(0.01)$& $1.01~(0.04)$&$ 0.82~(0.01)$& $0.95~(0.03)$\\
chameleon& $16.60$ & $2913.48$ & $0.231$ & $0.15~(0.04)$& $-0.06~(0.01)$ & $-0.06~(0.01) $& $-0.17~(0.03) $& $0.88~(0.02)$& $-0.16~(0.03)$ & $-0.15~(0.02)$\\
squirrel  & $41.55$ & $31888.02$& $0.222$ & $0.38~(0.05)$& $-0.26~(0.03)$ & $-0.24~(0.03)$ & $0.31~(0.80)$& $1.75~(0.07)$& $0.26~(0.70)$& $1.69~(0.56)$\\
arxiv-year & $7.13 $& $82.85 $ & $0.218$ & $-0.25~(0.01) $& $-0.27~(0.01)$ & $-0.40~(0.01)$ & $0.01~(0.01)$& $0.99~(0.01)$& $0.05~(0.01)$& $0.80~(0.01)$\\
Wisconsin& $2.05$ & $76.26 $ & $0.206 $& $0.95~(0.05)$& $-0.09~(0.04) $& $-0.05~(0.01)$ &$ 1.27~(0.25)$& $-0.94~(0.05)$ & $0.66~(0.07)$& $-0.64~(0.06)$\\
Cornell & $1.74 $& $58.47 $ & $0.132$ & $0.88~(0.05)$& $-0.17~(0.07)$ & $-0.07~(0.03) $& $1.04~(0.29)$& $-0.86~(0.11)$ & $0.80~(0.10)$& $-0.78~(0.08)$\\
Texas  & $1.71$ & $70.72 $ & $0.111$ & $0.93~(0.03)$& $-0.09~(0.04)$ & $-0.05~(0.01)$ &$ 1.17~(0.20)$& $-0.98~(0.05)$ & $0.65~(0.07)$& $-0.64~(0.07)$     \\ \hline      
\end{tabular}
}
\end{table}

\subsection{$k$-Core Centrality}
\begin{table}[h]
\centering
\caption{Graph Properties of the used datasets and the corresponding learned hyperparameters in GAGCN w/ K-Core}
\label{cgsO:tab:kcore_hyper}
\resizebox{\textwidth}{!}{%
\begin{tabular}{l|rrr|rrrrrrr}\toprule
Dataset &  \multicolumn{3}{c}{Graph Properties}  & \multicolumn{7}{c}{Hyperparameters}      \\ \midrule
 &   \multicolumn{1}{l}{Avg. K-core}  & \multicolumn{1}{l}{Avg. Count. Walks} & \multicolumn{1}{l}{homophily}  & \multicolumn{1}{r}{$e_1$} &  \multicolumn{1}{r}{$e_2$}  &  \multicolumn{1}{r}{$e_3$} & \multicolumn{1}{r}{$m_1$}  &  \multicolumn{1}{r}{$m_2$}&\multicolumn{1}{r}{$m_3$} &  \multicolumn{1}{r}{$a$} \\ \midrule
Physics  & $ 7.71 $ & $ 449.22 $ & $ 0.931$ & $ 0.38~(0.03) $ & $ -0.35~(0.01) $ & $ -0.35~(0.01) $ & $ 0.34~(0.02) $ & $ 1.28~(0.01) $ & $ 0.30~(0.02) $ & $ 1.35~(0.02) $ \\
Photo  & $ 16.97$ & $ 3204.098 $ & $ 0.827$ & $ 0.52~(0.02) $ & $ -0.31~(0.01) $ & $ -0.31~(0.01) $ & $ 0.73~(0.03) $ & $ 1.44~(0.02) $ & $ 0.59~(0.03) $ & $ 1.77~(0.05) $ \\
Cora   & $ 2.31 $ & $ 42.52 $ & $ 0.809$ & $ 0.34~(0.01) $ & $ -0.74~(0.00) $ & $ 0.24~(0.01) $ & $ 0.60~(0.01) $ & $ 1.55~(0.01) $ & $ 0.59~(0.01) $ & $ 0.68~(0.01) $ \\
CS   & $ 4.94 $ & $ 162.75 $ & $ 0.808$ & $ 0.41~(0.01) $ & $ -0.29~(0.01) $ & $ -0.29~(0.01) $ & $ 0.43~(0.01) $ & $ 1.39~(0.01) $ & $ 0.39~(0.01) $ & $ 1.47~(0.01) $ \\
PubMed & $ 2.39 $ & $ 75.43 $ & $ 0.802$ & $ 0.27~(0.00) $ & $ -0.31~(0.00) $ & $ -0.32~(0.00) $ & $ 0.34~(0.01) $ & $ 1.34~(0.01) $ & $ 0.34~(0.01) $ & $ 1.35~(0.01) $ \\
Computers  & $ 18.84$ & $ 6221.39$ & $ 0.777$ & $ 0.51~(0.02) $ & $ -0.28~(0.01) $ & $ -0.29~(0.01) $ & $ 0.78~(0.03) $ & $ 1.50~(0.01) $ & $ 0.66~(0.03) $ & $ 1.72~(0.05) $ \\
CiteSeer  & $ 1.73 $ & $ 18.91 $ & $ 0.735$ & $ 0.39~(0.01) $ & $ -0.26~(0.00) $ & $ -0.26~(0.00) $ & $ 0.45~(0.01) $ & $ 1.43~(0.01) $ & $ 0.44~(0.01) $ & $ 1.46~(0.00) $ \\
ogbn-arxiv  & $ 7.13 $ & $ 4898.16$ & $ 0.654$ & $ 0.27~(0.01) $ & $ -0.53~(0.01) $ & $ -0.55~(0.01) $ & $ -0.70~(0.01) $ & $ -1.04~(0.03) $ & $ 0.31~(0.02) $ & $ 0.70~(0.02) $ \\

deezer-europe & $ 3.57 $ & $ 106.16 $ & $ 0.525$ & $ -0.01~(0.03) $ & $ -0.51~(0.00) $ & $ -0.51~(0.00) $ & $ 0.02~(0.01) $ & $ 0.99~(0.01) $ & $ 0.02~(0.01) $ & $ 1.07~(0.01) $ \\
Penn94 & $ 33.68$ & $ 10662.08 $ & $ 0.470$ & $ -0.09~(0.30) $ & $ -0.39~(0.08) $ & $ -0.40~(0.08) $ & $ 0.28~(0.36) $ & $ 1.27~(0.16) $ & $ 0.05~(0.41) $ & $ 1.60~(0.15) $ \\
chameleon  & $ 16.60$ & $ 2913.48$ & $ 0.231$ & $ 0.15~(0.04) $ & $ -0.06~(0.01) $ & $ -0.06~(0.01) $ & $ -0.17~(0.02) $ & $ 0.88~(0.01) $ & $ -0.16~(0.02) $ & $ -0.15~(0.01)$ \\
squirrel  & $ 41.55$ & $ 31888.02 $ & $ 0.222$ & $ 0.46~(0.02) $ & $ -0.78~(0.01) $ & $ 0.24~(0.01) $ & $ -0.97~(0.05) $ & $ 1.78~(0.04) $ & $ -0.97~(0.05) $ & $ -1.08~(0.12)$ \\
arxiv-year  & $ 7.13 $ & $ 82.85 $ & $ 0.218$ & $ 0.34~(0.05) $ & $ -0.36~(0.01) $ & $ -0.41~(0.01) $ & $ -0.13~(0.06) $ & $ 1.03~(0.01) $ & $ -0.03~(0.02) $ & $ 0.80~(0.02) $ \\
Wisconsin  & $ 2.05$ & $ 76.26 $ & $ 0.206$ & $ 1.20~(0.02) $ & $ -0.05~(0.02) $ & $ -0.06~(0.02) $ & $ 1.48~(0.03) $ & $ -0.96~(0.04) $ & $ 0.54~(0.02) $ & $ -0.54~(0.03)$ \\
Cornell   & $ 1.74 $ & $ 58.47 $ & $ 0.132$ & $ 0.34~(0.05) $ & $ -0.36~(0.01) $ & $ -0.41~(0.01) $ & $ -0.13~(0.06) $ & $ 1.03~(0.01) $ & $ -0.03~(0.02) $ & $ 0.80~(0.02) $ \\
Texas  & $ 1.71$ & $ 70.72 $ & $ 0.111$ & $ 0.50~(0.06)$  & $ -0.02~(0.02) $ & $ -0.04~(0.02) $ & $ -0.87~(0.05) $ & $ 1.04~(0.05) $ & $ -0.85~(0.05) $ & $ -0.86~(0.05)  $ \\ \bottomrule       
\end{tabular}
}
\end{table}

\subsection{PageRank Centrality}
\begin{table}[h]
\centering
\caption{Graph Properties of the used datasets and the corresponding learned hyperparameters in GAGCN w/ PageRank}
\label{cgsO:tab:pagerank_hyper}
\resizebox{\textwidth}{!}{%
\begin{tabular}{l|rrr|rrrrrrr}\toprule
Dataset & \multicolumn{3}{c}{Graph Properties} & \multicolumn{7}{c}{Hyperparameters}     \\ \midrule
 &    \multicolumn{1}{l}{Avg. K-core}  & \multicolumn{1}{l}{Avg. Count. Walks} & \multicolumn{1}{l}{homophily}  & \multicolumn{1}{r}{$e_1$} &  \multicolumn{1}{r}{$e_2$}  &  \multicolumn{1}{r}{$e_3$} & \multicolumn{1}{r}{$m_1$}  &  \multicolumn{1}{r}{$m_2$}&\multicolumn{1}{r}{$m_3$} &  \multicolumn{1}{r}{$a$} \\ \midrule
   Physics  & $7.71$ & $449.22$ & $0.931$ & $0.51~(0.00)$ & $0.00~(0.00)$ & $0.00~(0.00)$ & $1.31~(0.06)$ & $1.00~(0.08)$ & $0.34~(0.06)$ & $0.33~(0.07)$  \\
Photo  & $16.97$ & $3204.098$ & $0.827$ & $0.53~(0.02)$ & $0.08~(0.01)$ & $0.08~(0.01)$ & $0.88~(0.01)$ & $0.85~(0.01)$ & $-0.12~(0.01)$ & $-0.13~(0.01)$ \\
Cora  & $2.31$ & $42.52$ & $0.809$ & $0.00~(0.00)$ & $-0.71~(0.02)$ & $0.11~(0.01)$ & $0.63~(0.01)$ & $1.49~(0.02)$ & $0.63~(0.01)$ & $0.67~(0.01) $ \\
CS  & $4.94$ & $162.75$ & $0.808$ & $0.00~(0.00)$ & $-0.10~(0.01)$ & $-0.10~(0.01)$ & $0.46~(0.04)$ & $1.38~(0.10)$ & $0.46~(0.04)$ & $1.49~(0.03)$  \\
PubMed  & $2.39$ & $75.43$ & $0.802$ & $0.51~(0.00)$ & $0.00~(0.00)$ & $0.00~(0.00)$ & $1.34~(0.02)$ & $1.28~(0.02)$ & $0.36~(0.02)$ & $0.38~(0.02)$  \\
Computers & $18.84$ & $6221.39$ & $0.777$ & $0.00~(0.00)$ & $-0.42~(0.00)$ & $-0.42~(0.00)$ & $-0.13~(0.02)$ & $0.84~(0.01)$ & $-0.13~(0.02)$ & $0.87~(0.02)$  \\
CiteSeer& $1.73$ & $18.91$ & $0.735$ & $0.00~(0.00)$ & $-0.14~(0.00)$ &$ -0.12~(0.00)$ &$ 0.44~(0.01)$ &$ 1.41~(0.00)$ & $0.44~(0.01) $& $1.47~(0.01) $ \\
ogbn-arxiv & 7.13 & $4898.16$ & $0.654$ & $0.00~(0.00)$ & $-0.89~(0.01) $& 0.11~(0.01) & $-0.22~(0.01) $& $0.78~(0.01)$ & $-0.22~(0.01)$ & $-0.22~(0.01)$ \\
deezer-europe  & $3.57$ & $106.16$ & $0.525$ & $0.57~(0.05)$ & $0.05~(0.01)$ & $0.05~(0.01)$ & $0.90~(0.03)$ & $0.90~(0.02)$ & $-0.10~(0.03)$ & $-0.10~(0.02)$ \\
Penn94 & $33.68$ & $10662.08 $& $0.470$ & $0.54~(0.01) $& $0.05~(0.01)$ & $0.05~(0.01)$ & $1.10~(0.01)$ & $-0.90~(0.01)$ & $0.10~(0.01)$ & $-0.10~(0.01) $\\
chameleon  & $16.60 $ & $2913.48$ & $0.231$ & $-0.01~(0.00)$ & $-0.94~(0.00)$ &$ 0.06~(0.01)$ & $0.27~(0.05)$ & -$0.88~(0.02)$ & $1.26~(0.05)$ & $-0.25~(0.05)$ \\
squirrel &$ 41.55$ & $31888.02$ & $0.222$ & $0.00~(0.00) $& $-0.43~(0.01) $& $-0.43~(0.01)$ & $0.14~(0.01)$ & $-0.86~(0.01)$ & $1.14~(0.01)$ & $-0.14~(0.01)$ \\
arxiv-year  & $7.13$ & $82.85$ & $0.218 $& $0.00~(0.00)$ & $-0.84~(0.03)$ & $0.06~(0.01)$ & $-0.04~(0.02)$ & $0.86~(0.03) $& $-0.04~(0.02)$ &$ -0.05~(0.03)$ \\
Wisconsin  & $2.05$ & $76.26$ & $0.206$ & $0.64~(0.02)$ & $0.10~(0.01)$ & $0.10~(0.01)$ & $1.68~(0.03)$ & $-1.01~(0.04)$ & $0.69~(0.02)$ & $-0.69~(0.02) $\\
Cornell & 1.74 & 58.47 & 0.132 & $0.64~(0.03)$ & $0.11~(0.01)$ & $0.11~(0.01)$ & $1.71~(0.02)$ & $-1.03~(0.05)$ & $0.71~(0.01$) & $-0.72~(0.01)$ \\
Texas  &$ 1.71$ & $70.72$ & $0.111$ & $0.68~(0.03)$ & $0.14~(0.03)$ & $0.14~(0.03)$ & $1.63~(0.04)$ & $-0.93~(0.06)$ & $0.64~(0.04)$ & $-0.63~(0.04)$     
  \\ \bottomrule       
\end{tabular}
}
\end{table}

\newpage
\subsection{Count of Walks Centrality}
\begin{table}[h]
\centering
\caption{Graph Properties of the used datasets and the corresponding learned hyperparameters in GAGCN w/ Count of walks.}
\label{cgsO:tab:count_walks_hyper}
\resizebox{\textwidth}{!}{%
\begin{tabular}{l|rrr|rrrrrrr}\toprule
Dataset &  \multicolumn{3}{c}{Graph Properties} & \multicolumn{7}{c}{Hyperparameters}      \\ \midrule
 &   \multicolumn{1}{l}{Avg. K-core}  & \multicolumn{1}{l}{Avg. Count. Walks} & \multicolumn{1}{l}{homophily}  & \multicolumn{1}{r}{$e_1$} &  \multicolumn{1}{r}{$e_2$}  &  \multicolumn{1}{r}{$e_3$} & \multicolumn{1}{r}{$m_1$}  &  \multicolumn{1}{r}{$m_2$}&\multicolumn{1}{r}{$m_3$} &  \multicolumn{1}{r}{$a$} \\ \midrule
Physics    & $ 7.71 $ & $ 449.22 $ & $ 0.931 $ & $ 0.95~(0.01) $ & $ -0.02~(0.02) $ & $ -0.02~(0.02) $ & $ 0.89~(0.02) $ & $ 0.96~(0.03) $ & $ -0.10~(0.01) $ & $ -0.09~(0.01)$ \\
Photo  & $ 16.97 $ & $ 3204.098 $ & $ 0.827 $ & $ -0.06~(0.01) $ & $ -0.07~(0.00) $ & $ -0.07~(0.00) $ & $ -0.05~(0.01) $ & $ 0.87~(0.00) $ & $ -0.04~(0.02) $ & $ -0.09~(0.01)$ \\
Cora & $ 2.31 $ & $ 42.52  $ & $ 0.809 $ & $ 0.36~(0.02) $ & $ 0.03~(0.01) $ & $ 0.02~(0.01) $ & $ 0.68~(0.02) $ & $ 1.37~(0.09) $ & $ 0.63~(0.02) $ & $ 0.64~(0.01)$ \\
CS  & $ 4.94 $ & $ 162.75 $ & $ 0.808 $ & $ 0.45~(0.02) $ & $ 0.03~(0.01) $ & $ 0.02~(0.01) $ & $ 0.62~(0.03) $ & $ 1.23~(0.04) $ & $ 0.47~(0.02) $ & $ 0.52~(0.02)$ \\
PubMed & $ 2.39 $ & $ 75.43  $ & $ 0.802 $ & $ 0.30~(0.02) $ & $ -0.15~(0.02) $ & $ -0.16~(0.02) $ & $ 0.60~(0.02) $ & $ 1.58~(0.03) $ & $ 0.56~(0.02) $ & $ 1.38~(0.04)$ \\
Computers & $ 18.84 $ & $ 6221.39 $ & $ 0.777 $ & $ -0.05~(0.01) $ & $ -0.07~(0.00) $ & $ -0.07~(0.00) $ & $ -0.05~(0.02) $ & $ 0.86~(0.01) $ & $ -0.04~(0.03) $ & $ -0.10~(0.02)$ \\
CiteSeer  & $ 1.73 $ & $ 18.91  $ & $ 0.735 $ & $ 0.25~(0.01) $ & $ -0.13~(0.01) $ & $ -0.14~(0.01) $ & $ 0.67~(0.01) $ & $ 1.63~(0.01) $ & $ 0.62~(0.01) $ & $ 1.63~(0.01)$ \\
ogbn-arxiv & $ 7.13 $ & $ 4898.16 $ & $ 0.654 $ & $ 0.12~(0.00) $ & $ -0.08~(0.00) $ & $ -0.19~(0.00) $ & $ 0.32~(0.00) $ & $ 1.57~(0.02) $ & $ 0.32~(0.00) $ & $ 0.93~(0.01)$ \\
deezer-europe & $ 3.57 $ & $ 106.16 $ & $ 0.525 $ & $ 0.26~(0.03) $ & $ -1.04~(0.03) $ & $ -0.07~(0.03) $ & $ 0.58~(0.04) $ & $ 0.88~(0.06) $ & $ 0.60~(0.04) $ & $ 0.67~(0.06)$ \\
Penn94   & $33.68$& $ 10662.08 $ & $ 0.470 $ & $ 0.30~(0.00) $ & $ -0.77~(0.03) $ & $ 0.06~(0.09) $ & $ 0.58~(0.00) $ & $ -0.46~(0.16) $ & $ 1.39~(0.01) $ & $ -0.31~(0.17)$ \\
chameleon  & $ 16.60 $ & $ 2913.48 $ & $ 0.231 $ & $ 0.32~(0.06) $ & $ -0.05~(0.01) $ & $ -0.05~(0.01) $ & $ -0.28~(0.08) $ & $ 0.89~(0.02) $ & $ -0.17~(0.03) $ & $ -0.14~(0.02)$ \\
squirrel  & $ 41.55 $ & $ 31888.02 $ & $ 0.222 $ & $ 0.23~(0.11) $ & $ -0.71~(0.10) $ & $ 0.35~(0.10) $ & $ -0.46~(0.46) $ & $ 1.28~(0.22) $ & $ -0.32~(0.33) $ & $ -0.29~(0.48)$ \\
arxiv-year& $ 7.13 $ & $ 82.85  $ & $ 0.218 $ & $ 0.23~(0.01) $ & $ -0.28~(0.01) $ & $ -0.16~(0.01) $ & $ -0.26~(0.01) $ & $ 0.99~(0.03) $ & $ -0.01~(0.04) $ & $ 0.84~(0.03)$ \\
Wisconsin & $ 2.05 $ & $ 76.26  $ & $ 0.206 $ & $ 0.40~(0.01) $ & $ -0.79~(0.07) $ & $ 0.18~(0.05) $ & $ 0.61~(0.01) $ & $ -1.38~(0.08) $ & $ 1.48~(0.01) $ & $ -0.69~(0.06)$ \\
Cornell  & $ 1.74 $ & $ 58.47  $ & $ 0.132 $ & $ 0.40~(0.01) $ & $ -0.21~(0.04) $ & $ -0.22~(0.04) $ & $ 0.59~(0.01) $ & $ -1.51~(0.06) $ & $ 1.47~(0.01) $ & $ -0.61~(0.05)$ \\
Texas  & $ 1.71 $ & $ 70.72  $ & $ 0.111 $ & $ 0.40~(0.01) $ & $ -0.72~(0.03) $ & $ 0.24~(0.03) $ & $ 0.58~(0.01) $ & $ -1.48~(0.05) $ & $ 1.47~(0.01) $ & $ -0.59~(0.03)
  $ \\ \bottomrule      
\end{tabular}
} 
\end{table}

\chapter{Appendix: Adaptive Depth Message Passing GNN}
\section{Dataset Statistics}\label{dynam:app:data_stats}
Characteristics and information about the datasets utilized in the node classification part of the study are presented in Table \ref{dynam:tab:data_statistics}, which provides statistics about each dataset, and  highlights the variety in their features, size, and edge homophily,

\begin{table}[h]

\caption{Statistics of the node classification datasets used in our experiments.}
\label{dynam:tab:data_statistics}
\begin{center}
\begin{small}
\resizebox{\columnwidth}{!}{%
\begin{tabular}{lrrrrr}
\toprule
Dataset & \#Features & \#Nodes & \#Edges & \#Classes & Edge Homophily \\
\midrule
Cora    & 1,433 & 2,708   & 5,208    & 7 & 0.809 \\
CiteSeer   & 3,703 & 3,327 & 4,552 & 6 & 0.735\\
PubMed    & 500 & 19,717 & 44,338 & 3 & 0.802\\
CS    & 6,805 & 18,333 & 81,894 & 15 & 0.808\\
chameleon  &   2,325  &   2,277  &   62,792  &   5  &   0.231\\
Cornell  &   1,703  &   183  &   557  &   5  &   0.132 \\
squirrel  &   2,089  &   5,201  &   396,846  &   5  &   0.222 \\
Wisconsin  &   1,703  &   251  &   916  &   5  &   0.206  \\
Texas  &   1,703  &   183  &   574  &   5  &   0.111 \\
Photo  &   745  &   7,650  &   238,162  &   8  &   0.827 \\
ogbn-arxiv  &   128  &   169,343  &   2,315,598  &   40  &   0.654  \\
Computers  &   767  &   13752  &   491,722  &   10  &   0.777 \\
\hline
\end{tabular}
}
\end{small}
\end{center}
\end{table}

\section{Node-Specific Depth Analysis in Graph Neural Networks}\label{dynam:app:motivation}
We generate synthetic graphs of size $N=5,000$. We select nodes belonging to sparse or dense regions in the original graph based on their core number. We consider only nodes with labels that are sufficiently present in both dense and sparse region. Last, we randomly select nodes, all by keeping the label distribution similar in both sparse and dense subgraphs. Yellow points indicate edges, while purple points represent non-edges. Notably, there are variations in edge density across different blocks: the first block is characterized by extreme sparsity, whereas the second block exhibits a much denser structure.

\begin{figure}[ht]
    \centering
    \includegraphics[width=\linewidth]{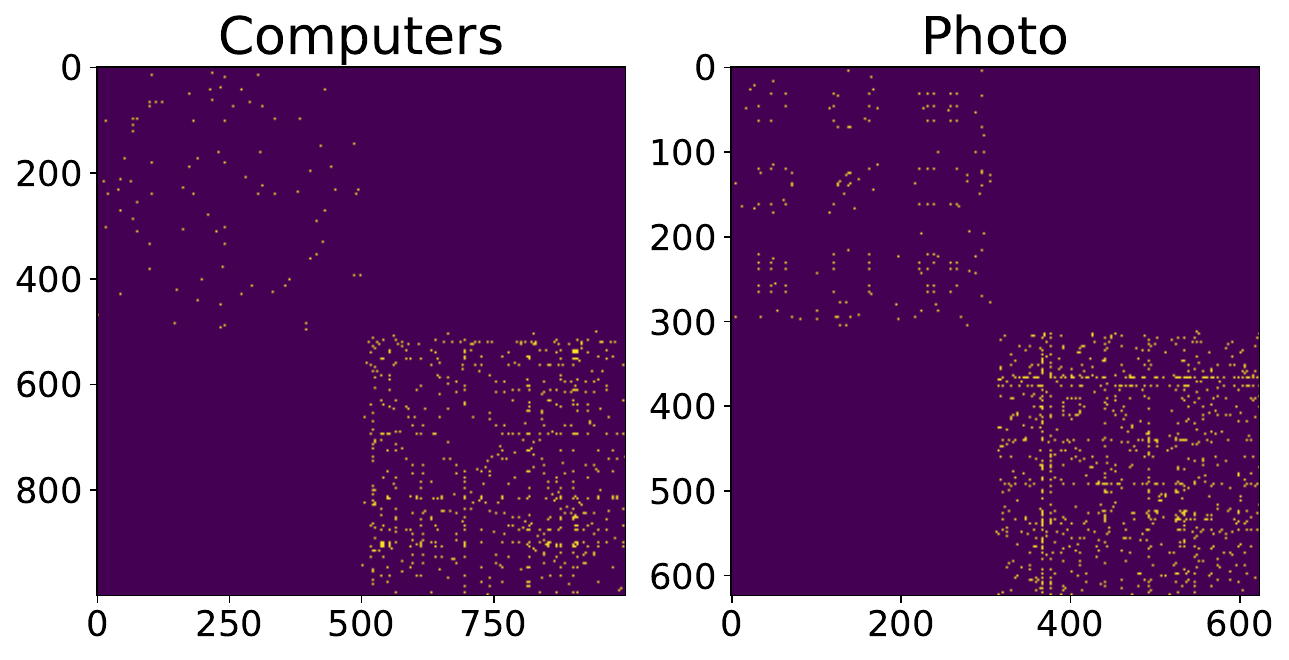}
    \caption[The synthetic graphs extracted from Computers and Photo]{The adjacency matrix of the synthetic graphs extracted from the real graphs Computers and Photo.}
    \label{dynam:fig:motivation_hybid}
\end{figure}

\section{Time Complexity} \label{dynam:app:time_comp}
Understanding the time complexity of the ADMP-GCN model is crucial for evaluating its practical efficiency and scalability. Table \ref{dynam:tab:time_complexity} reports the average training time, measured in seconds, for two distinct settings: ALM  and ST.

\begin{table}[h]
\centering
\caption{The average time needed for each training setting for different datasets. }
\resizebox{0.9\columnwidth}{!}{%
\begin{tabular}{llllllll}
\toprule
Model & Cora & squirel & chamelon & Computers  & Photo & Ogbn-arxiv  \\ \midrule
ADMP-GCN ALM & 14 & 36 & 15 & 51 & 26 & 266 \\
ADMP-GCN ST &32 & 88 & 40 & 175 & 87 & 882  \\ 
\bottomrule

\end{tabular}
}
\label{dynam:tab:time_complexity}
\end{table}

\section{Hyperparameter Configurations}\label{dynam:app:hyper}
For a more balanced comparison, however, we use the same training procedure for all the models. The hyperparameters in each dataset where performed using a Grid search on the classical GCN over the following search space:
\begin{itemize}
    \item Hidden size: $[16, 32, 64, 128, 256, 512],$,
    \item Learning rate: $[0.1, 0.01, 0.001],$
    \item Dropout probability: $[16, 32, 64, 128, 256, 512].$,
\end{itemize}

The number of layers was fixed to 2. The optimal hyperparameters can be found in Table \ref{dynam:tab:gcn_tuned_hyperparams}.

\begin{table}[h]
\caption{Hyperparameters used in our experiments.}
\label{dynam:tab:gcn_tuned_hyperparams}
\begin{center}
\begin{small}
\resizebox{0.95\columnwidth}{!}{%
\begin{tabular}{lccc}
\toprule
Dataset & Hidden Size & Learning Rate & Dropout Probability   \\
\midrule
Cora    & 64 & 0.01 & 0.8   \\
CiteSeer    &64 & 0.01 & 0.4   \\
PubMed     &64 & 0.01 & 0.2   \\\
CS     & 512 & 0.01 & 0.4 \\
arxiv-year   & 512 & 0.01  & 0.2 \\
chameleon   & 512 & 0.01  & 0.2 \\
cornell   & 512 & 0.01  & 0.2 \\
deezer-europe   & 512 & 0.01  & 0.2 \\
squirrel   & 512 & 0.01  & 0.2 \\
Wisconsin   &  512 & 0.01  & 0.2 \\
Texas   & 512 & 0.01  & 0.2 \\
Photo     & 512 & 0.01 & 0.6 \\
OGBN-Arxiv    & 512 & 0.01 & 0.5  \\
Computers     & 512 & 0.01 & 0.2 \\
Physics   & 512 & 0.01 & 0.4 \\
Penn94     & 64 & 0.01 & 0.2 \\

\bottomrule
\end{tabular}
}
\end{small}
\end{center}
\end{table}

\section{Supplementary Results of ADMP-GCN}\label{dynam:app:GCN}

We provide additional results on the ADMP-GCN in Tables \ref{dynam:tab:multi_task_2}, \ref{dynam:tab:oracle_2} and \ref{dynam:tab:results_with_baselines_2}. 

\begin{table}[ht]
\caption[Comparison of ADMP-GCN training paradigms ALM and ST]{Comparison of ADMP-GCN training paradigms ALM and ST. These paradigms are also compared to the single-task training setting to evaluate which approach most closely mimics the classical GCN under single-task training.  The best results for each dataset are \textbf{bolded}.}
\resizebox{\textwidth}{!}{%
\begin{tabular}{c|clccccccc}
\toprule
\#Layers & Training Paradigm & Model & Photo & Computers & Chameleon & Cornell & Wisconsin & Texas & Squirrel \\ \midrule

\multirow{3}{*}{\circled{0}} & \multirow{1}{*}{Single-task} & GCN & $70.18 {\scriptstyle \pm 3.39}$ & $56.83 {\scriptstyle \pm 2.92}$ & $33.55 {\scriptstyle \pm 0.00}$ & $40.54 {\scriptstyle \pm 0.00}$ & $70.59 {\scriptstyle \pm 0.00}$ & $64.86 {\scriptstyle \pm 0.00}$ & $26.42 {\scriptstyle \pm 0.00}$ \\ \cmidrule(lr){2-10}
& \multirow{2}{*}{Multi-task} & ADMP-GCN (ALM) & $68.23 {\scriptstyle \pm 3.11}$ & $56.61 {\scriptstyle \pm 2.07}$ & $30.70 {\scriptstyle \pm 0.00}$ & $40.54 {\scriptstyle \pm 0.00}$ & $52.94 {\scriptstyle \pm 0.00}$ & $64.86 {\scriptstyle \pm 0.00}$ & $22.60 {\scriptstyle \pm 0.06}$ \\
& & ADMP-GCN (ST) & $\mathbf{70.56 {\scriptstyle \pm 3.33}}$ & $\mathbf{57.94 {\scriptstyle \pm 3.62}}$ & $\mathbf{33.55 {\scriptstyle \pm 0.00}}$ & $\mathbf{40.54 {\scriptstyle \pm 0.00}}$ & $\mathbf{70.59 {\scriptstyle \pm 0.00}}$ & $64.86 {\mathbf{\scriptstyle \pm 0.00}}$ & $\mathbf{26.42 {\scriptstyle \pm 0.00}}$ \\ \midrule

\multirow{3}{*}{\circled{1}} & \multirow{1}{*}{Single-task} & GCN & $65.86 {\scriptstyle \pm 5.18}$ & $59.62 {\scriptstyle \pm 3.40}$ & $38.16 {\scriptstyle \pm 0.00}$ & $40.54 {\scriptstyle \pm 0.00}$ & $56.86 {\scriptstyle \pm 0.00}$ & $64.86 {\scriptstyle \pm 0.00}$ & $21.23 {\scriptstyle \pm 0.00}$ \\ \cmidrule(lr){2-10}
& \multirow{2}{*}{Multi-task} & ADMP-GCN (ALM) & $\mathbf{64.49 {\scriptstyle \pm 6.95}}$ & $\mathbf{57.60 {\scriptstyle \pm 6.63}}$ & $27.41 {\scriptstyle \pm 0.00}$ & $40.54 {\scriptstyle \pm 0.00}$ & $54.90 {\scriptstyle \pm 0.00}$ & $64.86 {\scriptstyle \pm 0.00}$ & $19.31 {\scriptstyle \pm 0.00}$ \\
& & ADMP-GCN (ST) & $64.03 {\scriptstyle \pm 3.63}$ & $56.61 {\scriptstyle \pm 5.44}$ & $\mathbf{38.16 {\scriptstyle \pm 0.00}}$ & $\mathbf{40.54 {\scriptstyle \pm 0.00}}$ & $\mathbf{56.86 {\scriptstyle \pm 0.00}}$ & $\mathbf{64.86 {\scriptstyle \pm 0.00}}$ & $\mathbf{21.23 {\scriptstyle \pm 0.00}}$ \\ \midrule

\multirow{3}{*}{\circled{2}} & \multirow{1}{*}{Single-task} & GCN & $82.32 {\scriptstyle \pm 2.97}$ & $69.05 {\scriptstyle \pm 2.60}$ & \textbf{$58.73 {\scriptstyle \pm 0.96}$} & \textbf{$54.59 {\scriptstyle \pm 2.91}$} & $57.06 {\scriptstyle \pm 1.85}$ & $62.16 {\scriptstyle \pm 1.21}$ & $33.40 {\scriptstyle \pm 0.68}$ \\ \cmidrule(lr){2-10}
& \multirow{2}{*}{Multi-task} & ADMP-GCN (ALM) & $48.10 {\scriptstyle \pm 6.28}$ & $34.06 {\scriptstyle \pm 10.87}$ & $40.39 {\scriptstyle \pm 0.95}$ & $44.32 {\scriptstyle \pm 1.32}$ & $56.86 {\scriptstyle \pm 1.52}$ & $\mathbf{62.16 {\scriptstyle \pm 0.00}}$ & $21.73 {\scriptstyle \pm 0.28}$ \\
& & ADMP-GCN (ST) & $\mathbf{85.80 {\scriptstyle \pm 0.43}}$ & $\mathbf{68.22 {\scriptstyle \pm 4.53}}$ & $\mathbf{58.77 {\scriptstyle \pm 1.08}}$ & $\mathbf{55.95 {\scriptstyle \pm 4.02}}$ & $\mathbf{56.86 {\scriptstyle \pm 1.52}}$ & $61.89 {\scriptstyle \pm 1.46}$ & $\mathbf{33.57 {\scriptstyle \pm 0.46}}$ \\ \midrule

\multirow{3}{*}{\circled{3}} & \multirow{1}{*}{Single-task} & GCN & $86.51 {\scriptstyle \pm 2.33}$ & \textbf{$75.32 {\scriptstyle \pm 4.18}$} & $58.40 {\scriptstyle \pm 2.46}$ & $45.95 {\scriptstyle \pm 3.42}$ & $43.33 {\scriptstyle \pm 2.23}$ & $44.59 {\scriptstyle \pm 4.05}$ & $36.29 {\scriptstyle \pm 0.73}$ \\ \cmidrule(lr){2-10}
& \multirow{2}{*}{Multi-task} & ADMP-GCN (ALM) & $79.78 {\scriptstyle \pm 4.17}$ & $53.40 {\scriptstyle \pm 9.51}$ & $49.67 {\scriptstyle \pm 1.03}$ & $38.92 {\scriptstyle \pm 5.43}$ & $46.47 {\scriptstyle \pm 2.16}$ & $50.00 {\scriptstyle \pm 3.25}$ & $29.65 {\scriptstyle \pm 2.22}$ \\
& & ADMP-GCN (ST) & $\mathbf{85.82 {\scriptstyle \pm 1.76}}$ & $\mathbf{73.69 {\scriptstyle \pm 3.67}}$ & $\mathbf{50.79 {\scriptstyle \pm 1.02}}$ & $\mathbf{48.38 {\scriptstyle \pm 3.07}}$ & $\mathbf{47.25 {\scriptstyle \pm 2.23}}$ & $\mathbf{57.30 {\scriptstyle \pm 5.38}}$ & $\mathbf{32.95 {\scriptstyle \pm 0.45}}$ \\ \midrule

\multirow{3}{*}{\circled{4}} & \multirow{1}{*}{Single-task} & GCN & $75.36 {\scriptstyle \pm 12.68}$ & $61.77 {\scriptstyle \pm 20.27}$ & $51.12 {\scriptstyle \pm 7.56}$ & $47.03 {\scriptstyle \pm 2.16}$ & $51.57 {\scriptstyle \pm 2.91}$ & $48.38 {\scriptstyle \pm 3.51}$ & \textbf{$37.73 {\scriptstyle \pm 0.80}$} \\ \cmidrule(lr){2-10}
& \multirow{2}{*}{Multi-task} & ADMP-GCN (ALM) & $82.37 {\scriptstyle \pm 4.86}$ & $62.61 {\scriptstyle \pm 5.71}$ & $49.63 {\scriptstyle \pm 1.90}$ & $44.59 {\scriptstyle \pm 4.23}$ & $48.04 {\scriptstyle \pm 2.67}$ & $52.43 {\scriptstyle \pm 4.86}$ & $28.99 {\scriptstyle \pm 1.40}$ \\
& & ADMP-GCN (ST) & $\mathbf{87.08 {\scriptstyle \pm 0.77}}$ & $\mathbf{72.35 {\scriptstyle \pm 5.44}}$ & $\mathbf{53.86 {\scriptstyle \pm 2.44}}$ & $\mathbf{51.35 {\scriptstyle \pm 3.20}}$ & $\mathbf{50.39 {\scriptstyle \pm 1.76}}$ & $\mathbf{57.84 {\scriptstyle \pm 5.01}}$ & $\mathbf{34.74 {\scriptstyle \pm 0.48}}$ \\ \midrule

\multirow{3}{*}{\circled{5}} & \multirow{1}{*}{Single-task} & GCN & $78.49 {\scriptstyle \pm 7.79}$ & $55.63 {\scriptstyle \pm 14.13}$ & $50.07 {\scriptstyle \pm 1.73}$ & $46.76 {\scriptstyle \pm 4.53}$ & $45.88 {\scriptstyle \pm 2.18}$ & $53.51 {\scriptstyle \pm 6.02}$ & $36.52 {\scriptstyle \pm 1.01}$ \\ \cmidrule(lr){2-10}
& \multirow{2}{*}{Multi-task} & ADMP-GCN (ALM) & $82.57 {\scriptstyle \pm 5.33}$ & $63.05 {\scriptstyle \pm 9.21}$ & $48.36 {\scriptstyle \pm 1.86}$ & $48.92 {\scriptstyle \pm 3.72}$ & $46.67 {\scriptstyle \pm 4.71}$ & $49.46 {\scriptstyle \pm 5.93}$ & $29.60 {\scriptstyle \pm 1.70}$ \\
& & ADMP-GCN (ST) & $\mathbf{86.37 {\scriptstyle \pm 0.85}}$ & $\mathbf{74.53 {\scriptstyle \pm 3.65}}$ & $\mathbf{53.86 {\scriptstyle \pm 1.39}}$ & $\mathbf{52.70 {\scriptstyle \pm 3.02}}$ & $\mathbf{50.59 {\scriptstyle \pm 1.47}}$ & $\mathbf{56.49 {\scriptstyle \pm 3.07}}$ & $\mathbf{32.11 {\scriptstyle \pm 1.14}}$ \\ 
\bottomrule

\end{tabular}
}
\label{dynam:tab:multi_task_2}
\end{table}

\begin{table}[ht]
\caption[Comparison of highest accuracy  for GCN and ADMP-GCN ST]{Comparison of highest accuracy ($\pm$ standard deviation) for GCN and ADMP-GCN ST, with the layer achieving the best accuracy indicated in brackets. The final row shows the \textit{Oracle Accuracy} for ADMP-GCN ST. Best results for each dataset are \textbf{bolded}.}
\resizebox{\textwidth}{!}{%
\begin{tabular}{llllllll}
\toprule
 Model & Photo & Computers & chamelon & Cornell  & Wisconsin & Texas  & Squirrel \\ \midrule

GCN & $86.51 {\scriptstyle \pm 2.33} \, [3]$ & $75.32 {\scriptstyle \pm 4.18} \, [3]$ & $58.73 {\scriptstyle \pm 0.96} \, [2]$ & $54.59 {\scriptstyle \pm 2.91} \, [2]$ & $70.59 {\scriptstyle \pm 0.00} \, [0]$ & $64.86 {\scriptstyle \pm 0.00} \, [0]$ & $37.73 {\scriptstyle \pm 0.80} \, [4]$ \\

ADMP-GCN DT & $87.08 {\scriptstyle \pm 0.77} \, [4]$ & $74.53 {\scriptstyle \pm 3.65} \, [5]$ & $58.77 {\scriptstyle \pm 1.08} \, [2]$ & $55.95 {\scriptstyle \pm 4.02} \, [2]$ & $70.59 {\scriptstyle \pm 0.00} \, [0]$ & $64.86 {\scriptstyle \pm 0.00} \, [0]$ & $34.74 {\scriptstyle \pm 0.48} \, [4]$ \\

ADMP-GCN ST - Oracle & $\mathbf{96.14} {\scriptstyle \pm \mathbf{0.81}}$ & $\mathbf{90.15} {\scriptstyle \pm \mathbf{0.97}}$ & $\mathbf{79.50} {\scriptstyle \pm \mathbf{1.31}}$ & $\mathbf{74.59} {\scriptstyle \pm \mathbf{3.24}}$ & $\mathbf{80.39} {\scriptstyle \pm \mathbf{0.00}}$ & $\mathbf{77.84} {\scriptstyle \pm \mathbf{2.36}}$ & $\mathbf{69.54} {\scriptstyle \pm \mathbf{0.65}}$ \\   \bottomrule

\end{tabular}
}
\label{dynam:tab:oracle_2}
\end{table}

\begin{table}[ht]
\centering
\caption[Comparison of ADMP-GCN training paradigms ALM and ST]{Comparison of ADMP-GCN training paradigms ALM and ST. These paradigms are also compared to the single-task training setting to evaluate which approach most closely mimics the classical GCN under single-task training. The best results for each dataset are \textbf{bolded}.}
\resizebox{\textwidth}{!}{%
\begin{tabular}{lllllllll}
\toprule
Model & Photo & Computers & chamelon & Cornell  & Wisconsin & Texas  & Squirrel \\ \midrule

JKNET-CAT & $87.92 {\scriptstyle \pm 1.98} ~[2]$ & $74.68 {\scriptstyle \pm 6.92}~[3]$ & $56.89 {\scriptstyle \pm 1.77}~[2]$ & $46.22 {\scriptstyle \pm 5.73}~[5]$ & $\underline{72.55 {\scriptstyle \pm 0.00}}~[0]$ & $64.86 {\scriptstyle \pm 0.00}~[0]$ & $\underline{41.26 {\scriptstyle \pm 0.88}}~[2]$ \\

JKNET-MAX & $88.02 {\scriptstyle \pm 2.21}~[2]$ &  $\mathbf{77.97{\scriptstyle \pm 2.57}}~[0]$ & $57.70 {\scriptstyle \pm 2.79}~[2]$ & $45.95 {\scriptstyle \pm 5.41}~[5]$ & $62.75 {\scriptstyle \pm 2.63}~[1]$ & $68.65 {\scriptstyle \pm 4.05}~[5]$ &  $\mathbf{41.36 {\scriptstyle \pm 0.59}}~[2]$ \\

JKNET-LSTM & $87.74 {\scriptstyle \pm 1.94}~[0]$ & $\underline{77.13 {\scriptstyle \pm 2.80}}~[1]$ & $53.07 {\scriptstyle \pm 4.47}~[2]$ & $43.78 {\scriptstyle \pm 3.78}~[2]$ & $62.94 {\scriptstyle \pm 2.23}~[1]$ & $64.86 {\scriptstyle \pm 0.00}~[0]$ & $41.32 {\scriptstyle \pm 0.58}~[5]$ \\

Residuals - GCNII & $86.99 {\scriptstyle \pm 2.47}~[2]$ & $62.83 {\scriptstyle \pm 19.61}~[2]$ &  $\underline{61.03 {\scriptstyle \pm 2.07}}~[2]$ & $55.14 {\scriptstyle \pm 8.48}~[5]$ & $70.59 {\scriptstyle \pm 0.00}~[0]$ & $64.86 {\scriptstyle \pm 0.00}~[0]$ & $35.90 {\scriptstyle \pm 1.25}~[5]$ \\

AdaGCN & $64.84 {\scriptstyle \pm 0.89}$ & $58.68 {\scriptstyle \pm 1.31}$ & $47.37 {\scriptstyle \pm 0.48}$ &  $\mathbf{70.00{\scriptstyle \pm 4.75}}$ &  $\mathbf{75.88 {\scriptstyle \pm 1.76}}$ &  $\mathbf{72.43 {\scriptstyle \pm 1.62}}$ & $29.15 {\scriptstyle \pm 0.77}$ \\

GPR-GNN &  $\mathbf{89.16 {\scriptstyle \pm 2.04}}~[2]$ & $77.43 {\scriptstyle \pm 3.32}~[2]$ & $63.29 \mathbf{{\scriptstyle \pm 0.91}}~[2]$ &  $\underline{62.70 {\scriptstyle \pm 4.65}}~[2]$ & $58.82 {\scriptstyle \pm 1.24}~[2]$ & $59.19 {\scriptstyle \pm 3.72}~[2]$ & $38.52 {\scriptstyle \pm 1.06}~[4]$ \\ \midrule

GCN & $86.51 {\scriptstyle \pm 2.33}$ & $75.32 {\scriptstyle \pm 4.18}$ & $58.73 {\scriptstyle \pm 0.96}$ & $54.59 {\scriptstyle \pm 2.91}$ & $70.59 {\scriptstyle \pm 0.00}$ & $64.86 {\scriptstyle \pm 0.00}$ & $37.73 {\scriptstyle \pm 0.80}$ \\

ADMP-GCN & $87.08 {\scriptstyle \pm 0.77}$ & $74.53 {\scriptstyle \pm 3.65}$ & $58.77 {\scriptstyle \pm 1.08}$ & $55.95 {\scriptstyle \pm 4.02}$ & $70.59 {\scriptstyle \pm 0.00}$ & $64.86 {\scriptstyle \pm 0.00}$ & $34.74 {\scriptstyle \pm 0.48}$ \\ \midrule

ADMP-GCN w/ \textit{Degree} & $88.15 {\scriptstyle \pm 1.52}$ & $75.53 {\scriptstyle \pm 2.09}$ & $58.75 {\scriptstyle \pm 0.91}$ & $44.32 {\scriptstyle \pm 4.22}$ & $63.92 {\scriptstyle \pm 1.30}$ & $56.49 {\scriptstyle \pm 1.89}$ & $35.07 {\scriptstyle \pm 1.13}$ \\

ADMP-GCN w/ \textit{$k$-core} & 88.31 ${\scriptstyle \pm 1.31}$ & $75.57 {\scriptstyle \pm 1.88}$ & $58.97 {\scriptstyle \pm 1.09}$ & $48.11 {\scriptstyle \pm 4.65}$ & $68.82 {\scriptstyle \pm 1.37}$ & $60.81 {\scriptstyle \pm 3.25}$ & $34.46 {\scriptstyle \pm 0.70}$ \\

ADMP-GCN w/ \textit{Walk Count} &  $\underline{88.42 {\scriptstyle \pm 1.51}}$ & 76.14 ${\scriptstyle \pm 2.08}$ & $58.29 {\scriptstyle \pm 1.22}$ & $51.89 {\scriptstyle \pm 4.32}$ & $67.45 {\scriptstyle \pm 1.80}$ & $\underline{68.65 {\scriptstyle \pm 2.16}}$ & $34.39 {\scriptstyle \pm 0.89}$ \\

ADMP-GCN w/ \textit{PageRank} & $88.21 {\scriptstyle \pm 1.45}$ & $75.50 {\scriptstyle \pm 2.23}$ & $58.44 {\scriptstyle \pm 1.23}$ & 55.14 ${\scriptstyle \pm 4.86}$ & $65.29 {\scriptstyle \pm 2.16}$ & $60.00 {\scriptstyle \pm 3.15}$ & 34.68 ${\scriptstyle \pm 0.66}$ \\ \bottomrule

\end{tabular}
}

\label{dynam:tab:results_with_baselines_2}
\end{table}

\section{Supplementary Results of ADMP-GIN}\label{dynam:app:GIN}

Further results on the ADMP-GIN can be found in Tables \ref{dynam:tab:multi_task_GIN}, \ref{dynam:tab:multi_task_GIN_2}, \ref{dynam:tab:oracle_GIN}, \ref{dynam:tab:oracle_GIN_2} and \ref{dynam:tab:results_with_baselines_GIN_2}. 

\begin{table}[ht]
\caption[Comparison of ADMP-GIN training paradigms ALM and ST]{Comparison of ADMP-GIN training paradigms ALM and ST. These paradigms are also compared to the single-task training setting to evaluate which approach most closely mimics the classical GIN under single-task training.  The best results for each dataset are \textbf{bolded}.}
\resizebox{\textwidth}{!}{%
\begin{tabular}{c|clcccccc}
\toprule
\# Layers & Training Paradigm & Model & Cora & CiteSeer & CS & PubMed & Genius & Ogbn-arxiv \\
\midrule

\multirow{3}{*}{\circled{0}} & \multirow{1}{*}{Single-task} & GIN & $56.40 {\scriptstyle \pm 0.06}$ & $57.14 {\scriptstyle \pm 0.09}$ & $87.17 {\scriptstyle \pm 1.41}$ & $72.49 {\scriptstyle \pm 0.08}$ & $80.78 {\scriptstyle \pm 1.03}$ & $48.87 {\scriptstyle \pm 0.04}$ \\ \cmidrule(lr){2-9}
& \multirow{2}{*}{Multi-task} & ADMP-GIN (ALM) & $\mathbf{58.00 {\scriptstyle \pm 0.00}}$ & $\mathbf{61.50 {\scriptstyle \pm 0.00}}$ & $86.36 {\scriptstyle \pm 0.78}$ & $\mathbf{73.20 {\scriptstyle \pm 0.00}}$ & $79.95 {\scriptstyle \pm 0.10}$ & $36.49 {\scriptstyle \pm 0.19}$ \\
& & ADMP-GIN (ST)  & $56.38 {\scriptstyle \pm 0.04}$ & $57.17 {\scriptstyle \pm 0.08}$ & $\mathbf{87.41 {\scriptstyle \pm 0.95}}$ & $72.47 {\scriptstyle \pm 0.14}$ & $\mathbf{80.47 {\scriptstyle \pm 0.91}}$ & $\mathbf{48.87 {\scriptstyle \pm 0.04}}$ \\ \midrule

\multirow{3}{*}{\circled{1}} & \multirow{1}{*}{Single-task} & GIN & $75.17 {\scriptstyle \pm 0.09}$ & $64.79 {\scriptstyle \pm 0.03}$ & $90.29 {\scriptstyle \pm 0.99}$ & $74.97 {\scriptstyle \pm 0.11}$ & $78.42 {\scriptstyle \pm 4.95}$ & $60.90 {\scriptstyle \pm 0.15}$ \\ \cmidrule(lr){2-9}
& \multirow{2}{*}{Multi-task} & ADMP-GIN (ALM) & $74.50 {\scriptstyle \pm 0.00}$ & $\mathbf{66.50 {\scriptstyle \pm 0.00}}$ & $88.66 {\scriptstyle \pm 1.18}$ & $\mathbf{75.40 {\scriptstyle \pm 0.00}}$ & $72.07 {\scriptstyle \pm 17.44}$ & $59.43 {\scriptstyle \pm 0.73}$ \\
& & ADMP-GIN (ST)  & $\mathbf{75.07 {\scriptstyle \pm 0.06}}$ & $64.80 {\scriptstyle \pm 0.00}$ & $\mathbf{90.82 {\scriptstyle \pm 1.15}}$ & $75.00 {\scriptstyle \pm 0.00}$ & $\mathbf{77.01 {\scriptstyle \pm 12.06}}$ & $\mathbf{60.85 {\scriptstyle \pm 0.01}}$ \\ \midrule

\multirow{3}{*}{\circled{2}} & \multirow{1}{*}{Single-task} & GIN & \textbf{$77.73 {\scriptstyle \pm 0.99}$} & \textbf{$65.23 {\scriptstyle \pm 1.45}$} & $87.93 {\scriptstyle \pm 0.71}$ & $76.05 {\scriptstyle \pm 1.14}$ & $78.89 {\scriptstyle \pm 0.48}$ & $16.23 {\scriptstyle \pm 9.60}$\\ \cmidrule(lr){2-9}
& \multirow{2}{*}{Multi-task} &ADMP-GIN (ALM) & $63.53 {\scriptstyle \pm 4.80}$ & $60.68 {\scriptstyle \pm 2.09}$ & $12.90 {\scriptstyle \pm 8.05}$ & $72.20 {\scriptstyle \pm 4.11}$ & $79.60 {\scriptstyle \pm 0.51}$ & $\mathbf{47.63 {\scriptstyle \pm 3.20}}$ \\
& & ADMP-GIN (ST)  & $\mathbf{78.07 {\scriptstyle \pm 0.68}}$ & $\mathbf{65.41 {\scriptstyle \pm 1.91}}$ & $\mathbf{87.47 {\scriptstyle \pm 2.18}}$ & $\mathbf{76.04 {\scriptstyle \pm 0.88}}$ & $72.99 {\scriptstyle \pm 13.48}$ & $10.13 {\scriptstyle \pm 8.00}$ \\ \midrule

\multirow{3}{*}{\circled{3}} & \multirow{1}{*}{Single-task} & GIN & $74.40 {\scriptstyle \pm 1.14}$ & $60.81 {\scriptstyle \pm 2.20}$ & $82.13 {\scriptstyle \pm 2.24}$ & $74.97 {\scriptstyle \pm 1.69}$ & $52.22 {\scriptstyle \pm 28.47}$ & $6.00 {\scriptstyle \pm 0.17}$ \\ \cmidrule(lr){2-9}
& \multirow{2}{*}{Multi-task} &ADMP-GIN (ALM) & $66.99 {\scriptstyle \pm 2.85}$ & $59.57 {\scriptstyle \pm 2.57}$ & $17.69 {\scriptstyle \pm 12.02}$ & $74.15 {\scriptstyle \pm 2.00}$ & $68.00 {\scriptstyle \pm 23.98}$ & $\mathbf{27.51 {\scriptstyle \pm 16.25}}$ \\
& & ADMP-GIN (ST)  & $\mathbf{76.28 {\scriptstyle \pm 1.11}}$ & $\mathbf{65.13 {\scriptstyle \pm 1.00}}$ & $\mathbf{83.60 {\scriptstyle \pm 3.72}}$ & $\mathbf{76.21 {\scriptstyle \pm 1.75}}$ & $\mathbf{80.04 {\scriptstyle \pm 0.09}}$ & $13.74 {\scriptstyle \pm 9.66}$ \\ \midrule

\multirow{3}{*}{\circled{4}} & \multirow{1}{*}{Single-task} & GIN & $67.68 {\scriptstyle \pm 4.07}$ & $57.54 {\scriptstyle \pm 2.92}$ & $47.37 {\scriptstyle \pm 17.10}$ & $73.62 {\scriptstyle \pm 1.63}$ & \textbf{$80.05 {\scriptstyle \pm 0.10}$} & $6.07 {\scriptstyle \pm 0.21}$ \\ \cmidrule(lr){2-9}
& \multirow{2}{*}{Multi-task} & ADMP-GIN (ALM) & $68.78 {\scriptstyle \pm 4.30}$ & $60.12 {\scriptstyle \pm 2.36}$ & $21.07 {\scriptstyle \pm 14.15}$ & $\mathbf{74.20 {\scriptstyle \pm 2.57}}$ & $\mathbf{79.36 {\scriptstyle \pm 1.18}}$ & $15.23 {\scriptstyle \pm 11.85}$ \\
& & ADMP-GIN (ST)  & $\mathbf{74.94 {\scriptstyle \pm 1.58}}$ & $\mathbf{65.21 {\scriptstyle \pm 1.54}}$ & $\mathbf{81.33 {\scriptstyle \pm 2.15}}$ & $\mathbf{76.46 {\scriptstyle \pm 1.04}}$ & $\mathbf{80.04 {\scriptstyle \pm 0.09}}$ & $\mathbf{16.02 {\scriptstyle \pm 10.78}}$ \\ \midrule

\multirow{3}{*}{\circled{5}} & \multirow{1}{*}{Single-task} & GIN & $32.48 {\scriptstyle \pm 10.47}$ & $54.76 {\scriptstyle \pm 2.32}$ & $18.56 {\scriptstyle \pm 11.32}$ & $67.98 {\scriptstyle \pm 5.80}$ & $80.05 {\scriptstyle \pm 0.10}$ & $6.19 {\scriptstyle \pm 0.39}$ \\ \cmidrule(lr){2-9}
& \multirow{2}{*}{Multi-task} &ADMP-GIN (ALM) & $67.46 {\scriptstyle \pm 4.34}$ & $59.74 {\scriptstyle \pm 1.95}$ & $29.16 {\scriptstyle \pm 13.84}$ & $73.94 {\scriptstyle \pm 2.42}$ & $79.99 {\scriptstyle \pm 0.10}$ & $14.61 {\scriptstyle \pm 12.11}$ \\
& & ADMP-GIN (ST)  & $\mathbf{71.34 {\scriptstyle \pm 2.09}}$ & $\mathbf{63.89 {\scriptstyle \pm 1.39}}$ & $\mathbf{79.10 {\scriptstyle \pm 1.84}}$ & $\mathbf{75.75 {\scriptstyle \pm 1.09}}$ & $79.57 {\scriptstyle \pm 1.40}$ & $\mathbf{15.72 {\scriptstyle \pm 12.13}}$ \\ 
\bottomrule

\end{tabular}
}
\label{dynam:tab:multi_task_GIN}
\end{table}

\begin{table}[ht]
\caption[Comparison of ADMP-GIN training paradigms ALM and ST]{Comparison of ADMP-GIN training paradigms ALM and ST. These paradigms are also compared to the single-task training setting to evaluate which approach most closely mimics the classical GIN under single-task training.  The best results for each dataset are \textbf{bolded}.}
\resizebox{\textwidth}{!}{%
\begin{tabular}{c|clccccccc}
\toprule

\#Layers & Training Paradigm & Model & Photo & Computers & Chameleon & Cornell & Wisconsin & Texas & Squirrel \\ \midrule

\multirow{3}{*}{\circled{0}} & \multirow{1}{*}{Single-task} &GIN & $70.19 {\scriptstyle \pm 2.91}$ & $56.83 {\scriptstyle \pm 3.89}$ & $33.55 {\scriptstyle \pm 0.00}$ & $40.54 {\scriptstyle \pm 0.00}$ & $70.59 {\scriptstyle \pm 0.00}$ & $64.86 {\scriptstyle \pm 0.00}$ & $26.42 {\scriptstyle \pm 0.00}$ \\ \cmidrule(lr){2-10}
& \multirow{2}{*}{Multi-task} &  ADMP-GIN \textit{ALM} & $\mathbf{69.16 {\scriptstyle \pm 3.27}}$ & $56.27 {\scriptstyle \pm 3.00}$ & $30.70 {\scriptstyle \pm 0.00}$ & $40.54 {\scriptstyle \pm 0.00}$ & $52.94 {\scriptstyle \pm 0.00}$ & $64.86 {\scriptstyle \pm 0.00}$ & $22.62 {\scriptstyle \pm 0.05}$ \\
& & ADMP-GCN \textit{ST}& $68.68 {\scriptstyle \pm 3.36}$ & $\mathbf{56.71 {\scriptstyle \pm 2.77}}$ & $\mathbf{33.55 {\scriptstyle \pm 0.00}}$ & $\mathbf{40.54 {\scriptstyle \pm 0.00}}$ & $\mathbf{70.59 {\scriptstyle \pm 0.00}}$ & $\mathbf{64.86 {\scriptstyle \pm 0.00}}$ & $\mathbf{26.42 {\scriptstyle \pm 0.00}}$ \\ \midrule

\multirow{3}{*}{\circled{1}} & \multirow{1}{*}{Single-task} & GIN & $82.32 {\scriptstyle \pm 2.03}$ & $70.90 {\scriptstyle \pm 2.64}$ & $61.03 {\scriptstyle \pm 0.10}$ & $40.54 {\scriptstyle \pm 0.00}$ & $56.86 {\scriptstyle \pm 0.00}$ & $64.86 {\scriptstyle \pm 0.00}$ & $47.65 {\scriptstyle \pm 0.27}$ \\ \cmidrule(lr){2-10}
& \multirow{2}{*}{Multi-task} &  ADMP-GIN \textit{ALM} & $78.79 {\scriptstyle \pm 2.84}$ & $65.28 {\scriptstyle \pm 2.17}$ & $56.75 {\scriptstyle \pm 0.09}$ & $40.54 {\scriptstyle \pm 0.00}$ & $54.90 {\scriptstyle \pm 0.00}$ & $64.86 {\scriptstyle \pm 0.00}$ & $44.12 {\scriptstyle \pm 0.39}$ \\
& & ADMP-GCN \textit{ST}& $\mathbf{83.35 {\scriptstyle \pm 1.67}}$ & $\mathbf{71.88 {\scriptstyle \pm 3.80}}$ & $\mathbf{60.96 {\scriptstyle \pm 0.00}}$ & $\mathbf{40.54 {\scriptstyle \pm 0.00}}$ & $\mathbf{56.86 {\scriptstyle \pm 0.00}}$ & $\mathbf{64.86 {\scriptstyle \pm 0.00}}$ & $\mathbf{47.65 {\scriptstyle \pm 0.00}}$ \\ \midrule

\multirow{3}{*}{\circled{2}} & \multirow{1}{*}{Single-task} &GIN & $83.86 {\scriptstyle \pm 2.19}$ & $63.39 {\scriptstyle \pm 10.30}$ & $63.57 {\scriptstyle \pm 1.01}$ & $61.62 {\scriptstyle \pm 2.65}$ & $54.71 {\scriptstyle \pm 2.70}$ & $64.86 {\scriptstyle \pm 2.96}$ & $19.84 {\scriptstyle \pm 1.59}$ \\ \cmidrule(lr){2-10}
& \multirow{2}{*}{Multi-task} & ADMP-GIN \textit{ALM} & $25.83 {\scriptstyle \pm 9.50}$ & $13.04 {\scriptstyle \pm 8.60}$ & $22.37 {\scriptstyle \pm 0.00}$ & $40.00 {\scriptstyle \pm 3.97}$ & $49.61 {\scriptstyle \pm 3.51}$ & $64.32 {\scriptstyle \pm 2.91}$ & $\mathbf{20.11 {\scriptstyle \pm 0.85}}$ \\
& & ADMP-GCN \textit{ST}& $\mathbf{68.44 {\scriptstyle \pm 23.78}}$ & $\mathbf{42.65 {\scriptstyle \pm 17.47}}$ & $\mathbf{62.30 {\scriptstyle \pm 2.18}}$ & $\mathbf{59.46 {\scriptstyle \pm 2.42}}$ & $\mathbf{53.33 {\scriptstyle \pm 2.60}}$ & $\mathbf{64.32 {\scriptstyle \pm 1.62}}$ & $19.31 {\scriptstyle \pm 0.00}$ \\ \midrule

\multirow{3}{*}{\circled{3}} & \multirow{1}{*}{Single-task} & GIN & $28.63 {\scriptstyle \pm 13.64}$ & $19.90 {\scriptstyle \pm 9.20}$ & $26.29 {\scriptstyle \pm 1.94}$ & $35.68 {\scriptstyle \pm 2.65}$ & $48.04 {\scriptstyle \pm 4.04}$ & $60.54 {\scriptstyle \pm 5.82}$ & $24.84 {\scriptstyle \pm 2.06}$ \\ \cmidrule(lr){2-10}
& \multirow{2}{*}{Multi-task} &  ADMP-GIN \textit{ALM} & $17.94 {\scriptstyle \pm 6.70}$ & $23.04 {\scriptstyle \pm 13.93}$ & $26.29 {\scriptstyle \pm 1.31}$ & $\mathbf{44.05 {\scriptstyle \pm 4.53}}$ & $50.20 {\scriptstyle \pm 5.13}$ & $64.59 {\scriptstyle \pm 7.69}$ & $19.95 {\scriptstyle \pm 2.40}$ \\
& & ADMP-GCN \textit{ST}& $\mathbf{71.32 {\scriptstyle \pm 19.80}}$ & $\mathbf{54.40 {\scriptstyle \pm 16.24}}$ & $\mathbf{48.84 {\scriptstyle \pm 2.76}}$ & $43.51 {\scriptstyle \pm 3.51}$ & $\mathbf{56.27 {\scriptstyle \pm 5.26}}$ & $\mathbf{64.59 {\scriptstyle \pm 14.58}}$ & $\mathbf{26.03 {\scriptstyle \pm 0.52}}$ \\ \midrule

\multirow{3}{*}{\circled{4}} & \multirow{1}{*}{Single-task} &GIN & $14.43 {\scriptstyle \pm 4.93}$ & $12.51 {\scriptstyle \pm 8.60}$ & $29.63 {\scriptstyle \pm 2.48}$ & $43.78 {\scriptstyle \pm 4.15}$ & $51.57 {\scriptstyle \pm 1.53}$ & $66.76 {\scriptstyle \pm 4.69}$ & $20.47 {\scriptstyle \pm 1.26}$ \\  \cmidrule(lr){2-10}
& \multirow{2}{*}{Multi-task} &  ADMP-GIN \textit{ALM} & $11.81 {\scriptstyle \pm 6.16}$ & $9.01 {\scriptstyle \pm 9.74}$ & $21.01 {\scriptstyle \pm 2.16}$ & $45.14 {\scriptstyle \pm 4.02}$ & $50.78 {\scriptstyle \pm 3.22}$ & $64.05 {\scriptstyle \pm 3.83}$ & $19.18 {\scriptstyle \pm 1.57}$ \\
& & ADMP-GCN \textit{ST}& $\mathbf{68.49 {\scriptstyle \pm 16.64}}$ & $\mathbf{52.61 {\scriptstyle \pm 19.82}}$ & $\mathbf{44.54 {\scriptstyle \pm 4.16}}$ & $\mathbf{45.14 {\scriptstyle \pm 4.84}}$ & $\mathbf{50.59 {\scriptstyle \pm 3.14}}$ & $\mathbf{68.92 {\scriptstyle \pm 5.70}}$ & $\mathbf{26.59 {\scriptstyle \pm 1.36}}$ \\ \midrule

\multirow{3}{*}{\circled{5}} & \multirow{1}{*}{Single-task} & GIN & $14.22 {\scriptstyle \pm 5.42}$ & $12.55 {\scriptstyle \pm 7.97}$ & $26.47 {\scriptstyle \pm 3.22}$ & $43.51 {\scriptstyle \pm 4.43}$ & $50.20 {\scriptstyle \pm 6.09}$ & $66.22 {\scriptstyle \pm 1.81}$ & $20.60 {\scriptstyle \pm 1.25}$ \\   \cmidrule(lr){2-10}
& \multirow{2}{*}{Multi-task} &  ADMP-GIN \textit{ALM} & $15.84 {\scriptstyle \pm 4.93}$ & $9.81 {\scriptstyle \pm 7.59}$ & $28.82 {\scriptstyle \pm 5.47}$ & $42.97 {\scriptstyle \pm 5.33}$ & $\mathbf{49.41 {\scriptstyle \pm 4.09}}$ & $65.14 {\scriptstyle \pm 5.05}$ & $19.77 {\scriptstyle \pm 1.52}$ \\
& & ADMP-GCN \textit{ST}& $\mathbf{68.48 {\scriptstyle \pm 13.80}}$ & $\mathbf{58.96 {\scriptstyle \pm 9.83}}$ & $\mathbf{43.31 {\scriptstyle \pm 3.01}}$ & $\mathbf{44.05 {\scriptstyle \pm 4.53}}$ & $45.69 {\scriptstyle \pm 5.19}$ & $\mathbf{67.57 {\scriptstyle \pm 9.44}}$ & $\mathbf{29.74 {\scriptstyle \pm 2.46}}$ \\ 
\bottomrule

\end{tabular}
}
\label{dynam:tab:m}
\label{dynam:tab:multi_task_GIN_2}
\end{table}

\begin{table}[ht]
\centering
\caption[Comparison of highest accuracy for GIN and ADMP-GIN ST]{Comparison of highest accuracy ($\pm$ standard deviation) for GIN and ADMP-GIN ST, with the layer achieving the best accuracy indicated in brackets. The final row shows the \textit{Oracle Accuracy} for ADMP-GIN ST. Best results for each dataset are \textbf{bolded}.}
\resizebox{0.85\textwidth}{!}{%
\begin{tabular}{lllllll}
\toprule
 Model & Cora & CiteSeer & CS & PubMed & Genius & Ogbn-arxiv \\  \midrule

GIN & $77.73 {\scriptstyle \pm 0.99} ~[2]$ & $65.23 {\scriptstyle \pm 1.45} ~[2]$ & $90.29 {\scriptstyle \pm 0.99} ~[1]$ & $76.05 {\scriptstyle \pm 1.14} ~[2]$ & $80.78 {\scriptstyle \pm 1.03} ~[0]$ & $60.90 {\scriptstyle \pm 0.15} ~[1]$ \\

ADMP-GIN (ST) & $78.07 {\scriptstyle \pm 0.68} ~[2]$ & $65.41 {\scriptstyle \pm 1.91} ~[2]$ & $90.82 {\scriptstyle \pm 1.15} ~[1]$ & $76.46 {\scriptstyle \pm 1.04} ~[4]$ & $80.47 {\scriptstyle \pm 0.91} ~[0]$ & $60.85 {\scriptstyle \pm 0.01} ~[1]$ \\

ADMP-GIN ST - Oracle & $\mathbf{90.76} {\scriptstyle \pm \mathbf{0.27}}$ & $\mathbf{81.87} {\scriptstyle \pm \mathbf{0.63}}$ & $\mathbf{97.64} {\scriptstyle \pm \mathbf{0.18}}$ & $\mathbf{92.73} {\scriptstyle \pm \mathbf{0.54}}$ & $\mathbf{92.07} {\scriptstyle \pm \mathbf{6.13}}$ & $\mathbf{71.23} {\scriptstyle \pm \mathbf{2.69}}$ \\   \bottomrule

\end{tabular}
}
\label{dynam:tab:oracle_GIN}
\end{table}

\begin{table}[ht]
\centering
\caption[Comparison of highest accuracy for GIN and ADMP-GIN ST]{Comparison of highest accuracy ($\pm$ standard deviation) for GIN and ADMP-GIN ST, with the layer achieving the best accuracy indicated in brackets. The final row shows the \textit{Oracle Accuracy} for ADMP-GIN ST. Best results for each dataset are \textbf{bolded}.}
\resizebox{0.85\textwidth}{!}{%
\begin{tabular}{llllllll}
\toprule
 Model & Photo & Computers & chamelon & Cornell  & Wisconsin & Texas  & Squirrel \\ \midrule

GIN & $83.86 {\scriptstyle \pm 2.19} \, [2]$ & $70.90 {\scriptstyle \pm 2.64} \, [1]$ & $63.57 {\scriptstyle \pm 1.01} \, [2]$ & $61.62 {\scriptstyle \pm 2.65} \, [2]$ & $70.59 {\scriptstyle \pm 0.00} \, [0]$ & $66.76 {\scriptstyle \pm 4.69} \, [4]$ & $47.65 {\scriptstyle \pm 0.27} \, [1]$ \\

ADMP-GIN \textit{ST} & $83.35 {\scriptstyle \pm 1.67} \, [1]$ & $71.88 {\scriptstyle \pm 3.80} \, [1]$ & $62.30 {\scriptstyle \pm 2.18} \, [2]$ & $59.46 {\scriptstyle \pm 2.42} \, [2]$ & $70.59 {\scriptstyle \pm 0.00} \, [0]$ & $68.92 {\scriptstyle \pm 5.70} \, [4]$ & $47.65 {\scriptstyle \pm 0.00} \, [1]$ \\

ADMP-GIN \textit{ST} - Oracle & $\mathbf{95.92} {\scriptstyle \pm \mathbf{1.03}}$ & $\mathbf{90.40} {\scriptstyle \pm \mathbf{2.66}}$ & $\mathbf{86.73} {\scriptstyle \pm \mathbf{0.90}}$ & $\mathbf{73.24} {\scriptstyle \pm \mathbf{4.75}}$ & $\mathbf{80.78} {\scriptstyle \pm \mathbf{1.18}}$ & $\mathbf{84.05} {\scriptstyle \pm \mathbf{3.07}}$ & $\mathbf{78.41} {\scriptstyle \pm \mathbf{0.89}}$ \\  \bottomrule

\end{tabular}
}
\label{dynam:tab:oracle_GIN_2}
\end{table}

\begin{table}[ht]
\centering
\caption[Classification accuracy for the baselines based on the GIN backbone]{Classification accuracy ($\pm$ standard deviation) on different node classification datasets for the baselines based on the GIN backbone. The higher the accuracy (in \%), the better the model. Highlighted are the \textbf{first}, \underline{second} best results. OOM means \textit{Out of memory}.}
\resizebox{\textwidth}{!}{%
\begin{tabular}{lllllllll}
\toprule
Model & Photo & Computers & chamelon & Cornell  & Wisconsin & Texas  & squirrel \\ \midrule
JKNET-MAX  & $82.48 {\scriptstyle \pm  2.95} ~[1]$ & $69.62 {\scriptstyle \pm  3.57} ~[1]$ & $59.61 {\scriptstyle \pm  2.24} ~[2]$ & $47.84 {\scriptstyle \pm  4.84} [4]$ & $57.84 {\scriptstyle \pm  2.19} ~[0]$ & $66.49 {\scriptstyle \pm 2.16} ~[5]$ & $23.19 {\scriptstyle \pm 3.55} ~[2]$\\ 
JKNET-LSTM   & $\mathbf{87.02 {\scriptstyle \pm 1.42} ~[2]}$ & $\mathbf{76.50 {\scriptstyle \pm 1.44} ~[2]}$   & $62.61 {\scriptstyle \pm 1.39} ~[2]$ & $50.54 {\scriptstyle \pm 3.64} ~[2]$ & $57.84 {\scriptstyle \pm 3.64} [4]$ & $70.81 {\scriptstyle \pm 3.15} ~[1]$ & $26.34 {\scriptstyle \pm 4.31} ~[0]$ \\
GPR-GIN &  $\underline{85.82 {\scriptstyle \pm  1.02} ~[2]}$  & $\underline{76.01 {\scriptstyle \pm 3.20} ~[3]}$ & $\mathbf{67.26 {\scriptstyle \pm 0.50} ~[2]}$  & $\mathbf{62.97 {\scriptstyle \pm 2.97} ~[5]} $ & $68.04 {\scriptstyle \pm 3.62} [4]$ & $\underline{73.78 {\scriptstyle \pm 2.72} ~[5]}$  & $46.78 {\scriptstyle \pm 1.80} ~[2]$ \\ \midrule
 GIN  &$ 83.86 {\scriptstyle \pm 2.19} ~[2]$ & $70.9 {\scriptstyle \pm 2.64} ~[1]$ & $63.57 {\scriptstyle \pm 1.01} ~[2]$ & $\underline{61.62 {\scriptstyle \pm 2.65} ~[2]}$  & $70.59 {\scriptstyle \pm 0.00} ~[0]$ & $66.76 {\scriptstyle \pm 4.69} [4]$ & $47.65 {\scriptstyle \pm 0.27} ~[1]$ \\ 
  ADMP-GIN  & $83.35 {\scriptstyle \pm 1.67} ~[1]$ & $71.88 {\scriptstyle \pm 3.8} ~[1]$ & $62.30 {\scriptstyle \pm 2.18} ~[2]$ & $59.46 {\scriptstyle \pm 2.42} ~[2]$ & $\underline{70.59 {\scriptstyle \pm 0.00} ~[0]}$ & $68.92 {\scriptstyle \pm 5.70} [4]$ & $\mathbf{47.65 {\scriptstyle \pm 0.00} ~[1]}$ \\ \midrule
ADMP-GIN w/ \textit{Degree}  & $84.15 {\scriptstyle \pm 1.17}$  & $74.55 {\scriptstyle \pm 2.89}$  & $64.43 {\scriptstyle \pm 1.47}$ & $46.49 {\scriptstyle \pm 4.65}$  & $66.47 {\scriptstyle \pm 2.83}$  & $69.46 {\scriptstyle \pm 3.21}$  & $\underline{47.65 {\scriptstyle \pm 0.00}}$   \\
ADMP-GIN w/ \textit{$\boldsymbol{k}$-core}  & $84.08 {\scriptstyle \pm 1.09}$  & $74.58 {\scriptstyle \pm 3.56}$  & $\underline{64.65 {\scriptstyle \pm 1.45}}$   & $46.76 {\scriptstyle \pm 4.84}$  & $\mathbf{70.78 {\scriptstyle \pm 4.68 }}$ & $71.89{\scriptstyle \pm 4.22}$  & $47.65 {\scriptstyle \pm 0.00}$  \\
ADMP-GIN w/ \textit{Walk Count}  & $84.03 {\scriptstyle \pm 1.31}$  & $73.85 {\scriptstyle \pm 3.70}$  & $63.38 {\scriptstyle \pm 1.53}$  &$ 58.11 {\scriptstyle \pm 6.07}$  & $63.14 {\scriptstyle \pm 3.90}$  & $\mathbf{74.86 {\scriptstyle \pm 2.97}} $ & $47.65 {\scriptstyle \pm 0.00}$  \\
ADMP-GIN w/ \textit{PageRank}  & $84.36 {\scriptstyle \pm 1.32}$  & $74.27 {\scriptstyle \pm 2.97}$  & $63.31 {\scriptstyle \pm 1.10}$  & $55.41 {\scriptstyle \pm 3.68}$  & $65.49 {\scriptstyle \pm  3.3}$  & $70.81 {\scriptstyle \pm 3.78}$  & $47.65 {\scriptstyle \pm 0.00}$  \\
\bottomrule

\end{tabular}
}
\label{dynam:tab:results_with_baselines_GIN_2}
\end{table}

\chapter{Appendix: Adversarial Robustness of GNNs}

\section{Proof Of Lemma \ref{lemma:worst_case}}\label{sec:proof_lemma}

\begin{proof}

Let $\mathcal{W}^{\alpha, \beta}_{\epsilon} = \{ (A_w, X_w) ; d^{\alpha, \beta}([G, X], [\tilde{G}, \tilde{X}]) \leq \epsilon  \}$ be the set of ``worst-case'' adversarial attacks within the attack budget $\epsilon$. We denote the expected vulnerability of a graph function $f$ on the set of worst-case adversarial examples as $Adv^{\alpha, \beta}_{\epsilon}[f]|_{\mathcal{W}^{\alpha, \beta}_{\epsilon}}$. 

By definition, we have $\mathcal{W}^{\alpha, \beta}_{\epsilon}  \subset \ B^{\alpha, \beta}(G,X ,\epsilon)$. Consequently, we have that
\begin{align}
Adv^{\alpha, \beta}_{\epsilon}[f]|_{\mathcal{W}^{\alpha, \beta}_{\epsilon}} & = \mathbb{P}_{(G, X) \sim \mathcal{D}_{\mathcal{G}, \mathcal{X}}} [ (\tilde{G}, \tilde{X}) \in \mathcal{W}^{\alpha, \beta}_{\epsilon}:   d_{\mathcal{Y}}(f(\tilde{G}, \tilde{X}), f(G, X)) > \sigma] \\
& \leq \mathbb{P}_{(G, X) \sim \mathcal{D}_{\mathcal{G}, \mathcal{X}}} [ (\tilde{G}, \tilde{X}) \in B^{\alpha, \beta}(G, X, \epsilon):   d_{\mathcal{Y}}(f(\tilde{G}, \tilde{X}), f(G, X)) > \sigma] \\
& \leq Adv^{\alpha, \beta}_{\epsilon}[f].\label{eq:worst_case}
\end{align}

Let the graph function $f$ be $((d^{\alpha, \beta}, \epsilon),( d_{\mathcal{Y}}, \gamma))$--robust, then $Adv^{\alpha, \beta}_{\epsilon}[f] \leq \gamma$. From \ref{eq:worst_case}, we have
\begin{center}
$\mathcal{W}-Adv^{\alpha, \beta}_{\epsilon}[f] \leq Adv^{\alpha, \beta}_{\epsilon}[f] \leq \gamma. $
\end{center}

We therefore, can conclude that if
    $f$ is $((d^{\alpha, \beta}, \epsilon),( d_{\mathcal{Y}}, \gamma))$--robust, then $f$ is $((d^{\alpha, \beta}, \epsilon),( d_{\mathcal{Y}}, \gamma))-\text{``worst-case'' robust.}$ 

\end{proof}

\section{Proof Of Proposition \ref{prop:equivalence}}\label{sec:proof_proposition}
\begin{proposition} \label{prop:equivalence}
Let $f: (\mathcal{G}, \mathcal{X}) \rightarrow \mathcal{Y}$ be a graph-based classifier subject to feature-based attacks and $\mathcal{X}$ be of dimension $D.$ Let $d_2^{0,1}$ be the graph distance corresponding to the spectral norm, $d_1^{0,1}$ corresponding to the $L_1$ norm and $d_p^{0,1}$ corresponding to the $L_p$ norm with $p>2$.
If $f$ is $((d_2^{0, 1}, \epsilon),( d_{\mathcal{Y}}, \gamma))$--robust, then $f$ is also $((d_p^{0, 1}, D^{-1/2}\epsilon),( d_{\mathcal{Y}}, \gamma))$--robust and $((d_1^{0, 1}, D^{-2}\epsilon),( d_{\mathcal{Y}}, \sqrt{D}\gamma))$--robust.
\end{proposition}



\begin{proof}
    
Let $\mathbb{R}^D$ the real finite-dimensional space. The set of standard considered norms of $\mathbb{R}^D$ for $x=(x_1, \ldots, x_D) \in \mathbb{R}^D$ is
$$\forall p >0, ~~  \lVert x  \lVert_p = \left ( \sum_{i=1}^{D} |x_i|^p \right )^{1/p}, \qquad \lVert x  \lVert_\infty = \max_{1\leq i\leq n}|x_i| .$$
As for any $x\in \mathbb{R}^D$, the mapping $p \mapsto \lVert x  \lVert_p $ is monotone decreasing \citep{raissouli2010various} , meaning that
\begin{equation*}
    q>p \geq  1 \Rightarrow  \lVert x  \lVert_q \leq  \lVert x  \lVert_p . \label{eq:equiv1}
\end{equation*} 

\label{eq:equv1}

Using \textit{Holder's inequality} with $s=q/p>1$, we have
\begin{align*}
   \left ( \lVert x  \lVert_p \right )^{p}  & = \sum_{i=1}^D |x_i|^p\\
                    & =  \sum_{i=1}^D |x_i|^p \cdot 1 \\
                    & \leq  \left ( \sum_{i=1}^D  \left (|x_i|^p \right )^{s} \right )^{\frac{1}{s}}  \left ( \sum_{i=1}^D  1 ^{\frac{s}{s-1}} \right )^{1-\frac{1}{s}} \\
                & =  \left ( \sum_{i=1}^D  \left (|x_i|^p \right )^{q/p} \right )^{\frac{p}{q}}  \left ( \sum_{i=1}^D  1 ^{\frac{q}{q-p}} \right )^{1-\frac{p}{q}} \\
                & = D^{1 - p/q} \left ( \lVert x  \lVert_q \right )^{p} .
\end{align*}

Thus, 
 \begin{equation}
     q>p \geq  1 \Rightarrow  \lVert x  \lVert_p \leq  D^{1/p - 1/q} \lVert x  \lVert_q . \label{eq:equiv2}
 \end{equation}

We additionally have that
\begin{align}
    \lVert x  \lVert_q & = \left ( \sum_{i=1}^{D} |x_i|^q  \right )^{1/q} , \nonumber \\
                & \leq \left (\sum_{i=1}^{D} \lVert x  \lVert_\infty^q  \right )^{1/q} \nonumber\\
                & \leq D^{1/q} \lVert x  \lVert_\infty \nonumber \\
                & \leq D^{1/q} \lVert x  \lVert_p . \label{eq:second}
\end{align}

From \ref{eq:equiv2} and \ref{eq:second}, we deduce that
 \begin{equation*}
     q>p\geq  1 \Rightarrow  D^{\frac{1}{q}-\frac{1}{p}} \lVert x  \lVert_p \leq   \lVert x  \lVert_q  \leq  D^{\frac{1}{q}} \lVert x  \lVert_p ,
 \end{equation*}

Consequently, for any matrix $N\in \mathbb{R}^{n \times D}$, we have the following
 \begin{align*}
     p>q \geq  1 & \Rightarrow \forall x\neq 0  \left\{
                                \begin{array}{l}
                                 D^{\frac{1}{q}-\frac{1}{p}}\lVert Nx  \lVert_p \leq   \lVert Nx  \lVert_q  \leq D^{\frac{1}{q}} \lVert Nx  \lVert_p  , \\ 
                                D^{\frac{1}{q}-\frac{1}{p}} \lVert x  \lVert_p \leq   \lVert x  \lVert_q  \leq  D^{\frac{1}{q}}\lVert x  \lVert_p , \\ 
                                    \end{array}
                                \right.  \\
            & \Rightarrow \forall x\neq 0  \left\{
                                \begin{array}{l}
                                 D^{\frac{1}{q}-\frac{1}{p}}\lVert Nx  \lVert_p \leq   \lVert Nx  \lVert_q  \leq D^{\frac{1}{q}} \lVert Nx  \lVert_p , \\ 
                               D^{-\frac{1}{q}}  \frac{1}{\lVert x  \lVert_p} \leq  \frac{1}{ \lVert x  \lVert_q}  \leq D^{-\frac{1}{q}+\frac{1}{p}}  \frac{1}{\lVert x  \lVert_p} , \\ 
                                    \end{array}
                                \right.  \\
             & \Rightarrow \forall x\neq 0 , ~~ D^{-\frac{1}{p}}  \frac{\lVert Nx  \lVert_p }{\lVert x  \lVert_p } \leq    \frac{\lVert Nx  \lVert_q }{\lVert x  \lVert_q }   \leq D^{\frac{1}{p}} \frac{\lVert Nx  \lVert_p }{\lVert x  \lVert_p } , \\
            & \Rightarrow  ~~ D^{-\frac{1}{p}}  \lVert N  \lVert_p \leq \lVert N \lVert_q  \leq D^{\frac{1}{p}} \lVert N  \lVert_p .
 \end{align*} 

 In particular, for $p=2$, we have 
\begin{equation}
    \forall q >2, \forall N \in \mathbb{R}^{n\times D} ~~~~  \lVert N  \lVert_2 \leq \sqrt{D} \lVert N \lVert_q  . \label{eq:norm_res1}
\end{equation}

Similar for $q=2$ and $p=1$, we have 
\begin{equation}
    \forall N \in \mathbb{R}^{n\times D} ~~~~  \lVert N  \lVert_2 \leq D\lVert N \lVert_1 .  \label{eq:norm_res2}
\end{equation}

The Inequalities \ref{eq:norm_res1} and \ref{eq:norm_res2} stay valid for the distance  related to the norm $L_p$, \ie 
\begin{align*}
    \forall X,\Tilde{X} \in \mathbb{R}^{n\times D} ~~,\forall p > 2, ~~~    d_2(X, \Tilde{X}) & \leq  \sqrt{D}  d_p(X, \Tilde{X}) , \\
    d_2(X, \Tilde{X}) & \leq D  d_1(X, \Tilde{X}) .
\end{align*}
Finally, for every input graph $G, X$, we have
\begin{align*}
  \forall p>2,~~  \{(G', X') \mid  d_p([G, X], [\tilde{G}, \tilde{X}]) <\frac{1}{\sqrt{D}} \epsilon \} \subset \{(G', X') \mid  d_2([G, X], [\tilde{G}, \tilde{X}]) < \epsilon \} ,  \\
  \{(G', X') \mid  d_1([G, X], [\tilde{G}, \tilde{X}]) < \frac{1}{D}\epsilon \} \subset \{(G', X') \mid  d_1([G, X],[ \tilde{G}, \tilde{X}]) < \epsilon \} .
\end{align*}

Therefore, the $((d_2^{0, 1}, \epsilon),( d_{\mathcal{Y}}, \gamma))$-robustness of a graph function $f$ implies the $((d_p^{0, 1}, D^{-1/2}\epsilon),( d_{\mathcal{Y}}, \gamma))$--robustness and $((d_1^{0, 1}, D^{-1}\epsilon),( d_{\mathcal{Y}}, \sqrt{D}\gamma))$--robustness.



\end{proof}

\section{Proof Of Theorem \ref{theo:main_result} and \ref{theo:structural_perturbtation}}\label{sec:proof_theorem}

\begin{proof} Let's consider a graph-function $f$ that is based on $L$ layers of GCN. The GCN message-passing propagation can be written for a node $u$ as
\begin{equation}
    h_u^{(\ell)} = \sigma^{(\ell)} (\underset{v \in \mathcal{N}(u) \bigcup \{ u \}}{\Sigma} \frac{W^{(\ell)} h_v^{(\ell-1)}}{ \sqrt{(1 + d_u)(1 + d_v)} }),
\end{equation}
where $W^{(\ell)} \in \mathbb{R}^{d_{\ell-1} \times d_{\ell}}$ is the learnable weight matrix with $d_{\ell}$ being the embedding dimension of layer $\ell$ and $\sigma^{(\ell)}$ is the activation function of $\ell$-th layer. We recall that $h^{(0)} = X \in \mathbb{R}^{n\times d}$ is set to the initial node features.

Let $X$ the original node features and denote by $X'$ the perturbed adversarial features. Let's consider a node $u \in V$, we denote by $h_u$ its representation in the clean graph and $h'_{u}$ its representation in the attacked graph , we note that the activation functions $(\sigma^{(\ell)})_{1 \leq \ell \leq L}$ are \textit{nonexpensive}. Since we only consider node feature based adversarial attacks, then by definition the original node $u$ and its corresponding node in the corrupted graph have the same neighborhood, we can therefore write

\begin{align*}
    &\lVert h_u^{(L)} - {h'}_{u}^{(L)} \lVert = \lVert h_u^{(\ell)} - {h'}_{u}^{(\ell)} \lVert
  \\
 & = \lVert \sigma^{(\ell)} (\underset{v \in \mathcal{N}(u) \bigcup \{ u \}}{\Sigma} \frac{W^{(\ell)} h_v^{(\ell-1)}}{ \sqrt{(1 + d_u)(1 + d_v)} }) - \sigma^{(\ell)} (\underset{v' \in \mathcal{N}(u) \bigcup \{ u \}}{\Sigma} \frac{W^{(\ell)} {h'}_{v'}^{(\ell -1 )}}{ \sqrt{(1 + d_{u})(1 + d_{v'})} }) \lVert \\
   & \leq \lVert W^{(\ell)} \lVert \lVert \underset{v \in \mathcal{N}(u) \bigcup \{ u \}}{\Sigma} \frac{ h_v^{(\ell-1)}}{ \sqrt{(1 + d_u)(1 + d_v)} } - \underset{v' \in \mathcal{N}(u) \bigcup \{ u \}}{\Sigma} \frac{{h'}_{v'}^{(\ell-1)}}{ \sqrt{(1 + d_{u})(1 + d_{v'})} } \lVert \\
   & \leq \lVert W^{(\ell)} \lVert \lVert \underset{v \in \mathcal{N}(u) \bigcup \{ u \}}{\Sigma} \frac{ h_v^{(\ell-1)} - {h'}_{v}^{(\ell-1)}}{ \sqrt{(1 + d_u)(1 + d_v)} } \lVert \\
   & = \lVert W^{(\ell)} \lVert \lVert \underset{v \in \mathcal{N}(u) \bigcup \{ u \}}{\Sigma} \frac{1}{\sqrt{(1 + d_u)(1 + d_v)}} [ \sigma^{(\ell-1)} (\underset{j \in \mathcal{N}(v) \bigcup \{ v \}}{\Sigma} \frac{W^{(\ell-1)} h_j^{(\ell-2)}}{ \sqrt{(1 + d_v)(1 + d_j)} }) - \\
   & \hspace{19em} \sigma^{(\ell-1)} (\underset{j \in \mathcal{N}(v) \bigcup \{ v \}}{\Sigma} \frac{W^{(\ell-1)} {h'}_{j}^{(\ell-2)}}{ \sqrt{(1 + d_{v})(1 + d_{j})} })  ] \lVert \\
   & \leq \lVert W^{(\ell)} \lVert \lVert W^{(\ell-1)} \lVert \lVert \underset{v \in \mathcal{N}(u) \bigcup \{ u \}}{\Sigma} \frac{1}{\sqrt{(1 + d_u)(1 + d_v)}} \underset{j \in \mathcal{N}(v) \bigcup \{ v \}}{\Sigma} \frac{h_j^{(\ell-1)} - {h'}_j^{(\ell-1)}}{ \sqrt{(1 + d_v)(1 + d_j)} }  \lVert \\
\end{align*}

By iteration on the same process, we get the following result:

\begin{align*}
    \lVert h_u^{(L)} - {h'}_{u'}^{(L} \lVert
   & \leq \prod_{l=1}^L \lVert W^{(l)}\lVert_{2} \lVert \underset{v \in \mathcal{N}(u) \bigcup \{ u \}}{\Sigma} \underset{j \in \mathcal{N}(v) \bigcup \{ v \}}{\Sigma} \ldots \\
   & \hspace{3em} \underset{z \in \mathcal{N}(y) \bigcup \{ y \}}{\Sigma} \frac{X_u - X'_{u}}{\sqrt{(1 + d_u)} (1 + d_w) (1 + d_j) \ldots  (1 + d_y) \sqrt{(1 + d_z)}}   \lVert \\
   & \leq \prod_{l=1}^L \lVert W^{(l)}\lVert \lVert \hat{w}_u \lVert \epsilon
\end{align*}

with $\hat{w}_u$ being the sum of normalized walks of length $(L-1)$ starting from node $u$. 

Let's now consider the final output of the model which represents in the case of node classification the individual output of each node. 

\begin{equation*}
    \lVert f(A, X) - f(A, X') \lVert = \lVert \begin{bmatrix}
  \vdots \\
  h_u^{(L)} - {h'}_{u}^{(L)} \\
  \vdots \\
\end{bmatrix} \lVert
\end{equation*}

Based on the previous formulation, we have the following results:

\begin{equation*}
    \lVert f(A, X) - f(A, X') \lVert_1 \leq \epsilon \prod_{l=1}^L \lVert W^{(l)}\lVert_{1}  \sum_{u \in \mathcal{V}}  \hat{w}_u 
\end{equation*}

\begin{equation*}
    \lVert f(A, X) - f(A, X') \lVert_{\infty} = \epsilon \prod_{l=1}^L \lVert W^{(l)}\lVert_{\infty}  \max_{u \in \mathcal{V}}  \hat{w}_u  
\end{equation*}

Therefore, we computed the Lipschitz constant of our considered model $f$. To link this constant to our robustness definition in Equation \ref{equation:robustness_definition_1}, we use the Markov inequality as follows

\begin{align*}
     Adv_{\epsilon}^{\alpha, \beta}[f] & = \mathbb{P}_{(G, X) \sim \mathcal{D}_{\mathcal{G}, \mathcal{X}}} [ (\tilde{G}, \tilde{X}) \in B_{\alpha, \beta}(G, X, \epsilon) \hspace{.5em}  \text{ s.t.}  \hspace{.5em}  d_{\mathcal{Y}}(f(\tilde{G}, \tilde{X}), f(G, X)) > \sigma]  \\
     & \leq  \frac{1}{\sigma} \mathbb{E}_{\substack{(G, X) \sim \mathcal{D}_{\mathcal{G}, \mathcal{X}} ,\\(G', X') \in B_{\alpha, \beta}((G,X) ,\epsilon)}}
        \left [\lVert f(A,X) - f(A,X') \lVert_{\infty}  \right ]   \\
    & \leq \frac{1}{\sigma}  \left (   \prod_{l=1}^L \lVert W^{(l)}\lVert_{op} \right )  \epsilon \max_{u \in \mathcal{V}} \mid \hat{w}_u \mid_1  .  
\end{align*}
Thus,  the classifier $f$ is $((d^{0, 1}, \epsilon),( d_\infty, \gamma))$--robust robust with 
$$\gamma  = \prod_{i=1}^{L} \lVert W^{(i)} \rVert  \epsilon \max_{u \in \mathcal{V}} \hat{w}_u /\sigma.$$

And we also have that the classifier $f$ is $((d^{0, 1}, \epsilon),( d_1, \gamma))$--robust with 
$$\gamma  = \prod_{i=1}^{L} \lVert W^{(i)} \rVert  \epsilon (\sum_{u \in \mathcal{V}} \hat{w}_u ) /\sigma.$$

\end{proof}



\begin{theorem}

Let $f: (\mathcal{G}, \mathcal{X}) \rightarrow \mathcal{Y}$ denote a graph function composed of $L$ GCN layers, where $W^{(i)}$ denotes the weight matrix of the $i$-th layer. Further, let $d^{1,0}$ be a graph distance. For structural attacks with a budget $\epsilon$, the function $f$ is $((d^{1, 0}, \epsilon),( d_2, \gamma))$--robust with
\begin{equation*}
\gamma = \prod_{i=1}^{L} \lVert W^{(i)} \rVert \lVert X \rVert \epsilon (1 + L  \prod_{i=1}^{L} \lVert W^{(i)} \lVert )/\sigma.    
\end{equation*}

\end{theorem}

\begin{proof}

Following the same analogy, let's consider the case of structural perturbations. We only consider that the activation functions $(\sigma^{(\ell)})_{1 \leq \ell \leq L}$ are \textit{nonexpensive}.
In this setting, the graph shift operator (the normalized adjacency matrix in our work) is the one to be edited and the input features are shared between the two graphs. Let $\Tilde{A}$ the clean adjacency and $h^{(i)}$ its hidden represented in the $i$-th layer and $\Tilde{A'}$, ${h'}^{(i)}$,  with  the attacked/perturbed one. 

Let's first consider the following 

\begin{align*}
    \lVert {h'}^{(l)}\rVert 
 & =  \lVert \sigma^{(l)}(\Tilde{A_2}{h'}^{(l-1)}W^{(l)}) \rVert \\
   & \leq \lVert \tilde{A_2} \rVert \lVert W^{(l)} \rVert \lVert {h'}^{(l-1)} \rVert \\
   & \leq \lVert W^{(l)} \rVert \lVert {h'}^{(l-1)} \rVert . 
\end{align*}

By recursion, we have: $\lVert {h'}^{(L)}\rVert  \leq \prod_{i=1}^{L} \lVert W^{(i)} \rVert \lVert X \rVert .$

From another side, we have the following

\begin{align*}
    \lVert \sigma^{(l)} (\tilde{A} h_1^{(l)} W^{(l)} ) 
 - \sigma^{(l)} (\tilde{A'} {h'}_2^{(l)} W^{(l)} ) \lVert_2
 & \leq \lVert \tilde{A} h^{(l)} W^{(l)}  -\tilde{A'} {h'}_2^{(l)} W^{(l)} \lVert_2 \\
   & \leq \lVert W^{(l)} \rVert \lVert \tilde{A} h^{(l)} - \tilde{A'} {h'}^{(l)} + \tilde{A} {h'}^{(l)} - \tilde{A} {h'}^{(l)} \lVert_2  \\
   & \leq \lVert W^{(l)} \rVert \lVert \tilde{A} (h^{(l)} - {h'}^{(l)}) - {h'}^{(l)} (\tilde{A} - \tilde{A}) \lVert_2  \\
   & \leq \lVert W^{(l)} \rVert ( \lVert \tilde{A} \lVert \lVert(h^{(l)} - {h'}^{(l)}) \lVert + \lVert {h'}^{(l)}\lVert \lVert(\tilde{A} - \tilde{A'}) \lVert_2) \\
   & \leq \lVert W^{(l)} \rVert \lVert(h^{(l)} - {h'}^{(l)}) \lVert + \lVert W^{(l)} \rVert \lVert {h'}^{(l)}\lVert \epsilon . \\
\end{align*}

By recursion, we get the following (By considering $\forall i > 1 : \lVert W^{(i)} \rVert \geq 1$)

\begin{align*}
    \lVert \Phi(A_1,X) -\Phi(A_2,X) \lVert_2
 & \leq \prod_{i=1}^{L} \lVert W^{(i)} \rVert \lVert X \rVert \epsilon + L  (\prod_{i=1}^{L} \lVert W^{(i)} \rVert)^2 \lVert X \rVert \epsilon  \\
 & \leq \prod_{i=1}^{L} \lVert W^{(i)} \rVert \lVert X \rVert \epsilon (1 + L  \prod_{i=1}^{L} \lVert W^{(i)} \lVert ) . \\
\end{align*}

Similarly to the previous proof, we deduce that the classifier $f$ is $((d^{1, 0}, \epsilon),( d_2, \gamma))$--robust where

\begin{center}
    $\gamma = \prod_{i=1}^{L} \lVert W^{(i)} \rVert \lVert X \rVert \epsilon (1 + L  \prod_{i=1}^{L} \lVert W^{(i)} \lVert )/ \sigma  .$
\end{center}
\end{proof}

\section{Proof Of Theorem \ref{theo:gin_results}}\label{sec:proof_generalization}

\begin{proof}

Let's now consider a graph-function $f$ that is based on $L$ GIN-layers (with a parameter $\zeta = 0,$ that is usually denoted by $\epsilon$ in the literature). The GIN message-passing propagation process can be written for a node $u$ as:

\begin{equation*}
    h_u^{(\ell + 1)} = T^{(\ell + 1)}( (1 + \zeta) h_u^{(\ell)} +  \underset{v \in \mathcal{N}(u)}{\Sigma} h_v^{(\ell)})
\end{equation*}

where $T$ denotes a Neural Networks (a MLP) for example and $\zeta$ denotes the parameter of the GIN.  We recall that $h^{(0)} = X \in \mathbb{R}^{n\times d}$ is set to the initial node features.

Let $X$ the original node features and $\tilde{X}$ the perturbed features. Let's consider a node $u \in V$, we denote by $h_u$ its representation in the clean graph and $h'_{u}$ its representation in the attacked graph , we note that we consider the activation functions to be \textit{nonexpensive}. Since we only consider node feature based adversarial attacks, then by definition the original node $u$ and its corresponding corrupted node have the same neighborhood, we can therefore write

\begin{align*}
    \lVert h_u^{(L)} - {h'}_{u}^{(L)} \lVert
 & = \lVert T^{(\ell)}( (1 + \zeta) h_u^{(\ell-1)} +  \underset{v \in \mathcal{N}(u)}{\Sigma} h_v^{(\ell-1)}) - T^{(\ell)}( (1 + \zeta) {h'}_{u}^{(\ell)} +  \underset{v \in \mathcal{N}(u)}{\Sigma} {h'}_v^{(\ell)}) \lVert \\
   & \leq \lVert W^{(\ell)} \lVert \lVert (1 + \zeta) (h_u^{(\ell-1)} - {h'}_{u}^{(\ell-1)}) + \underset{v \in \mathcal{N}(u)}{\Sigma} (h_v^{(\ell-1)} - {h'}_v^{(\ell-1)}) \lVert 
\end{align*}

We assume that the input feature space $\mathcal{H}_0$ is bounded, thus each hidden space $\mathcal{H}_i$ of the iterative process of message passing is bounded and let $B = \underset{\ell \leq L}{\max} B_{\ell}$ be its global maximum bound. From the previous result, we have

\begin{align*}
    \lVert h_u^{(\ell)} - {h'}_{u}^{(\ell)} \lVert
    \leq \lVert W^{(\ell)} \lVert \lVert (1 + \zeta) (h_u^{(\ell-1)} - {h'}_{u}^{(\ell-1)}) + B \times deg(u)  \lVert 
\end{align*}

By iteration over the previous process, and by considering $\forall i > 1 : \lVert W^{(i)} \rVert \geq 1$,  we can write

\begin{align*}
    \lVert h_u^{(L)} - {h'}_{u}^{(L)} \lVert
   & \leq B \times deg(u) \times \prod_{l=1}^L \lVert W^{(l)}\lVert \sum_{l=1}^{L} (1+\zeta)^{l} + \prod_{l=1}^L \lVert W^{(l)}\lVert (1+\zeta)^L  \lVert h_u^{(0)} - {h'}_{u}^{(0)} \lVert  \\
   & \leq B \times deg(u) \times \prod_{l=1}^L \lVert W^{(l)}\lVert \sum_{l=1}^{L} (1+\zeta)^{l} + \prod_{l=1}^L \lVert W^{(l)}\lVert (1+\zeta)^L  \epsilon  \\
   & \leq \prod_{l=1}^L \lVert W^{(l)}\lVert [B \times deg(u) \sum_{l=1}^{L} (1+\zeta)^{l} + (1+\zeta)^L  \epsilon]  
\end{align*}

Since we consider the GIN-parameter $\zeta \approx 0$, we can deduce from the previous result that
\begin{equation*}
    \lVert h_u^{(\ell + 1)} - {h'}_{u'}^{(\ell + 1)} \lVert \leq \prod_{l=1}^L \lVert W^{(l)}\lVert [B \times L \times deg(u) + \epsilon]
\end{equation*}

Let's now consider the final output of the model which represents in the case of node classification the individual output of each node. 
\begin{equation*}
    \lVert f(A, X) - f(A, X') \lVert = \lVert \begin{bmatrix}
  \vdots \\
  h_u^{(L)} - {h'}_{u}^{(L)} \\
  \vdots \\
\end{bmatrix} \lVert
\end{equation*}

Based on the previous formulation, we have the following results

\begin{equation*}
    \lVert f(A, X) - f(A, X') \lVert_{\infty} = \prod_{l=1}^L \lVert W^{(l)}\lVert_{\infty}  [B \times L \times \max_{u \in \mathcal{V}}  deg(u) + \epsilon] 
\end{equation*}

Similar to the previous proof of Theorem \ref{theo:main_result}, we connect the Lipschitz constant of our considered model $f$ to the robustness definition in Equation \ref{equation:robustness_definition_1} through Markov inequality and we get the following

$$\gamma  = \prod_{l=1}^L \lVert W^{(l)}\lVert_{\infty}  [B \times L \times \max_{u \in \mathcal{V}}  deg(u) + \epsilon]  /\sigma.$$

\end{proof}

\section{Generalizing to Any Graph Neural Network}
\label{appendix:generalization_GNN}
While this work focuses on GCNs, it can be easily extended to other GNN architecture. Given the iterative nature of GNNs, they can be viewed as a composition of multiple continuously differentiable functions, $f_i : \mathcal{H}_{i} \rightarrow\mathcal{H}_{i+1}$, where $\mathcal{H}_{i}$ represents the input space of the $i-th$ function. The Lipschitz constant, as shown in \cite{pmlr-v139-zhao21e} is given by $C = \prod_{i} \lVert \triangledown f_i  \lVert$. The functions $\{f_i\}_i$ can be grouped into two categories : \textbf{(i)} The set $\mathcal{L}$ of functions $f_i$ that are linear with a corresponding parameter $W_i$. ~ \textbf{(ii)} The set of functions that are not linear. We can hence decompose $C'$ according to the nature of the function as follows

\begin{equation*}
C' = \Big( \prod_{f_i \notin \mathcal{L}} \lVert \triangledown f_i  \rVert \Big)\Big( \prod_{f_i \in \mathcal{L}} \lVert W_i  \rVert \Big).
\end{equation*}

We can assume the input feature space $\mathcal{H}_0$ to be bounded; this a realistic assumption. Thus, each hidden space $\mathcal{H}_{i}$ is also bounded, and since $\{\triangledown f_i\}_{f_i \notin \mathcal{L}}$ are continuous functions on bounded spaces, there exists an upper bound $C_i$ for $ \lVert \triangledown f_i  \lVert $. Therefore, there exists $C>0$, such that, 

\begin{equation*}
C' \leq C \prod_{f_i \in \mathcal{L}} \lVert W_i  \lVert.   
\end{equation*}
We return to the previous case by simply taking $\epsilon' = \epsilon C/C'.$ Furthermore, following the assumption on the input feature space being bounded, it is possible to derive an upper bound on the GNN's robustness when subject to both structural and feature-based attacks simultaneously (See Appendix~\ref{appendix:simultaneous_attacks}).

\section{Vulnerability Upper-bound When Dealing With Both Structural and Node Features Attacks } 
\label{appendix:simultaneous_attacks}

In this section, we derive the upper-bound for a GCN's vulnerability $Adv_{\alpha, \beta, \epsilon}[f]$ when dealing with both structural and feature-based attacks simultaneously.

\begin{align*}
     Adv_{\alpha, \beta, \epsilon}[f] & \leq  \frac{1}{\sigma} \mathbb{E}_{\substack{(G_1, X_1) \sim \mathcal{D}_{\mathcal{G}, \mathcal{X}} ,\\(G_2, X_2) \in B_{\alpha, \beta}((G_1, X_1) ,\epsilon)}}
        \left [\lVert f(A_1,X_1) - f(A_2,X_2) \lVert_2  \right ]  \\
         & \leq \frac{1}{\sigma} \mathbb{E}_{\substack{(G_1, X_1) \sim \mathcal{D}_{\mathcal{G}, \mathcal{X}} ,\\(G_2, X_2) \in B_{\alpha, \beta}((G_1, X_1) ,\epsilon)}}
        \left [\lVert f(A_1,X_1) -  f(A_1,X_2) + f(A_1,X_2) - f(A_2,X_2) \lVert_2  \right ]    \\
    & \leq \frac{1}{\sigma} \mathbb{E}_{\substack{(G_1, X_1) \sim \mathcal{D}_{\mathcal{G}, \mathcal{X}} ,\\(G_2, X_2) \in B_{\alpha, \beta}((G_1, X_1) ,\epsilon)}}
        \left [\lVert f(A_1,X_1) -  f(A_1,X_2)\lVert_2    + \lVert f(A_1,X_2) - f(A_2,X_2) \lVert_2  \right ]   \\
    & \leq \frac{1}{\sigma} \mathbb{E}_{\substack{(G_1, X_1) \sim \mathcal{D}_{\mathcal{G}, \mathcal{X}} ,\\(G_2, X_2) \in B_{\alpha, \beta}((G_1, X_1) ,\epsilon)}}
        \left [\lVert f(A_1,X_1) -  f(A_1,X_2)\lVert_2  \right ]  + \\
        & \hspace*{3cm}\mathbb{E}_{\substack{(G_1, X_1) \sim \mathcal{D}_{\mathcal{G}, \mathcal{X}} ,\\(G_2, X_2) \in B_{\alpha, \beta}((G_1, X_1) ,\epsilon)}}
        \left [\lVert f(A_1,X_2) - f(A_2,X_2) \lVert_2  \right ]. \label{} \\
\end{align*}

We already established an upper-bound in Appendix \ref{sec:proof_theorem} for each term of the previous inequality. Therefore, we can have a new upper-bound for $Adv_{\alpha, \beta, \epsilon}[f]$

\begin{align*}
     Adv_{\alpha, \beta, \epsilon}[f] & \leq  \frac{1}{\sigma}  \left ( \prod_{i=1}^{L} \lVert W^{(i)} \rVert  \epsilon (\sum_{u \in \mathcal{V}} \hat{w_u} )^2  \right )  \epsilon +  \frac{1}{\sigma} \left ( \prod_{i=1}^{L} \lVert W^{(i)} \rVert \lVert X \rVert \epsilon (1 + L  \prod_{i=1}^{L} \lVert W^{(i)} \lVert )  \right ) . 
\end{align*}

Following the assumption on the input feature space being bounded, i.e. $\exists B >0, \forall X ~~~\lVert X \rVert  \leq B$, the inquality become

\begin{align*}
     Adv_{\alpha, \beta, \epsilon}[f] & \leq  \frac{1}{\sigma}  \left (   \prod_{i=1}^{L} \lVert W^{(i)} \rVert \right )    \left (  (\sum_{u \in \mathcal{V}} \hat{w_u} )^2   +  B (1 + L  \prod_{i=1}^{L} \lVert W^{(i)} \lVert ) \right ) \epsilon .
\end{align*}

Thus,  the classifier $f$ is $d^{0,1}$-$(\epsilon, \gamma)$ robust where 
$$\gamma  = \left (   \prod_{i=1}^{L} \lVert W^{(i)} \rVert \right )    \left (  (\sum_{u \in \mathcal{V}} \hat{w_u} )^2   +  B (1 + L  \prod_{i=1}^{L} \lVert W^{(i)} \lVert ) \right )  \epsilon/\sigma.$$

\subsection{Experimental Validation}
We assessed the efficacy of our proposed method against both Structural and Node-feature-based adversarial attacks simultaneously. This evaluation involved a combined evasion attack approach, wherein two attacks were integrated. Initially, we generated the perturbed adjacency matrix by employing a surrogate GCN model along with the "Mettack" (utilizing the ‘Meta-Self’ strategy) \citep{zugner2019adversarial}. Subsequently, we perturbed the node features using a random attack strategy involving the injection of Gaussian noise $\mathcal{N}(0, \mathbf{I})$ into the features. This perturbation was controlled by a scaling parameter $\psi$, which we identified as effective in prior experiments. For the structural perturbations, we set the perturbation budget to $0.1 E$, while for the node features, $\psi$ was set to $5.0$. We compared our method to the same considered defense benchmarks that were used in Section \ref{sec:experiments}.

\begin{table}[]
\centering
\caption[Attacked classification accuracy  after both structural attacks and node feature attacks application]{Attacked classification accuracy ($\pm$ standard deviation) of the models on different benchmark node classification datasets after both structural attacks and node feature attacks application.}

\resizebox{0.95\textwidth}{!}{%

\begin{tabular}{llllllll}
\hline
\multicolumn{1}{c}{Dataset} & \multicolumn{1}{c}{GCN} & \multicolumn{1}{c}{GCN-Jaccard} & \multicolumn{1}{c}{RGCN} & \multicolumn{1}{c}{GNN-SVD} & \multicolumn{1}{c}{GNNGuard} & \multicolumn{1}{c}{ParsevalR} & \multicolumn{1}{c}{GCORN} \\ \hline
Cora                        & 76.7   $\pm$ 1.2        & 76.7 $\pm$ 0.4                  & 78.1 $\pm$ 0.6           & 65.9 $\pm$ 4.0              & 61.0 $\pm$ 0.4               & 77.6 $\pm$ 1.1               & \textbf{79.8 $\pm$ 0.7}   \\
CiteSeer                    & 68.7 $\pm$ 0.3          & 71.5 $\pm$ 0.3                  & 68.9 $\pm$ 0.8           & 68.3 $\pm$ 0.5              & 65.0 $\pm$ 0.3               & 70.9 $\pm$ 0.6               & \textbf{74.2 $\pm$ 0.3}   \\
PubMed                      & 47.0 $\pm$ 0.4          & 49.4 $\pm$ 0.8                  & 51.2 $\pm$ 0.4           & 47.7 $\pm$ 0.5              & 40.3 $\pm$ 0.3               & 49.7 $\pm$ 1.1               & \textbf{53.9 $\pm$ 0.8}   \\
CoraML                      & 63.3 $\pm$ 0.0          & 32.2 $\pm$ 1.5                  & 59.7 $\pm$ 1.4           & 53.8 $\pm$ 0.8              & 32.2 $\pm$ 1.5               & 65.4 $\pm$ 0.2               & \textbf{77.4 $\pm$ 1.1}   \\ \hline
\end{tabular}
}
\end{table}

\section{Time and Complexity Analysis} \label{sec:time_complexity_analysis}
Drawing from the original paper \citep{bjork_paper}, it has been established that the Bjorck orthonormalization method will always converge, provided that the condition $\lVert W^TW - I \rVert_2 < 1$ is met. In this section, we will begin by examining the connection between the selected order, the number of iterations and the resulting performance and then present an empirical time analysis of the proposed GCORN on various datasets.

\subsection{On the Effect Of Order/Iterations}
As per the projection equation~\ref{equation:bjork}, it appears that a higher order/iteration generally results in a more precise projection of the weight matrix. In our study, we gauge the accuracy of the approximation by assessing the model's performance in the absence of adversarial attacks, as well as under attack. Although our experiments demonstrate that employing the first order with a restricted number of iterations can produce satisfactory outcomes, we believe that an hyper-parameter analysis should be conducted. Consequently, we evaluate the influence of modifying the order (while maintaining a constant number of iterations) on the resulting accuracy in Table~\ref{tab:order_iter_analysis}.

\begin{table}[h]
\small
\caption[Performance of GCN and our proposed GCORN model]{Performance of GCN and our proposed GCORN model, for different used approximation orders, on the Cora dataset.}
\label{tab:order_iter_analysis}
\vskip 0.01in
\begin{center}
\begin{small}
\resizebox{0.8\textwidth}{!}{
\begin{tabular}{lcccc}
\toprule
 & GCN & GCORN(1 ord) & GCORN(2 ord) & GCORN(3 ord) \\
\midrule
Training Time (in s)    & 2.8 $\pm$ 0.01 & 4.8 $\pm$ 0.07 & 8.7 $\pm$ 0.07 & 10.9 $\pm$ 0.08 \\
Accuracy w/o attack    & 79.2 $\pm$ 1.6 & 78.8 $\pm$ 1.3 & 79.8 $\pm$ 0.9 & 80.8 $\pm$ 1.1 \\
Accuracy w. attack    & 68.4 $\pm$ 1.9 & 77.1 $\pm$ 2.1 & 78.3 $\pm$ 1.1 & 78.6 $\pm$ 0.4 \\

\bottomrule
\end{tabular}
}
\end{small}
\end{center}
\vskip -0.1in
\end{table}

Table~\ref{tab:order_iter_analysis} shows both clean and attacked accuracy of the GCN and our proposed GCORN and the corresponding standard deviations for different orders. We observe that using higher order approximations generally leads to enhanced accuracy and robustness, albeit at the expense of increased running time. Therefore, the choice of order hyperparameter employed in the approximation must be carefully tuned for each application to find the best trade-off between robustness,accuracy and training time. In our experiments, we found that an approximation of order 1 yielded adequate and satisfactory results.

\subsection{On Training Time}

In line with the above, we also investigate the training time of our model compared to that of a standard GCN. To conduct this experiment, we held the order and number of iterations constant and utilized identical architectures and hyperparameters for both models (see Appendix~\ref{sec:dataset_implementation_details}).

\begin{table}[h]
\small
\caption[Mean training time analysis of a our GCORN]{Mean training time analysis (in s) of a our GCORN in comparison to the other benchmarks.}
\label{tab:time_analysis}
\vskip 0.01in
\begin{center}
\begin{small}
\begin{tabular}{lccccc}
\toprule
Dataset & GCN & GCN-K & AIRGNN & RGCN & GCORN  \\
\midrule
Cora    & 2.8 & 1.8 & 2.6 & 3.2 & 4.8  \\
CiteSeer    & 2.4 & 5.8 & 2.9 & 2.4 & 4.6  \\
PubMed    & 5.9 & 8.9 & 7.4 & 14.5 & 7.3  \\
CS    & 6.1 & 12.1 & 12.4 & 13.8 & 15.5  \\
Ogbn-Arxiv & 77.8 & 185.8 & 68.1 & 161.6 & 78.4  \\

\bottomrule
\end{tabular}
\end{small}
\end{center}
\vskip -0.1in
\end{table}

The mean training time are presented in Table \ref{tab:time_analysis}, indicating a trade-off between the improved robustness provided by the GCORN model and its slightly longer training time compared to the GCN model and other available methods.

\section{More Details About The Estimation Of Our Robustness Measure}  \label{appendix:prob_eval}
 
We begin by providing the pseudocode of the algorithm used to estimate our robustness measure.

    \begin{algorithm}[H]
    \small
    \textbf{Inputs: }Sphere Radius :  $\epsilon > 0,$ 
    Number of Samples $L_{max},$ 
    Number of Input Graphs $|\mathcal{D}|$\; 
    Initialize $Adv$ = 0\; 
    \ForEach{$[G_i,X_i] \in \mathcal{D}$}{
        Initialize $Adv_i$ = 0\; 
        \ForEach{$l = 1,\ldots,L_{max}$}{
            1. Sample a distance $r\in [0,\epsilon]$ from the prior distribution $p_\epsilon$ (see Appendix \ref{appendix:proof_lemma})\;
            2. Uniformly sample $Z_l \in \mathbb{R}^{n \times K}$ from $S_r$ (see Appendix \ref{appendix:prob_eval})\;
            3. Choose $\tilde{X}_l = X_i + Z_l$\;
            4. Update \\$~~~~~~~Adv_i \leftarrow Adv_i + \mathbf{1}\{d_{\mathcal{Y}}(f(\tilde{G}_l, \tilde{X}_l), f(G, X))>\sigma\}$\;
        }
        $Adv =Adv +  Adv_i/L_{max}$;
    }
    Return $Adv/|\mathcal{D}| $ \;
    \caption{Estimation of  $Adv^{\alpha, \beta}_{\epsilon}[f].$ }\label{algo:prob_eval}
    \end{algorithm}

We now detail how to uniformly sample from the set $S_r$ defined by, 
$$S_r = \{ Z\in \mathbb{R}^{n\times D}  ~ \mid ~~~\max_{1\leq i \leq n} \lVert Z_i\lVert_p = r \},$$
where $r \in [0, \epsilon]$. The set $S_r$ can also be defined in the following way,

\begin{align*}
  Z \in S_r   &\Leftrightarrow  \max_{1\leq i \leq n} \lVert  Z_i\lVert_p = r  \\
    &\Leftrightarrow \left\{
                    \begin{array}{ll}
                      \exists i_0 \in \{1,\ldots,n\} &\lVert  Z_{i_0}\lVert_p = r ,\\ 
                     \forall i \in \{1,\ldots,n\} &\lVert  Z_i\lVert_p \leq  r .
                    \end{array}
                \right.  
 \end{align*}

  Since the position of $i_0$ within $\{1,\ldots,n\}$ do not matter in the previous equivalence, we can uniformly sample $i_0 \in \{1,\ldots,n\}$, such that $r_{i_0} = \lVert  Z_{i_0}\lVert_p = r$ and that the other index $i \in \{1,\ldots,n\}\setminus\{i_0\} $ should satisfy $r_i= \lVert  Z_i\lVert_p \leq r $, i.e., $Z_i \in \mathbb{B}^{K}(r) = \{x \in \mathbb{R}^K| \lVert x \lVert_p \leq r  \}$. We will use another time \textit{Stratified Sampling} where we start by sampling $\{r_i\}_{i\neq i_0}$ within $[0, r]$. Using the Lemma \ref{lem:sampling_radius}, we can directly sample $\{r_i\}_{i\neq i_0}$ from the probability distribution 

 $$\forall i \in \{1,\ldots,n\}\setminus\{i_0\}, ~~ p(r_i) = K\frac{1}{r} \left ( \frac{r_i}{r}  \right )^{K-1} . $$

Once we know the values of $\{r_i\}_{i}$, the problem boils down to sampling $n$ vectors $\{Z_i\}_{1\leq i \leq n }$ from $\mathbb{R}^K$, such that 

\begin{equation}
    \forall i\in\{1, \ldots , n\}, ~~~~ \lVert Z_i  \lVert_p = r_i   .
    \label{eq:constraint_lp}
\end{equation} 

For that, we have to consider the two cases $p< \infty$ and $p=\infty$  

\subsection{Case Where $p<\infty$}
In this case, Equation \ref{eq:constraint_lp} can be written as follows

 \begin{align}
     \forall i\in\{1, \ldots , n\}, ~~~~ \lVert  Z_i \lVert_p =  r_i & \Leftrightarrow  \forall i\in\{1, \ldots , n\}, ~~~~  \sum_{j=1}^{K} |Z_{i,j}|^p = r_i^p  , \nonumber \\
      & \Leftrightarrow  \forall i\in\{1, \ldots , n\}, ~~~~  \sum_{j=1}^{K} \left (  \frac{|Z_{i,j}|}{r_i} \right )^{p} = 1 . \label{eq:sample}
 \end{align}

To satisfy Equation \eqnref{eq:sample}, we randomly uniformly partition the $[0,1]$ into $K$ parts. To do so, for each $i \in \{1, \ldots , n\}$, we randomly sample $D-1$ element from uniform distribution $\mathcal{U}[0,1]$; Let's denote $[p^{(i)}_1, \ldots ,p^{(i)}_{K-1}]$ the sorted list of sampled elements. We directly choose 
\begin{equation*}
      \mathcal{O}_{i,j} =  \left\{
    \begin{array}{ll}
              1-p^{(i)}_{j-1} & \text{if } j = K,\\
              p^{(i)}_j & \text{if } j = 0 ,\\ 
             p^{(i)}_j-p^{(i)}_{j-1} &  \text{otherwise .}  
            \end{array}
    \right.
\end{equation*}

The matrix $\mathcal{O} = \left ( \mathcal{O}_{i,j}  \right )_{1\leq i \leq n,1\leq j \leq D}$ satisfy 

$$\forall i\in\{1, \ldots , n\}, ~~~~  \sum_{j=1}^{K}  \mathcal{O}_{i,j} = 1 . $$
Figure \ref{fig:GraphConstructions} gives insight into the previously introduced concept of randomly partitioning in the case of the set $[0,1]$ into $K=4$.

Based on the previous elements, we can choose $Z$ such that 
\begin{align}
    \forall i,j, ~~~~~ \left (  \frac{|Z_{i,j}|}{r_i} \right )^{p} = \mathcal{O}_{i,j} & \Leftrightarrow   \forall i,j, ~~~~~ |Z_{i,j}| =  r_i \times \mathcal{O}_{i,j} ^{1/p} . \label{eq:O_Z}
\end{align}

\begin{figure*}[t]
    \centering
    \resizebox{0.7\columnwidth}{!}{%
    \tikzset{every picture/.style={line width=0.75pt}} 

\begin{tikzpicture}[x=0.75pt,y=0.75pt,yscale=-1,xscale=1]

\draw    (98.71,120.14) -- (492.5,120.75) ;
\draw    (150.41,114.96) -- (150.56,125.73) ;
\draw    (98.71,108.14) -- (98.71,132.14) ;
\draw    (353.92,114.76) -- (354.08,125.53) ;
\draw    (399.69,115.38) -- (399.85,126.15) ;
\draw    (492.5,108.75) -- (492.5,132.75) ;
\draw   (99.57,119.86) .. controls (99.52,124.53) and (101.82,126.89) .. (106.49,126.94) -- (114.78,127.03) .. controls (121.45,127.1) and (124.75,129.47) .. (124.7,134.14) .. controls (124.75,129.47) and (128.11,127.18) .. (134.78,127.26)(131.78,127.22) -- (143.06,127.35) .. controls (147.73,127.4) and (150.09,125.1) .. (150.14,120.43) ;
\draw   (150.71,120.43) .. controls (150.7,125.1) and (153.03,127.43) .. (157.7,127.44) -- (242.56,127.56) .. controls (249.23,127.57) and (252.56,129.9) .. (252.55,134.57) .. controls (252.56,129.9) and (255.89,127.57) .. (262.56,127.58)(259.56,127.58) -- (347.42,127.7) .. controls (352.09,127.71) and (354.42,125.38) .. (354.43,120.71) ;
\draw   (354.43,121) .. controls (354.37,125.67) and (356.67,128.03) .. (361.34,128.09) -- (366.91,128.16) .. controls (373.58,128.25) and (376.88,130.62) .. (376.82,135.29) .. controls (376.88,130.62) and (380.24,128.33) .. (386.91,128.41)(383.91,128.38) -- (392.48,128.49) .. controls (397.15,128.54) and (399.51,126.24) .. (399.57,121.58) ;
\draw   (399.86,122.14) .. controls (399.87,126.81) and (402.21,129.13) .. (406.88,129.12) -- (436.02,129.03) .. controls (442.69,129.01) and (446.03,131.33) .. (446.04,136) .. controls (446.03,131.33) and (449.35,128.99) .. (456.02,128.97)(453.02,128.98) -- (485.16,128.87) .. controls (489.83,128.86) and (492.15,126.52) .. (492.14,121.85) ;

\draw (92.83,134.17) node [anchor=north west][inner sep=0.75pt]   [align=left] {$\displaystyle 0$};
\draw (487.33,132.33) node [anchor=north west][inner sep=0.75pt]   [align=left] {$\displaystyle 1$};
\draw (142,93) node [anchor=north west][inner sep=0.75pt]   [align=left] {$\displaystyle p_{1}$};
\draw (345.33,92.33) node [anchor=north west][inner sep=0.75pt]   [align=left] {$\displaystyle p_{2}$};
\draw (389.33,91.67) node [anchor=north west][inner sep=0.75pt]   [align=left] {$\displaystyle p_{4}$};
\draw (114.67,134) node [anchor=north west][inner sep=0.75pt]   [align=left] {$\displaystyle \mathcal{O}_{1}$};
\draw (244,134.67) node [anchor=north west][inner sep=0.75pt]   [align=left] {$\displaystyle \mathcal{O}_{2}$};
\draw (370,134) node [anchor=north west][inner sep=0.75pt]   [align=left] {$\displaystyle \mathcal{O}_{3}$};
\draw (437.33,136.67) node [anchor=north west][inner sep=0.75pt]   [align=left] {$\displaystyle \mathcal{O}_{4}$};

\end{tikzpicture}
    }
    \caption[A toy example on how to randomly partition the set ${[0,1]}$]{
    A toy example on how to randomly partition the set $[0,1]$ into $K=4$ parts such as the sum of the parts lengths is 1. 
    We first uniformly sample $3$ elements from $[0,1]$ and reorder them $[p_1,p_2,p_3]$. 
    And subsequently, we consider $\mathcal{O}_1 = [0,p_1], ~~\mathcal{O}_2= [p_4,p_3], ~~\mathcal{O}_3 = [p_4,p_3], ~~\mathcal{O}_4 = [1,p_4]$.
}
    \label{fig:GraphConstructions}
\end{figure*}

Equation \ref{eq:O_Z} is invariant with respect to the sign of $Z_{i,j}$, hence, we use the following
$$  \forall i,j, ~~~~~ Z_{i,j} = u_{i,j} \times   r_i \times \mathcal{O}_{i,j} ^{1/p} , $$
where $u_{i,j}$ is sampled from the discrete uniform distribution $\mathcal{U}_{\{-1,1\}}$.

We recall that for $p=1$, the distance $d^{0,1}([G, X], [\Tilde{G} , \Tilde{X}]) = \max_{i \in \{1,\ldots, n\}} \lVert X_{i}-\Tilde{X}_{i}\lVert_p$ defined in Equation \ref{eq:prob_eval_distance} matches with the infinity matrix distance $d_\infty: (A,B) \mapsto \sum_{X\neq 0} \frac{\left \| AX-BX \right \|_\infty }{\left \| X \right \|_\infty } $ as shown in \cite{nagisa2022p}. Therefore, from the Theorem \ref{theo:main_result},  the upper-bound of the expected vulnerability $Adv^{\alpha, \beta}_{\epsilon}[f]$ is $\gamma  = \prod_{i=1}^{L} \lVert W^{(i)} \rVert_\infty  \epsilon \hat{w}_G /\sigma$.

\subsection{Case Where $p=\infty$}
In this case, Equation \ref{eq:constraint_lp} can be written as follows

 \begin{align}
     \forall i\in\{1, \ldots , n\}, ~~~~ \lVert  Z_i \lVert_\infty =  r_i & \Leftrightarrow  \forall i\in\{1, \ldots , n\}, ~~~~  \max_{1 \leq j \leq K} |Z_{i,j}|= r_i \nonumber \\
      & \Leftrightarrow  \forall i\in\{1, \ldots , n\}, ~~~~  \left\{
                    \begin{array}{ll}
                      \exists j_0 \in \{1,\ldots,K\} &   |Z_{i,j_0}| = r_i , \\ 
                     \forall j \in \{1,\ldots,K\} &|Z_{i,j_0}| \leq  r_i.
                    \end{array}
                \right.  
 \end{align}
 
For each $i\in\{1, \ldots , n\}$, we randomly select $j_0  \in \{1,\ldots,K\}$ to satisfy $|Z_{i,j_0}| = r_i$. For the other $j\neq j_0$, we uniformly sample the value of $|Z_{i,j}| $ from $\mathcal{U}_{[0,r_i]}$. These equations are invariant with respect to the sign of $Z_{i,j}$, therefore, we choose, 
$$  \forall i,j, ~~~~~ Z_{i,j} = u_{i,j} \times   |Z_{i,j}| , $$
where $u_{i,j}$ is sampled from the discrete uniform distribution $\mathcal{U}_{\{-1,1\}}$.

\section{Proof of Lemma \ref{lem:sampling_radius}}
\label{appendix:proof_lemma}

\begin{proof}
Using the following equivalence 
$$\exists  Z \in  \mathbb{R}^{n,K} \max_{1\leq i \leq n} \lVert Z_i\lVert_p = r   \Leftrightarrow  \exists T \in \mathbb{R}^K, ~~  \lVert T \lVert_p = r .$$
We deduce that the density $p_\epsilon$ do not depend on $n$, and therefore we can set $n=1$ for this proof. Therefore,
$$\mathcal{B}_\epsilon = \left \{  Z \in \mathbb{R}^{ K}| \lVert Z \lVert_p \leq \epsilon  \right \} .$$

We start by calculating the volume of the $K-$sphere of radius $r$ for any real finite-dimensional space $\mathbb{R}^K$ where $K>2$ is an integer.
$$\mathbb{S}^{K}(r) = \{x \in \mathbb{R}^K| \lVert x \lVert_p = r  \} .$$
Let's denote the volume of $\mathbb{S}^{K}(r)$ by $\mathcal{V}^{K}(r)$.\\
Using \textit{Fubini Theorem}, we have 

\begin{align}
  \mathcal{V}^{K}(r) & = \int_{-r}^{r}    \mathcal{V}^{K-1} \left ( \left ( r^p - |x|^p\right )^{1/p}\right ) dx  \nonumber\\
  & = r \int_{-1}^{1}    \mathcal{V}^{K-1}\left (  r \left ( 1 - |x|^p   \right )^{1/p}\right ) dx  \nonumber\\
  & = r \int_{-1}^{1}   r^{K-1}  \mathcal{V}^{K-1}\left ( \left ( 1 - |x|^p   \right )^{1/p} \right)  dx   \nonumber\\
    & = r^{K}\int_{-1}^{1}     \mathcal{V}^{K-1}\left ( \left ( 1 - |x|^p   \right )^{1/p} \right ) dx   \nonumber\\
      & = r^{K}    \mathcal{V}^{K} (  1 ) . \nonumber\\
\end{align}

Thus, the surface $L^p$ ball in $\mathbb{R}^{K}$ of radius $r$ is 

\begin{align}
  \mathcal{S}^{K}(r)  & = \frac{d \mathcal{V}^{K}(r)}{d r}   \\
                & = K r^{K-1} \mathcal{V}^{K} (  1 ) .
\end{align}

Consequently the density distribution of $\mathbf{R}^{(p)}$ could be written as
\begin{align}
   p_{\epsilon}(r)  & = \frac{\mathcal{S}_{K}^{p}(r)}{\mathcal{V}_{K}^{p}(\epsilon)}\mathbf{1}\{ 0 \leq r \leq \epsilon \}   \nonumber\\
  & = \frac{1}{\mathcal{V}_{K}^{p}(\epsilon)}\frac{d \mathcal{V}_{K}^{p}(r)}{d r}\mathbf{1}\{ 0 \leq r \leq \epsilon \}   \nonumber \\
                & = a \frac{1}{\epsilon} \left ( \frac{r}{\epsilon}\right )^{K-1}\mathbf{1}\{ 0 \leq r \leq \epsilon \}.
\end{align}
We note that the previous quantity doesn't depend on $p$.
\end{proof}

\section{Additional Results} 
\label{appendix:additional_results}
\subsection{Node Classification}

\begin{figure}[t]
\begin{center}
\includegraphics[width=\columnwidth]{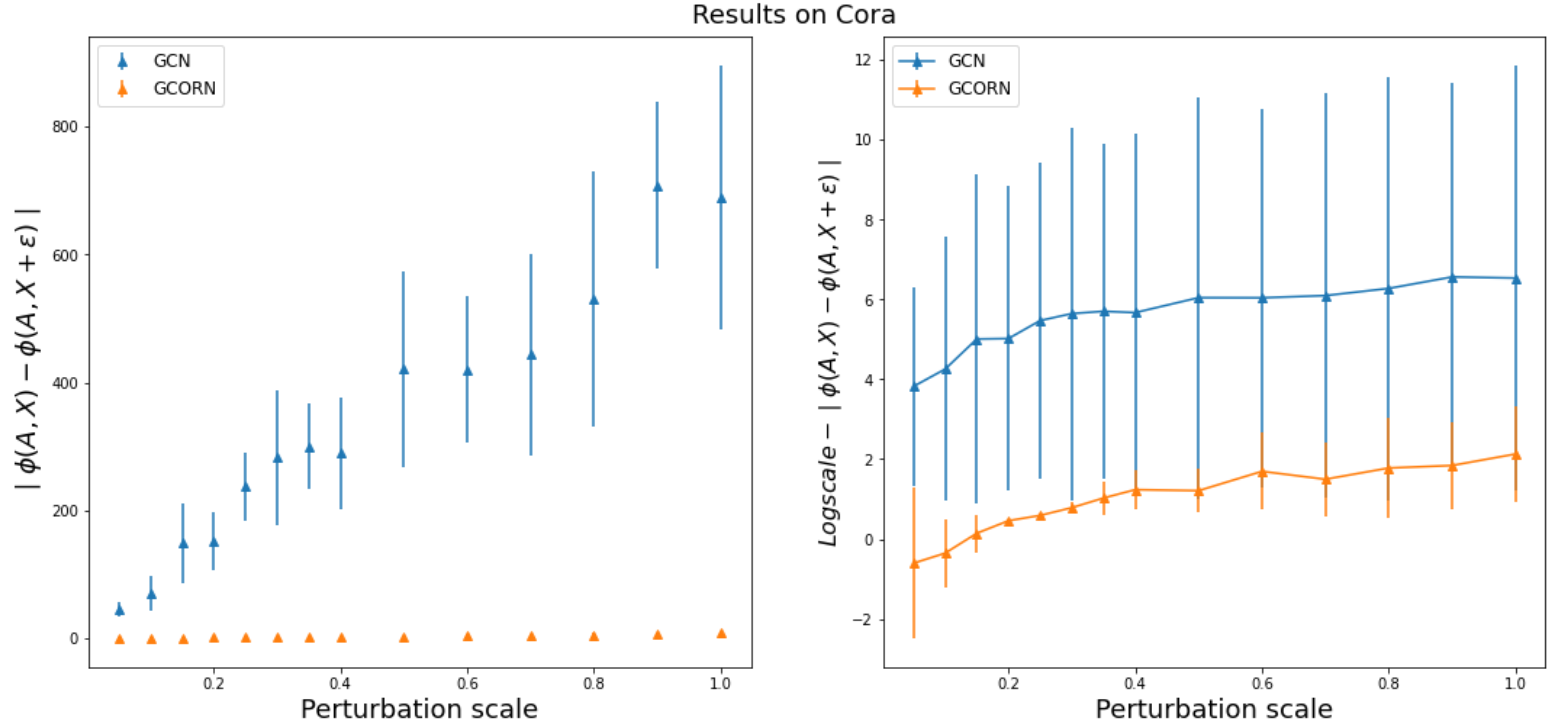}
\includegraphics[width=\columnwidth]{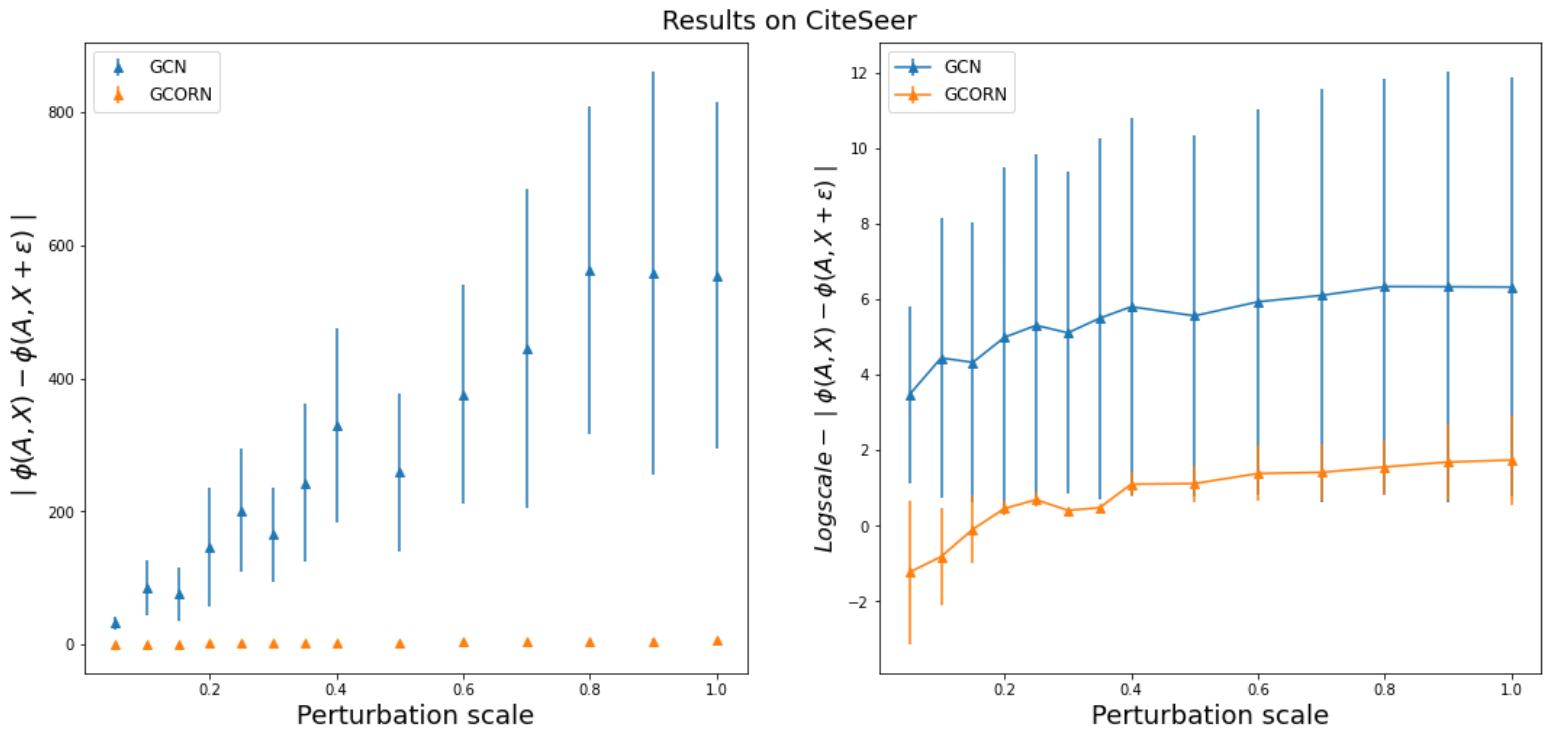}
\caption[Difference in output for the GCN and our GCORN]{Difference (in average and standard deviation) in output for the GCN and our GCORN when subject to random perturbations with different attack budgets for both (a) Cora and (b) CiteSeer. The right-hand side plots are on the log-scale.}
\label{fig:plot_diff_nrom}
\end{center}
\end{figure}

To further understand the relationship between robustness and input sensitivity, from which we derived the upper bound for GNNs and specifically GCNs, we empirically validate the trade-off between robustness and input perturbation. We hence compare the difference in the output of each model when subjected to random perturbations with varying attack budgets. 
The results of this study are presented in Figure \ref{fig:plot_diff_nrom}. The figure displays the results for both Cora and CiteSeer datasets in both absolute (left figure) and log-scale (right figure) format. The results demonstrate that the GCN is highly sensitive to perturbations in comparison to the proposed approach in terms of the difference in output. These findings provide validation for the stability of GCORN against input perturbations, and by extension, adversarial attacks.

We additionally considered two defense techniques benchmarks: \textbf{(1)} ElasticGNN \citep{elasticnet} which proposes a novel and general message passing scheme into GNNs to enhance the local smoothness adaptivity of GNNs via $\ell_1$-based graph smoothing and \textbf{(2)} EvenNet \citep{evennet} which proposes a spectral GNN corresponding to even-polynomial graph filter with the main idea being that ignoring odd-hop neighbors improves the underlying robustness. For this additional evaluation, we consider similar attacks techniques as the one from Table \ref{tab:results_node_classification}.

\begin{table}[h]
\centering
\caption[Attacked classification accuracy after the attack application]{Attacked classification accuracy ($\pm$ standard deviation) of the models on different benchmark node classification dataset after the attack application. }

\begin{tabular}{llccc}
\hline
Attack                                                                           & Dataset  & ElasticGNN     & EvenNet        & GCORN          \\ \hline
\multirow{3}{*}{\begin{tabular}[c]{@{}l@{}}Random \\ ($\psi =0.5$)\end{tabular}} & Cora     & 72.5 $\pm$ 2.4 & 69.7 $\pm$ 1.8 & 77.1 $\pm$ 1.8 \\
                                                                                 & CiteSeer & 58.9 $\pm$ 2.3 & 48.6 $\pm$ 1.6 & 67.8 $\pm$ 1.4 \\
                                                                                 & PubMed   & 70.5 $\pm$ 0.6 & 68.8 $\pm$ 1.6 & 73.1 $\pm$ 1.1 \\ \hline
\multirow{3}{*}{\begin{tabular}[c]{@{}l@{}}Random \\ ($\psi =0.5$)\end{tabular}} & Cora     & 54.4 $\pm$ 2.4 & 48.9 $\pm$ 2.1 & 57.6 $\pm$ 1.9 \\
                                                                                 & CiteSeer & 43.8 $\pm$ 2.0 & 39.7 $\pm$ 4.2 & 57.3 $\pm$ 1.7 \\
                                                                                 & PubMed   & 56.8 $\pm$ 2.1 & 58.5 $\pm$ 3.2 & 65.8 $\pm$ 1.4 \\ \hline
\multirow{3}{*}{PGD}                                                             & Cora     & 64.7 $\pm$ 1.6 & 57.6 $\pm$ 3.6 & 71.1 $\pm$ 1.4 \\
                                                                                 & CiteSeer & 50.8 $\pm$ 3.2 & 37.1 $\pm$ 3.3 & 65.6 $\pm$ 1.4 \\
                                                                                 & PubMed   & 64.6 $\pm$ 1.5 & 61.0 $\pm$ 4.1 & 72.3 $\pm$ 1.3 \\ \hline
\end{tabular}
\end{table}

\subsection{Graph Classification}

In this section, we present the result of our \textit{GCORN} on the graph classification task using both empirical and our proposed probabilistic evaluation. Details about the experimental setting for this task are provided in Appendix~\ref{sec:dataset_implementation_details}. Note that for the random attack, we used $\psi = 0.5$ and for the gradient-based, we used a budget $\delta = 0.2$. Table \ref{tab:results_graph_classification} reports the average accuracy and the corresponding standard deviation for both clean and attacked accuracy.

\begin{table*}[t]
	\centering
	\caption[Classification accuracy before (clean) and after the attack.]{Classification accuracy ($\pm$ standard deviation) of the models on different benchmark graph classification dataset before (clean) and after the attack. The higher the accuracy (in \%) the better.}
	\label{tab:results_graph_classification}
    \resizebox{0.8\textwidth}{!}{
    \begin{tabular}{p{0.10\textwidth}cccccc}
        \toprule
        \multirow{2}{*}{Attack} & \multicolumn{2}{c}{PROTEINS} & \multicolumn{2}{c}{D\&D} & \multicolumn{2}{c}{NCI1}\\
        
        \cmidrule(lr){2-3} \cmidrule(lr){4-5} \cmidrule(lr){6-7}
        & \(\mathbf{GCN}\)  & \(\mathbf{GCORN}\) & \(\mathbf{GCN}\) & \(\mathbf{GCORN}\) & \(\mathbf{GCN}\)  & \(\mathbf{GCORN}\)   \\
        \midrule
        Clean & 73.4 $\pm$ 2.8 &  74.1 $\pm$ 1.9 & 75.8 $\pm$ 3.6 & 76.4 $\pm$ 4.1  & 75.7 $\pm$ 2.2 & 74.8 $\pm$ 1.7 \\
        Random & 66.7 $\pm$ 2.5 & 71.7 $\pm$ 2.8 &  70.9 $\pm$ 3.2 & 75.1 $\pm$ 4.8 & 67.1 $\pm$ 2.6 & 71.8 $\pm$ 2.1 \\=
        PGD & 56.7 $\pm$ 2.8  & 65.4 $\pm$ 3.1 & 61.8 $\pm$ 4.1  & 68.6 $\pm$ 4.3  & 54.9 $\pm$ 2.9  & 62.4 $\pm$ 2.1 \\

        \toprule   
        \end{tabular}
        }

\end{table*}

\section{Experimental Results on GIN} \label{sec:experimental_setup_for_gin}
To showcase our method's versatility beyond the considered GCN framework and confirm the GIN-related results outlined in Theorem \ref{theo:gin_results}, we conducted similar experiments to those presented in Table \ref{tab:results_node_classification}. However, this time, we employed GIN as the message-passing scheme. In this perspective, we used the same previously considered node features attacks, notably: \textbf{(i)} The baseline random attack injecting Gaussian noise $\mathcal{N}(0, \mathbf{I})$ to the features with a scaling parameter $\psi$ controlling the attack budget;  
\textbf{(ii)} The white-box Proximal Gradient Descent \citep{pgd_paper}, which is a gradient-based approach to the adversarial optimization task with a chosen attack budget of 15\%.

\begin{table}[h]
\centering
\caption[Attacked classification accuracy using GIN after the attack application.]{Attacked classification accuracy ($\pm$ standard deviation) of the models on different benchmark node classification dataset using Graph Isomorphism Network after the attack application. }
\begin{tabular}{llccc}
\hline
                                      & Model & Cora                    & CiteSeer                & PubMed                  \\ \hline
\multirow{2}{*}{Random ($\psi =0.5$)} & GIN   & 67.8 $\pm$ 1.3          & 50.1 $\pm$ 0.3          & 72.8 $\pm$ 0.4          \\
                                      & GIORN & \textbf{71.8 $\pm$ 0.8} & \textbf{60.8 $\pm$ 0.5} & \textbf{74.3 $\pm$ 0.7} \\ \hline
\multirow{2}{*}{Random ($\psi =1.0$)} & GIN   & 54.5 $\pm$ 0.9          & 45.1 $\pm$ 0.4          & 67.3 $\pm$ 0.6          \\
                                      & GIORN & \textbf{66.7 $\pm$ 1.2} & \textbf{54.4 $\pm$ 0.8} & \textbf{68.9 $\pm$ 1.1} \\ \hline
\multirow{2}{*}{PGD}                  & GIN   & 61.4 $\pm$ 1.4          & 44.1 $\pm$ 0.5          & 70.6 $\pm$ 0.4          \\
                                      & GIORN & \textbf{69.5 $\pm$ 0.6} & \textbf{58.3 $\pm$ 0.6} & \textbf{73.1 $\pm$ 1.6} \\ \hline
\end{tabular}
\end{table}

\section{Datasets and Implementation Details} \label{sec:dataset_implementation_details}

For all the used models, the same number of layers, hyperparameters, and activation functions were used. The models were trained using the cross-entropy loss function with the Adam optimizer, the number of epochs and learning rate were kept similar for the different approaches across all experiments.

\subsection{Node Classification}
Characteristics and information about the datasets utilized in the node classification part of the study are presented in Table \ref{tab:data_statistics}. As outlined in the main paper, we conduct experiments on a set of citation networks, including Cora, CiteSeer, and PubMed \citep{dataset_node_classification}, as well as the Amazon Co-author network of authors from the Computer Science (CS) domain \citep{cs_data}. For Cora, CiteSeer, and PubMed, we adhere to the train/valid/test splits provided by the datasets. For the CS dataset, we adopt the same methodology as used in \citet{yang_2016}, by randomly selecting 20 nodes from each class to form the training set and 500/1000 nodes from the remaining for the validation and test sets. 

\begin{table}[t]
\caption{Statistics of the node classification datasets used in our experiments.}
\label{tab:data_statistics}
\begin{center}
\begin{small}
\begin{tabular}{lcccc}
\toprule
Dataset & \#Features & \#Nodes & \#Edges & \#Classes \\
\midrule
Cora    & 1433 & 2708 & 5208 & 7 \\
CiteSeer    & 3703 & 3327 & 4552 & 6 \\
PubMed    & 500 & 19717 & 44338 & 3 \\
CS    & 6805 & 18333 & 81894 & 15 \\
OGBN-Arxiv    & 128 & 31971 & 71669 & 40 \\
\bottomrule
\end{tabular}
\end{small}
\end{center}
\end{table}

In all of the experiments, the models employed a 2-layer convolutional architecture (consisting of two iterations of message passing and updating) stacked with a Multi-Layer Perception (MLP) as a readout. The intent was to compare the models in an iso-architectural setting, to ensure a fair evaluation of their robustness. All experiments were conducted using the Adam optimizer \citep{kingma_adam} and the same hyperparameters, including a learning rate of 1e-2, 300 epochs, and a hidden feature dimension of 16. For the OGB dataset, we used 512 as a hidden feature dimension. To account for the impact of random initialization, each experiment was repeated 10 times, and the mean and standard deviation of the results were reported. We finally note that for the AIRGNN, we set $K=2$ so as to have the same number of propagation as the other benchmarks. For our proposed GCORN, we tuned the iteration numbers for the different datasets and this was also done for the Parseval Regularization (ParselR) to find the best regularization parameter.

\subsection{Graph Classification}
We additionally emperically evaluated our proposed method GCORN using both probabilistic evaluation and the classical experimental evaluation. We used benchmark datasets derived from bioinformatics and chemoinformatics (PROTEINS, NCI1, D\&D) \citep{morris2020tudataset}. The framework proposed by \citet{errica2020fair} was used to evaluate the performance of the models on this task. We therefore performed a $10$-fold cross-validation using the same folds as provided by the paper to obtain an estimate of the generalization performance of each method. 

\begin{table}[t]
\caption{Statistics of the graph classification datasets used in our experiments.}
\label{tab:data_statistics}
\vskip 0.15in
\begin{center}
\begin{small}
\begin{tabular}{lcccc}
\toprule
Dataset & \#Graphs & \#Nodes & \#Edges & \#Classes \\
\midrule
DD    & 1178 & 284.32 & 715.66 & 2 \\
NCI1    & 4110 & 29.87 & 32.30 & 2 \\
PROTEINS    & 1113 & 39.06 & 72.82 & 2 \\

\bottomrule
\end{tabular}
\end{small}
\end{center}
\vskip -0.1in
\end{table}

Similar to node classification, we employed a 2-layer convolutional architecture (consisting of two iterations of message passing and updating) stacked with a Multi-Layer Perception (MLP) as a readout using the Adam Optimizer. 

\subsection{Implementation Details}

Our implementation is available in the supplementary materials (and will be publicly available afterwards). It is built using the open-source library \textit{PyTorch Geometric} (PyG) under the MIT license \citep{Fey/Lenssen/2019}. We leveraged the publicly available implementation of the different benchmarks from their available repositories : From GCN-k \footnote{https://github.com/ChangminWu/RobustGCN}, for AIRGNN \footnote{https://github.com/lxiaorui/AirGNN}, for GNNGuard \footnote{https://github.com/mims-harvard/GNNGuard} and for RGCN we used the implementation from the DeepRobust package. Note that we additionally utilized the PyTorch DeepRobust package\footnote{https://github.com/DSE-MSU/DeepRobust} to implement the adversarial attacks used in this study. The experiments have been run on both a NVIDIA A100 GPU and a RTX A6000 GPU.

\subsection{Implementation Details of our Empirical Robustness Estimation}
 We considered the input distance defined in Equation \ref{eq:prob_eval_distance} with $p=2$, we fixed the radius $\epsilon = 10$ and the number of sampling per graph at $L_{max} = 100$.
For each graph $G$, the output distance $d_{\mathcal{Y}}$ has been re-scaled to the interval $[0,1]$ by normalization using a factor of $2 \sqrt{N_G}$ where $N_G$ is the number of nodes in the graph $G$.


\chapter{Appendix: A Post-hoc Approach With Conditional Random Fields}
\section{Proof of Lemma \ref{crf:lem:mean_field_expected}}
\label{crf:appendix:proof_lema}
\textbf{Lemma \ref{crf:lem:mean_field_expected}}
By solving the system of equations in \eqref{crf:eq:new_objective}, we can get the optimal distribution $Q^*$ as follows:
 \begin{equation*}
    \forall a \in V^{\text{\textit{CRF}}}, ~~ Q(\widetilde{Y}_a) \propto \exp \left \{  \mathbb{E}_{-a}  \left [ \log~P\left ( \widetilde{Y} |Y, V^{\text{\textit{CRF}}}, E^{\text{\textit{CRF}}}  \right ) \right ]  \right \}.
 \end{equation*}

\begin{proof}
The $\mathcal{KL}$-divergence includes the true posterior $P (\widetilde{Y} |Y, V^{\text{\textit{CRF}}}, E^{\text{\textit{CRF}}})$  which is exactly the unknown value. We can rewrite the $\mathcal{KL}$-divergence as:
\begin{align*}
    \mathcal{KL}\left ( Q |P  \right ) & = \int  Q(\widetilde{Y})  \log \frac{Q(\widetilde{Y}) }{ P (\widetilde{Y} |Y, V^{\text{\textit{CRF}}}, E^{\text{\textit{CRF}}})} d\widetilde{Y} \\
    & = \int  Q(\widetilde{Y}) \log\frac{Q(\widetilde{Y})P(Y, V^{\text{\textit{CRF}}}, E^{\text{\textit{CRF}}}) }{ P (\widetilde{Y} ,Y, V^{\text{\textit{CRF}}}, E^{\text{\textit{CRF}}})}d\widetilde{Y}  \\
    & =  \int Q(\widetilde{Y})  \left ( \log~P(Y, V^{\text{\textit{CRF}}}, E^{\text{\textit{CRF}}}) +  \log \frac{Q(\widetilde{Y}) }{ P (\widetilde{Y} ,Y, V^{\text{\textit{CRF}}}, E^{\text{\textit{CRF}}})} \right )d\widetilde{Y}  \\
    & = \log~P(Y, V^{\text{\textit{CRF}}}, E^{\text{\textit{CRF}}})  \int Q(\widetilde{Y})d\widetilde{Y} - \int  Q(\widetilde{Y}) \log \frac{ P (\widetilde{Y} ,Y, V^{\text{\textit{CRF}}}, E^{\text{\textit{CRF}}})}{Q(\widetilde{Y}) }   d\widetilde{Y}. 
\end{align*}
Since $\int Q(\widetilde{Y})d\widetilde{Y}  = 1 $, we conclude that: 
\begin{equation*}
    \mathcal{KL}\left ( Q |P  \right ) = \log~P(Y, V^{\text{\textit{CRF}}}, E^{\text{\textit{CRF}}})   - \int  Q(\widetilde{Y}) \log \frac{ P (\widetilde{Y} ,Y, V^{\text{\textit{CRF}}}, E^{\text{\textit{CRF}}})}{Q(\widetilde{Y}) }   d\widetilde{Y}.
\end{equation*}
We are minimizing the $\mathcal{KL}-$divergence over $Q$, therefore the term $log~P(Y, V^{CRF}, E^{CRF}) $ can be ignored. The second term is the \textit{This is the negative ELBO}. We know that the $\mathcal{KL}-$divergence is not negative. Thus, $\log~P(Y, V^{\text{\textit{CRF}}}, E^{\text{\textit{CRF}}}) \geq \text{ELBO}(Q)$  justifying the name \textit{Evidence lower bound (ELBO)}.

Therefore, the main objective is to optimize the ELBO in the mean field variational inference, i.e., choose the variational factors that maximize ELBO:

\begin{equation}\label{crf:eq:elbo_decom}
\text{ELBO}(Q)=\int  Q(\widetilde{Y}) \log \frac{ P (\widetilde{Y} ,Y, V^{\text{\textit{CRF}}}, E^{\text{\textit{CRF}}})}{Q(\widetilde{Y}) }   d\widetilde{Y} = \mathbb{E}_{Q}\left [ \log~  P (\widetilde{Y} ,Y, V^{\text{\textit{CRF}}}, E^{\text{\textit{CRF}}})\right ]-\mathbb{E}_{Q}\left [ \log~  Q(\widetilde{Y})\right ].
\end{equation} 

We will employ \textit{coordinate ascent inference}, where we iteratively optimize each variational distribution while keeping the others constant.

We assume that the set of CRF nodes is finite, i.e.,  $\left | V^{\text{\textit{CRF}}} \right | < \infty$, which is a realistic assumption if we consider the set of all GNN inputs used during inference. 
If $\left | V^{\text{\textit{CRF}}} \right | = m$, we can order the elements $V^{\text{\textit{CRF}}}$ in a a specific order $i=1, \ldots, m$. Thus, using the chain rule, we decompose the probability $P (\widetilde{Y} ,Y, V^{\text{\textit{CRF}}}, E^{\text{\textit{CRF}}})$ as follows
\begin{align*}
    P (\widetilde{Y} ,Y, V^{\text{\textit{CRF}}}, E^{\text{\textit{CRF}}}) & =   P (\widetilde{Y}_{1:m} ,Y_{1:m}, V^{\text{\textit{CRF}}}, E^{\text{\textit{CRF}}}) \\
    & =  P(Y_{1:m}, V^{\text{\textit{CRF}}}, E^{\text{\textit{CRF}}}) \prod_{i=1}^{m}  P (\widetilde{Y}_{i} |Y_{1:(i-1)}, V^{\text{\textit{CRF}}}, E^{\text{\textit{CRF}}}).
\end{align*}
Using the independence of the mean field approximation, we also have
\begin{equation*}
  \mathbb{E}_{Q}\left [ \log~  Q(\widetilde{Y})\right ] = \sum_{i=1}^{m} \mathbb{E}_{Q_j}\left [ \log~  Q(\widetilde{Y}_j)\right ].
\end{equation*}
Now, we have the expression of the two terms appearing in ELBO in \eqref{crf:eq:elbo_decom}:
\begin{equation*}
    \text{ELBO}(Q)  = \log~ P(Y_{1:m}, V^{\text{\textit{CRF}}}, E^{\text{\textit{CRF}}}) + \sum_{i=1}^{m} \mathbb{E}_{Q} \left [ \log~P (\widetilde{Y}_{i} |Y_{1:(i-1)}, V^{\text{\textit{CRF}}}, E^{\text{\textit{CRF}}}) \right ] - \mathbb{E}_{Q_j} \left [ \log~Q (\widetilde{Y}_{i}) \right ].
\end{equation*}
The above decomposition is valid for any ordering of the GNN inputs. Thus, for a fixed GNN input $a \in V^{\text{\textit{CRF}}}$, if we consider $a$ as the last variable $m$ of the list, we can consider the ELBO as a function of $Q (\widetilde{Y}_{a}) = Q (\widetilde{Y}_{m})$
\begin{align*}
    \text{ELBO}(Q(\widetilde{Y}_a)) & = \text{ELBO}(Q(\widetilde{Y}_k)) \\
    & = \mathbb{E}_{Q} \left [ \log~P (\widetilde{Y}_{m} |Y_{1:(m-1)}, V^{\text{\textit{CRF}}}, E^{\text{\textit{CRF}}}) \right ] - \mathbb{E}_{Q_m} \left [ \log~Q (\widetilde{Y}_{m} ) \right ]~~ + ~~\text{\textit{const}} \\
    & =  \int Q (\widetilde{Y}_{m} )  \mathbb{E}_{Q_{\neq  m}} \left [ \log~P (\widetilde{Y}_{m} |Y_{\neq  m}, V^{\text{\textit{CRF}}}, E^{\text{\textit{CRF}}}) \right ]     d\widetilde{Y}_{m}  - \int Q (\widetilde{Y}_{m} )  log~Q (\widetilde{Y}_{m} ) d\widetilde{Y}_{m}  \\
    & =  \int Q (\widetilde{Y}_{a} )  \mathbb{E}_{Q_{\neq  a}} \left [ \log~P (\widetilde{Y}_{a} |Y_{\neq  a}, V^{\text{\textit{CRF}}}, E^{\text{\textit{CRF}}}) \right ]     d\widetilde{Y}_{a}  - \int Q (\widetilde{Y}_{a} )  \log~Q (\widetilde{Y}_{a} ) d\widetilde{Y}_{a}, 
\end{align*}
where $\neq m$ means all indices except the $m^{th}$.
Now, we take the derivative of the ELBO with respect to $Q(\widetilde{Y}_a)$:

\begin{equation*}
    \frac{d~\text{ELBO}}{dQ(\widetilde{Y}_a)} =  \mathbb{E}_{Q_{\neq  a}} \left [ \log~P (\widetilde{Y}_{a} |Y_{\neq  a}, V^{\text{\textit{CRF}}}, E^{\text{\textit{CRF}}}) \right ] - \log~Q (\widetilde{Y}_{a} ) -1 = 0.
\end{equation*}
Therefore, 
\begin{equation*}
    \forall a \in V^{\text{\textit{CRF}}}, ~~ Q(\widetilde{Y}_a) \propto \exp \left \{  \mathbb{E}_{-a}  \left [ \log~P\left ( \widetilde{Y} |Y, V^{\text{\textit{CRF}}}, E^{\text{\textit{CRF}}}  \right ) \right ]  \right \}.
\end{equation*}
\end{proof}

\section{Asymptotic Behavior of CRF Neighborhood Size}
\label{crf:appendix:crf_neighbors}
\subsection{Proof of Lemma \ref{crf:lem:size_CRF}}
\textbf{Lemma \ref{crf:lem:size_CRF}}
For any integer $r$ in  $\{0,\ldots, \frac{n(n+1)}{2}\}$, the number of CRF neighboors  $\left | \mathcal{N}^{\text{\textit{CRF}}}(a)   \right |$ for any $a\in V^{\text{\textit{CRF}}}$, i.e., the set of graphs $[\widetilde{\mathbf{A}}, \widetilde{\mathbf{X}}]$ with a Hamming distance smaller or equal than $r$, for each $a \in V^{\text{\textit{CRF}}}$, we have the following lower bound: 
\begin{equation}
 \frac{2^{H(\epsilon)n(n+1)/2}}{\sqrt{4n(n+1)\epsilon (1-\epsilon)}} \leq \left | \mathcal{N}^{\text{\textit{CRF}}}(a)   \right |,
\end{equation}
where $0\leq \epsilon=\frac{2r}{n(n+1)}\leq 1$ and $H(\cdot)$ is the binary entropy function, i.e., 
$H(\epsilon) = -\epsilon \log_2(\epsilon) - (1-\epsilon) \log_2(1-\epsilon).$

\begin{proof}
    We use Stirling's formula;
    \begin{equation}
        \forall s, ~~ s! = \sqrt{2\pi s}s^s e^{-s} \exp\left ( \frac{1}{12s} - \frac{1}{360s^3} + \ldots \right ).
    \end{equation}

    Thus, for $r$ in  $\{0,\ldots, \frac{n(n+1)}{2}\}$ and $L=\frac{n(n+1)}{2}$, we write

        \begin{align}
       \binom{\frac{n(n+1)}{2}}{r} & =  \binom{L}{r}  \\
     & =   \frac{ L \!}{ r !   ( L - r  ) !   } \\
     & \geq \frac{  \sqrt{2\pi L}L^L e^{-L} \exp\left [-1/12r -1/12(L-r)  \right ]     }{\sqrt{2\pi r}r^r e^{-r }  \sqrt{2\pi (L-r)} (L-r)^{(L-r)} e^{-(L-r)}}.  
    \end{align}
For $L\geq 4$, for any $r\in \{1, \ldots , L\}$, we always have $L-r \geq 3$ or $r\geq$, thus,
    \begin{equation}
        \frac{1}{12r} + \frac{1}{12(L-r)} \leq \frac{1}{12} + \frac{1}{36} =  \frac{1}{9}. 
    \end{equation}

Therefore,
        \begin{equation}
        \exp\left ( - \frac{1}{12r} - \frac{1}{12(L-r)} \right ) \geq \frac{1}{12} + \frac{1}{36} = e^{-1/9} \geq \frac{1}{2} \sqrt{\pi}.
    \end{equation}
We can then therefore derive a lower bound for $\binom{\frac{n(n+1)}{2}}{r}$ :
    \begin{align}
     \binom{\frac{n(n+1)}{2}}{r} & =  \binom{L}{r}  \\
     & \geq \frac{  \sqrt{2\pi L}L^L e^{-L} \frac{1}{2} \sqrt{\pi}   }{\sqrt{2\pi r}r^r e^{-r }  \sqrt{2\pi (L-r)}(L-r)^(L-r) e^{-(L-r)}    }  \\
    & = \frac{  \sqrt{ L}L^L  }{\sqrt{8 r}r^r   \sqrt{ (L-r)}(L-r)^{(L-r)}    } \\
    & = \sqrt{\frac{L}{8 r(L-r)}}\frac{  L^L  }{r^r (L-r)^{(L-r)}    } \\
    & = \frac{1}{\sqrt{8  L \epsilon  (1-\epsilon)}}\frac{  1}{  \epsilon^r (1-\epsilon)^{L-r}  } \\
    & = \frac{1}{\sqrt{8  L \epsilon  (1-\epsilon)}} \epsilon^{-L\epsilon}(1-\epsilon)^{L(1-\epsilon)} \\
    & = \frac{1}{\sqrt{8  L \epsilon  (1-\epsilon)}} 2^{L H(\epsilon)}.
    \end{align}

    For $d$ in  $\{0,\ldots, \frac{n(n+1)}{2}\}$, the number of possible graphs with a Hamming distance equal to $d$ from a graph $a = [G,x]$ is $\binom{L}{d}$.
    Thus,
    \begin{align}
        \left | \mathcal{N}^{\text{\textit{CRF}}}(a)   \right | &  = \sum_{d=0}^{r} \binom{L}{d} \\
        & \geq \binom{L}{r} \\
        & \geq \frac{1}{\sqrt{8  L \epsilon  (1-\epsilon)}} 2^{L H(\epsilon)} \\
        & = \frac{1}{\sqrt{4N(N+1)\epsilon (1-\epsilon)}}2^{H(\epsilon)N(N+1)/2}. 
    \end{align}
\end{proof}

\subsection{Empirical Investigation of the Lower Bound}

We empirically investigate the evolution of the lower-bound stated in Lemma \ref{crf:lem:size_CRF} as a function of ratio $\epsilon = \epsilon(r)$. As noticed, the number of neighbors increases exponentially as the radius increases. This motivates the need for sampling strategies to reduce the size of the CRF.

\begin{figure}[h]
    \centering
    \includegraphics[width=.8\textwidth]{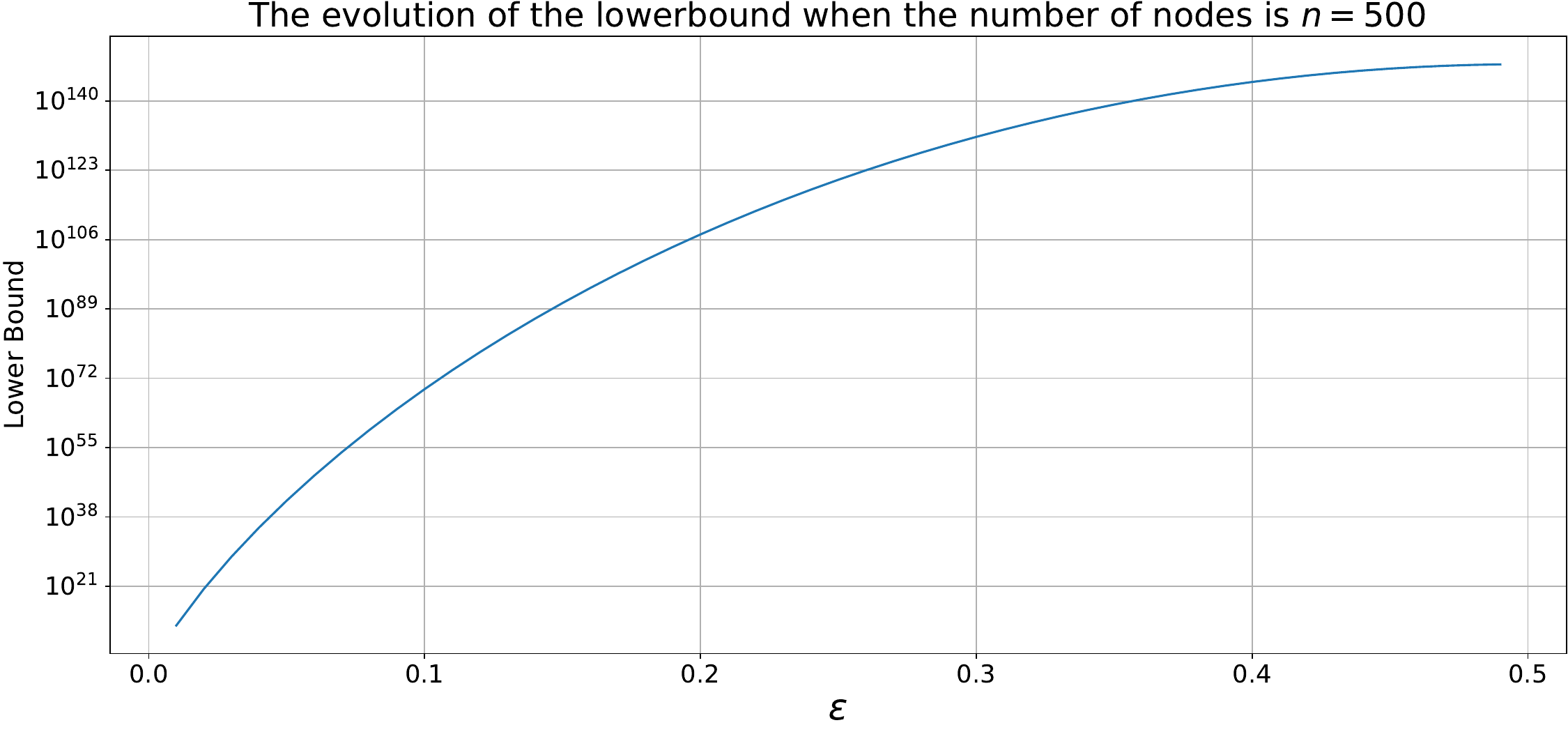}
    \caption[The radius on the lower bound stated in Lemma \ref{crf:lem:size_CRF}]{The effect of the radius on the lower bound stated in Lemma \ref{crf:lem:size_CRF}.}
    \label{crf:fig:lower_bound}
\end{figure}

\section{The Set of Hyperparameters Used to Construct the CRF}
\label{crf:appendix:crf_hyperparam}
In Tables \ref{crf:tab:param_CRF_features} and \ref{crf:tab:param_CRF_structure}, we present the hyperparameters used to run the inference of RobustCRF for each dataset. The number of iterations was fixed to 2. $p_r$ and we compute the radius as a floor function, \ie ~$r = \left \lfloor p_r \times m \right \rfloor$, where $m = |E|$ is the number of existing edges in the original graph.
\begin{table}[h]
\caption{The optimal RobustCRF's hyperparameters for the feature-based attacks.}
\label{crf:tab:param_CRF_features}
\begin{center}
\begin{small}
\begin{tabular}{lcccc}
\toprule
Hyperparameter & Cora & CiteSeer  & PubMed & CS \\
\midrule
$r$   &  0.1 & 0.9 & 0.3 &0.3 \\
$\sigma$ & 0.9 & 0.8 & 0.9 &0.5  \\
\bottomrule
\end{tabular}
\end{small}
\end{center}
\vskip -0.1in
\end{table}
\begin{table}[h]
\caption{The optimal RobustCRF's hyperparameters for the structure-based attacks.}
\label{crf:tab:param_CRF_structure}
\begin{center}
\begin{small}
\begin{tabular}{lcccc}
\toprule
Hyperparameters & Cora & CoraML  & CiteSeer & PolBlogs \\
\midrule
$p_r$   &   0.02 & 0.02 & 0.04 &0.005 \\
$\sigma$ & 0.05 & 0.2 & 0.05 &0.05  \\
\bottomrule
\end{tabular}
\end{small}
\end{center}
\vskip -0.1in
\end{table}

\section{Experimental Setup}\label{crf:app:empric}
\textbf{Datasets.} For our experiments, we focus on node classification within the general perspective of node representation learning. We use the citation networks Cora, CoraML, CiteSeer, and PubMed \citep{dataset_node_classification}. We additionally consider the co-authorship network CS \citep{cs_data}, the blog and citation graph PolBlogs \citep{adamic2005political}, and the non-homophilous dataset Texas \citep{lim2021new}. More details and statistics of the datasets can be found in Table \ref{crf:tab:data_statistics}. For the CS dataset, we randomly selected 20 nodes from each class to form the training set and 500/1000 nodes for the validation and test sets \citep{yang_2016}. For all the remaining datasets, we adhere to the public train/valid/test splits provided by the datasets. 

\textbf{Implementation Details.} We used the PyTorch Geometric (PyG) open-source library, licensed under MIT \citep{Fey/Lenssen/2019}. Additionally, for adversarial attacks in this study, we used the DeepRobust package \footnote{https://github.com/DSE-MSU/DeepRobust}. The experiments were conducted on an RTX A6000 GPU.  For the structure-based CRF, we leveraged the sampling strategy detailed in Section \ref{crf:proposed_method}. The set of hyperparameters for each dataset can be found in Appendix \ref{crf:appendix:crf_hyperparam}. We compute the similarity $g_{ab}$  between two inputs $a=[G,X]$ and $b=[\tilde{G}, \tilde{X}]$ using the cosine similarity for the features based attacks, namely $\text{CosSim}(X,\tilde{X})$, while for the structural attacks, we use the prior distribution $g_{ab}=\binom{r}{d}\frac{1}{2^r}$, where $d$ is the value of the Hamming distance between the original graph $a$ and its sampled neighbor $b$. We note that our code is provided in the supplementary materials and will be made public upon publication.

\begin{table}
\caption{Statistics of the node classification datasets used in our experiments.}
\centering
\resizebox{0.5\textwidth}{!}{%
\begin{tabular}{lcccc}
\toprule
Dataset & \#Features & \#Nodes & \#Edges & \#Classes \\
\midrule
Cora    & 1,433 & 2,708 & 5,208 & 7 \\
CoraML    &  300 &  2,995& 8,226 & 7\\
CiteSeer & 3,703 & 3,327 & 4,552 & 6 \\
PubMed   & 500 & 19,717 & 44,338 & 3 \\
CS    & 6,805 & 18,333 & 81,894 & 15 \\
PolBlogs    & - & 1,490 & 19,025 & 2 \\
Texas    & 1,703 & 183 & 309 &  5\\
OGBN-Arxiv & 128 & 31,971 & 71,669 &  40\\
\bottomrule
\end{tabular}
}
\label{crf:tab:data_statistics}
\end{table}

\section{Time and Complexity of RobustCRF}
\label{crf:appendix:time_robustcrf}
In Table \ref{crf:tab:order_iter_analysis}, we present the average time (in seconds) required for RobustCRF inference. As observed, the inference time increases exponentially with larger values of $K$ and $L$. Nevertheless, empirical evidence suggests that a small number of samples is sufficient to improve GNN robustness. In our experiments, we specifically used $L=5$ samples and set the number of iterations to 2.
\begin{table}[h]
\caption{Inference time for different values of the number of iterations/samples.}
\label{crf:tab:order_iter_analysis}
\begin{center}
\begin{tabular}{lccc}
\toprule
Num Samples $L$ & 0 Iter & 1 Iter & 2 Iter \\
\midrule
5  & $0.26 \pm 0.52 $ & $1.99 \pm 0.54 $&  $16.40\pm 2.04 $\\
10 & $0.20 \pm 0.41 $ & $3.24 \pm 0.47 $&  $58.28\pm 1.15 $\\
20 & $0.22 \pm 0.44 $ & $5.86 \pm 0.53 $&  $224.86\pm 0.71 $\\

\bottomrule
\end{tabular}
\end{center}
\end{table}

\section{Resuls on OGBN-Arxiv} \label{crf:app:ogbn}
To further evaluate the effectiveness of RobustCRF, we conducted experiments on the OGBN-Arxiv dataset, a large-scale Open Graph Benchmark (OGB) dataset commonly used for evaluating node classification in graph-based learning models. The dataset consists of a citation network where nodes represent ArXiv papers and edges denote citation links \cite{hu2020ogb}. In Table \ref{crf:tab:ogbn}, we present the attacked classification accuracy of the GCN, a baseline and the proposed RobustCRF on the OGBN-Arxiv dataset.

\begin{table}[t]
\caption{Attacked classification accuracy  on the OGBN-Arxiv dataset.}
\label{crf:tab:ogbn}
\begin{center}
\begin{tabular}{lccc}
\toprule
Dataset & GCN & NoisyGCN & RobustCRF  \\
\midrule
Clean    
    & $\mathbf{60.41 {\scriptstyle \pm 0.15}}$  
    & $59.97 {\scriptstyle \pm 0.11}$ 
    & $60.28 {\scriptstyle \pm 0.15}$  \\

Random    
    & $58.97 {\scriptstyle \pm 0.24}$  
    & $58.71 {\scriptstyle \pm 0.10}$  
    & $\mathbf{59.03 {\scriptstyle \pm 0.26}}$ \\

PGD    
    & $50.24 {\scriptstyle \pm 0.42}$ 
    & $\mathbf{50.26 {\scriptstyle \pm 0.37}}$ 
    & $50.10 {\scriptstyle \pm 0.56}$ \\

\bottomrule
\end{tabular}
\end{center}
\end{table}

\chapter{Appendix: Improving Generalization in GNNs through Data
Augmentation}
\section{Proof of Theorem \ref{gen:thm:rademacher}}\label{gen:appendix:proof_lemma} 
In this section, we provide a detailed proof of Theorem \ref{gen:thm:rademacher}, aiming to derive a theoretical upper bound for both the generalization gap and the Rademacher complexity.\\
\textbf{Theorem \ref{gen:thm:rademacher}}
Let $\ell$ be a classification loss function with $\text{L}_{\text{Lip}}$ as a Lipschitz constant and $\ell(\cdot,\cdot)\in [0,1].$ Then, with a probability at least $1-\delta$ over the samples $\mathcal{D}_{\text{train}}$, we have, 
\begin{equation*} \mathbb{E}_{\mathcal{G}  \sim \mathcal{D}}\left [ \ell(\mathcal{G},\hat{\theta}_{\text{aug}}) \right ] - \mathbb{E}_{\mathcal{G}  \sim \mathcal{D}}\left [ \ell(\mathcal{G},\theta_{\star}) \right ] \leq 2 \mathcal{R}(\ell_{\text{aug}} ) +  5\sqrt{\frac{2\log(4/\delta)}{N}} + 2 \text{L}_{\text{Lip}}  \mathbb{E}_{\mathcal{G} \sim \mathcal{D},\widetilde{\mathcal{G}}\sim A_\lambda} \left [ \left \| \widetilde{\mathcal{G}}- \mathcal{G} \right \| \right ]. \end{equation*}
Moreover, we have,
$$\mathcal{R}(\ell_{\text{aug}} )  \leq \mathcal{R}(\ell ) +  \max_{n}  \text{L}_{\text{Lip}}    \mathbb{E}_{\widetilde{\mathcal{G}}_{n}^{m} \sim A_\lambda} \left [ \left \| \widetilde{\mathcal{G}}_{n}^{m}  -\mathcal{G}_n \right \| \right ].  
$$
\begin{proof}

We will decompose $  \mathbb{E}_{\mathcal{G}  \sim \mathcal{D}}\left [ \ell(\mathcal{G},\hat{\theta}_{\text{aug}}) \right ] - \mathbb{E}_{\mathcal{G}  \sim \mathcal{D}}\left [ \ell(\mathcal{G},\theta_{\star}) \right ] $ into a finite sum of 5 terms as follows,
$$  \mathbb{E}_{\mathcal{G}  \sim \mathcal{D}}\left [ \ell(\mathcal{G},\hat{\theta}_{\text{aug}}) \right ] - \mathbb{E}_{\mathcal{G}  \sim \mathcal{D}}\left [ \ell(\mathcal{G},\theta_{\star}) \right ] = u_1 + u_2 + u_3 + u_4 + u_5 $$
where,
\begin{align*}
 u_1 & =  \mathbb{E}_{\mathcal{G}  \sim \mathcal{D}}\left [ \ell(\mathcal{G},\hat{\theta}_{\text{aug}}) \right ] - \mathbb{E}_{\mathcal{G}  \sim \mathcal{D}}\left [ \mathbb{E}_{\widetilde{\mathcal{G}}  \sim A_\lambda }\left [ \ell(\widetilde{\mathcal{G}},\hat{\theta}_{\text{aug}}) \right ] \right ], \\
 u_2 & = \mathbb{E}_{\mathcal{G}  \sim \mathcal{D}}\left [ \mathbb{E}_{\widetilde{\mathcal{G}}  \sim A_\lambda }\left [ \ell(\widetilde{\mathcal{G}},\hat{\theta}_{\text{aug}}) \right ] \right ] -  \frac{1}{N} \sum_{n=1}^{N} \mathbb{E}_{\widetilde{\mathcal{G}}_n^m  \sim A_\lambda }\left [ \ell(\widetilde{\mathcal{G}}_n^m,\hat{\theta}_{\text{aug}}) \right ],  \\
u_3 & = \frac{1}{N} \sum_{n=1}^{N} \mathbb{E}_{\widetilde{\mathcal{G}}_n^m  \sim A_\lambda }\left [ \ell(\widetilde{\mathcal{G}}_n^m,\hat{\theta}_{\text{aug}}) \right ] -  \frac{1}{N} \sum_{n=1}^{N} \mathbb{E}_{\widetilde{\mathcal{G}}_n^m  \sim A_\lambda }\left [ \ell(\widetilde{\mathcal{G}}_n^m,\hat{\theta}_{\star}) \right ]       , \\
 u_4 & =  \frac{1}{N} \sum_{n=1}^{N} \mathbb{E}_{\widetilde{\mathcal{G}}_n^m  \sim A_\lambda }\left [ \ell(\widetilde{\mathcal{G}}_n^m,\hat{\theta}_{\star}) \right ]      - \mathbb{E}_{\mathcal{G}  \sim \mathcal{D}}\left [ \mathbb{E}_{\widetilde{\mathcal{G}}  \sim A_\lambda }\left [ \ell(\widetilde{\mathcal{G}},\theta_{\star}) \right ] \right ],\\
 u_5 & = \mathbb{E}_{\mathcal{G}  \sim \mathcal{D}}\left [ \mathbb{E}_{\widetilde{\mathcal{G}}  \sim A_\lambda }\left [ \ell(\widetilde{\mathcal{G}},\theta_{\star}) \right ] \right ] - \mathbb{E}_{\mathcal{G}  \sim \mathcal{D}}\left [ \ell(\mathcal{G},\theta_{\star}) \right ].      
\end{align*}

We upperbound each of the terms in the sum. We get,
\begin{align*}
  u_1 + u_5 & = \mathbb{E}_{\mathcal{G}  \sim \mathcal{D}}\left [ \ell(\mathcal{G},\hat{\theta}_{\text{aug}}) \right ] - \mathbb{E}_{\mathcal{G}  \sim \mathcal{D}}\left [ \mathbb{E}_{\widetilde{\mathcal{G}}  \sim A_\lambda }\left [ \ell(\widetilde{\mathcal{G}},\hat{\theta}_{\text{aug}}) \right ] \right ]   + \mathbb{E}_{\mathcal{G}  \sim \mathcal{D}}\left [ \mathbb{E}_{\widetilde{\mathcal{G}}  \sim A_\lambda }\left [ \ell(\widetilde{\mathcal{G}},\theta_{\star}) \right ] \right ] - \mathbb{E}_{\mathcal{G}  \sim \mathcal{D}}\left [ \ell(\mathcal{G},\theta_{\star}) \right ] \\
  & \leq \left | \mathbb{E}_{\mathcal{G}  \sim \mathcal{D}}\left [ \ell(\mathcal{G},\hat{\theta}_{\text{aug}}) \right ] - \mathbb{E}_{\mathcal{G}  \sim \mathcal{D}}\left [ \mathbb{E}_{\widetilde{\mathcal{G}}  \sim A_\lambda }\left [ \ell(\widetilde{\mathcal{G}},\hat{\theta}_{\text{aug}}) \right ] \right ] \right |   + \left | \mathbb{E}_{\mathcal{G}  \sim \mathcal{D}}\left [ \mathbb{E}_{\widetilde{\mathcal{G}}  \sim A_\lambda }\left [ \ell(\widetilde{\mathcal{G}},\theta_{\star}) \right ] \right ] - \mathbb{E}_{\mathcal{G}  \sim \mathcal{D}}\left [ \ell(\mathcal{G},\theta_{\star}) \right ] \right |\\
  & \leq 2 \sup_{\theta \in \Theta} \left | \mathbb{E}_{\mathcal{G}  \sim \mathcal{D}}\left [ \mathbb{E}_{\widetilde{\mathcal{G}}  \sim A_\lambda }\left [ \ell(\widetilde{\mathcal{G}},\theta) \right ] \right ] - \mathbb{E}_{\mathcal{G}  \sim \mathcal{D}}\left [ \ell(\mathcal{G},\theta) \right ] \right | \\
  & \leq 2 \sup_{\theta \in \Theta} \left |  \mathbb{E}_{\mathcal{G}  \sim \mathcal{D}}\left [ \mathbb{E}_{\widetilde{\mathcal{G}}  \sim A_\lambda }\left [ \ell(\widetilde{\mathcal{G}},\theta) \right ] - \ell(\mathcal{G},\theta) \right ]  \right | \\
  & \leq 2 \sup_{\theta \in \Theta} \left |  \mathbb{E}_{\mathcal{G}  \sim \mathcal{D}}\left [ \mathbb{E}_{\widetilde{\mathcal{G}}  \sim A_\lambda }\left [ \ell(\widetilde{\mathcal{G}},\theta)  - \ell(\mathcal{G},\theta) \right ] \right ]  \right | \\
  & \leq  2 \text{L}_{\text{Lip}} \sup_{\theta \in \Theta}  \mathbb{E}_{\mathcal{G}  \sim \mathcal{D}} \mathbb{E}_{\widetilde{\mathcal{G}}  \sim A_\lambda }\left [ \left \| \widetilde{\mathcal{G}} - \mathcal{G} \right \| \right ] .
\end{align*}

For the term $u_4$, we apply McDiarmid's inequality. Since the classification loss satisfy $\ell(\cdot)\in [0,1]$, we get for $k \in \{0,\ldots,N\}$,

For all $ \{ (\mathcal{G}_n, y_n)  \}_{n=1}^{N}, \{ (\mathcal{G^\prime}_n, y^\prime_n)  \}_{n=1}^{N},\theta,$ such that $\forall n \neq k, \quad \mathcal{G}_n = \mathcal{G^\prime}_n$ and $ \mathcal{G}_k \neq \mathcal{G^\prime}_k$:
    
\begin{align*}
\left |\frac{1}{N} \sum_{n=1}^{N} \mathbb{E}_{\widetilde{\mathcal{G}}  \sim A_\lambda }\left [ \ell(\mathcal{G}_n,\theta) \right ] - \frac{1}{N}\sum_{n=1}^{N} \mathbb{E}_{\widetilde{\mathcal{G}}  \sim A_\lambda }\left [ \ell(\mathcal{G}_n^\prime,\theta) \right ] \right |
& = \frac{1}{N} \left | \sum_{n=1}^{N} \mathbb{E}_{\widetilde{\mathcal{G}}  \sim A_\lambda }\left [ \ell(\mathcal{G}_n,\theta) - \ell(\mathcal{G}_n^\prime,\theta) \right ] \right | \\
& \leq \frac{1}{N} \left |  \mathbb{E}_{\widetilde{\mathcal{G}}  \sim A_\lambda }\left [ \ell(\mathcal{G}_k,\theta) - \ell(\mathcal{G}_k^\prime,\theta) \right ] \right | \\
&\leq 2/N.
\end{align*}

The first equality is obtained by your claim that $\forall n \neq k, \quad \mathcal{G}_n = \mathcal{G^\prime}_n$ and $ \mathcal{G}_k \neq \mathcal{G^\prime}_k$, the last inequality is obtained by the fact that $\ell(\cdot) \in [0,1]$.

Thus,
\begin{align*}
    \forall t>0, \quad \mathbb{P} \left (  u_4\geq t \right ) & = \mathbb{P} \left (   \frac{1}{N} \sum_{n=1}^{N} \mathbb{E}_{\widetilde{\mathcal{G}}_n^m  \sim A_\lambda }\left [ \ell(\widetilde{\mathcal{G}}_n^m ,\theta_{\star}) \right ]  - \mathbb{E}_{\mathcal{G}  \sim \mathcal{D}}\left [ \mathbb{E}_{\widetilde{\mathcal{G}}  \sim A_\lambda }\left [ \ell(\widetilde{\mathcal{G}},\theta_{\star}) \right ] \right ] \geq t \right )\\
    & \leq \exp \left ( - \frac{2 t^2}{\sum_{n=1}^{N} 4/N^2  } \right ) \\
    & = \exp \left ( - \frac{N t^2}{2} \right ).
\end{align*}

Therefore, for $\delta \in ]0,1]$, and for $t=\sqrt{2 \log(1/\delta)/N }$, i.e. $exp \left ( - \frac{N  t^2}{2}\right )  = \delta.$, we have,

$$ \mathbb{P} \left (  u_4\geq \sqrt{2 \log(1/\delta)/N } \right ) \leq \delta.$$

Therefore,

$$\mathbb{P} \left (  u_4< \sqrt{\frac{2 \log(1/\delta)}{N}}\right ) = 1- \mathbb{P} \left (  u_4\geq \sqrt{\frac{2 \log(1/\delta)}{N}} \right )\geq 1- \delta.$$
Thus, with a probability of at least $1-\delta$,
$$u_4 \leq \sqrt{\frac{2 \log(1/\delta)}{N}} < \sqrt{\frac{2\log(4/\delta)}{N}}.$$

Moreover, Rademacher complexity holds for $u_2$, 

$$u_2 = \mathbb{E}_{\mathcal{G}  \sim \mathcal{D}}\left [ \mathbb{E}_{\widetilde{\mathcal{G}}  \sim A_\lambda }\left [ \ell(\widetilde{\mathcal{G}},\hat{\theta}_{\text{aug}}) \right ] \right ] -  \frac{1}{N} \sum_{n=1}^{N} \mathbb{E}_{\widetilde{\mathcal{G}}_n^m  \sim A_\lambda }\left [ \ell(\widetilde{\mathcal{G}}_n^m,\hat{\theta}_{\text{aug}}) \right ] \leq  2 \mathcal{R}(\ell_{\text{aug}} ) + 4 \sqrt{\frac{2\log(4/\delta)}{N}}.$$

The above inequality tells us that the true risk $\mathbb{E}_{\mathcal{G}  \sim \mathcal{D}}\left [ \mathbb{E}_{\widetilde{\mathcal{G}}  \sim A_\lambda }\left [ \ell(\widetilde{\mathcal{G}},\hat{\theta}_{\text{aug}}) \right ] \right ]$ is bounded by the empirical risk $\frac{1}{N} \sum_{n=1}^{N} \mathbb{E}_{\widetilde{\mathcal{G}}_n^m  \sim A_\lambda }\left [ \ell(\widetilde{\mathcal{G}}_n^m,\hat{\theta}_{\text{aug}}) \right ] $ plus a term depending on the Rademacher complexity of the augmented hypothesis class and an additional term that decreases with the size of the sample $N$.

Additionally, since $\hat{\theta}_{\text{aug}}$ is the optimal parameter for the loss $\frac{1}{N} \sum_{n=1}^{N} \mathbb{E}_{\widetilde{\mathcal{G}}_n^m  \sim A_\lambda }\left [ \ell(\widetilde{\mathcal{G}}_n^m,\hat{\theta}) \right ] $, thus,
$$u_3\leq 0.$$

By summing all the inequalities, we conclude that, 
$$  \mathbb{E}_{\mathcal{G}  \sim \mathcal{D}}\left [ \ell(\mathcal{G},\hat{\theta}_{\text{aug}}) \right ]  -   \mathbb{E}_{\mathcal{G}  \sim \mathcal{D}}\left [ \ell(\mathcal{G},\theta_{\star}) \right ]   < 2 \mathcal{R}(\ell_{\text{aug}} ) + 5 \sqrt{\frac{2 \log(4/\delta)}{N}} + 2 \text{L}_{\text{Lip}}  \mathbb{E}_{\mathcal{G}  \sim \mathcal{D}}\mathbb{E}_{\widetilde{\mathcal{G}}_n^m  \sim A_\lambda }\left [ \left \| \widetilde{\mathcal{G}}_n^m -\mathcal{G}_n \right \|  \right ].$$
Part 2 of the proof.
\begin{align*}
\mathcal{R}(\ell_{\text{aug}} )-\mathcal{R}(\ell) &  = \mathbb{E}_{\epsilon_n \sim P_{\epsilon}}\left [  \sup_{\theta \in \Theta} \left | \frac{1}{N} \sum_{n=1}^{N} \epsilon_n \ell_{\text{aug}}(\mathcal{G}_n, \theta)  \right |-  \sup_{\theta \in \Theta} \left | \frac{1}{N} \sum_{n=1}^{N} \epsilon_n \ell(\mathcal{G}_n, \theta)  \right |  \right ] \\
& \leq \mathbb{E}_{\epsilon_n \sim P_{\epsilon}}\left [  \sup_{\theta \in \Theta} \left | \frac{1}{N} \sum_{n=1}^{N} \epsilon_n \ell_{\text{aug}}(\mathcal{G}_n, \theta) -  \frac{1}{N} \sum_{n=1}^{N} \epsilon_n \ell(\mathcal{G}_n, \theta)  \right |  \right ] \\
& = \mathbb{E}_{\epsilon_n \sim P_{\epsilon}}\left [  \sup_{\theta \in \Theta} \left | \frac{1}{N} \sum_{n=1}^{N} \epsilon_n \left ( \ell_{\text{aug}}(\mathcal{G}_n, \theta) -   \ell(\mathcal{G}_n, \theta)  \right ) \right |  \right ]\\
& \leq \mathbb{E}_{\epsilon_n \sim P_{\epsilon}}\left [  \sup_{\theta \in \Theta}  \frac{1}{N} \sum_{n=1}^{N} \left | \epsilon_n \left ( \ell_{\text{aug}}(\mathcal{G}_n, \theta) -   \ell(\mathcal{G}_n, \theta)  \right ) \right |  \right ]\\
& \leq  \sup_{\theta \in \Theta}  \frac{1}{N} \sum_{n=1}^{N} \left |  \ell_{\text{aug}}(\mathcal{G}_n, \theta) -   \ell(\mathcal{G}_n, \theta)   \right | \\
& = \sup_{\theta \in \Theta}  \frac{1}{N} \sum_{n=1}^{N} \left |  \mathbb{E}_{\mathcal{G}_n^m \sim A_\lambda} \left [ \ell(\mathcal{G}_n^\lambda, \theta) -   \ell(\mathcal{G}_n, \theta) \right ]   \right |\\
& \leq  \max_{n \in \{1, \ldots,N\}}  \text{L}_{\text{Lip}}    \mathbb{E}_{\mathcal{G}_n^m \sim A_\lambda} \left [ \left \| \widetilde{\mathcal{G}}_n^m -\mathcal{G}_n \right \| \right ].   
\end{align*}

\end{proof}

\section{Proof of Proposition \ref{gen:prop:sampling_ineq}} \label{gen:proposition:KL_proof}
\textbf{Proposition \ref{gen:prop:sampling_ineq}}
Let $\delta_\mathcal{D}$ denote the discrete distribution of the training graph representations. Suppose we sample new augmented graph representations from a distribution $Q_\lambda$  defined on $\mathbb{R}^d$. Then, the following inequality holds,
\begin{equation*}\mathbb{E}_{\mathbf{h} \sim \delta_\mathcal{D},\widetilde{\mathbf{h}} \sim Q_\lambda} \left[ \| \mathbf{h} - \widetilde{\mathbf{h}} \| \right]  
\leq   \sqrt{2} \cdot \sup_{\substack{\mathbf{h} \sim \delta_\mathcal{D} \\ \widetilde{\mathbf{h}} \sim Q_\lambda}} \| \mathbf{h} - \widetilde{\mathbf{h}} \| \left( \sqrt{ \mathcal{KL}(\delta_\mathcal{D} \parallel Q_\lambda) } + \sqrt{2} \right).\end{equation*} 
where $\mathcal{KL}(\delta_\mathcal{D} \parallel Q_\lambda)$ is  is the Kullback-Leibler divergence from $\delta_\mathcal{D}$ to $Q_\lambda$.

\begin{proof}
We have,

\begin{align*}
    \mathbb{E}_{\mathbf{h} \sim \delta_{G},\widetilde{\mathbf{h}} \sim Q_\lambda} \left [ \|\mathbf{h} - \widetilde{\mathbf{h}}\|  \right ] & = \int_{h \in \mathbb{R}^d}  \int_{\widetilde{h} \in \mathbb{R}^d} \|\mathbf{h} - \widetilde{\mathbf{h}}\| \delta_{\mathcal{D}} ( \mathbf{h}) Q_\lambda( \widetilde{\mathbf{h}} ) \mathop{d\mathbf{h}} \mathop{d\widetilde{\mathbf{h}}},
\end{align*}

where $\delta_\mathcal{D}: \mathbf{h}\mapsto \frac{1}{N} \sum_{n=1}^{N} \delta_{\mathbf{h}_{\mathcal{G}_n}}(\mathbf{h})$, and $\delta$ is the Dirac distribution.

\begin{align*}
    \mathbb{E}_{\mathbf{h} \sim \delta_\mathcal{D}, \widetilde{\mathbf{h}} \sim Q_\lambda} \left [ \|\mathbf{h} - \widetilde{\mathbf{h}}\|  \right ]  & =\int_{h \in \mathbb{R}^d}  \int_{\widetilde{h} \in \mathbb{R}^d} \|\mathbf{h} - \widetilde{\mathbf{h}}\| \delta_{\mathcal{D}} ( \mathbf{h}) Q_\lambda( \widetilde{\mathbf{h}} ) \mathop{d\mathbf{h}} \mathop{d\widetilde{\mathbf{h}}} \\
    &  \leq C\int_{\mathbb{R}^d} \int_{\mathbb{R}^d} \left|  \delta_{\mathcal{D}} ( \mathbf{h})  -  Q_\lambda( \widetilde{\mathbf{h}} )\right| \delta_{\mathcal{D}} ( \mathbf{h}) Q_\lambda( \widetilde{\mathbf{h}} )    \mathop{d\mathbf{h}} \mathop{d\widetilde{\mathbf{h}}},
\end{align*}

where 
$$C = \sup_{ \substack{\mathbf{h} \sim \delta_\mathcal{D}\\
                  \widetilde{h} \sim \mathcal{Q}\\
                  h \neq \widetilde{h} }} \frac{\|\mathbf{h} - \widetilde{\mathbf{h}}\| \delta_{\mathcal{D}} ( \mathbf{h}) Q_\lambda( \widetilde{\mathbf{h}} )  }{\left|  \delta_{\mathcal{D}} ( \mathbf{h})  -  Q_\lambda( \widetilde{\mathbf{h}} ) \right|}.$$

\begin{align*}
    \mathbb{E}_{\mathbf{h} \sim \delta_\mathcal{D}, \widetilde{h} \sim \mathcal{Q}} \left [ \|\mathbf{h} - \widetilde{\mathbf{h}}\|  \right ]  & \leq C\int_{h \in \mathbb{R}^d} \int_{\widetilde{h} \in \mathbb{R}^d} \left|  \delta_{\mathcal{D}} ( \mathbf{h})  -  Q_\lambda( \widetilde{\mathbf{h}} )\right|    \mathop{d\mathbf{h}} \mathop{d\widetilde{\mathbf{h}}}  \\ 
    & \leq C\int_{h \in \mathbb{R}^d} \int_{\substack{\widetilde{h} \in \mathbb{R}^d \\ h = \widetilde{h}}} \left|  \delta_{\mathcal{D}} ( \mathbf{h})  -  Q_\lambda( \widetilde{\mathbf{h}} )\right|     \mathop{d\mathbf{h}} \mathop{d\widetilde{\mathbf{h}}}  + C\int_{h \in \mathbb{R}^d} \int_{\substack{\widetilde{h} \in \mathbb{R}^d \\ h \neq \widetilde{h}}} \left|  \delta_{\mathcal{D}} ( \mathbf{h})  -  Q_\lambda( \widetilde{\mathbf{h}} )\right| \mathop{d\mathbf{h}} \mathop{d\widetilde{\mathbf{h}}}.  
\end{align*}
We first get an upperbound for the first term in the right side of the inquality,
\begin{align*}
    \int_{h \in \mathbb{R}^d} \int_{\substack{\widetilde{h} \in \mathbb{R}^d  h = \widetilde{h}}} \left|  \delta_{\mathcal{D}} ( \mathbf{h})  -  Q_\lambda( \widetilde{\mathbf{h}} )\right|     \mathop{d\mathbf{h}} \mathop{d\widetilde{\mathbf{h}}}  & =\int_{h \in \mathbb{R}^d} \left( \int_{\substack{\widetilde{h} \in \mathbb{R}^d h = \widetilde{h}}} \left|  \delta_{\mathcal{D}} ( \mathbf{h})  -  Q_\lambda( \widetilde{\mathbf{h}} )\right|   \mathop{d\widetilde{\mathbf{h}}} \right)mathop{dh} \\
     & =\int_{h \in \mathbb{R}^d} \left( \int_{\substack{\widetilde{h} \in \mathbb{R}^d h = \widetilde{h}}} \left|  \delta_{\mathcal{D}} ( \mathbf{h})  -  Q( h)\right|    \mathop{d\widetilde{\mathbf{h}}} \right) \mathop{d\mathbf{h}}  \\
     & =\int_{h \in \mathbb{R}^d} \left( \int_{\widetilde{h} \in \mathbb{R}^d} \delta_h(\widetilde{h})\mathop{d\widetilde{\mathbf{h}}} \right) \left|  \delta_{\mathcal{D}} ( \mathbf{h})  -  Q( h)\right|    \mathop{d\mathbf{h}} \\
     & =\int_{h \in \mathbb{R}^d} \left|  \delta_{\mathcal{D}} ( \mathbf{h})  -  Q( h)\right|   \mathop{d\mathbf{h}} \\
    & \leq \sqrt{\int_{h \in \mathbb{R}^d} \left|  \delta_{\mathcal{D}} ( \mathbf{h})  -  Q( h)\right|^2  \mathop{d\mathbf{h}}},  ~~ \text{~~ using Jensen's Inequaliy.}
\end{align*}


The term $\int_{h \in \mathbb{R}^d} \left|  \delta_{\mathcal{D}} ( \mathbf{h})  -  Q( h)\right|  \mathop{d\mathbf{h}} $ corresponds to the Total Variation Distance (TD) $d_{TV}$ between $P_G$ and $Q$ \citep{yao2024new}, i.e.
$$d_{TV}(P_G, Q) =2 \sup_{A \subset \mathbb{R}^d} \left| P_G(A) - Q(A) \right| = \int_{\mathbb{R}} \left| P_G(x) - Q(x) \right| dx = \| P_G - Q \|_1.$$

Using Pinsker’s Inequality, we have,
$$\int_{h \in \mathbb{R}^d} \left|  \delta_{\mathcal{D}} ( \mathbf{h})  -  Q( h)\right|  \mathop{d\mathbf{h}} \leq \sqrt{2\mathcal{KL}(\delta_\mathcal{D} \parallel Q_\lambda)}$$

For the second term in the upperboud, we have,
\begin{align*}
    \int_{h \in \mathbb{R}^d} \int_{\substack{\widetilde{h} \in \mathbb{R}^d \\ h \neq \widetilde{h}}} \left|  \delta_{\mathcal{D}} ( \mathbf{h})  -  Q_\lambda( \widetilde{\mathbf{h}} )\right|  \delta_{\mathcal{D}} ( \mathbf{h})   \mathop{d\mathbf{h}} \mathop{d\widetilde{\mathbf{h}}}   & = \int_{\substack{\widetilde{h} \in \mathbb{R}^d \\ h \neq \widetilde{h}}} \left| \frac{1}{N} \sum_{n=N}^{N} \delta_{h_{\mathcal{G}_n}}(h)     -  Q_\lambda( \widetilde{\mathbf{h}} )\right|  \left( \frac{1}{N}   \sum_{n=N}^{N} \delta_{h_{\mathcal{G}_n}}(h)   \right) \mathop{d\mathbf{h}} \mathop{d\widetilde{\mathbf{h}}} \\
    & = \frac{1}{N}   \sum_{n=1}^{N} \int_{\substack{\widetilde{h} \in \mathbb{R}^d \\ \widetilde{h} \neq h_{\mathcal{G}_n}}} \left| \frac{1}{N}  -  Q_\lambda( \widetilde{\mathbf{h}} )\right|   \mathop{d\mathbf{h}} \mathop{d\widetilde{\mathbf{h}}}  \leq 2,
\end{align*}
because distributions are bounded between 0 and 1

Let now upperbound the constant $C$. We have $\forall a,b \in \mathbb{R}^{+}, ~ ab \leq \frac{1}{2}(a-b)^2  $. Therefore, 

\begin{align*}
    C & = \sup_{ \substack{\mathbf{h} \sim \delta_\mathcal{D}\\
                  \widetilde{h} \sim \mathcal{Q}\\
                  h \neq \widetilde{h} }} \frac{\|\mathbf{h} - \widetilde{\mathbf{h}}\| \delta_{\mathcal{D}} ( \mathbf{h}) Q_\lambda( \widetilde{\mathbf{h}} )  }{\left|  \delta_{\mathcal{D}} ( \mathbf{h})  -  Q_\lambda( \widetilde{\mathbf{h}} ) \right|} \\
        & \leq  \frac{1}{2}\sup_{ \substack{\mathbf{h} \sim \delta_\mathcal{D}\\
                  \widetilde{h} \sim \mathcal{Q}\\
                  h \neq \widetilde{h} }} \|\mathbf{h} - \widetilde{\mathbf{h}}\| \left|  \delta_{\mathcal{D}} ( \mathbf{h})  -  Q_\lambda( \widetilde{\mathbf{h}} ) \right| \\
        & \leq  \frac{1}{2}\sup_{ \substack{\mathbf{h} \sim \delta_\mathcal{D}\\
                  \widetilde{h} \sim \mathcal{Q}\\
                  h \neq \widetilde{h} }} \|\mathbf{h} - \widetilde{\mathbf{h}}\| 2 ~~\text{Because distributions are bounded between 0 and 1}\\\\
        & = \sup_{ \substack{\mathbf{h} \sim \delta_\mathcal{D}\\
                  \widetilde{h} \sim \mathcal{Q} }} \|\mathbf{h} - \widetilde{\mathbf{h}}\|
\end{align*}

Therefore,
$$ \mathbb{E}_{\mathbf{h} \sim \delta_\mathcal{D}} \mathbb{E}_{ \widetilde{h} \sim \mathcal{Q}} \left [ \|\mathbf{h} - \widetilde{\mathbf{h}}\|  \right ]   
\leq  \sqrt{2} \sup_{ \substack{\mathbf{h} \sim \delta_\mathcal{D}\\
                   \widetilde{h} \sim \mathcal{Q} }} \|\mathbf{h} - \widetilde{\mathbf{h}}\|   \left(  \sqrt{\mathcal{KL}(\delta_\mathcal{D} \parallel Q_\lambda)}  + \sqrt{2}  \right). $$

\end{proof}

\section{Proof of Theorem \ref{gen:thm:closed_formula}}
\label{gen:app:proof_thm_infl}
In this section, we present the detailed proof of Theorem \ref{gen:thm:closed_formula}, which allows us to e perform an in-depth theoretical analysis of our augmentation strategy through the lens of influence functions.

\textbf{Theorem \ref{gen:thm:closed_formula}}
Given a test graph $\mathcal{G}_k$ from the test set, let $\hat{\theta} = \arg\min_\theta \mathcal{L}$ be the GNN parameters that minimize the objective function in \eqref{gen:eq:standard_loss}. The impact of upweighting the objective function $\mathcal{L}$ to $\mathcal{L}_{n,m}^{\text{aug}} = \mathcal{L} + \epsilon_{n,m} \ell(\widetilde{\mathcal{G}}_{n}^{m}, \theta)$, where $\widetilde{\mathcal{G}}_{n}^{m}$ is an augmented graph candidate of the training graph $\mathcal{G}_n$ and $\epsilon_{n,m}$ is a sufficiently small perturbation parameter, on the model performance on the test graph $\mathcal{G}_k^{test}$ is given by
$$
\frac{d \ell(\mathcal{G}_k^{\text{test}}, \hat{\theta}_{\epsilon_{n,m}})}{d \epsilon_{n,m}} = - \nabla_\theta \ell(\mathcal{G}_k^{test}, \hat{\theta})\mathbf{H}_{\hat{\theta}}^{-1} \nabla_\theta \ell(\widetilde{\mathcal{G}}_{n}^{m}, \hat{\theta} ),
$$
where $\hat{\theta}_{\epsilon_{n,m}} = \argmin_{\theta} \mathcal{L}_{n,m}^{\text{aug}}$ denotes the parameters that minimize the upweighted objective function $\mathcal{L}_{n,m}^{\text{aug}}$ and $\mathbf{H}_{\hat{\theta}}  =\nabla_{\theta}^2\mathcal{L}(\hat{\theta})$ is the Hessian Matrix of the loss w.r.t. the model parameters.

\begin{proof}
Let $\widetilde{\mathcal{G}}_n^m$ be an augmented graph candidate of the training graph $\mathcal{G}_n$ and $\epsilon_{n,m}$ is a sufficiently small perturbation parameter. The parameters $\hat{\theta}$ and 
$\hat{\theta}$ and $\hat{\theta}_{\epsilon_{n,m}}$  the parameters that minimize the empirical risk on the train set, i.e., 
\begin{align*}
    \hat{\theta} & = \arg\min_\theta \mathcal{L}, \\
    \hat{\theta}_{\epsilon_{n,m}} & =\arg\min_\theta \mathcal{L}_{n,m}^{\text{aug}} = \arg\min_\theta \mathcal{L} + \epsilon_{n,m} \ell(\widetilde{\mathcal{G}}_n^m, \theta).
\end{align*}

Therefore, we examine its first-order optimality conditions,

\begin{align}
   0 & =  \nabla_{\hat{\theta}}  \mathcal{L} \label{gen:eq:grad_normal}\\
   0 & = \nabla_{\hat{\theta}_{\epsilon_{n,m}}} \left( \mathcal{L} + \epsilon_{n,m} \ell(\widetilde{\mathcal{G}}_n^m\mathcal{G}_n^m, \theta) \right). \label{gen:eq:grad_aug}
\end{align}
\end{proof}

Using Taylor Expansion, we now develop Eq. \eqref{gen:eq:grad_aug}. We have  $\lim_{\epsilon_{n,m} \rightarrow 0} \hat{\theta}_{\epsilon_{n,m}} = \hat{\theta}$, thus,

\begin{equation*}
    0  \simeq \left[ \nabla_{\hat{\theta}}  \mathcal{L}(\hat{\theta}) + \epsilon_{n,m}  \nabla_{\hat{\theta}} \ell(\widetilde{\mathcal{G}}_n^m, \hat{\theta})  \right] + \left[ \nabla_{\hat{\theta}}^2  \mathcal{L}(\hat{\theta}) + \epsilon_{n,m}  \nabla_{\hat{\theta}}^2 \ell(\widetilde{\mathcal{G}}_n^m, \hat{\theta})  \right] \left( \hat{\theta}_{\epsilon_{n,m}} -\hat{\theta} \right) .
\end{equation*}
Therefore, 

$$\hat{\theta}_{\epsilon_{n,m}} -\hat{\theta}   = - \left[ \nabla_{\hat{\theta}}^2  \mathcal{L}(\hat{\theta}) + \epsilon_{n,m}  \nabla_{\hat{\theta}}^2 \ell(\widetilde{\mathcal{G}}_n^m, \hat{\theta})  \right]^{-1} \left[ \nabla_{\hat{\theta}}  \mathcal{L}(\hat{\theta}) + \epsilon_{n,m}  \nabla_{\hat{\theta}} \ell(\widetilde{\mathcal{G}}_n^m, \hat{\theta})  \right].  $$
Dropping the $\circ(\epsilon_{n,m})$ terms, and using the Equation \ref{gen:eq:grad_normal}, i.e. $\nabla_{\hat{\theta}}  \mathcal{L} = 0$, we conclude that,
$$\frac{\hat{\theta}_{\epsilon_{n,m}} -\hat{\theta} }{\epsilon_{n,m}}  = - \left[ \nabla_{\hat{\theta}}^2  \mathcal{L}(\hat{\theta}) \right]^{-1}  \nabla_{\hat{\theta}} \ell(\widetilde{\mathcal{G}}_n^m, \hat{\theta}) .  $$
Therefore, 

$$ \frac{d \hat{\theta}_{\epsilon_{n,m}}}{d \epsilon_{n,m}} \simeq \frac{\hat{\theta}_{\epsilon_{n,m}} -\hat{\theta} }{\epsilon_{n,m}}  = - \left[ \nabla_{\hat{\theta}}^2  \mathcal{L}(\hat{\theta}) \right]^{-1}  \nabla_{\hat{\theta}} \ell(\widetilde{\mathcal{G}}_n^m, \hat{\theta}).   $$

\begin{align*}
 \frac{d \ell(\mathcal{G}_k^{test}, \hat{\theta}_{\epsilon_{n,m}})}{d \epsilon_{n,m}}   & =  \frac{d \ell(\mathcal{G}_k^{test}, \hat{\theta}_{\epsilon_{n,m}})}{d\hat{\theta}_{\epsilon_{n,m}}}  
 \frac{d \hat{\theta}_{\epsilon_{n,m}}}{d \epsilon_{n,m}} \\
 & = - \nabla_\theta \ell(\mathcal{G}_k^{test}, \hat{\theta})\mathbf{H}_{\hat{\theta}}^{-1} \nabla_\theta \ell(\widetilde{\mathcal{G}}_n^m, \hat{\theta} ).
\end{align*}

\section{Mathematical Expressions of GCN and GIN} \label{gen:app:gnn_definitions}

In this section, we provide concise definitions of two widely used GNN architectures: Graph Convolutional Networks (GCN) and Graph Isomorphism Networks (GIN). These architectures differ in how they aggregate and combine information from neighboring nodes in a graph.

\paragraph{Graph Convolutional Network (GCN) \cite{kipf2017semisupervised}.} The GCN updates node embeddings by aggregating normalized features from their neighbors. Specifically, for a node $v\in \mathcal{V}$, its feature vector $\mathbf{h}_v^{(t)}$ at layer $t$ is computed as,
\begin{equation*}
    \mathbf{h}_v^{(t)} = \sigma \Bigg( \sum_{u \in \mathcal{N}(v) \cup \{v\}} \frac{1}{\sqrt{\deg(v) \deg(u)}} \mathbf{W}^{(t)} \mathbf{h}_u^{(t-1)} \Bigg),
\end{equation*}
where $\mathbf{h}_u^{(t-1)}$ is the feature vector of node $u$ at layer $t-1$, $\mathbf{W}^{(t)}$ is a trainable weight matrix for layer $t$, and $\sigma(\cdot)$ is a non-linear activation function, such as ReLU.

\paragraph{Graph Isomorphism Network (GIN) \cite{xu2018how}.} GIN is designed to match the expressiveness of the Weisfeiler-Lehman (WL) graph isomorphism test. It updates a node’s embedding by aggregating its own feature with those of its neighbors, followed by a a multi-layer perceptron (MLP). The update rule at each message passing layer $t$ is 

\begin{equation*}
    \mathbf{h}_v^{(t)} = \text{MLP}^{(t)} \Big( (1 + \epsilon^{(t)}) \cdot \mathbf{h}_v^{(t-1)} + \sum_{u \in \mathcal{N}(v)} \mathbf{h}_u^{(t-1)} \Big),
\end{equation*}

where $\epsilon^{(t)}$ is a fixed or learnable scalar. GIN allows for more expressive feature transformations compared to methods with fixed aggregation schemes, enabling it to distinguish a broader range of graph structures.

\paragraph{Matrix forms and comparison.}
The node representations $\mathbf{h}_v^{(t)}$ at each layer can be expressed in matrix form as $\mathbf{H}^{(t)} \in \mathbb{R}^{p \times d_t}$, where $p$ is the number of nodes in the graph, $d$ is hidden dimension in the $t-$th layer, and $\mathbf{H}^{(t)}$ is the concatenation of all node representations $\mathbf{h}_v^{(t)}$ for $v \in \mathcal{V}$. For GCN, the update rule can be written as:

\begin{equation*}
    \mathbf{H}^{(t)} = \sigma\Bigl(\widetilde{\mathbf{D}}^{-1/2} \widetilde{\mathbf{A}} \widetilde{\mathbf{D}}^{-1/2} \mathbf{H}^{(t-1)} \mathbf{W}^{(t)}\Bigr),
\end{equation*}

where $\widetilde{\mathbf{A}} = \mathbf{A} + \mathbf{I}$ is the adjacency matrix with self-loops, and $\widetilde{\mathbf{D}}$ is its diagonal degree matrix. In contrast, for GIN, the update rule is given by:

\begin{equation*}
    \mathbf{H}^{(t)} = \mathrm{MLP}^{(t)}\Bigl((1 + \epsilon^{(t)})  \mathbf{H}^{(t-1)} + \mathbf{A} \mathbf{H}^{(t-1)}\Bigr).
\end{equation*}

If the same MLP is used, comprising a learnable linear layer followed by a ReLU activation, the only difference between GCN and GIN lies in the \textit{graph shift operator}: GCN uses the degree-normalized operator $\widetilde{\mathbf{D}}^{-1/2} \widetilde{\mathbf{A}} \widetilde{\mathbf{D}}^{-1/2}$, while GIN uses $(1 + \epsilon^{(t)})\mathbf{I} + \mathbf{A}$.

Therefore, when node features are taken as constant and only the graph structure is considered, the difference in their Lipschitz behavior can be traced back to the function that maps the adjacency matrix $A$ (and degrees) to these respective shift operators. In other words, the Lipschitz constant difference arises from whether the adjacency information is normalized ($\widetilde{\mathbf{D}}^{-1/2}\,\widetilde{\mathbf{A}}\,\widetilde{\mathbf{D}}^{-1/2}$) or as ($(1 + \epsilon^{(t)})\mathbf{I} + \mathbf{A}$).

\paragraph{Lipschitz constants of GCN and GIN.} Under constant node features, the Lipschitz constant of each architecture can be decomposed into two parts: one accounting for the sensitivity of the mapping from the adjacency matrix to the respective \emph{graph shift operator}, i.e., degree-normalized $\widetilde{\mathbf{D}}^{-1/2}\,\widetilde{\mathbf{A}}\,\widetilde{\mathbf{D}}^{-1/2}$ for GCN vs. $(1 + \epsilon)\mathbf{I} + \mathbf{A}$ for GIN, and the other capturing all shared learnable transformations. 

Let $\ell_{\mathrm{GCN}}$ and $\ell_{\mathrm{GIN}}$ be respectively  the graph shift operators $A \mapsto \widetilde{\mathbf{D}}^{-1/2}\,\widetilde{\mathbf{A}}\,\widetilde{\mathbf{D}}^{-1/2}$ and 
$A \mapsto (1 + \epsilon)\mathbf{I} + \mathbf{A}$, and let $\ell_{\mathrm{params}}$ represent the product of Lipschitz factors arising from the shared functions. Then, 
\begin{align*}
 L_{\mathrm{GCN}}& =\ell_{\mathrm{GCN}}\times\ell_{\mathrm{params}}    \\
  L_{\mathrm{GIN}}& =\ell_{\mathrm{GIN}}\times\ell_{\mathrm{params}} 
\end{align*}

It becomes clear that the difference between GCN and GIN Lipschitz constants depends solely on whether the graph adjacency is normalized, since $\ell_{\mathrm{params}}$ is common to both. It's straightforward that $\ell_{\mathrm{GIN}} \leq 1$ since the corresponding graph shift operator is just a translation. Let now derive an upperbound for $\ell_{\mathrm{GIN}}.$ Let $\mathbf{A}_1, \mathbf{A}_2 \in \mathbb{R}^{p \times p}$ be adjacency matrices of graphs on $n$ nodes. For each adjacency matrix $A_i$, we define the diagoanl digree matrix, $\mathbf{D}_i = \mathrm{diag}\bigl(\text{deg}_1^{(i)}, \dots, \text{deg}_p^{(i)}\bigr)$ where $\forall j \leq p,~~\text{deg}_j^{(i)} = \sum_{k=1}^n (\mathbf{A}_i)_{j k}.$ We have,
\begin{align*}
D_1^{-\tfrac12}\,\mathbf{A}_1\,\mathbf{D}_1^{-\tfrac12} - \mathbf{D}_2^{-\tfrac12}\,\mathbf{A}_2\,\mathbf{D}_2^{-\tfrac12} & = \mathbf{D}_1^{-\tfrac12}\,\mathbf{A}_1\,\mathbf{D}_1^{-\tfrac12} - \mathbf{D}_1^{-\tfrac12}\,\mathbf{A}_2\,\mathbf{D}_1^{-\tfrac12} +  \mathbf{D}_1^{-\tfrac12}\,\mathbf{A}_2\,\mathbf{D}_1^{-\tfrac12}
  \;-\; 
  \mathbf{D}_2^{-\tfrac12}\,\mathbf{A}_2\,\mathbf{D}_2^{-\tfrac12}
  \\
  &=   \underbrace{\mathbf{D}_1^{-\tfrac12}\,\bigl(\mathbf{A}_1 - \mathbf{A}_2\bigr)\,\mathbf{D}_1^{-\tfrac12}}_{\text{(i)}} + 
  \underbrace{\bigl[\mathbf{D}_1^{-\tfrac12} - \mathbf{D}_2^{-\tfrac12}\bigr]\;\mathbf{A}_2\;\mathbf{D}_1^{-\tfrac12} + 
  \mathbf{D}_2^{-\tfrac12}\,\mathbf{A}_2\,\bigl[\mathbf{D}_1^{-\tfrac12} - \mathbf{D}_2^{-\tfrac12}\bigr]}_{\text{(ii)}}.
\end{align*}

We can upperbound each of (i), and (ii). For (i), we have the following upperbound,
\begin{equation*}
      \bigl\|\mathbf{D}_1^{-\tfrac12}\,\bigl(\mathbf{A}_1 - \mathbf{A}_2\bigr)\,\mathbf{D}_1^{-\tfrac12}\bigr\| \leq   \|\mathbf{D}_1^{-\tfrac12}\|^2 \,\|\mathbf{A}_1 - \mathbf{A}_2\|
  \leq   \frac{1}{\delta_{1,\text{min}}}\,\|\mathbf{A}_1 - \mathbf{A}_2\|,
\end{equation*}
where $\delta_{i,\text{min}} = \min_j \text{deg}_j^{(i)}.$ For the second (ii), if we consider for example $\|\|$ as the $\text{L}_1$ norm, then,

\begin{align*}
    \|(ii)\| & \leq   \|\mathbf{D}_1^{-1/2} - \mathbf{D}_2^{-1/2}\| \|\mathbf{A}_2\| \bigl(\|\mathbf{D}_1^{-1/2}\| +\|\mathbf{D}_2^{-1/2}\|\bigr), \\
    & \leq \frac{2}{\text{min}(\delta_{1,\text{min}},\delta_{2,\text{min}})^{1/2}} \|\mathbf{D}_1^{-1/2} - \mathbf{D}_2^{-1/2}\| \|\mathbf{A}_2\| \\
    & \leq \frac{2M}{\text{min}(\delta_{1,\text{min}},\delta_{2,\text{min}})^{1/2}} \|\mathbf{D}_1^{-1/2} - \mathbf{D}_2^{-1/2}\|  \\ 
    & \leq \frac{2M}{\text{min}(\delta_{1,\text{min}},\delta_{2,\text{min}})^{5/2}} \|\mathbf{D}_1 - \mathbf{D}_2\| \\
    & \leq \frac{2M}{\text{min}(\delta_{1,\text{min}},\delta_{2,\text{min}})^{5/2}} \|\mathbf{A}_1 \mathbf{1}_p - \mathbf{A}_2\mathbf{1}_p\|\\
    & \leq \frac{2M p}{\text{min}(\delta_{1,\text{min}},\delta_{2,\text{min}})^{5/2}} \|\mathbf{A}_1  - \mathbf{A}_2\|,
\end{align*}

where $M$ is the maximum norm of the adjacency matrix in the graph dataset, and $\mathbf{1}_p \in \mathbb{R}^p$ is the vector of ones.

Putting both inequalities together, we can come up with an upperbound for $\ell_{\mathrm{GCN}}$ that depends on the minimum degree in the dataset.

\section{Configuration Models}\label{gen:app:config_models}
In this section, we present a novel adaptation of configuration models as a graph data augmentation technique for GNN. Configuration models \cite{yang2013networks} enable the generation of randomized graphs that maintain the original degree distribution. We can, therefore, leverage this strategy to improve the generalization of GNNs. Below, we present the steps involved in our approach to using Configuration Models for Graph Data Augmentation:
\begin{enumerate}
    \item \textbf{Extract edges:} For each training graph $\mathcal{G}_n$, we first extract the complete set of edges $\mathcal{E}_n$. 
    \item \textbf{Stub creation:} Using a Bernoulli distribution with parameter $r \in [0,1]$, we randomly select a subset of candidate edges and \textit{break} them to create \textit{stubs} (half-edges).
    \item \textbf{Stub pairing:} We then randomly pair these stubs to form new edges, creating a randomized graph structure with the same degree distribution. 
\end{enumerate}
Table \ref{gen:tab:config_models} shows the performance of this approach on the two GNN backbones, GCN and GIN.
\begin{table}[ht]
\centering
\caption[Classification accuracy for the data augmentation baselines based on the GIN backbone]{Classification accuracy ($\pm$ standard deviation) on different benchmark node classification datasets for the data augmentation baselines based on the GIN backbone. The higher the accuracy (in \%) the better the model.}
\resizebox{0.8\columnwidth}{!}{%
\begin{tabular}{lllllll}
\toprule 
Model & IMDB-BIN  & IMDB-MUL & MUTAG & PROTEINS  & DD \\ \midrule

Config Models w/ GCN   & $71.70  {\scriptstyle \pm 3.16}$ & $48.40  {\scriptstyle \pm 3.88}$ & $74.97  {\scriptstyle \pm 6.77}$ &  $70.08  {\scriptstyle \pm 4.93}$ & $69.01  {\scriptstyle \pm 3.44}$ \\ 
Config Models w/ GIN  & $71.70  {\scriptstyle \pm 4.24}$ & $49.00  {\scriptstyle \pm 3.44}$  & $81.43  {\scriptstyle \pm 10.05}$ & $68.34  {\scriptstyle \pm 5.30} $ & $71.61  {\scriptstyle \pm 5.93}$  \\
 \bottomrule

\end{tabular}
}

\label{gen:tab:config_models}
\end{table}

As noticed, the configuration model-based graph augmentation method performs competitively with the baselines and even outperforms them in certain cases. This underscores the importance of Theorem \ref{gen:thm:rademacher}. When compared to our approach \method, the latter gives better results across different datasets and GNN backbones. This difference is primarily due to the configuration model based approach being model-agnostic, whereas \method~leverages the model's weights and architecture, as explained in Section \ref{gen:sub_sec:influence} and supported by Theorem \ref{gen:thm:closed_formula}. 

\section{Ablation Study}\label{gen:app:ablation_study}

To provide additional comparison and motivate the use of GMMs with the EM algorithm within \method, we expanded our evaluation to include additional methods for modeling the distribution of the graph representations. Specifically, the comparison includes:
\begin{itemize}
    \item \textbf{GMM w/ Variational Bayesian Inference (VBI):}  We specifically compared the Expectation-Maximization (EM) algorithm, discussed in the main paper, with the Variational Bayesian (VB) estimation technique for parameter estimation of each Gaussian Mixture Model (GMM) \citep{tzikas2008variational} for both the GCN and GIN models. The objective of including this baseline is to explore alternative approaches for fitting GMMs to the graph representations.
    \item \textbf{Kernel Density Estimation (KDE):} KDE is a Neighbor-Based Method and a non-parametric approach to estimating the probability density \citep{hardle2004nonparametric}. KDE estimates the probability density function by placing a kernel function (e.g., Gaussian) at each data point. The sum of these kernels approximates the underlying distribution. Sampling can be done using techniques like Metropolis-Hastings. The purpose of using KDE as a baseline is to evaluate alternative distributions different from the Gaussian Mixture Model~(GMM).
    \item \textbf{Copula-Based Methods:} We model the dependence structure between variables using copulas, while marginal distributions are modeled separately. We sample from marginal distributions and then transform them using the copula \citep{nelsen2006introduction}.
    \item \textbf{Generative Adversarial Network (GAN):} GANs are powerful generative models that learn to approximate the data distribution through an adversarial process between two neural networks. To evaluate the performance of deep learning-based generative approaches for modeling graph representations, we included tGAN, a GAN architecture specifically designed for tabular data \citep{yang2012robust}. We particularly train tGAN on the graph representations and then sample new graph representations from the generator.
\end{itemize}

\begin{table}[h]
\centering
\caption[Ablation study on the density estimation scheme - GCN]{Ablation study on the density estimation scheme for learned GCN representations in \method. }
\label{gen:tab:ablation_GCN}
\resizebox{0.7\columnwidth}{!}{%
\begin{tabular}{lllllll}
\toprule
Model & IMDB-BIN  & IMDB-MUL & MUTAG & PROTEINS  & DD \\ \midrule

GMM w/ EM  & $\mathbf{ 71.00  {\scriptstyle \pm  4.40}}$ & $\mathbf{49.82  {\scriptstyle \pm  4.26}}$  & $\mathbf{76.05  {\scriptstyle \pm  6.47}}$ & $\mathbf{70.97  {\scriptstyle \pm  5.07}}$ &  $\mathbf{71.90  {\scriptstyle \pm  2.81}}$  \\
GMM w/ VBI  & $\mathbf{71.00  {\scriptstyle \pm  4.21} }$&  $49.53  {\scriptstyle \pm  4.26}$  & $\mathbf{76.05  {\scriptstyle \pm  6.47}}$ & $\mathbf{70.97  {\scriptstyle \pm  4.52}}$ & $71.64  {\scriptstyle \pm  2.90}$ \\
KDE  &$55.90  {\scriptstyle \pm 10.29}$ & $39.53  {\scriptstyle \pm  2.87}$ & $66.64  {\scriptstyle \pm 6.79}$ & $59.56  {\scriptstyle \pm 2.62}$ & $58.66  {\scriptstyle \pm 3.97}$ \\
Copula & $69.80  {\scriptstyle \pm  4.04}$ & $47.13  {\scriptstyle \pm 3.45}$ & $74.44  {\scriptstyle \pm  6.26}$ & $65.04  {\scriptstyle \pm  3.37}$ & $65.70  {\scriptstyle \pm  3.04}$ \\
GAN & $70.60  {\scriptstyle \pm  3.41}$ & $48.80  {\scriptstyle \pm 5.51}$ &  $75.52  {\scriptstyle \pm 4.96}$& $69.98  {\scriptstyle \pm  5.46}$ & $66.26  {\scriptstyle \pm  3.72}$ \\
 \bottomrule

\end{tabular}
}
\end{table}

\begin{table}[h]
\centering
\caption[Ablation study on the density estimation scheme - GIN]{Ablation study on the density estimation scheme for learned GIN representations in \method.}
\label{gen:tab:ablation_GIN}
\resizebox{0.7\columnwidth}{!}{%
\begin{tabular}{lllllll}
\toprule
Model & IMDB-BIN  & IMDB-MUL & MUTAG & PROTEINS  & DD \\ \midrule

GMM w/ EM  &$ \mathbf{71.70  {\scriptstyle \pm  4.24}}$ & $\mathbf{49.20  {\scriptstyle \pm  2.06}}$ & $\mathbf{88.83  {\scriptstyle \pm  5.02}}$ & $\mathbf{71.33  {\scriptstyle \pm  5.04}} $ &$\mathbf{68.61  {\scriptstyle \pm 4.62}}$ \\
GMM w/ VBI  & $71.40  {\scriptstyle \pm  2.65}$ & $47.80  {\scriptstyle \pm  2.22}$ & $88.30  {\scriptstyle \pm  5.19}$ & $70.25  {\scriptstyle \pm  4.65}$ & $67.82  {\scriptstyle \pm 4.96}$\\
KDE  & $69.10  {\scriptstyle \pm  3.93}$ & $41.46  {\scriptstyle \pm 3.02}$ & $77.60  {\scriptstyle \pm 6.83}$ &  $60.37  {\scriptstyle \pm  3.04}$ & $67.48  {\scriptstyle \pm  6.18}$ \\
Copula & $70.60  {\scriptstyle \pm  2.61}$ & $47.60  {\scriptstyle \pm 2.29}$ & $88.30  {\scriptstyle \pm  5.19}$ & $70.16  {\scriptstyle \pm  4.55}$ & $67.91  {\scriptstyle \pm 4.90}$ \\
GAN & $70.50  {\scriptstyle \pm 3.80}$ & $48.40  {\scriptstyle \pm 1.71}$ & $\mathbf{88.83  {\scriptstyle \pm  5.02}}$ & $\mathbf{71.33  {\scriptstyle \pm  5.55}}$ & $67.74  {\scriptstyle \pm 4.82}$ \\
 \bottomrule

\end{tabular}
}
\end{table}

We compare these approaches for both the GCN and GIN models in Tables \ref{gen:tab:ablation_GCN} and \ref{gen:tab:ablation_GIN}, respectively. As noticed, GMM with EM consistently outperforms the alternative methods across most datasets in terms of accuracy. The VBI method, an alternative approach for estimating GMM parameters, yields comparable 
performance to the EM algorithm. This consistency across datasets highlights the effectiveness and robustness of GMMs in capturing the underlying data distribution.

In certain cases, particularly with the GIN model, we observed competitive performance from the GAN approach, which, unlike GMM, requires additional training. Hence, GMMs provide a more straightforward and efficient solution.



 \section{Training and Augmentation time} \label{gen:app:time}
We compare the data augmentation times of our approach and the baselines in Table \ref{gen:tab:complexity}. In addition to outperforming the baselines on most datasets, our approach offers an advantage in terms of time complexity. The training time of baseline models varies depending on the augmentation strategy used, specifically, whether it involves pairs or individual graphs. Even in cases where a graph augmentation has a low computational cost for some baselines, training can still be time-consuming as multiple augmented graphs are required to achieve satisfactory test accuracy. For instance, methods like DropEdge, DropNode, and SubMix, while computationally simple, require generating multiple augmented samples at each epoch, thereby increasing the overall training time. Following the framework of \cite{yoo2022model}, we must sample several augmented graphs for each training graph at every epoch to achieve optimal results. In contrast, \method~ introduces a more efficient approach by generating only one augmented graph per training instance, which is reused across all epochs. This design ensures a balance between computational efficiency and augmentation effectiveness, reducing the overall training burden while maintaining strong performance. The only baseline that is more time-efficient than our approach is GeoMix; however, our method consistently outperforms GeoMix across all settings, as shown in Tables \ref{gen:tab:results_gcn} and~\ref{gen:tab:results_gin}.
\begin{table}[h]
\centering
\caption[Mean training and augmentation time in second]{Mean training and augmentation time in seconds of our model in comparison to the other benchmarks.}
\begin{tabular}{l|llrrr}
\toprule
& Time & Model  & IMDB-BIN  & MUTAG   &  DD \\ \midrule
\multirow{5}{*}{\textcircled{1}} & \multirow{7}{*}{Aug. Time} & Vanilla  & - & - & -\\
&   & DropEdge  & 0.02 & 0.01 & 0.01  \\\
&   & DropNode & 0.01 & 0.02 & 0.01  \\
 &   & SubMix & 1.27 & 0.23 &  0.45 \\
 &   & $\mathcal{G}$-Mixup   & 0.74 & 0.11 & 4.26\\
&   & GeoMix  & 2,344.12 & 73.52  & 1,005.35  \\ 
&   & \method  & 2.87 & 0.51 & 3.25 \\ \hline
\multirow{5}{*}{\textcircled{2}} & \multirow{7}{*}{Train. Time} & Vanilla  &765.96 & 99.32 & 428.10  \\
&   & DropEdge  & 892.14 & 596.82 & 3,037.30  \\
&   & DropNode & 884.71 & 803.63 & 3,325  \\
 &   & SubMix & 1,711.01 & 1,487.03 &  2,751.92 \\
 &   & $\mathcal{G}$-Mixup  &  148.71 & 28.14 & 177.55 \\
&   & GeoMix  & 89.01 & 101.82  & 123.41 \\ 
&   & \method & 774.47 & 101.56  & 438.39  \\ \bottomrule

\end{tabular}
\label{gen:tab:complexity}
\end{table}

\begin{table}[h]
\centering
\caption[Classification accuracy for the data augmentation baselines using the GCN backbone]{Classification accuracy ($\pm$ std) on different benchmark datasets for the data augmentation baselines using the GCN backbone. Higher accuracy (in \%) is better. }
\resizebox{0.7\columnwidth}{!}{%
\begin{tabular}{lccccc}
\toprule
Model & IMDB-BIN  & IMDB-MUL & MUTAG & PROTEINS  & DD \\ 
\midrule
DropEdge  & $71.40  {\scriptstyle \pm 3.69}$ & $48.47  {\scriptstyle \pm 2.36}$ & $73.88  {\scriptstyle \pm 7.98}$  &   $67.56  {\scriptstyle \pm 4.50}$ &  $66.04  {\scriptstyle \pm 4.35}$ \\
DropNode  & $71.80  {\scriptstyle \pm 4.11}$ & $48.60  {\scriptstyle \pm 3.45}$ & $72.78  {\scriptstyle \pm 8.01}$ &  $67.83  {\scriptstyle \pm 4.75}$ &  $66.55  {\scriptstyle \pm 3.97}$ \\
SubMix  & $72.50   {\scriptstyle \pm 3.90}$  & $47.93   {\scriptstyle \pm 3.58}$  & $71.72 {\scriptstyle \pm 10.59}$  & $59.74  {\scriptstyle \pm 2.83}$ & $62.25  {\scriptstyle \pm 3.29}$\\
\bottomrule
\end{tabular}
}
\label{gen:tab:results_gcn_classic_baselines}
\end{table}

\begin{table}[h]
\centering
\caption[Classification accuracy for the data augmentation baselines using the GIN backbone]{Classification accuracy ($\pm$ std) on different benchmark datasets for the data augmentation baselines using the GIN backbone. Higher accuracy (in \%) is better. }
\resizebox{0.7\columnwidth}{!}{%
\begin{tabular}{lccccc}
\toprule
Model & IMDB-BIN  & IMDB-MUL & MUTAG & PROTEINS  & DD \\ 
\midrule
DropEdge  & $71.70  {\scriptstyle \pm 3.25}$ & $43.67  {\scriptstyle \pm 6.58}$ & $72.36  {\scriptstyle \pm 10.40}$  &   $66.31  {\scriptstyle \pm 5.49}$ &  $70.20  {\scriptstyle \pm 2.85}$ \\
DropNode  & $71.00  {\scriptstyle \pm 4.49}$ & $42.67  {\scriptstyle \pm 5.54}$ & $75.00  {\scriptstyle \pm 6.67}$ &  $64.60  {\scriptstyle \pm 4.52}$ &  $70.80  {\scriptstyle \pm 3.31}$ \\
SubMix  & $71.10    {\scriptstyle \pm 4.15}$  &  $42.80    {\scriptstyle \pm 3.93}$ & $84.53 {\scriptstyle \pm 6.66}$  & $60.10  {\scriptstyle \pm 3.78}$ & $70.03  {\scriptstyle \pm 2.61}$ \\
\bottomrule
\end{tabular}
}
\label{gen:tab:results_gin_classic_baselines}
\end{table}

Furthermore, the performance of simple augmentation baselines such as DropEdge, DropNode, and SubMix significantly drops when using \textit{only one augmentation} per training graph for all the epochs, which is the framework adopted in \method, as shown in Tables~\ref{gen:tab:results_gcn_classic_baselines} and~\ref{gen:tab:results_gin_classic_baselines}. Tables~\ref{gen:tab:results_gcn_classic_baselines} and~\ref{gen:tab:results_gin_classic_baselines}, these baselines experience a significant drop in accuracy under this constraint, emphasizing their reliance on generating diverse augmentations at each epoch to maintain strong performance, c.f. the results in Tables \ref{gen:tab:results_gcn} and~\ref{gen:tab:results_gin}. 

The only graph augmentation baseline with comparable or better time complexity than \method is $\mathcal{G}$-Mixup. However, \method consistently outperforms $\mathcal{G}$-Mixup in most cases, as shown by the results in Tables~\ref{gen:tab:results_gcn} and~\ref{gen:tab:results_gin}.

\section{Augmentation Strategies}\label{gen:app:aug_strat}
We present several graph data augmentation strategies, each defined by the augmentation function  $A_{\lambda}$. For simplicity, we denote it $A_{\lambda}(\mathcal{G}_n)$  instead of the more explicit $A_{\lambda}(\mathcal{G}_n,y_n)$.

\textbf{Edge Perturbation.} Randomly adding or removing edges. Given a graph $\mathcal{G}_n = (\mathcal{V}_n, \mathcal{E}_n, \mathbf{X}_n)$, edge perturbation is defined as $A_{\lambda}(\mathcal{G}_n) = (\mathcal{V}_n, \mathcal{E}_n \cup \mathcal{E}_\text{add} \setminus \mathcal{E}_\text{remove}, \mathbf{X}_n),$ where edges in  $\mathcal{E}_\text{add}$ and  $\mathcal{E}_\text{remove}$  are sampled from a Bernouli distribution $\mathcal{P}(\lambda) = \mathcal{B}(\lambda)$ with probability $\lambda$.

\textbf{Node Feature Perturbation.} Augmenting node features by introducing noise or masking some features. This is given by $A_{\lambda}(\mathcal{G}_n) = (\mathcal{V}_n, \mathcal{E}_n, \mathbf{X}_n + \lambda \mathbf{Z}),$ where $\mathbf{Z} \sim \mathcal{N}(0, I)$  for Gaussian noise addition. 

\textbf{Subgraph Sampling.} Extracting subgraphs. A common approach is $k$-hop neighborhood sampling, where  $A_{\lambda}(\mathcal{G}_n) = (\mathcal{V}_n', \mathcal{E}_n', \mathbf{X}_n'),$
with $\mathcal{V}_n' \subseteq \mathcal{V}_n$  and  $\mathcal{E}_n'$ being edges induced by the $k$-hop neighbors of randomly selected nodes selected based on a prior distribution $\mathcal{P}(\lambda)$. 

\textbf{\method.}
In our approach, the augmented hidden representations $\mathbf{h}_{\widetilde{\mathcal{G}}} = A_\lambda(\mathbf{h}_{\mathcal{G}})$  corresponds to a sampled vector from the GMM  distribution $\mathcal{P}_c$ that was previously fit on the hidden representations $\mathcal{H}_c =\{\mathbf{h}_{\mathcal{G}}~~\mid y_n = c \}$ of the graphs in the training set with the same class $c$. Formally, $\mathbf{h}_{\widetilde{\mathcal{G}}}= A_{\lambda_c}( \{\mathcal{H}_c \mid \mathcal{G} \in c\}) = \lambda_c$, where $\lambda_c$ are the parameter of the GMM distribution $\mathcal{P}_c$.

It is important to note that in some augmentation strategies, the augmented graphs $\widetilde{\mathcal{G}}_{n}^{m}$ may not explicitly depend on the specific training graph $\mathcal{G}_n$. Instead, they may be sampled or generated based on other factors, such as a general graph distribution or global augmentation rules. This flexibility allows the augmentation framework to capture a broader range of variations while maintaining consistency with the original training data.

\section{Graph Distance Metrics}\label{gen:app:metrics}
Let us consider the graph structure space $(\mathbb{A}, \lVert \cdot \rVert_{\mathbb{A}})$ and the feature space $(\mathbb{X}, \lVert \cdot \rVert_{\mathbb{X}})$, where $\lVert \cdot \rVert_{\mathbb{G}}$ and $\lVert \cdot \rVert_{\mathbb{X}}$ denote the norms applied to the graph structure and features, respectively. When considering only structural changes with fixed node features, the distance between two graphs $\widetilde{\mathcal{G}},\mathcal{G}$ is defined as
  \begin{equation}\label{gen:eq:norm_1}
      \left \| \widetilde{\mathcal{G}} - \mathcal{G} \right \|= \lVert \mathbf{A} - \mathbf{\widetilde{A}} \rVert_{\mathbb{G}},
  \end{equation} where $\mathbf{\widetilde{A}},\mathbf{A}$ are respectively the adjacency matrix of $\widetilde{\mathcal{G}},\mathcal{G},$ and the norm $\lVert\cdot \rVert_{\mathbb{G}}$ can be for example the Frobenius or spectral norm. If both structural and feature changes are considered, the distance extends to:
 \begin{equation}\label{gen:eq:norm_2}
     \left \| \widetilde{\mathcal{G}} - \mathcal{G} \right \| = \alpha \lVert A - \widetilde{A} \rVert_{\mathbb{A}} +\beta \lVert \mathbf{X}- \widetilde{\mathbf{X}} \rVert_{\mathbb{X}},
 \end{equation}
 where $\widetilde{\mathbf{X}}, \mathbf{X}$ are the node feature matrices of $\widetilde{\mathcal{G}},\mathcal{G}$ respectively, and $\alpha, \beta$ are hyperparameters controlling the contribution of structural and feature differences. 

In most baseline graph augmentation techniques, such as $\mathcal{G}$-Mixup, SubMix, and DropNode, the alignment between nodes in the original graph $\mathcal{G}$ and the augmented graph $\widetilde{\mathcal{G}}$ is known.  However, in cases where the node alignment is unknown, we must take into account node permutations. The distance between the two graphs is then defined as
\begin{equation} \label{gen:eq:norm_3}
    \left \| \widetilde{\mathcal{G}} - \mathcal{G} \right \|  = \min_{P \in \Pi} \left( \alpha \lVert A - P \widetilde{A} P^T \rVert_{\mathbb{A}}  + \beta \lVert \mathbf{X}- P \widetilde{\mathbf{X}} \rVert_{\mathbb{X}} \right), \qquad \qquad 
\end{equation}
where $\Pi$ is the set of permutation matrices. The matrix $P$ corresponds to a permutation matrix used to order nodes from different graphs. By using Optimal Transport, we find the minimum distance over the set of permutation matrices, which corresponds to the optimal matching between nodes in the two graphs. This formulation represents the general case of graph distance, which has been used in the literature \citep{abbahaddou2024bounding}.

A common choice for measuring the distance between two graphs is the norm applied to their adjacency matrices. One widely used norm is the Frobenius norm, defined as  
$ \lVert A - \widetilde{A} \rVert_F = \sqrt{\sum_{i,j} (A_{ij} - \widetilde{A}_{ij})^2} $,  
which captures element-wise differences between the adjacency matrices. Another commonly used norm is the spectral norm, defined as  
$ \lVert A - \widetilde{A} \rVert_2 = \sigma_{\max}(A - \widetilde{A}) $,  
where $ \sigma_{\max} $ denotes the largest singular value of the difference matrix. 

\section{Gaussian Mixture Models}\label{gen:app:gmm}

GMMs are probabilistic models used for modeling complex data by representing them as a mixture of multiple Gaussian distributions. The probability density function \( p(\mathbf{x}) \) of a data point \( \mathbf{x} \) in a GMM with \( K \) Gaussian components is given by:
\begin{equation}
    p(\mathbf{x}) := \sum_{k=1}^K \pi_k \mathcal{N}(\mathbf{x} \mid \boldsymbol{\mu}_k, \boldsymbol{\Sigma}_k), \label{gen:eq:GMM_format}
\end{equation}
where $\pi_k$ is the weight of the $k$-th Gaussian component, with $ \pi_k \geq 0 $ and $\sum_{k=1}^K \pi_k = 1$, and $\mathcal{N}(\mathbf{x} \mid \boldsymbol{\mu}_k, \boldsymbol{\Sigma}_k) $ is the Gaussian probability density function for the $k$-th component, defined as,
\begin{equation*}
        \mathcal{N}(\mathbf{x} \mid \boldsymbol{\mu}_k, \boldsymbol{\Sigma}_k) := \frac{1}{(2 \pi)^{d/2} \det(\boldsymbol{\Sigma}_k)^{1/2}} \times  \exp \left(  -\frac{1}{2} (\mathbf{x} - \boldsymbol{\mu}_k)^\top \boldsymbol{\Sigma}_k^{-1} (\mathbf{x} - \boldsymbol{\mu}_k) \right),
\end{equation*}
where $\boldsymbol{\mu}_k $ and $\boldsymbol{\Sigma}_k$ are respectively the mean vectors and the covariance vectors of the $k$-th  Gaussian component, and $d$ the dimensionality of $\mathbf{x}$. The parameters of a GMM are typically estimated using the EM algorithm \citep{dempster1977maximum}, which alternates between estimating the membership probabilities of data points for each Gaussian component (Expectation step) and updating the parameters of the Gaussian distributions (Maximization step). GMMs are a powerful tool in statistics and machine learning and are used for various purposes, including clustering and density estimation \citep{ozertem2011locally,naim2012convergence,zhang2021gaussian}.

\section{Experimental Setup} \label{gen:app:dataset_impl}
\textbf{Datasets.} We evaluate our model on five widely used datasets from the GNN literature, specifically IMDB-BIN, IMDB-MUL, PROTEINS, MUTAG, and DD, all sourced from the TUD Benchmark \citep{Morris+2020}. These datasets consist of either molecular or social graphs. Detailed statistics for each dataset are provided in Table \ref{gen:tab:data_statistics}. We split
the dataset into train/test/validation set by 80\%/10\%/10\%
and use 10-fold cross-validation for evaluation following the recent work of \citet{zeng2024graph}. If a dataset does not contain node features, we follow the standard practice in GNN literature by using one-hot encoding of node degrees as input features.
\begin{table}[h]
\caption{Statistics of the graph classification datasets used in our experiments.}
\label{gen:tab:data_statistics}
\begin{center}
\begin{small}
\begin{tabular}{lccccc}
\toprule
Dataset & \#Graphs&  \#Features & Avg. Nodes & Avg. Edges & \#Classes \\
\midrule
IMDB-BIN    & 1,000 & - & 19.77 & 96.53  & 2\\
IMDB-MUL    & 1,500 & - & 13.00 & 65.94 & 3\\
MUTAG    & 188 & 7  &17.93  & 19.79 & 2\\
PROTEINS    & 1,113 & 3 & 39.06 & 72.82 & 2 \\
DD    & 1,178& 82 & 284.32 & 715.66 & 2 \\

\bottomrule
\end{tabular}
\end{small}
\end{center}
\end{table}

\textbf{Baselines.} We benchmark the performance of our approach against the state-of-the-art graph data augmentation strategies. In particular, we
consider the DropNode \citep{you2020graph}, DropEdge \citep{rong2019dropedge}, SubMix \citep{yoo2022model}, $\mathcal{G}$-Mixup \citep{han2022g}  and GeoMix \citep{zeng2024graph}.

\textbf{Implementation Details.}  We used the PyTorch Geometric (PyG) open-source library, licensed under MIT \citep{Fey/Lenssen/2019}. The experiments were conducted on an RTX A6000 GPU. For the datasets from the TUD Benchmark, we used a size base split.  We utilized two GNN architectures, GIN and GCN, both consisting of two layers with a hidden dimension of 32. The GNN was trained on graph classification tasks for 300 epochs with a learning rate of $10^{-2}$ using the Adam optimizer \cite{kingma_adam}. To model the graph representations of each class, we fit a GMM using the EM algorithm, running for 100 iterations or until the average lower bound gain dropped below $10^{-3}$. The number of Gaussians used in the GMM is provided in Table \ref{gen:tab:gmm_num}.  After generating new graph representations from each GMM, we fine-tuned the post-readout function for 100 epochs, maintaining the same learning rate of $10^{-2}$.

\textbf{Computation of Influence Scores.} Computing and inverting the Hessian matrix of the empirical risk is computationally expensive, with a complexity of  $\mathcal{O}(N \times p^2 + p^3)$, where $p=|\theta|$ is the number of parameters in the GNN. To mitigate the cost of explicitly calculating the Hessian matrix, we employ implicit Hessian-vector products (iHVPs), following the approach outlined in \citet{koh2017understanding}.

\begin{table}[h]

\caption[The optimal number of Gaussian distributions in the GMM ]{The optimal number of Gaussian distributions in the GMM for each pair of dataset and GNN backbone.}
\label{gen:tab:gmm_num}
\vskip 0.15in
\begin{center}
\begin{small}
\begin{tabular}{lccccc}
\toprule
Model & IMDB-BIN & IMDB-MUL & MUTAG & PROTEINS & DD\\
\midrule
GCN   & 40 & 50 & 10 & 10 & 2\\
GIN   & 50 & 5 & 2 & 2 & 50 \\

\bottomrule
\end{tabular}
\end{small}
\end{center}
\vskip -0.1in
\end{table}

\section{Fisher-Guided GMM Augmentation}\label{gen:app:fisher_gmm_experiment}
\begin{figure}[h]
    \centering
    \includegraphics[width=0.5\linewidth]{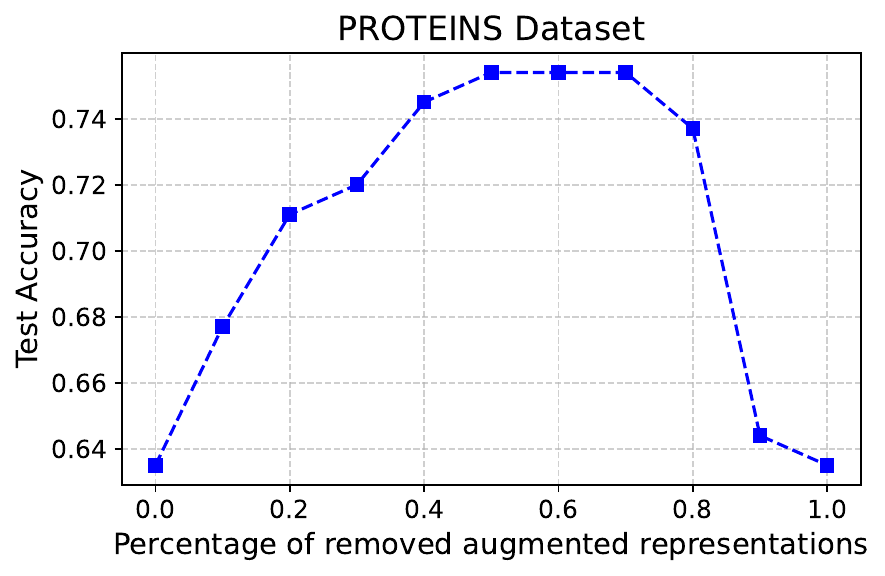}
    \caption{Effect of Filtering Augmented Representations on Test Accuracy}
    \label{gen:fig:fisher_proteins}
\end{figure}

Figure \ref{gen:fig:fisher_proteins} illustrates the impact of removing augmented representations on test accuracy in the Fisher-Guided GMM Augmentation experiment. At the beginning, removing augmented graphs with low or negative influence scores improves generalization.  The highest test accuracy is reached when a significant portion of low-quality augmentations has been removed while retaining high-influence ones. This indicates that a well-selected augmentation subset enhances model performance. As more augmentations are removed, the overall diversity of the training set decreases. Since data augmentation generally helps the model generalize better, excessive removal reduces its effectiveness. At 100\% removal, augmentation is entirely disabled, meaning the model is trained only on the original dataset, i.e., the reference case, leading to a significant drop in accuracy.

\section{Softmax Confidence and Entropy Distributions}\label{gen:app:softmax_saturation}
\begin{figure*}[ht]
    \centering
    \includegraphics[width=\textwidth]{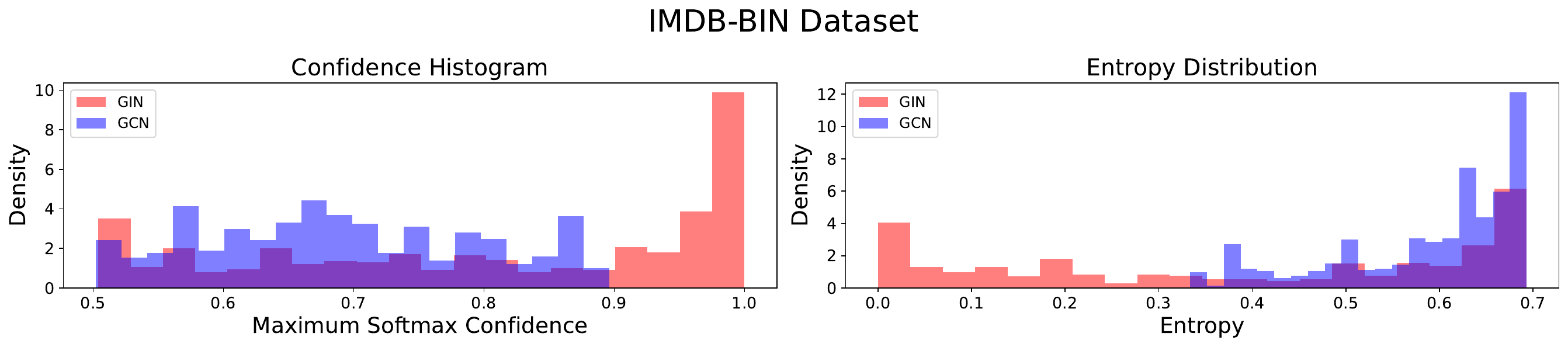}
    \caption[Softmax Confidence and Entropy Distributions for IMDB-BIN]{Softmax Confidence and Entropy Distributions for the IMDB-BIN Dataset.} 
    \label{gen:fig:saturation_IMDB_B}
\end{figure*}

\begin{figure*}[ht]
    \centering
    \includegraphics[width=\textwidth]{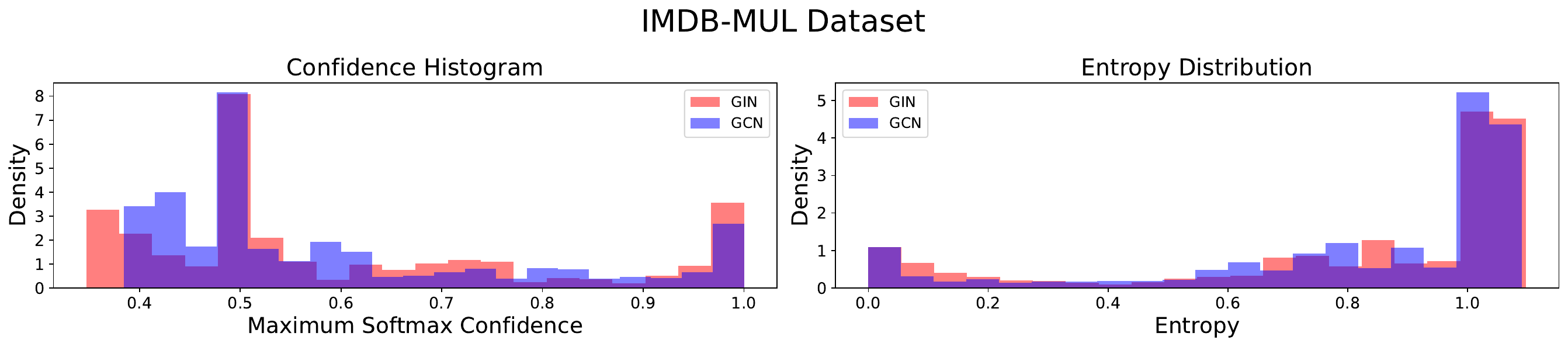}
    \caption[Softmax Confidence and Entropy Distributions for IMDB-MUL]{Softmax Confidence and Entropy Distributions for the IMDB-MUL Dataset.} 
    \label{gen:fig:saturation_IMDB_M}
\end{figure*}

\begin{figure*}[ht]
    \centering
    \includegraphics[width=\textwidth]{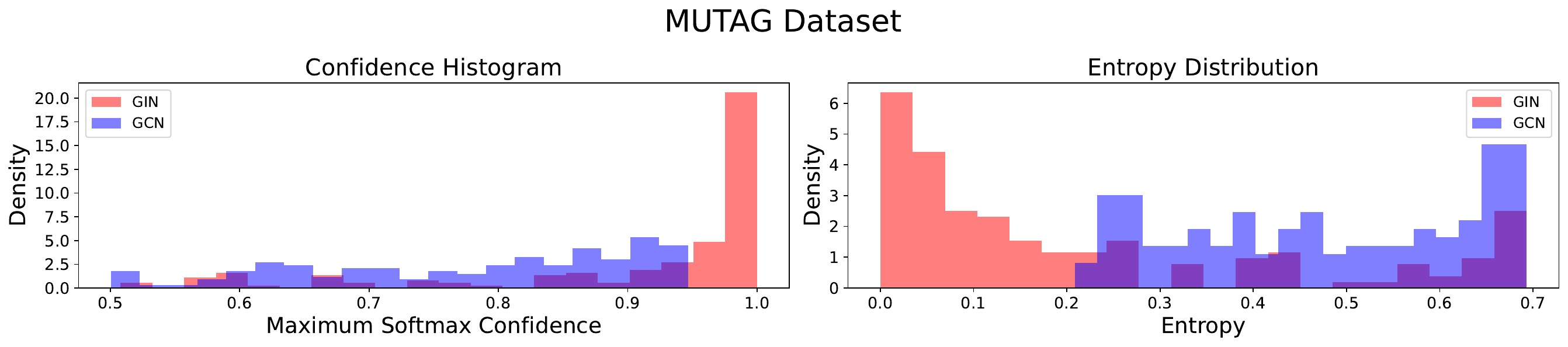}
    \caption[Softmax Confidence and Entropy Distributions for MUTAG]{Softmax Confidence and Entropy Distributions for the MUTAG Dataset.} 
    \label{gen:fig:saturation_MUTAG}
\end{figure*}

\begin{figure*}[ht]
    \centering
    \includegraphics[width=\textwidth]{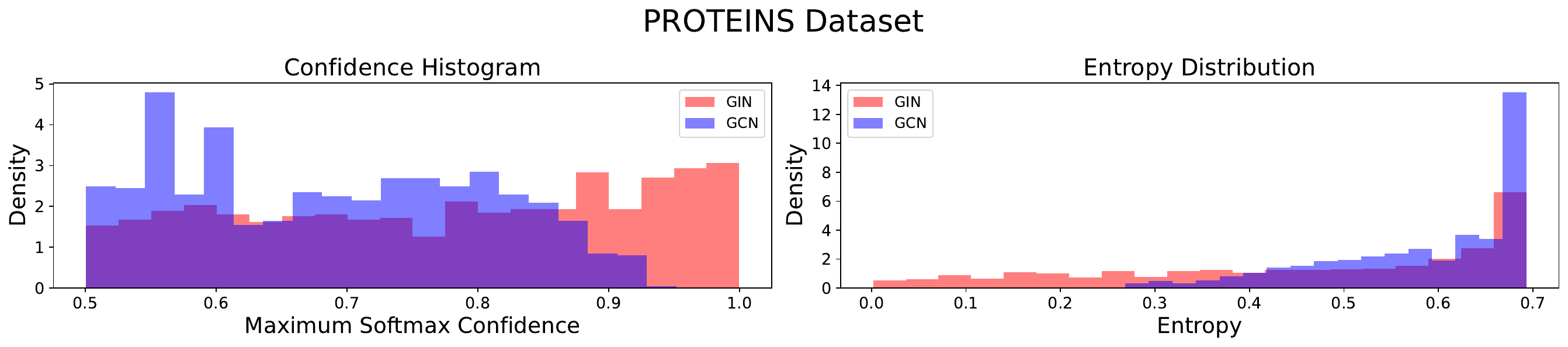}
    \caption[Softmax Confidence and Entropy Distributions for PROTEINS]{Softmax Confidence and Entropy Distributions for the PROTEINS Dataset.} 
    \label{gen:fig:saturation_PROTEINS}
\end{figure*}

\begin{figure*}[h]
    \centering
    \includegraphics[width=\textwidth]{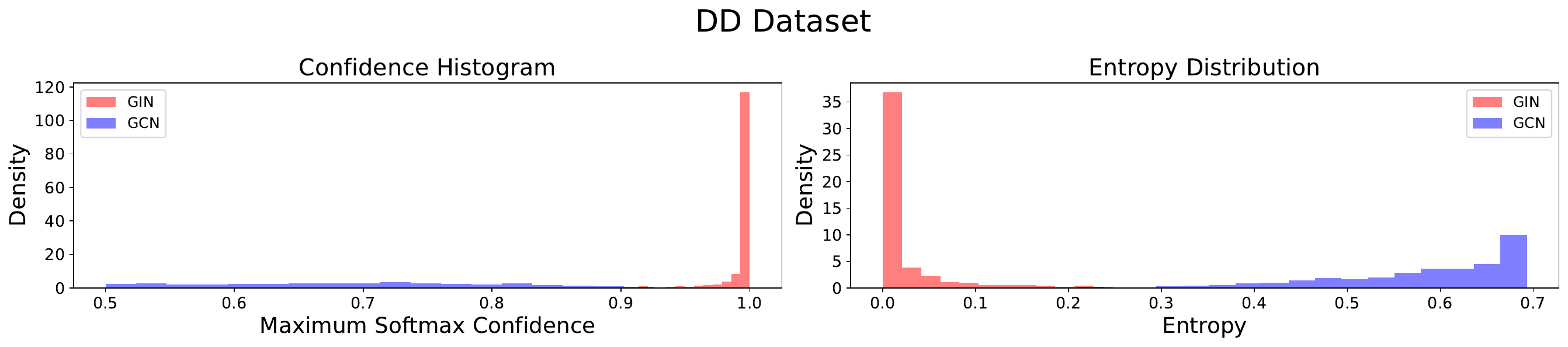}
    \caption[Softmax Confidence and Entropy Distributions for DD]{Softmax Confidence and Entropy Distributions for the DD Dataset.} 
    \label{gen:fig:saturation_DD}
\end{figure*}

One of the critical challenges in training GNN is Softmax saturation, where the model produces confident predictions. This high confidence leads to vanishing gradients, making the influence score of the augmented graphs converge to zero. To analyze this phenomenon, we examine the Softmax confidence and entropy distributions for GCN and GIN across the different graph classification datasets. In Figure \ref{gen:fig:saturation_IMDB_B}, \ref{gen:fig:saturation_IMDB_M}, \ref{gen:fig:saturation_MUTAG}, \ref{gen:fig:saturation_PROTEINS} and \ref{gen:fig:saturation_DD}, we present histograms illustrating the distribution of maximum Softmax confidence and entropy for models trained on the original DD, IMDB-BIN, IMDB-MUL, MUTAG, and PROTEINS datasets. Each dataset panel contains two histograms, the Confidence Histogram and the Entropy Distribution.  The \textit{Confidence Histogram} illustrates the distribution of the maximum Softmax confidence scores assigned by the model to the predicted class. A higher confidence value indicates that the model is more certain about its classification decision. The \textit{Entropy Distribution}  provides a measure of uncertainty in the model's predictions, computed as: $H(y) = -\sum_{c} y_c ~\log(y_c)$, where, for each class $c$, $y_c$ is the predicted probability corresponding to this class. Lower entropy values reflect high-confidence predictions and higher entropy values indicate more significant uncertainty. 

Specifically, in the DD dataset, we observe an extreme case of Softmax saturation in GIN, where almost all predictions collapse to near-maximum confidence.  The entropy histogram further reinforces the Softmax saturation in GIN on DD, where entropy values are heavily skewed towards zero, meaning the model rarely assigns significant probability mass outside the predicted class. Compared to other dataset-model settings, this effect is particularly pronounced in GIN trained on DD. In contrast, GCN exhibits a wider confidence distribution, maintaining a more balanced uncertainty. 

\myemptypage
\end{document}